\documentclass[11pt]{article}
\usepackage[a4paper, total={6.5in, 9in}]{geometry}

\usepackage[american]{babel}

\usepackage{amsmath,amsfonts,amssymb,amsthm}
\usepackage{subfiles}
\usepackage{enumitem, tabularx, booktabs, ragged2e}
\usepackage{listings, lstautogobble}
\usepackage{alltt}
\usepackage{tikz}
\usepackage{tabularx}
\usepackage{xcolor, soul}
\usepackage{blkarray}
\usepackage[bottom]{footmisc}
\usepackage{rotating}
\usepackage{wasysym}
\usepackage{rotating}
\usepackage{graphicx}
\graphicspath{{images/}{./img/}}
\usepackage{verbatimbox}
\usepackage{tablefootnote}
\usepackage{algorithm, algpseudocode}
\newcolumntype{L}{>{\centering\arraybackslash}m{1cm}}
\usepackage[colorlinks=true,citecolor=codegreen,linkcolor=darkbrown,urlcolor=blue,breaklinks]{hyperref}
\usepackage{cleveref}
\usepackage{diagbox}

\usepackage{natbib} % has a nice set of citation styles and commands
\usepackage{bibentry}

\usepackage[toc,page,header]{appendix}
\usepackage{minitoc}

\usepackage{xcolor}
\definecolor{orange}{rgb}{1,0.5,0}
\definecolor{blue}{rgb}{0.22, 0.58, 0.82}
\definecolor{green}{rgb}{0.2, 0.65, 0.32}
\definecolor{red}{rgb}{0.91, 0.26, 0.2}
\definecolor{purple}{rgb}{0.46, 0.21, 0.68}
\definecolor{bluegray}{rgb}{0.04,0,0.7}
\definecolor{darkbrown}{rgb}{0.40,0.2,0.05}
\definecolor{forestgreen}{RGB}{34, 139, 34}
\makeatletter
\def\mathcolor#1#{\@mathcolor{#1}}
\def\@mathcolor#1#2#3{%
	\protect\leavevmode
	\begingroup
	\color#1{#2}#3%
	\endgroup
}
\makeatother

\makeatletter
\def\namedlabel#1#2{\begingroup
    #2%
    \def\@currentlabel{#2}%
    \phantomsection\label{#1}\endgroup
}
\makeatother

\allowdisplaybreaks

\usetikzlibrary{arrows.meta, arrows}

\usepackage[labelformat=simple]{subcaption}

\usetikzlibrary{automata,positioning}

\newcommand{\abs}[1]{\left\lvert#1\right\rvert}

\DeclareMathOperator*{\argmax}{argmax} % no space, limits underneath in displays

\newcommand{\G}[1][G]{\mathcal{#1}}
\newcommand{\g}[1][G]{\mathcal{#1}}

\newcommand{\incomp}{\not\lessgtr}
\newcommand{\bV}{\mathbf{V}}
\newcommand{\bE}{\mathbf{E}}
\theoremstyle{plain}% default
\newtheorem{innercustomgeneric}{\customgenericname}
\providecommand{\customgenericname}{}

\definecolor{codegreen}{rgb}{0,0.6,0}
\definecolor{codegray}{rgb}{0.5,0.5,0.5}
\definecolor{codepurple}{rgb}{0.58,0,0.82}
\definecolor{backcolour}{rgb}{0.95,0.95,0.92}
\lstdefinestyle{mystyle}{
    backgroundcolor=\color{backcolour},   
    commentstyle=\color{codegreen},
    keywordstyle=\color{magenta},
    numberstyle=\tiny\color{codegray},
    stringstyle=\color{codepurple},
    basicstyle=\ttfamily\footnotesize,
    breakatwhitespace=false,         
    breaklines=true,                 
    captionpos=b,                    
    keepspaces=true,                 
    numbers=left,                    
    numbersep=5pt,                  
    showspaces=false,                
    showstringspaces=false,
    showtabs=false,                  
    tabsize=2
}
\newcommand{\BlackBox}{\rule{1.5ex}{1.5ex}}  % end of proof
\ifdefined\proof
    \renewenvironment{proof}{\par\noindent{\bf Proof\ }}{\hfill\BlackBox\\[2mm]}
\else
    \newenvironment{proof}{\par\noindent{\bf Proof\ }}{\hfill\BlackBox\\[2mm]}
\fi
\newtheorem{example}{Example} 
\newtheorem{theorem}{Theorem}
\newtheorem{lemma}[theorem]{Lemma} 
\newtheorem{proposition}[theorem]{Proposition} 
\newtheorem{remark}[theorem]{Remark}
\newtheorem{corollary}[theorem]{Corollary}
\newtheorem{definition}[theorem]{Definition}

\newenvironment{proofsketch}[1][]{\begin{trivlist}
   \item[\hskip \labelsep {\bfseries Proof Sketch of #1.}]}{\hfill{}$\BlackBox$\end{trivlist}}
      \newenvironment{proofof}[1][]{\begin{trivlist}
   \item[\hskip \labelsep {\bfseries Proof of #1.}]}{\hfill{}$\BlackBox$\end{trivlist}}

\newcommand{\skipitems}[1]{%
  \addtocounter{\@enumctr}{#1}%
}
\definecolor{blue-violet}{rgb}{0.54, 0.17, 0.89}
\definecolor{antiquefuchsia}{rgb}{0.57, 0.36, 0.51}
\definecolor{amethyst}{rgb}{0.6, 0.4, 0.8}
\definecolor{blue-violet}{rgb}{0.54, 0.17, 0.89}
\definecolor{ao}{rgb}{0.0, 0.5, 0.0}
\definecolor{blue(ncs)}{rgb}{0.0, 0.53, 0.74}
\definecolor{dgreen}{rgb}{0.12, 0.3, 0.17}
\definecolor{cadmiumgreen}{rgb}{0.0, 0.42, 0.24}
\definecolor{darkolivegreen}{rgb}{0.33, 0.42, 0.18}
\definecolor{dartmouthgreen}{rgb}{0.05, 0.5, 0.06}

\newcommand{\tild}{\raise.17ex\hbox{ $\scriptstyle\sim$ }}

\DeclareMathOperator{\De}{De}

\DeclareMathOperator{\An}{An}
\DeclareMathOperator{\PossAn}{PossAn}
\DeclareMathOperator{\PossDe}{PossDe}
\DeclareMathOperator{\Pa}{Pa}
\DeclareMathOperator{\pa}{pa}
\DeclareMathOperator{\Adj}{Adj}

\newcommand{\mb}[1]{\mathbf{#1}}
\newcommand{\mc}[1]{\mathcal{#1}}

\newcommand{\msepp}{\perp_{m}}

\newcommand{\vars}[1][V]{\mathbf{#1}}
\newcommand{\e}[1][E]{\mathbf{#1}}

\newcommand{\pstar}[1][p]{{#1}^{*}}

\newcommand{\bulletbullet}{
  \setlength{\unitlength}{1mm}
  \begin{picture}(5,1)(0,0)
    \put(0.2,0){$\bullet$}
    \put(1.3,1){\line(1,0){2.4}}
    \put(2.8,0){$\bullet$}
  \end{picture}
}

\newcommand{\bulletcirc}{
  \setlength{\unitlength}{1mm}
  \begin{picture}(5,1)(0,0)
    \put(0.2,0){$\bullet$}
    \put(1.1, 1){\line(1,0){2.4}}
    \put(4, 1){\circle{1}}
  \end{picture}
}

\newcommand{\circbullet}{
  \setlength{\unitlength}{1mm}
  \begin{picture}(5,1)(0,0)
    \put(1,1){\circle{1}}
    \put(1.5,1){\line(1,0){2.4}}
    \put(2.9,0){$\bullet$}
  \end{picture}
}

\newcommand{\bulletarrow}{
  \setlength{\unitlength}{1mm}
  \begin{picture}(5,1)(0,0)
    \put(0.2,0){$\bullet$}
    \put(1,0){$\rightarrow$}
  \end{picture}
}

\newcommand{\arrowbullet}{
  \setlength{\unitlength}{1mm}
  \begin{picture}(5,1)(0,0)
    \put(0.2,0){$\leftarrow$}
    \put(3,0){$\bullet$}
  \end{picture}
}

\newcommand{\circarrow}{
  \setlength{\unitlength}{1mm}
  \begin{picture}(5,1)(0,0)
    \put(1,1){\circle{1}}
    \put(1.2,0){$\rightarrow$}
  \end{picture}
}

\newcommand{\arrowcirc}{
  \setlength{\unitlength}{1mm}
  \begin{picture}(5,1)(0,0)
    \put(0.2,0){$\leftarrow$}
    \put(4.3,1){\circle{1}}
  \end{picture}
}

\newcommand{\circcirc}{
  \setlength{\unitlength}{1mm}
  \begin{picture}(5,1)(0,0)
    \put(1,1){\circle{1}}
    \put(1.5,1){\line(1,0){2}}
    \put(4,1){\circle{1}}
  \end{picture}
}

\definecolor{blue}{RGB}{70, 130, 180}
\definecolor{orange}{rgb}{1,0.5,0}
\definecolor{green}{rgb}{0.2, 0.65, 0.32}
\definecolor{red}{rgb}{0.91, 0.26, 0.2}
\definecolor{purple}{rgb}{0.46, 0.21, 0.68}
\definecolor{bluegray}{rgb}{0.04,0,0.7}
\definecolor{darkbrown}{rgb}{0.40,0.2,0.05}
\definecolor{forestgreen}{RGB}{34, 139, 34}

\title{Towards Complete Causal Explanation with Expert Knowledge}
\author{Aparajithan Venkateswaran\thanks{Work done while at University of Washington}\ (apara.vnkat@gmail.com) \\ Microsoft\\
              Redmond, WA 98052, USA \and 
       Emilija Perkovi\'c  (perkovic@uw.edu) \\
        Department of Statistics\\
              University of Washington\\
              Seattle, WA 98195, USA}
\date{}

\begin{document}

\maketitle

\begin{abstract}
We study the problem of restricting a Markov equivalence class of maximal ancestral graphs (MAGs) to only those MAGs that contain certain edge marks, which we refer to as expert or orientation knowledge. Such a restriction of the Markov equivalence class can be uniquely represented by a \textit{restricted essential ancestral graph}.
Our contributions are several-fold. First, we prove certain properties for the entire Markov equivalence class including a conjecture from \cite{ali2009markov}. Second, we present several new sound graphical orientation rules for adding orientation knowledge to an essential ancestral graph. We also show that some orientation rules of \cite{zhang2008completeness} are not needed for restricting the Markov equivalence class with orientation knowledge. Third, we provide an algorithm for including this orientation knowledge and show that in certain settings the output of our algorithm is a \textit{restricted essential ancestral graph}. Finally, outside of the specified settings, we provide an algorithm for checking whether a graph is a restricted essential graph and discuss its runtime. This work can be seen as a generalization of \cite{meek1995causal} to settings which allow for latent confounding.
\end{abstract}

\section{Introduction}

\begin{figure}[!t]
\tikzstyle{every edge}=[draw,>=stealth',->,thick]
\newcommand\dagvariant[1]{\begin{tikzpicture}[xscale=.4,yscale=0.4]
\node (d) at (0,0) {};
\node (a) at (0,2) {};
\node (c) at (2,0) {};
\node (b) at (2,2) {};
\draw (a) edge [-] (b);
\draw (a) edge [-] (c);
\draw (d) edge [-] (a);
\draw (d) edge [-] (c);
\draw (a) edge [-] (c);
\draw #1;
\end{tikzpicture}}

\tikzstyle{every edge}=[draw,>=stealth',->,thick]
\newcommand\dagempty[1]{\begin{tikzpicture}[xscale=.4,yscale=0.4]
\node (d) at (0,0) {};
\node (a) at (0,.5) {};
\node (c) at (2,0) {};
\node (b) at (2,.5) {};
\draw #1;
\end{tikzpicture}}

\centering
\begin{subfigure}{.23\textwidth}
\tikzstyle{every edge}=[draw,>=stealth',->]
  \centering
\begin{tikzpicture}[->,>=latex,shorten >=1pt,auto,node distance=0.8cm,scale=.8,transform shape]
  \tikzstyle{state}=[inner sep=1pt, minimum size=12pt]
\tikzstyle{every edge}=[draw,>=stealth',->]
  % rule 1
  \node[state] (D) at (0,0) {\Large $D$};
  \node[state] (A) at (0,2) {\Large $A$};
  \node[state] (C) at (2,0) {\Large  $C$};
  \node[state] (B) at (2,2) {\Large  $B$};
  \node[state] (L1) at (1,3) {\Large  $L_1$};
  \node[state] (L2) at (-1,1) {\Large  $L_2$};

  \draw (L1) edge [->] (B);
\draw (L1) edge [->] (A);
  \draw (A) edge [<-] (B);
\draw 		(A) edge  [->] (C);
 \draw   	(B) edge [->] (C);
 \draw    	(C) edge [->] (D);
\draw    	(A) edge [->] (L2);
\draw    	(L2) edge [->] (D);
\end{tikzpicture}
\caption{}
  \label{mpdag1}
\end{subfigure}
\unskip
\vrule
\hspace{0.2cm}
\begin{subfigure}{.23\textwidth}
\vspace{1cm}
\tikzstyle{every edge}=[draw,>=stealth',->]
  \centering
\begin{tikzpicture}[->,>=latex,shorten >=1pt,auto,node distance=0.8cm,scale=.8,transform shape]
  \tikzstyle{state}=[inner sep=1pt, minimum size=12pt]
\tikzstyle{every edge}=[draw,>=stealth',->]
  % rule 1
  \node[state] (D) at (0,0) {\Large $D$};
  \node[state] (A) at (0,2) {\Large $A$};
  \node[state] (C) at (2,0) {\Large  $C$};
  \node[state] (B) at (2,2) {\Large  $B$};

  \draw (A) edge [<-] (B);
\draw 		(A) edge  [->] (C);
 \draw   	(B) edge [->] (C);
 \draw    	(C) edge [->] (D);
\draw    	(A) edge [->] (D);
\end{tikzpicture}
\caption{}
  \label{mpdag2}
\end{subfigure}
%\unskip
\vrule
\hspace{0.2cm}
\begin{subfigure}{.23\textwidth}
\tikzstyle{every edge}=[draw,>=stealth',->]
  \centering
\begin{tikzpicture}[->,>=latex,shorten >=1pt,auto,node distance=0.8cm,scale=.8,transform shape]
  \tikzstyle{state}=[inner sep=1pt, minimum size=12pt]
\tikzstyle{every edge}=[draw,>=stealth',->]
  % rule 1
  \node[state] (D) at (0,0) {\Large $D$};
  \node[state] (A) at (0,2) {\Large $A$};
  \node[state] (C) at (2,0) {\Large  $C$};
  \node[state] (B) at (2,2) {\Large  $B$};

  \draw (A) edge [o-o,color=gray] (B);
\draw 		(A) edge  [o-o,color=gray] (C);
 \draw   	(B) edge [o-o,color=gray] (C);
 \draw    	(C) edge [o-o,color=gray] (D);
\draw    	(A) edge [o-o,color=gray] (D);
\end{tikzpicture}
\caption{}
  \label{mpdag3}
  \end{subfigure}
  \vrule
\hspace{0.2cm}
\begin{subfigure}{.23\textwidth}
\vspace{1cm}
\tikzstyle{every edge}=[draw,>=stealth',->]
  \centering
\begin{tikzpicture}[->,>=latex,shorten >=1pt,auto,node distance=0.8cm,scale=.8,transform shape]
  \tikzstyle{state}=[inner sep=1pt, minimum size=12pt]

  % rule 1
  \node[state] (D) at (0,0) {\Large $D$};
  \node[state] (A) at (0,2) {\Large $A$};
  \node[state] (C) at (2,0) {\Large  $C$};
  \node[state] (B) at (2,2) {\Large  $B$};

  \draw (A) edge [o-o,color=gray] (B);
\draw 		(A) edge  [o-o,color=gray] (C);
 \draw   	(B) edge [o->,line width=1pt,color=black] (C);
 \draw    	(C) edge [->,line width=1pt,color=blue] (D);
\draw    	(A) edge [->,line width=1pt,color=blue] (D);
\end{tikzpicture}
\caption{}
\end{subfigure}
\begin{subfigure}{\textwidth}
\vspace{1cm}
\tikzstyle{every edge}=[draw,>=stealth',->,thick]
\begin{center}
\dagvariant{
(a)  edge [<->, color = gray]  (b)
(a)  edge [->, color = blue]  (d)
(a) edge [<->, color = gray] (c)
(c) edge [->, color = blue] (d)
(c) edge [<->, color = black] (b)
}
\dagvariant{
(a)  edge [<->, color = gray]  (b)
(a)  edge [->, color = blue]  (d)
(a) edge [<-, color = gray] (c)
(c) edge [->, color = blue] (d)
(c) edge [<->, color = black] (b)
}
\dagvariant{
(a)  edge [<-, color = gray]  (b)
(a)  edge [->, color = blue]  (d)
(a) edge [<->, color = gray] (c)
(c) edge [->, color = blue] (d)
(c) edge [<->, color = black] (b)
}
\dagvariant{
(a)  edge [<-, color = gray]  (b)
(a)  edge [->, color = blue]  (d)
(a) edge [<-, color = gray] (c)
(c) edge [->, color = blue] (d)
(c) edge [<->, color = black] (b)
}
\dagvariant{
(a)  edge [->, color = gray]  (b)
(a)  edge [->, color = blue]  (d)
(a) edge [->, color = gray] (c)
(c) edge [->, color = blue] (d)
(c) edge [<->, color = black] (b)
}
\dagvariant{
(a)  edge [<->, color = gray]  (b)
(a)  edge [->, color = blue]  (d)
(a) edge [->, color = gray] (c)
(c) edge [->, color = blue] (d)
(c) edge [<->, color = black] (b)
}
\dagvariant{
(a)  edge [->, color = gray]  (b)
(a)  edge [->, color = blue]  (d)
(a) edge [<->, color = gray] (c)
(c) edge [->, color = blue] (d)
(c) edge [<->, color = black] (b)
}
\dagvariant{
(a)  edge [->, color =gray]  (b)
(a)  edge [->, color = blue]  (d)
(a) edge [->, color =gray] (c)
(c) edge [->, color = blue] (d)
(b) edge [->, color = black] (c)
} 
\dagvariant{
(a)  edge [<-, color = gray]  (b)
(a)  edge [->, color = blue]  (d)
(a) edge [->, color =gray] (c)
(c) edge [->, color = blue] (d)
(b) edge [->, color = black] (c)
}
\dagvariant{
(a)  edge [<-, color = gray]  (b)
(a)  edge [->, color = blue]  (d)
(a) edge [<-, color = gray] (c)
(c) edge [->, color = blue] (d)
(b) edge [->, color = black] (c)
}
\dagvariant{
(a)  edge [<->, color = gray]  (b)
(a)  edge [->, color = blue]  (d)
(a) edge [->, color = gray] (c)
(c) edge [->, color = blue] (d)
(c) edge [<-, color = black] (b)
}
\dagvariant{
(a)  edge [<->, color = gray]  (b)
(a)  edge [->, color = blue]  (d)
(a) edge [<->, color = gray] (c)
(c) edge [->, color =blue] (d)
(c) edge [<-, color = black] (b)
}
\dagvariant{
(a)  edge [<-, color = gray]  (b)
(a)  edge [->, color = blue]  (d)
(a) edge [<->, color = gray] (c)
(c) edge [->, color = blue] (d)
(c) edge [<-, color = black] (b)
}\\
\end{center}
\begin{center}
\dagvariant{
(a)  edge [->, color = gray]  (b)
(d)  edge [->, color = gray]  (a)
(a) edge [->, color = gray] (c)
(d) edge [->, color = gray] (c)
(c) edge [->, color = gray] (b)
}
\dagvariant{
(a)  edge [->, color = gray]  (b)
(a)  edge [->, color = gray]  (d)
(a) edge [->, color = gray] (c)
(d) edge [->, color = gray] (c)
(c) edge [->, color = gray] (b)
}
\dagvariant{
(a)  edge [->, color = gray]  (b)
(a)  edge [->, color = gray]  (d)
(a) edge [->, color = gray] (c)
(c) edge [->, color = gray] (d)
(c) edge [->, color = gray] (b)
}
\dagvariant{
(a)  edge [->, color = gray]  (b)
(d)  edge [->, color = gray]  (a)
(c) edge [->, color = gray] (a)
(d) edge [->, color = gray] (c)
(c) edge [->, color = gray] (b)
}
\dagvariant{
(a)  edge [->, color = gray]  (b)
(d)  edge [->, color = gray]  (a)
(c) edge [->, color = gray] (a)
(c) edge [->, color = gray] (d)
(c) edge [->, color = gray] (b)
}
\dagvariant{
(a)  edge [->, color = gray]  (b)
(a)  edge [->, color = gray]  (d)
(c) edge [->, color = gray] (a)
(c) edge [->, color = gray] (d)
(c) edge [->, color = gray] (b)
}
\dagvariant{
(b)  edge [->, color = gray]  (a)
(a)  edge [->, color = gray]  (d)
(c) edge [->, color = gray, color = gray] (a)
(c) edge [->, color = gray] (d)
(c) edge [->, color = gray] (b)
}
\dagvariant{
(a)  edge [<->, color = gray]  (b)
(a)  edge [->, color = gray]  (d)
(a) edge [<->, color = gray] (c)
(c) edge [->, color = gray] (d)
(c) edge [->, color = gray] (b)
}
\dagvariant{
(a)  edge [<->, color = gray]  (b)
(a)  edge [->, color = gray]  (d)
(a) edge [<-, color = gray] (c)
(c) edge [->, color = gray] (d)
(c) edge [->, color = gray] (b)
}
\dagvariant{
(a)  edge [<-, color = gray]  (b)
(a)  edge [->, color = gray]  (d)
(a) edge [<-, color = gray] (c)
(c) edge [->, color = gray] (d)
(c) edge [->, color = gray] (b)
}
\dagvariant{
(a)  edge [<-, color = gray]  (b)
(a)  edge [->, color = gray]  (d)
(a) edge [<-, color = gray] (c)
(c) edge [->, color = gray] (d)
(c) edge [<->, color = gray] (b)
}
\dagvariant{
(a)  edge [->, color = gray]  (b)
(a)  edge [<->, color = gray]  (d)
(a) edge [<->, color = gray] (c)
(c) edge [<->, color = gray] (d)
(c) edge [->, color = gray] (b)
}
\dagvariant{
(a)  edge [->, color = gray]  (b)
(a)  edge [<-, color = gray]  (d)
(a) edge [<->, color = gray] (c)
(c) edge [<->, color = gray] (d)
(c) edge [->, color = gray] (b)
}
\\
\end{center}
\begin{center}
\dagvariant{
(a)  edge [->, color = gray]  (b)
(a)  edge [<-, color = gray]  (d)
(a) edge [<->, color = gray] (c)
(c) edge [<-, color = gray] (d)
(c) edge [->, color = gray] (b)
}
\dagvariant{
(a)  edge [->, color = gray]  (b)
(a)  edge [<->, color = gray]  (d)
(a) edge [<->, color = gray] (c)
(c) edge [->, color = gray] (d)
(c) edge [->, color = gray] (b)
}
\dagvariant{
(a)  edge [->, color = gray]  (b)
(a)  edge [<->, color = gray]  (d)
(a) edge [<-, color = gray] (c)
(c) edge [<->, color = gray] (d)
(c) edge [->, color = gray] (b)
}
\dagvariant{
(a)  edge [->, color = gray]  (b)
(a)  edge [<-, color = gray]  (d)
(a) edge [<-, color = gray] (c)
(c) edge [<->, color = gray] (d)
(c) edge [->, color = gray] (b)
}
\dagvariant{
(a)  edge [->, color = gray]  (b)
(a)  edge [<->, color = gray]  (d)
(a) edge [<-, color = gray] (c)
(c) edge [->, color = gray] (d)
(c) edge [->, color = gray] (b)
}
\dagvariant{
(a)  edge [->, color = gray]  (b)
(a)  edge [->, color = gray]  (d)
(a) edge [<->, color = gray] (c)
(c) edge [<->, color = gray] (d)
(c) edge [->, color = gray] (b)
}
\dagvariant{
(a)  edge [->, color = gray]  (b)
(a)  edge [->, color = gray]  (d)
(a) edge [<->, color = gray] (c)
(c) edge [->, color = gray] (d)
(c) edge [->, color = gray] (b)
}
\dagvariant{
(a)  edge [->, color = gray]  (b)
(a)  edge [<->, color = gray]  (d)
(a) edge [->, color = gray] (c)
(c) edge [<->, color = gray] (d)
(c) edge [->, color = gray] (b)
}
\dagvariant{
(a)  edge [->, color = gray]  (b)
(a)  edge [->, color = gray]  (d)
(a) edge [->, color = gray] (c)
(c) edge [<->, color = gray] (d)
(c) edge [->, color = gray] (b)
}\\
\end{center}
\caption{}
\end{subfigure} 
\caption{(a) DAG $\g[D]$, (b) MAG $\g[M]$, (c)  essential ancestral graph $\g$, (d) restricted essential ancestral graph $\g^{\prime}$, and  (e) the Markov equivalence class of MAG $\g[M]$.} 
\label{fig:example_MEC_MAG}
\end{figure}
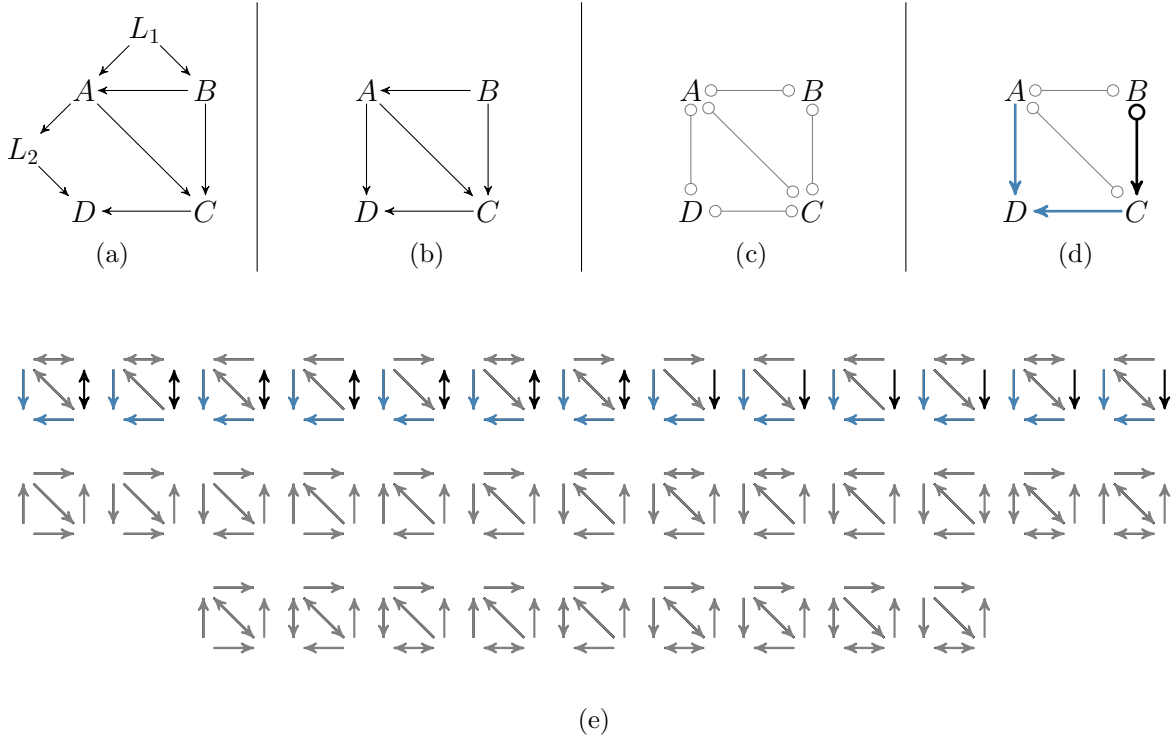

We consider proper restrictions of a Markov equivalence class of maximal ancestral graphs (MAGs). MAGs are probabilistic and causal graphical models on sets of observed random variables when certain variables in the causal system are unobserved. An example MAG $\g[M]$ is given in Figure \ref{fig:example_MEC_MAG}(b). Nodes in  $\g[M]$ index random variables and edges represent causal and probabilistic relationships between the variables (see definitions in Section \ref{sec:prelim}). MAG $\g[M]$ represents, for instance, the directed acyclic graph (DAG) $\g[D]$ in Figure \ref{fig:example_MEC_MAG}(a), where variables $X_{L_1}$ and $X_{L_2}$ are unobserved. $\g[M]$ is a simple graph that preserves causal (ancestral) relationships between the observed variables in $\g[D]$ \citep{richardson2002ancestral}. As a consequence of preserving causal relationships among observed variables while keeping a simple graph, a directed edge $B \to A$ in $\g[M]$ does not, generally, exclude the presence of unobserved confounding such as $A \leftarrow L_1 \to B$ in DAG $\g[D].$ We assume that the unobserved variables do not induce selection bias, so the MAGs we consider are mixed graphs containing directed ($\to$) and bidirected ($\leftrightarrow$) edges \citep{zhang2008causal}.

MAGs additionally preserve graphical separation relationships (m-separations,  \citealp{richardson2002ancestral}) between the observed variables in the underlying DAG model. Under certain assumptions, these m-separations can be interpreted as conditional independence relationships between the variables represented by the nodes. All MAGs representing the same set of m-separations form a  Markov equivalence class. For instance, the Markov equivalence class of $\g[M]$ is given in \ref{fig:example_MEC_MAG}(e).  
Any Markov equivalence class of MAGs can be uniquely represented by a partial mixed graph which we refer to as an essential ancestral graph \citep{zhang2008completeness}. 
 An essential ancestral graph $\g$ representing the Markov equivalence class in Figure \ref{fig:example_MEC_MAG}(e) is given in Figure \ref{fig:example_MEC_MAG}(c).
Generally, an essential ancestral graph may contain edges of the form $\circcirc$, $\circarrow$ in addition to $\to$ and $\leftrightarrow$. The circle edge mark, $\circ$, on an edge $A \circcirc B$ indicates that we are unsure whether $X_A$ causes $X_B$ ($A \to B$) or $X_A$ does not cause $X_B$ ($A \leftarrow B$ or $A \leftrightarrow B$). An edge of the form $A \circarrow B$ in an essential ancestral graph indicates that $X_B$ does not cause $X_A$, but we are unsure whether $X_A$ causes $X_B$ ($A \to B$), or $X_A$ does not cause $X_B$ ($A \leftrightarrow B$). Hence, causal relationships are not identified in $\g$. 

Under certain assumptions, we can learn an essential ancestral graph from conditional independence constraints present in data through a causal discovery algorithm  \cite[e.g.,][]{spirtes2000causation,zhang2008completeness,Colombo2012,ClaassenMooijHeskes13,triantafillou2016score,ogarrio2016hybrid,tsirlis2018scoring,bernstein2020ordering,rantanen2021maximal}. Subsequently, we can try to estimate a causal effect by using the learned essential ancestral graph \citep{tian2002general,tian2003studies,huang2006pearl,shpitser2008complete, maathuis2015generalized,perkovic2018complete,jaber2019causal}. However, as certain variables are unobserved and certain causal relationships may not be identified in the essential ancestral graph,  causal effect identification is often impossible in this setting.

Causal identification may be possible if one can restrict the Markov equivalence class to only certain member graphs. Hence, in this work, we consider using expert knowledge in the form of information on specific edge orientations (also called orientation knowledge), to obtain a proper restriction of the Markov equivalence class.  
For instance, consider again Figure \ref{fig:example_MEC_MAG} and suppose we have expert knowledge that $X_C$ does not cause $X_B$. This knowledge implies that $B \to C$ or $B \leftrightarrow C$ should be in the true MAG. Hence,  we want to restrict the Markov equivalence class in  \ref{fig:example_MEC_MAG}(e)  to only those MAGs that satisfy this expert knowledge.
The MAGs in Figure \ref{fig:example_MEC_MAG}(e) that satisfy either $B \to C$, or $B \leftrightarrow C$ are given in the first row of Figure \ref{fig:example_MEC_MAG}(e). Therefore, this orientation knowledge restricts the size of the Markov equivalence class from 35 to 13. 

This expert knowledge can be represented by adding $B \circarrow C$ to $\g$. Furthermore, in addition to containing $B \to C$, or $B \leftrightarrow C$, all MAGs in the first row of Figure \ref{fig:example_MEC_MAG}(e) also contain $A \to D$ and $C \to D$. A unique summary graph describing all invariant edge orientations in these MAGs is given in graph $\g^{\prime}$ in Figure \ref{fig:example_MEC_MAG}(d). We call $\g^{\prime}$ a restricted essential ancestral graph.  
Hence, simply adding orientation knowledge ($B \circarrow C$) to $\g$ is insufficient to identify the restricted essential ancestral graph due to the additional edge orientations implied by this knowledge.

Indeed, in the presence of latent variables, current causal discovery methods are either unable to fully utilize orientation knowledge to restrict the Markov equivalence class \citep{andrews2020completeness} or are limited to only a small set of observed variables \citep{hyttinen2014constraint, hyttinen2015calculus,tikka2019identifying,tikka2021causal}. Some recent work on this topic has explored specific kinds of expert knowledge. For instance, local expert knowledge \citep{mooij2020joint,wang2022sound,wang2023estimating, wang2024new} where all circle edge marks incident to a particular node $A$ are specified by an expert. This knowledge can arise from having data on an experiment where an outside intervention sets the variable $X_A$ to a fixed value. Another line of work considers specific forms of tiered expert knowledge \citep{andrews2020completeness}, where an expert imposes a causal ordering between certain partitions of variables. Our work aims to consider a more flexible class of expert knowledge where information about edge mark orientations on existing edges can be specified. 
Furthermore, we would like our approach to be unrestricted by the size of the observed variable set. 

A similar line of work exists under the assumption of no latent variables. In this setting, causal discovery algorithms can be deployed to learn the essential graph representing the Markov equivalence class of DAGs  \citep[e.g.,][]{spirtes2000causation,Chickering02-optimal,TsamardinosEtAl06}. Similarly, a causal effect will not always be identifiable given an essential graph in this setting. Still, one can incorporate various kinds of expert knowledge to help improve causal identification \citep{meek1995causal,shimizu2006linear,hoyer08,hauserBuehlmann12,wang2017permutation,ernestroth2016}. The process of incorporating expert knowledge to obtain a restricted essential graph is well understood in this setting. Furthermore,  the addition of  expert knowledge may lead to more causal identification results \citep{perkovic17,perkovic20,smucler2020efficient,guo2021efficient,guo2021minimal,bang23a,laplante2023conditional,bang2025constraint,laplante2026identification,laplante2026data}. 
In terms of practical significance, causal discovery with expert knowledge has already been successfully applied to some real-world settings including studies on childhood obesity \citep{foraita2024longitudinal}, diabetes \citep{wang2020prior}, Alzheimer's pathophysiology \citep{shen2020challenges}, life-course epidemiology \citep{petersen2021data}, postoperative recovery \citep{lee2023causal}, and bird species abundance \citep{bystrova2024causal}. We expect our contributions will lead to more such applied work in the future.

The structure of the main text is as follows. Preliminaries are given in Section \ref{sec:prelim}. Section \ref{sec:eq-class} then reviews several existing Markov equivalence characterizations of MAGs. We reconcile these characterizations and prove a result previously conjectured by \citet{ali2009markov}. We also provide a new algorithm for constructing an essential ancestral graph corresponding to a given MAG in Algorithm \ref{alg:mag2pag} (\texttt{MAGtoEssentialAncestralGraph}). 
Then, in Section \ref{sec:bg-knowledge}, we define consistent expert knowledge, sound, and complete edge orientations. Section \ref{sec:new-rules} contains definitions of several new edge orientation rules needed in the presence of orientation knowledge. 
Section \ref{sec:completeness} then presents the \texttt{{addOrientationKnowledge}} algorithm (Algorithm \ref{alg:mpag}), which shows how to incorporate orientation knowledge. We, furthermore, prove certain properties of the restricted Markov equivalence class. In Section \ref{section:partial-completeness}, we show that Algorithm \ref{alg:mpag}  is complete in specific settings  (Theorems 
\ref{thm:chordal-completeness}, \ref{thm:3}, and \ref{thm:chordal-subgraph-completeness}). Outside of these settings, 
we provide algorithm \texttt{verifyCompleteness} (Algorithm \ref{alg:verify-completeness2}) in Section \ref{section:full-complete} which can verify whether a partial mixed graph is a restricted essential ancestral graph. Our theoretical results (Lemmas \ref{lemma:nocycle3} and \ref{lemma:no-new-mcps} and Theorem \ref{thm:chordal-completeness}), afford Algorithm \ref{alg:verify-completeness2} a faster runtime compared to a brute force approach. We discuss the specific runtime of Algorithm \ref{alg:verify-completeness2} through a simulation study in Section \ref{sec:simulation}. Our code is available in our \texttt{R} package, \texttt{expertOrientR}, on GitHub (\url{https://github.com/AparaV/expertOrientR}). Even though we obtain no general completeness results, our simulation study has not produced an example of incompleteness for the new set of edge orientation rules. We provide concluding remarks in Section \ref{sec:discussion}.

\section{Preliminaries}
\label{sec:prelim}

Some graphical preliminaries are deferred to supplement \ref{sec:supp-prelim}. 

\begin{figure}
        % \vspace{1cm}
   \centering
    \begin{subfigure}{.45\textwidth}
          \tikzstyle{every edge}=[draw,>=stealth',semithick]
        \vspace{1cm}
        \centering
        \begin{tikzpicture}[->,>=stealth',shorten >=1pt,auto,node distance=1.2cm,scale=0.75,transform shape,font = {\Large\bfseries\sffamily}]
	\tikzstyle{state}=[inner sep=5pt, minimum size=5pt]

	\node[state] (A) at (-4,0) {$A$};
	\node[state] (B) at (-2,0) {$B$};
	\node[state] (C) at (0,0) {$C$};
	\node[state] (D) at (2,0) {$D$};

	\draw[<->]  (A) edge (B);
	\draw[<->]  (B) edge (C);
	\draw[<->]  (C) edge (D);
    
   \draw[<-, out=45, in=135] (A) edge (C);
   \draw[<-, out=225, in=315] (D) edge (B);

  \end{tikzpicture}
\caption{}
% \label{fig:r13a} 
    \end{subfigure}
    \vrule
    \begin{subfigure}{.45\textwidth}
        \centering
       \tikzstyle{every edge}=[draw,>=stealth',semithick]
	  \begin{tikzpicture}[->,>=stealth',shorten >=1pt,auto,node distance=1.2cm,scale=0.75,transform shape,font = {\Large\bfseries\sffamily}]
	\tikzstyle{state}=[inner sep=5pt, minimum size=5pt]

	\node[state] (A) at (-4,0) {$A$};
	\node[state] (B) at (-2,0) {$B$};
	\node[state] (C) at (0,0) {$C$};
	\node[state] (D) at (2,0) {$D$};
    \node[state] (E) at (4,0) {$E$};
    \node[state] (F) at (4,2) {$F$};
    \node[state] (G) at (4,-2) {$G$};

	\draw[o->]  (A) edge (B);
	\draw[<->]  (B) edge (C);
	\draw[<->]  (C) edge (D);
	\draw[<->]  (D) edge (E);
	\draw[->]  (E) edge (F);
	\draw[<->]  (E) edge (G);

	\draw[->]  (B) edge (F);
	\draw[->]  (C) edge (F);
	\draw[->]  (D) edge (F);
    
	\draw[->]  (B) edge (G);
	\draw[->]  (C) edge (G);
	\draw[->]  (D) edge (G);

  \end{tikzpicture}
\caption{}
% \label{fig:r13b}
    \end{subfigure}
    
    \caption{(a) $\langle A, B, C, D \rangle$ is an inducing path. (b) $\langle A, B, C, D, E, F \rangle$ is a discriminating path where $E$ is not a discriminating collider. However, $\langle A, B, C, D, E, G \rangle$ is a discriminating path where $E$ is the discriminating collider.}
    \label{fig:def-examples}
\end{figure}

\textbf{Nodes and edges.} Graph $\g= (\vars,\e) $ consists of  nodes $ \vars=\left\lbrace V_{1},\dots,V_{p}\right\rbrace$ and edges $ \e $. We  consider simple graphs that contain at most one edge between any pair of nodes. Two nodes are \textit{adjacent} if they are connected by an edge. Every edge has two edge marks that are either an arrowhead, tail, or circle.  An arrowhead or tail edge marks are called \textit{invariant} and circle edge marks are called \textit{variant}. Edges can be \emph{directed} $\rightarrow$,  \emph{bi-directed}  $\leftrightarrow$, \emph{non-directed} $\circcirc$, or \emph{partially directed} $\circarrow$. We use $\bullet$ as a stand-in for any of the allowed edge marks. An edge is \textit{into} (\textit{out of}) a node $A$ if the edge has an arrowhead (tail) at $A$.

\textbf{Directed paths, possibly directed paths, and cycles.} Path $p = \langle V_1, \dots, V_k\rangle, k>1$ is  \textit{directed} from $V_1$ to $V_k$, if $V_i \to V_{i+1}$ is on $p$ for all $i \in \{1, \dots k -1\}$. Path $p$ is  \textit{possibly directed} from $V_1$ to $V_k$ if there is no edge $V_i \arrowbullet V_{j}$, for $1\le i < j \le k$ in $\g$. 
A directed path from $V_1$ to $V_k$ together with $V_k \to V_1$ forms a \emph{directed cycle} of \emph{length} $k$.
A directed path from $V_1$ to $V_k$ together with $V_k \bulletarrow V_1$ forms an \emph{almost directed cycle} of \emph{length} $k$.

\textbf{Ancestral relationships.} {If $A\to B$, then $A$ is a \textit{parent} of $B$, and $B$ is a child of $A$. If there is a (possibly) directed path from $A$ to $B$, then $A$ is an (\textit{possible}) \textit{ancestor} of $B$, and $B$ is a (\textit{possible}) \textit{descendant}  of $A$. We assume every node is a (possible) descendant and (possible) ancestor of itself. Sets of parents, descendants, ancestors, and adjacencies of $A$ in~$\g$ are denoted by $\Pa(A,\g)$, $\De(A,\g)$ and $\An(A,\g)$, $\Adj(A,\g)$ respectively.
Sets of possible descendants and possible ancestors of $A$ in $\g$ are denoted by $\PossDe(A,\g)$ and $\PossAn(A,\g)$.
For a set of nodes $\mathbf{A} \subseteq \mathbf{V}$, we let $\Pa(\mathbf{A},\g) = \cup_{A \in \mathbf{A}} \Pa(A,\g)$, with analogous definitions for $\Adj(\mb{A},\g)$, $\De(\mathbf{A},\g)$, $\An(\mathbf{A},\g)$, $\PossDe(\mathbf{A},\g)$ and $\PossAn(\mathbf{A},\g)$.}

\textbf{Definite status paths, collider paths.} If a path $p$ contains $V_i \bulletarrow V_j \arrowbullet V_k$ as a subpath, then $V_j$ is a \textit{collider} on $p$. A path $\langle V_{i},V_{j},V_{k} \rangle$ is an \emph{(un)shielded triple} if $ V_{i} $ and $ V_{k}$ are (not) adjacent. A path is \textit{unshielded} if all successive triples on the path are unshielded. A node $V_{j}$ is a \textit{definite non-collider} on a path $p$ if there is at least one edge out of $V_{j}$ on $p$, or if $V_{i} \bulletcirc V_j \circbullet V_k$ is a subpath of $p$ and $\langle V_i, V_j, V_k \rangle$ is an unshielded triple. 
A node is of \textit{definite status} on a path $p$ if it is a collider or a definite non-collider on $p$. Path $p$ is of definite status if every non-endpoint node on $p$ is of definite status \citep{zhang2008causal}.
 A \textit{collider path} $p$, is a path such that $|p|\ge 2$ and such that every non-endpoint node on $p$ is a collider. A collider path $p = \langle V_1, \dots, V_k \rangle, k \ge 3$  is called a \textit{minimal collider path} in $\g = (\mb{V,E})$, if $V_1 \notin \Adj(V_k, \g)$ and no subsequence of $p$ is also a collider path \citep{zhao2005markov}.

\textbf{Discriminating and inducing paths.} Path $p = \langle A, Q_1, \dots Q_k, B \rangle, k \ge 2$  is a \textit{discriminating path} \citep{zhang2008completeness} for $Q_k$ in $\g$ if  (i)  $p(A, Q_k)$ is a collider path in $\g$, and (ii) $A \notin \Adj(B, \g)$, and (iii) $Q_i \in \Pa(B, \g)$ for all  $i \in \{1, \dots, k-1 \}$.
If $p = \langle A, Q_1, \dots Q_k, B \rangle, k \ge 2$ is a discriminating path for $Q_k$ and $Q_k$ is a collider on $p$, we say that $p$ is a \textit{discriminating collider path} and that $Q_k$ is a \textit{collider discriminated by} path $p$.
A path $p = \langle A, Q_1, \dots , Q_k, B \rangle, k\ge 2$ is an \textit{inducing path} in a graph $\g$ if (i) $A \notin \Adj(B, \g)$, and (ii) $p$ is a collider path in $\g$, and
    (iii) $Q_i \in \An(\{A,B\}, \g)$, for all  $i \in \{1, \dots, k \}$. {We illustrate examples of these paths in Figure \ref{fig:def-examples}.}

\textbf{Blocking, d-separation, and m-separation.}
A definite status path \textit{p} between nodes $A$ and $B$ is \textit{m-connecting}, or \textit{open} given $\mathbf{C}$ ($A,B \notin \mathbf{C}$) if every definite non-collider on $p$ is not in $\mathbf{C}$, and every collider on $p$ has a descendant in $\mathbf{C}$ \citep{richardson2002ancestral,zhang2008causal}. Otherwise, $\mathbf{C}$ \textit{blocks} $p$. If $\mathbf{C}$ blocks all definite status paths between $A$ and $B$, we say that $A$ and $B$ are m-separated given $\mathbf{C}$ in $\g$ \citep{richardson2002ancestral}. Otherwise, $A$ and $B$ are m-connected given $\mathbf{C}$ in $\g$. For pairwise disjoint subsets $\mathbf{A}$, $\mathbf{B}$, and $\mathbf{C}$ of $\vars$ in $\g$, we say that $\mathbf{A}$ and {$\mathbf{B}$} are m-separated given $\mathbf{C}$ in $\g$, and write $\mathbf{A} \msepp \mathbf{B} | \mathbf{C}$, if $A$ and $B$ are m-separated given $\mathbf{C}$ in $\g$ for any $A \in \mathbf{A}$ and $B\in \mathbf{B}$. Otherwise, $\mathbf{A}$ and $\mathbf{B}$ are m-connected given $\mathbf{C}$ in $\g$ and  we write $\mathbf{A} \not \msepp \mathbf{B} | \mathbf{C}$. The concepts of m-separation and m-connection subsume the concepts of d-separation and d-connection \cite{pearl1986fusion} and we use m-separation instead of d-separation throughout. 

\textbf{Directed, mixed, and partial mixed graphs.}  A graph $\g =(\mathbf{V,E})$  is  \textit{directed} if it only contains directed edges. A graph $\g =(\mathbf{V, E})$  is a \textit{mixed graph} if it only contains directed and bidirected edges.
A graph $\g = (\mathbf{V,E})$ is a \textit{partial mixed graph} if it contains non-directed ($\circcirc$), partially directed ($\circarrow$), directed, and bidirected edges. 

\textbf{Induced subgraph, skeleton.} Let $\mathbf{A} \subseteq \mathbf{V}$  for graph $\g = (\mathbf{V},\mathbf{E})$, then the $\mathbf{A}$ \textit{induced subgraph} of $\g$, labeled $\g_{\mathbf{A}}$ is a graph that consists of vertices $\mathbf{A}$ and all edges in $\mathbf{E}$ for which both endpoints are in $\mathbf{A}$. A \textit{skeleton} of a graph $\g = (\mathbf{V,E})$ is graph $\g_{\text{skel}} = (\mathbf{V}, \mathbf{E}^{\prime})$, where $\mathbf{E}^{\prime}$ is constructed from $\mathbf{E}$ by replacing each edge with a non-directed edge $\circcirc$.
For a partial mixed graph $\g$, the subgraph of $\g$ consisting of all $\circcirc$ edges is called the  \textit{circle component} of $\g$ and labeled as $\g_{C}$.

\textbf{Acyclic, ancestral, and maximal graphs.}
Graph  $\g = (\mathbf{V,E})$ is \textit{acyclic} if it does not contain directed cycles, and $\g$ is \textit{ancestral} if it also does not contain almost directed cycles.
An ancestral mixed graph $\g = (\mathbf{V,E})$ is \textit{maximal} if for any pair of non-adjacent nodes $V_1, V_2 \in \mathbf{V}$, there exists a node set $\mathbf{S}$, $V_1, V_2 \notin \mathbf{S}$ such that $V_1 \msepp V_2 \mid \mathbf{S}$ in $\g$.
Equivalently, an ancestral mixed graph is \textit{maximal} if it does not contain an inducing path $p = \langle A, Q_1 , \dots , Q_k, B \rangle$, $k \ge 2$, such that $A$ and $B$ are not adjacent (Theorem 4.2 of \citealp{richardson2002ancestral}). A directed acyclic graph (DAG) $\g[D] = (\mb{V,E})$ with unobserved variables $\mb{L}, \mb{L} \subset \mb{V}$, can be uniquely \textit{represented} by a maximal ancestral mixed graph (MAG) $\g[M] = (\mb{O, E'})$ on the observed variables $\mb{O} = \mb{V} \setminus \mb{L}$ that preserves the ancestral and m-separation relationships among the observed variables \citep[page~981~in][]{richardson2002ancestral}.  If a DAG $\g[D]$ can be represented by a MAG $\g[M]$, we also say that $\g[M]$ \textit{represents} $\g[D]$.
A directed edge $B \to A$ in a DAG implies $B$ is a direct cause of $A$. A directed edge $B \to A$ in a MAG $\g[M]$ implies the presence of a causal path $B \to \dots \to A$ in every DAG $\g[D]$ which $\g[M]$ represents, and also does not generally exclude the option of a latent common cause of $B$ and $A$ in $\g[D]$ (except in the case of ``visible'' edges, see \citealp{zhang2008causal}).

\textbf{Markov equivalence class, essential ancestral graphs.} 
Several MAGs can encode the same m-separation relationships. Such MAGs form a \textit{Markov equivalence class}.
The Markov equivalence class of MAGs can be uniquely represented by a partial mixed graph which we refer to as the \textit{essential ancestral graph}. Other works have also referred to this graph as a \emph{partial ancestral graph (PAG)} \citep{richardson2002ancestral, ali2009markov}. An essential ancestral {graph} {is a partial mixed graph.} Any invariant edge mark {(arrowhead or tail edge mark)} in an essential ancestral graph $\g$ corresponds to that same edge mark in every MAG in the Markov equivalence class described by $\g$. Additionally, for every circle mark $A \circbullet B$ in an essential ancestral graph $\g$, the Markov equivalence class described by $\g$ contains a MAG with $A \arrowbullet B$ and a MAG with $A \to B$  \citep{zhang2008completeness}. 

\textbf{Markov and faithfulness assumptions.} A joint probability density $f(x_\mathbf{v})$ for a random vector $X_\mathbf{V}$ is \textit{Markov} to a graph $\g = (\mathbf{V,E})$ if every m-separation in $\g$ implies a conditional independence the probability distribution defined by $f(x_\mathbf{v})$. Conversely, a graph $\g$ is said to be \textit{faithful} to joint probability density $f(x_\mathbf{v})$ if every m-connection in $\g$ implies a conditional dependence in the distribution $f(x_\mathbf{v})$.

\textbf{Do-intervention.} We label an outside intervention that sets a variable $X_i$ to a fixed value $x_i$ uniformly across the population as $do(X_i = x_i)$, or $do(x_i)$ for short, also called a \textit{do-intervention} \citep{pearl2000causality}. A probability distribution of random variables under an intervention will then be referred to as an \textit{interventional distribution}, while all other distributions will be labeled as \textit{observational}. 

\begin{definition}[Causal DAG, c.f.\ Definition 1.3.1 of \citealp{pearl2000causality}]\label{def:causal-dag} Let $X_\mathbf{V}$ be a random vector and let $\g[D] = (\mathbf{V,E})$ be a DAG on vertices $\mathbf{V}$. Furthermore, let $f(x_{\mathbf{v}})$ be a joint density for $X_{\mathbf{V}}$ and let $f_{do(x_i)}(x_{\mathbf{v'}})$ be a density of the random vector $X_{\mathbf{V'}}$, $\mathbf{V'}= \mathbf{V} \setminus \{i\}$, $i \in \mathbf{V}$, after an intervention $do(x_i)$. DAG $\g[D]$ is then causal for $X_{\mathbf{V}}$ if the following hold
\begin{align}
f(x_{\mathbf{v}}) = \prod_{j\in \mathbf{V}} f_{j}(x_j | x_{\pa(j,\g[D])}) \quad  \mathrm{ and } \quad  f_{do(x_i)}(x_{\mathbf{v'}})=\prod_{j \in \mb{V'}}f_j(x_j|x_{\pa(j, \g[D])}).
\label{eq11}
\end{align}
\end{definition}

The factorization of $f(x_{\mathbf{v}})$ in Equation \eqref{eq11} follows from the Markov assumption and the rules of m-separation, while the factorization of the interventional distribution, $f_{do(x_i)}(x_{\mathbf{v'}})$ is known as the \textit{g-formula} of \cite{robins1986new}, or the \textit{truncated factorization formula} \citep{pearl2000causality}. The g-formula is crucial in identifying and estimating causal effects from observational data, as it bridges the observational and interventional worlds. 

\textbf{Causal MAGs and causal essential ancestral graphs.} A MAG is \textit{causal} if it represents a causal DAG and an essential ancestral graph is \textit{causal} if the Markov equivalence class represented by this essential ancestral graph includes the causal MAG. Similar characterizations as in Definition \ref{def:causal-dag} cannot always be obtained directly for causal MAGs and essential ancestral graphs due to identifiability issues stemming from unobserved confounding. We do not discuss these difficulties in more detail but instead refer interested readers to works of \cite{zhang2008causal,jaber2019causal,mooij2020joint} and \cite{wang2023estimating} for a more in-depth exploration of this problem.

\section{Characterizing the Markov Equivalence Class}
\label{sec:eq-class}
 
There are several ways to characterize Markov equivalent MAGs. For instance, \citet{spirtes1996polynomial} characterize Markov equivalence through discriminating paths: MAG $\g[M]_{1}$ is Markov equivalent to MAG $\g[M]_2$ if $\g[M]_1$, and $\g[M]_2$ share the same adjacencies and unshielded colliders, and if a path $\langle V_1, \dots,V_{k-1},  V_k \rangle, k >3$ is a discriminating path from $V_1$ to $V_k$ for $V_{k-1}$ in both $\g[M]_1$ and $\g[M]_2$, then the $V_{k-1}$ is either a collider on both of these paths or a non-collider on both of these paths. \cite{ali2009markov} build on this work to provide another characterization using so-called colliders with order. 
Yet another characterization is given by
\citet{zhao2005markov}, who prove that all Markov equivalent MAGs share the same adjacencies and minimal collider paths. 

We favor \cite{zhao2005markov}'s characterization of Markov equivalence but also show how to bridge the \citet{spirtes1996polynomial} and \citet{zhao2005markov} characterizations through results in this section.
First, in Theorem \ref{thm:ali-conjecture}, we show that \textit{any} collider 
 $Q_k$, $k\ge 2$ discriminated by some path $\langle A, Q_1, \dots, Q_k,B  \rangle$ in a MAG $\g[M]$ is invariant across the Markov equivalence class. Meaning that $Q_k$ is a collider on path $\langle A, Q_1, \dots, Q_k, B \rangle$  in every MAG that is Markov equivalent to $\g[M]$, regardless of whether $\langle A, Q_1, \dots, Q_k, B \rangle$ is a discriminating path. This property was previously conjectured by \citet{ali2009markov}.

\begin{theorem}
\label{thm:ali-conjecture}
Suppose that $p = \langle A, Q_1, \dots, Q_{k-1}, Q_k, B \rangle$, $k \ge 2$ forms a discriminating path for $Q_k$ from $A$ to $B$ in MAG $\G[M] = (\mb{V,E})$, and that $\langle Q_{k-1}, Q_k, B \rangle$ is a collider. Then, $ \langle Q_{k-1}, Q_k, B \rangle$ is a collider in every MAG $\G[M]^* = (\mb{V,E'})$ that is Markov equivalent to $\G[M]$.
\end{theorem}

Next, we consider obtaining an essential ancestral graph $\g$ for a given MAG $\g[M].$ \cite{zhang2008completeness} proved that one can obtain an essential ancestral graph $\g$ from $\g[M]$ by taking the skeleton of $\g[M]$ called $\g_{skel}$, adding arrowhead edge marks to $\g_{skel}$ that make up the non-endpoints of an unshielded collider in $\g[M]$ and then exhaustively completing the following  set of orientation rules \citep{spirtes2000causation,zhang2008completeness}:

\begin{enumerate}[leftmargin = 1.5cm, label=  R\arabic*]
     \item\label{R1} If $A\bulletarrow B \circbullet C$ is in $\g = (\mathbf{V,E})$ for some nodes $A,B,C \in \mathbf{V}$, and $A \notin \Adj(C, \g)$ then orient $B \to C$.
    \item\label{R2} If $A \to B \bulletarrow C$ or $A \bulletarrow B \to C$ and $A \bulletcirc C$, then orient $A \bulletarrow C$.
    \item\label{R3} If $A \bulletarrow B \arrowbullet C$, $A \bulletcirc D \circbullet C$, $A \notin \Adj(C, \g)$ and $D \bulletcirc B$ is in $\g$, then orient $D \bulletarrow B$.
    \item[\namedlabel{R4zhang}{Zhang-R4}] If $p = \langle A, Q_1, \dots , Q_{k-1}, Q_k, B \rangle$ is a discriminating path for $Q_k$ in $\g$, and if $Q_k \circbullet B$ is in $\g$; then if $Q_k$ is in any m-separating set for $A$ and $B$ in $\g[M]$, orient {$Q_{k-1} \leftrightarrow Q_k \to B$; otherwise, orient $Q_k \leftrightarrow B$}.
    \setcounter{enumi}{7}
    \item\label{R8} If $A \to B \to C$ and $A \circarrow C$ is in $\g$ then orient $A \to C$.
    \item\label{R9} If $A \circarrow C$ is in $\g$ and $p = \langle A, B, D, \dots, C\rangle$ is an unshielded possibly directed path in $\g$ such that $B \notin \Adj(C,\g)$, then orient $A \to C$.
    \item\label{R10} If $A \circarrow C$ and $B \to C \leftarrow D$ are in $\g$, and if there are unshielded possibly directed paths $p_1 = \langle A, M_{11}, \dots , M_{1l} = B\rangle, l \ge 1$ and $p_2 =  \langle A , M_{21}, \dots, M_{2r}=D\rangle, r \ge 1$ and if $M_{11} \neq M_{21}$ and $M_{11} \notin \Adj(M_{21}, \g)$, then orient $A \to C$. 
\end{enumerate}

Above, we leave out orientation rules R5-R7 of \cite{zhang2008completeness}, as they only apply in the presence of selection bias.
{Motivated by Theorem \ref{thm:ali-conjecture} and \citet{zhao2005markov}'s characterization of Markov equivalence, we next introduce orientation rule \ref{R4new} and the \texttt{MAGtoEssentialAncestralGraph} algorithm (Algorithm \ref{alg:mag2pag}).} 

\begin{enumerate}[leftmargin = 1.5cm, label=  Zhao-R\arabic*]
     \setcounter{enumi}{3}
    \item\label{R4new} If $\langle A, Q_1, \dots , Q_{k-1}, Q_k, B \rangle, k \ge 2,$ is a discriminating path for $Q_k$ and if $Q_k \circbullet B$ is in $\g$; then orient $Q_k \to B$.
\end{enumerate}

{ We note that \ref{R4zhang} and \ref{R4new} are not equivalent in general: \ref{R4new} never orients a bidirected edge, whereas \ref{R4zhang} can. Nevertheless, the two rules are interchangeable within Algorithm \ref{alg:mag2pag}. This is because line \ref{step3algoPAG} of Algorithm \ref{alg:mag2pag} already orients all arrowheads corresponding to colliders on minimal collider paths before any orientation rule is applied; the colliders that \ref{R4zhang} would discriminate are thus already identified, and only \ref{R1} and \ref{R2} are needed to complete the remaining orientations.} %Hence either rule yields the same essential ancestral graph when used in Algorithm \ref{alg:mag2pag}, even though the two rules are not equivalent in full generality.}

Algorithm \ref{alg:mag2pag} takes as input MAG $\g[M]$ and returns the corresponding essential ancestral graph $\g$. This is proven in  Theorem \ref{thm:mag2pag}.  Instead of using the process of \cite{zhang2008completeness}, Algorithm \ref{alg:mag2pag} proceeds by obtaining the skeleton of $\g[M]$ called $\g_{skel},$ orienting those arrowheads in $\g_{skel}$ that correspond to non-endpoints on minimal collider paths in $\g[M]$ and completing orientation rules \ref{R1}-\ref{R3}, \ref{R4new}, \ref{R8}-\ref{R10}.

\begin{algorithm}[!t]
\caption{MAGtoEssentialAncestralGraph}
\label{alg:mag2pag}
\begin{algorithmic}[1]
    \Require MAG $\g[M] = (\mathbf{V,E})$.
    \Ensure Partial mixed graph $\g = (\mathbf{V,E'})$.
    \State Let $\g_{skel}$ denote the skeleton of $\g[M]$
    \State Let $\g = \g_{skel}$
    \State In $\g$, orient as arrowheads those edge marks that correspond to colliders on minimal collider paths in $\g[M]$ \label{step3algoPAG}
    \State Close orientations according to  \ref{R1}-\ref{R3}, \ref{R4new}, \ref{R8}-\ref{R10} in $\g$
    \State \Return $\g$
\end{algorithmic}
\end{algorithm}

\begin{theorem}\label{thm:mag2pag} 
Let $\g[M] = (\mathbf{V,E})$ be a MAG and let $\g = (\mb{V,E'})$ be the output of Algorithm \ref{alg:mag2pag} applied to $\g[M]$, that is,  $\g = $ \texttt{MAGtoEssentialAncestralGraph}$(\g[M])$. Then $\g$ is the essential ancestral graph of $\g[M]$.
\end{theorem}

One may be concerned that the process of finding minimal collider paths employed by Algorithm \ref{alg:mag2pag} is intractable. For this reason, we now present Lemma \ref{lemma:MCP-format}, which solidifies the connection between the different characterizations of Markov equivalence. 
We say that orientations in a graph are closed under a particular operation, if applying that operation does not change the orientations in the graph.

 \begin{lemma}\label{lemma:MCP-format} 
Let $\g = (\mathbf{V,E})$ be an ancestral partial mixed graph.
Furthermore, suppose 
edge orientations in $\g$ are closed under \ref{R1}, \ref{R2}, \ref{R4new}. 
Let $p = \langle P_1, P_2, \dots , P_k \rangle, k \ge 3$ be a minimal collider path in $\g$. Then for every $i \in \{2, \dots , k-1\}$, one of the following holds:
\begin{enumerate}[label = (\roman*)]
    \item\label{caseMCP1:1} $P_{i-1} \bulletarrow P_i \arrowbullet P_{i+1}$ and $P_{i-1} \notin \Adj(P_{i+1}, \g)$, or
    \item\label{caseMCP1:2}  $\exists \ l \in \{1, \dots , i-2\}$,  such that $P_{l} \bulletarrow P_{l+1} \leftrightarrow \dots \leftrightarrow P_i \arrowbullet P_{i+1}$ is a  discriminating collider path from $P_{l}$ to $P_{i+1}$ for $P_i$, or
     \item\label{caseMCP1:3} $\exists \ r \in \{i+2, \dots k\}$ such that {$P_{i-1} \bulletarrow P_i \leftrightarrow \dots \leftrightarrow P_{r-1} \arrowbullet P_{r}$}
     % $P_{r} \bulletarrow P_{r -1} \leftrightarrow \dots \leftrightarrow P_{i+1} \leftrightarrow P_i \arrowbullet P_{i-1}$
     is a discriminating collider path from $P_{r}$ to $P_{i-1}$ for $P_i$.   
\end{enumerate}
\end{lemma}

According to Lemma \ref{lemma:MCP-format}, every non-endpoint node on a minimal collider path $p$ is either an unshielded collider on $p$ or a collider that is discriminated by a subpath of $p$. So to find minimal collider paths in a MAG $\g[M]$ it suffices to determine the unshielded colliders and colliders discriminated by a path in $\g[M]$  (Theorem \ref{thm:ali-conjecture}). Finding unshielded colliders is relatively straightforward. Additionally,  \cite{wienobst2022new} recently introduced an algorithm that finds colliders discriminated by a path given a MAG $\g[M] = (\mathbf{V,E})$ in $O(|\mathbf{V}|^{3})$ worst-case runtime. This allows relatively tractable implementations of Algorithm \ref{alg:mag2pag}.  
 
\section{Expert Knowledge and Restricted Essential Ancestral Graphs}
\label{sec:bg-knowledge}

We now focus on restricting a Markov equivalence class of MAGs with expert knowledge in the form of specific edge marks. We first introduce some notation and definitions, starting with defining a representing graph.

\begin{definition}[Representing Graphs]\label{def:represent}
A MAG $\g[M] = (\mathbf{V,E})$ is \textit{represented} by a partial mixed graph $\g = (\mathbf{V,E'})$, or $\g$ represents $\g[M]$ if
\begin{enumerate}[label = (\roman*)]
    \item $\g$ and $\g[M]$ have the same skeleton and the same minimal collider paths and
    \item  every invariant edge mark in $\g$ is identical  in $\g[M]$.
\end{enumerate}
We use $[\g]$ to denote the set of MAGs represented by $\g$.
\end{definition}

If $\g$ is an essential ancestral graph, then $[\g]$ is the Markov equivalence class of MAGs represented by $\g$. We now define expert knowledge we consider, which we call orientation knowledge. 

\begin{definition}[Orientation knowledge]\label{def:bgknolwedge}
A piece of orientation knowledge $\langle \langle A, B \rangle \rangle$ on edge $\langle A, B \rangle$  is of one of the following forms:  $A \to B$, $A \leftarrow B$, $A \bulletarrow B$, or $A \arrowbullet B$. A set of orientation knowledge made up of pieces of orientation knowledge will be denoted by a calligraphic letter, most often $\mathcal{K}$. 
\end{definition}

Orientation knowledge $A \bulletarrow B$ implies that the edge mark at $B$ on edge $\langle A, B \rangle $ needs to be an arrowhead but does not imply anything about the edge mark at $A$. Information about a bidirected edge $A \leftrightarrow B$  would be represented using two pieces of orientation knowledge $A \bulletarrow B$ and $B \bulletarrow A$, that is, with the following set of orientation knowledge $\mathcal{K} = \{A \bulletarrow B,  B \bulletarrow A\}$. 

In Definition \ref{def:consistentbg} below, we also note that only certain sets of orientation knowledge $\mathcal{K}$ are consistent with a partial mixed graph $\g$. If $\g$ is an essential ancestral graph, then such consistent $\mathcal{K}$ can be used to restrict the Markov equivalence class $[\g]$ (Definition \ref{def:res-marko-eq-class}). Furthermore, there may be a restricted essential ancestral graph $\g^{\prime}$ which represents such a \textit{restricted Markov equivalence class} (Definition \ref{def:res-ess-graph}).

\begin{definition}[Consistent Orientation Knowledge]\label{def:consistentbg} A set of orientation knowledge  $\mathcal{K}$ is \textit{consistent} with a  partial mixed graph $\g = (\mathbf{V,E})$ if there is a MAG $\g[M]= (\mathbf{V,E'})$ represented by $\g$ such that for every piece of orientation knowledge in $\mathcal{K}$:
\begin{enumerate}[label = (\roman*)]
    \item if $A \to B$ is in $\g[K]$, then  $A \to B$ is in $\g[M]$, and
    \item if $A \leftarrow B$ is in $\g[K]$, then $A \leftarrow B$ is in $\g[M]$, and
    \item if $A \bulletarrow B$ is in $\g[K]$ then $A \to B$ or $A \leftrightarrow B$ is in $\g[M]$, and
    \item if $A \arrowbullet B$ is in $\g[K]$ then $A \leftarrow B$ or $A \leftrightarrow B$ is in $\g[M]$.
\end{enumerate}
\end{definition}

\begin{definition}[Restricted Markov equivalence class]
\label{def:res-marko-eq-class}
Let  $\g = (\mathbf{V,E})$ be an essential ancestral graph and $\mathcal{K}$ be some orientation knowledge consistent with $\g$. Then $[\g]_{\mathcal{K}}$ is a restriction of the Markov equivalence class of MAGs represented by $\g$ to exactly those MAGs for which $\mathcal{K}$ is a set of consistent orientation knowledge. We call $[\g]_{\mathcal{K}}$ a restricted Markov equivalence class, or more precisely, the $\mathcal{K}$-restricted Markov equivalence class. 
\end{definition}

\begin{definition}[Restricted essential ancestral graph]
\label{def:res-ess-graph}
Let  $\g = (\mathbf{V,E})$ be an essential ancestral graph and let $\mathcal{K}$ be orientation knowledge consistent with $\g$. Additionally, let $[\g]_{\mathcal{K}}$ be the  $\mathcal{K}$-restricted Markov equivalence class. Then $\g^{\prime} = (\mathbf{V,E'})$ is a restricted essential ancestral graph, or, more precisely, {the} $\mathcal{K}$-restricted essential ancestral graph  if
\begin{enumerate}[label = (\roman*)]
    \item\label{cond1:ress-ess} $\g^{\prime}$ has the same skeleton and the same minimal collider paths as $\g$,
    \item\label{cond2:ress-ess} a non-circle edge mark in $\g^{\prime}$ is \textit{invariant} across the $[\g]_{\mathcal{K}}$, and
    \item\label{cond3:ress-ess} for any circle edge mark in $\g^{\prime}$ there is at least one MAG in  $[\g]_{\mathcal{K}}$ such that this circle is replaced by a tail, and one MAG in $[\g]_{\mathcal{K}}$ where this circle is replaced by an arrowhead.   
\end{enumerate}
\end{definition}

If $\g^{\prime}$ is {the} restricted essential ancestral graph for $[\g]_{\mathcal{K}}$, then by construction of $\g^{\prime}$, $[\g^{\prime}] = [\g]_{\mathcal{K}}$. {Note that for any consistent orientation knowledge $\mc{K}$, the $\mc{K}$-restricted essential  ancestral graph is unique. However, it is possible that different sets of consistent orientation knowledge may lead to the same restricted essential ancestral graph.} An essential ancestral graph can be seen as {the} $\emptyset$-restricted essential ancestral graph.

Next, consider  MAG $\g[M]$ in Figure \ref{fig:example_MEC_MAG}(b) and its corresponding essential ancestral graph $\g$ in Figure \ref{fig:example_MEC_MAG}(c). Note that $\mathcal{K} = \{B \bulletarrow C\}$ is a consistent set of orientation knowledge with respect to $\g$, since there are multiple MAGs in $[\g]$ that contain this knowledge, see Figure \ref{fig:example_MEC_MAG}(e). These MAGs form the restricted Markov equivalence class $[\g]_{\mathcal{K}}$ and are given in the first row of Figure \ref{fig:example_MEC_MAG}(e). Furthermore, we can confirm that the partial mixed graph $\g^{\prime}$ in Figure \ref{fig:example_MEC_MAG}(d) is {the} restricted essential ancestral graph for $\g$ and orientation knowledge $\mathcal{K}$, as it satisfies all three conditions of Definition \ref{def:res-ess-graph}. 

For examples of partial mixed graphs that satisfy some but not all properties of a restricted essential ancestral graph, consider partial mixed graphs $\g_{1}$ and $\g_2$  in Figure \ref{fig:example_not_RMEC}(a)  and (b) respectively. Both $\g_{1}$ and $\g_{2}$ satisfy conditions \ref{cond1:ress-ess} and \ref{cond2:ress-ess} but not condition \ref{cond3:ress-ess} of Definition \ref{def:res-ess-graph} relative to $\g$ and $\mathcal{K}$ as they are both missing $A \to D$ edge orientation present in $\g^{\prime}$. Now consider $\g_{3}$ in Figure \ref{fig:example_not_RMEC}(c), which can be obtained from $\g$ by adding   orientation knowledge $\mathcal{K}_{1} = \{B \bulletarrow C, C \to D, D \to A\}$. There is no MAG represented by $\g_{3}$ as can be verified from Figure \ref{fig:example_MEC_MAG}(e). Moreover, $\mathcal{K}_1$ is not consistent with $\g$.  
The orientations in graphs $\g_{1}$ and $\g_{2}$ can be called sound but not complete, while the orientations in graph $\g_{3}$ are not sound per the following definition.

\begin{figure}[!t]

\tikzstyle{every edge}=[draw,>=stealth',->,thick]
\newcommand\dagvariant[1]{\begin{tikzpicture}[xscale=.45,yscale=0.45]
\node (d) at (0,0) {};
\node (a) at (0,2) {};
\node (c) at (2,0) {};
\node (b) at (2,2) {};
\draw (a) edge [-] (b);
\draw (a) edge [-] (c);
\draw (d) edge [-] (a);
\draw (d) edge [-] (c);
\draw (a) edge [-] (c);
\draw #1;
\end{tikzpicture}}

\tikzstyle{every edge}=[draw,>=stealth',->,thick]
\newcommand\dagempty[1]{\begin{tikzpicture}[xscale=.4,yscale=0.4]
\node (d) at (0,0) {};
\node (a) at (0,.5) {};
\node (c) at (2,0) {};
\node (b) at (2,.5) {};
\draw #1;
\end{tikzpicture}}

\centering
\begin{subfigure}{.31\textwidth}
\tikzstyle{every edge}=[draw,>=stealth',->]
  \centering
\begin{tikzpicture}[->,>=latex,shorten >=1pt,auto,node distance=0.8cm,scale=.8,transform shape]
  \tikzstyle{state}=[inner sep=1pt, minimum size=12pt]
\tikzstyle{every edge}=[draw,>=stealth',->]
  % rule 1
  \node[state] (D) at (0,0) {\Large $D$};
  \node[state] (A) at (0,2) {\Large $A$};
  \node[state] (C) at (2,0) {\Large  $C$};
  \node[state] (B) at (2,2) {\Large  $B$};

  \draw (A) edge [o-o,color=gray] (B);
\draw 		(A) edge  [o-o,color=gray] (C);
 \draw   	(B) edge [o->,line width=1pt,color=black] (C);
 \draw    	(C) edge [o-o, color=gray] (D);
\draw    	(A) edge [o-o,color=gray] (D);
\end{tikzpicture}
\caption{}
  \label{mpdag4}
  \end{subfigure}
  \vrule
\hspace{0.2cm}
\begin{subfigure}{.31\textwidth}
\vspace{1cm}
\tikzstyle{every edge}=[draw,>=stealth',->]
  \centering
\begin{tikzpicture}[->,>=latex,shorten >=1pt,auto,node distance=0.8cm,scale=.8,transform shape]
  \tikzstyle{state}=[inner sep=1pt, minimum size=12pt]

  \node[state] (D) at (0,0) {\Large $D$};
  \node[state] (A) at (0,2) {\Large $A$};
  \node[state] (C) at (2,0) {\Large  $C$};
  \node[state] (B) at (2,2) {\Large  $B$};

  \draw (A) edge [o-o,color=gray] (B);
\draw 		(A) edge  [o-o,color=gray] (C);
 \draw   	(B) edge [o->,line width=1pt,color=black] (C);
 \draw    	(C) edge [->,line width=1pt,color=black] (D);
\draw    	(A) edge [o-o,color=gray] (D);
\end{tikzpicture}
\caption{}
\end{subfigure}
  \vrule
\hspace{0.2cm}
\begin{subfigure}{.31\textwidth}
\vspace{1cm}
\tikzstyle{every edge}=[draw,>=stealth',->]
  \centering
\begin{tikzpicture}[->,>=latex,shorten >=1pt,auto,node distance=0.8cm,scale=.8,transform shape]
  \tikzstyle{state}=[inner sep=1pt, minimum size=12pt]

  \node[state] (D) at (0,0) {\Large $D$};
  \node[state] (A) at (0,2) {\Large $A$};
  \node[state] (C) at (2,0) {\Large  $C$};
  \node[state] (B) at (2,2) {\Large  $B$};

  \draw (A) edge [o-o,color=gray] (B);
\draw 		(A) edge  [o-o,color=gray] (C);
 \draw   	(B) edge [o->,line width=1pt,color=black] (C);
 \draw    	(C) edge [->,line width=1pt,color=black] (D);
\draw    	(A) edge [<-,line width=1pt,color=black] (D);
\end{tikzpicture}
\caption{}
\end{subfigure}
\caption{Partial mixed graphs (a)  ${\g}_{1}$, (b)  ${\g}_{2}$, and (c)  $\g_{3}$.}
\label{fig:example_not_RMEC}
\end{figure}

\begin{definition}[Sound and Complete Orientations]
\label{def:completeness}
Let $\g = (\mb{V,E})$ be an essential ancestral graph and $\g^{\prime} = (\mb{V,E'})$ be a partial mixed graph such that $\g$ and $\g^{\prime}$ have the same skeleton and minimal collider paths. Suppose additionally that the set of invariant edge marks in $\g$ is a subset of the invariant edge marks in $\g^{\prime}$. We say that orientations in $\g^{\prime}$ are \textit{sound} if there is at least one MAG $\g[M]$ in $[\g]$ such that invariant edge marks in $\g^{\prime}$ are a subset of edge marks in $\g[M].$ 
We say that the orientations in $\g^{\prime}$ are \textit{complete} if for every $A \circbullet B$ edge in $\g$, there are two MAGs $\g[M]_1$ and $\g[M]_2$ represented by $\g^{\prime}$ containing the edges $A \to B$ and $A \arrowbullet B$ respectively such that $\g[M]_1, \g[M]_2 \in [\g]$.
\end{definition}

It follows from Definitions \ref{def:consistentbg} and \ref{def:completeness} that including consistent orientation knowledge $\mc{K}$ into an essential ancestral graph guarantees soundness in the resulting partial mixed graph $\g^{\prime}$. However, to ensure completeness, additional orientations may need to be inferred after incorporating $\mathcal{K}$.
For instance, the graphs $\g_1, \g_2$ in Figure \ref{fig:example_not_RMEC} are sound but not complete for their respective orientation knowledge. Their corresponding complete (and sound) graph is the restricted essential ancestral graph $\g^{\prime}$ in Figure \ref{fig:example_MEC_MAG} containing inferred orientations $\{A \to D, C \to D\}$ for $\g_1$ and just $\{A \to D\}$ for $\g_2$.
We now turn our attention to these inferred orientations. One immediate result that follows from previous work \citep{zhang2008completeness} and our Theorem \ref{thm:mag2pag} is that any orientation that stems from completing orientation rules \ref{R1}-\ref{R3}, \ref{R4new}, \ref{R8}-\ref{R10} after adding $\g[K]$ to $\g$ is also sound (see also Theorem 1 of \citealp{andrews2020completeness}, Theorem 20 of \citealp{mooij2020joint} and  Theorem 2 of \citealp{wang2022sound}).

\begin{corollary}\label{cor:old-rules-sound}
Let $\g^{\prime} = (\mb{V,E'})$ be a restricted essential ancestral graph. Then orientations of $\g^{\prime}$ are closed under \ref{R1}- \ref{R3}, \ref{R4new}, and \ref{R8}-\ref{R10}.
\end{corollary}

\section{Additional Orientation Rules}
\label{sec:new-rules}

For certain types of tiered and local expert knowledge $\mathcal{K}$ consistent with an essential ancestral graph $\g$ (meaning there exists a MAG in the Markov equivalence class of $[\g]$ that satisfies this expert knowledge), \cite{andrews2020completeness,mooij2020joint} and \cite{wang2022sound} show that the known list of orientation rules suffices to obtain {the} $\mathcal{K}$-restricted essential ancestral graph.   However, these orientation rules are insufficient for completeness in generality, as none of them would lead to the conclusion that $A \to D$ should be present in the restricted essential ancestral graph $\g^{\prime}$ in Figure \ref{fig:example_MEC_MAG}(d) after adding $B \circarrow C$ to $\g$ in Figure \ref{fig:example_MEC_MAG}(c).
In this section, we present several new graphical orientation rules that are distinct from \ref{R1}-\ref{R3}, \ref{R4new}, and \ref{R8}-\ref{R10}.  
We start with the rule motivated by $\g^{\prime}$ in Figure \ref{fig:example_MEC_MAG}(d), which we refer to as \ref{R11}. Note that \ref{R11} can be considered a generalization of R4 of \citet{meek1995causal}. 

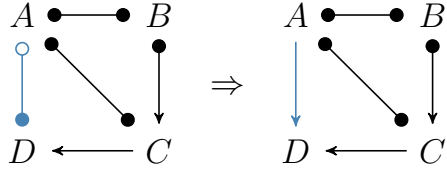
\begin{figure}[!t]
\centering
\tikzstyle{every edge}=[draw,>=stealth',semithick]
	\begin{tikzpicture}[->,>=stealth',shorten >=1pt,auto,node distance=1.2cm,scale=.9,transform shape,font = {\Large\bfseries\sffamily}]
	\tikzstyle{state}=[inner sep=5pt, minimum size=5pt]

	\node[state] (A) at (0,2) {$A$};
	\node[state] (B) at (2,2) {$B$};
	\node[state] (D) at (0,0) {$D$};
	\node[state] (C) at (2,0) {$C$};

	\draw[*->]  (B) edge (C);
	\draw[->] (C) edge (D);
	\draw[blue,*-o] (D) edge (A);
	\draw[*-*] (A) edge (B);
	\draw[*-*] (A) edge (C);
          
	\coordinate [label=above:$\Rightarrow$] (L4) at (3,0.7);

	\node[state] (A) at (4,2) {$A$};
	\node[state] (B) at (6,2) {$B$};
	\node[state] (D) at (4,0) {$D$};
	\node[state] (C) at (6,0) {$C$};

	\draw[*->]  (B) edge (C);
	\draw[->] (C) edge (D);
	\draw[blue,<-] (D) edge (A);
	\draw[*-*] (A) edge (B);
	\draw[*-*] (A) edge (C);

  \end{tikzpicture}
\caption{Representation of \ref{R11} in Theorem \ref{thm:meek}.}
\label{fig:R11compressed}
\end{figure}

\begin{theorem}
\label{thm:meek}
Let $A, B, C, D$ be distinct nodes in a partial mixed graph $\g = (\mathbf{V,E})$. 
\begin{enumerate}[label = R11]
    \item\label{R11} Suppose that the partial mixed graph on the left side of Figure \ref{fig:R11compressed} is an induced subgraph of $\g$. Then, in all MAGs represented by $\g$, the edge $A \circbullet D$ is oriented as $A \to D$. 
\end{enumerate}
\end{theorem}

Another new rule, \ref{R12}, is given in Theorem \ref{thm:rule12}. 
A graphical representation of \ref{R12} is given in Figure \ref{fig:R12-example} and explored in Example \ref{ex:r12}. 

\begin{figure}[!t]
\centering

\begin{subfigure}{.2\textwidth}
\tikzstyle{every edge}=[draw,>=stealth',semithick]
	\begin{tikzpicture}[->,>=stealth',shorten >=1pt,auto,node distance=1.2cm,scale=.9,transform shape,font = {\Large\bfseries\sffamily}]
	\tikzstyle{state}=[inner sep=5pt, minimum size=5pt]

	\node[state] (A) at (0,2) {$A$};
	\node[state] (B) at (2,2) {$B$};
	\node[state] (D) at (0,0) {$D$};
	\node[state] (C) at (2,0) {$C$};

	\draw[gray,o->]  (B) edge (C);
	\draw[gray,<-o] (C) edge (D);
	\draw[gray,o-o] (D) edge (A);
	\draw[gray,o-o] (A) edge (B);
	\draw[gray,o->] (A) edge (C);
  \end{tikzpicture}
\caption{}
  \end{subfigure}
  \vrule
  \begin{subfigure}{.55\textwidth}
  \centering
\tikzstyle{every edge}=[draw,>=stealth',semithick]
	\begin{tikzpicture}[->,>=stealth',shorten >=1pt,auto,node distance=1.2cm,scale=.9,transform shape,font = {\Large\bfseries\sffamily}]
	\tikzstyle{state}=[inner sep=5pt, minimum size=5pt]

	\node[state] (A) at (0,2) {$A$};
	\node[state] (B) at (2,2) {$B$};
	\node[state] (D) at (0,0) {$D$};
	\node[state] (C) at (2,0) {$C$};

	\draw[<->]  (B) edge (C);
	\draw[<-] (C) edge (D);
	\draw[gray,o-o] (D) edge (A);
	\draw[gray,o-o] (A) edge (B);
	\draw[gray,o->] (A) edge (C);
          
	\coordinate [label=above:$\Rightarrow$] (L4) at (3,0.7);

	\node[state] (A) at (4,2) {$A$};
	\node[state] (B) at (6,2) {$B$};
	\node[state] (D) at (4,0) {$D$};
	\node[state] (C) at (6,0) {$C$};

	\draw[<->]  (B) edge (C);
	\draw[<-] (C) edge (D);
	\draw[gray,o-o] (D) edge (A);
	\draw[blue,o->] (A) edge (B);
	\draw[gray,o->] (A) edge (C);

  \end{tikzpicture}
\caption{}
  \end{subfigure}
\caption{(a) Essential ancestral graph $\g$, (b) Representation of \ref{R12} in Example \ref{ex:r12}.}
\label{fig:R12-example}
\end{figure}

\begin{theorem} 
\label{thm:rule12}
Let  $V_1, \dots, V_i, V_{i+1}, i>2$ be distinct nodes in partial mixed graph $\g = (\mathbf{V,E})$. 
\begin{enumerate}[label = R12] 
\item\label{R12} Suppose there is an unshielded path of the form $V_1 \circcirc V_2 \circcirc \dots V_{i-1} \circbullet V_i$, $i>2$, as well as a path $V_1 \leftrightarrow V_{i+1} \leftarrow V_i$ in $\g$. Then, all MAGs represented by $\g$ contain $V_1 \arrowbullet V_2$.
\end{enumerate}
\end{theorem}

\begin{example}
\label{ex:r12}
Consider the essential ancestral graph $\g = (\mb{V,E})$ in Figure \ref{fig:R12-example}(a). Suppose we have expert knowledge $\mathcal{K}$ that $X_D$ is a cause of $X_C$ and $X_B$  is not a cause of $X_C$, that is, $\mathcal{K} = \{D \to C, B \arrowbullet C\}$. We add $\mathcal{K}$ to $\g$ to form the graph on the left-hand side of Figure \ref{fig:R12-example}(b). However, the orientations in this graph are not closed under \ref{R12} due to paths $B \circcirc A \circcirc D$ and $D \to C \leftrightarrow B$.  Hence, we orient $A \circarrow B$ to obtain the graph on the right-hand-side of Figure \ref{fig:R12-example}(b). {This is a restricted essential ancestral graph, which can be seen by enumerating the restricted Markov equivalence class $[\g]_{\mc{K}}$ to verify condition (iii) of Definition \ref{def:res-ess-graph}. Theorem \ref{thm:3}, presented later, can also be used to verify  that the graph on the right-hand-side of Figure \ref{fig:R12-example}(b) is a restricted essential ancestral graph.} 
\end{example}

\ref{R12} was also  concurrently discovered  by \cite{wang2024new}. \cite{wang2024new} state \ref{R12} slightly differently, but both versions of \ref{R12} lead to the same orientations when applied together with \ref{R1}-\ref{R4}, \ref{R8}-\ref{R11} (a consequence of Lemma \ref{lemma:not-allowed-paths} in the Supplement \ref{supp:r13}).

We now introduce \ref{R13}, which was initially discovered by \cite{wang2024new}. We simplify the statement of the orientation rule of \cite{wang2024new} below and show that our simplified version leads to equivalent orientations in Section \ref{supp:r13}.
We also reproduce an example of \cite{wang2024new} in Figure \ref{fig:r13} and Example \ref{ex:r13} below.

\begin{theorem} \label{thm:rule13star} Let  $A,B,C,D,V_1,\dots, V_{k}, k>1,$ be distinct nodes in  partial mixed graph $\g = (\mathbf{V,E})$.
\begin{enumerate}[label = R13]
    \item\label{R13}  If {the} edge $A \circbullet B$ path $C \leftrightarrow A \leftrightarrow D$, and unshielded path $C \arrowcirc V_1 \circcirc \dots \circcirc V_k \circarrow D$,  are in $\g$ and if there are unshielded possibly directed paths $\langle A, B, \dots, V_i \rangle $ in $\g$, for all $i \in \{1, \dots, k \}$, then $A \arrowbullet  B$ is present in all MAGs represented by $\g$. 
\end{enumerate}
\end{theorem}

\begin{figure}
        \vspace{1cm}
   \centering
    \begin{subfigure}{.3\textwidth}
          \tikzstyle{every edge}=[draw,>=stealth',semithick]
        \vspace{1cm}
        \centering
        \begin{tikzpicture}[->,>=stealth',shorten >=1pt,auto,node distance=1.2cm,scale=0.75,transform shape,font = {\Large\bfseries\sffamily}]
	\tikzstyle{state}=[inner sep=5pt, minimum size=5pt]

	\node[state] (A) at (-.3,0) {$A$};
	\node[state] (B) at (2,0) {$B$};
	\node[state] (D) at (2,-2) {$D$};
	\node[state] (C) at (2,2) {$C$};
        \node[state] (E) at (4,1) {$E$};
        \node[state] (F) at (4,-1) {$F$};

	\draw[gray, o-o]  (A) edge (B);
 	\draw[gray, o->]  (A) edge (C);
	\draw[gray, o->] (A) edge (D);
	\draw[gray, o->] (B) edge (C);
	\draw[gray, o->] (B) edge (D);

        \draw[gray, <-o] (C) edge (E);
        \draw[gray, <-o] (D) edge (F);
        \draw[gray, o-o] (B) edge (E);
        \draw[gray, o-o] (B) edge (F);
        \draw[gray, o-o] (F) edge (E);

   \draw[gray, <->, out= 235, in =125] (C) edge (D);

  \end{tikzpicture}
\caption{}
\label{fig:r13a} 
    \end{subfigure}
    \vrule
    \begin{subfigure}{.64\textwidth}
        \centering
       \tikzstyle{every edge}=[draw,>=stealth',semithick]
	  \begin{tikzpicture}[->,>=stealth',shorten >=1pt,auto,node distance=1.2cm,scale=0.75,transform shape,font = {\Large\bfseries\sffamily}]
	\tikzstyle{state}=[inner sep=5pt, minimum size=5pt]

	\node[state] (A) at (-2,0) {$A$};
	\node[state] (B) at (0.3,0) {$B$};
	\node[state] (D) at (0.3,-2) {$D$};
	\node[state] (C) at (0.3,2) {$C$};
        \node[state] (E) at (2.3,1) {$E$};
        \node[state] (F) at (2.3,-1) {$F$};

	\draw[gray, o-o]  (A) edge (B);
 	\draw[<->]  (A) edge (C);
	\draw[<->] (A) edge (D);
	\draw[gray,  o->] (B) edge (C);
	\draw[gray, o->] (B) edge (D);

        \draw[gray, <-o] (C) edge (E);
        \draw[gray, <-o] (D) edge (F);
        \draw[gray, o-o] (B) edge (E);
        \draw[gray, o-o] (B) edge (F);
        \draw[gray, o-o] (F) edge (E);
	
   \draw[gray, <->, out= 235, in =125] (C) edge (D);
   
   	\coordinate [label=above:$\Rightarrow$] (L4) at (3.2,-.3);

	\node[state] (A) at (4.2,0) {$A$};
	\node[state] (B) at (6.5,0) {$B$};
	\node[state] (D) at (6.5,-2) {$D$};
	\node[state] (C) at (6.5,2) {$C$};
        \node[state] (E) at (8.5,1) {$E$};
        \node[state] (F) at (8.5,-1) {$F$};

	\draw[blue, <-o]  (A) edge (B);
 	\draw[<->]  (A) edge (C);
	\draw[<->] (A) edge (D);
	\draw[gray, o->] (B) edge (C);
	\draw[gray, o->] (B) edge (D);

        \draw[gray, <-o] (C) edge (E);
        \draw[gray, <-o] (D) edge (F);
        \draw[gray, o-o] (B) edge (E);
        \draw[gray, o-o] (B) edge (F);
        \draw[gray, o-o] (F) edge (E);

   \draw[gray, <->, out= 235, in =125] (C) edge (D);

  \end{tikzpicture}
\caption{}
\label{fig:r13b}
    \end{subfigure}
    \caption{(a) Essential ancestral graph $\g$, (b) Representation of \ref{R13} used in Example \ref{ex:r13}.}
    \label{fig:r13}
\end{figure}

\begin{example}\label{ex:r13}
    Consider the essential ancestral graph $\g = (\mathbf{V,E})$ in Figure \ref{fig:r13}(a). Suppose we have expert knowledge $\mathcal{K}$ that $X_A$ does not cause $X_C$ or $X_D$, that is, $\mathcal{K} = \{A \arrowbullet C, A \arrowbullet D\}$. Once $\mathcal{K}$ is added to $\g$, as seen in left-hand-side of Figure \ref{fig:r13}(b), \ref{R13} implies that $A \circcirc B$ should be turned into $A \arrowcirc B$.  This is due to path $C \leftrightarrow A \leftrightarrow B$, unshielded path $C \arrowcirc E \circcirc F \circarrow D$ and possibly directed unshielded paths $A \circcirc B \circcirc E$, $A \circcirc B \circcirc F$. Once $A \arrowcirc B$ is added, we obtain {the} restricted essential ancestral graph on the right-hand side of Figure \ref{fig:r13}(b).

    To better understand \ref{R13}, consider what would happen if we added $A \to B$ to the graph on the left-hand side of Figure \ref{fig:r13}(b). Then $A \to B \circcirc E$, $A \to B \circcirc F$ and  \ref{R1}, would further imply $B \to E$ and $B \to F$. Furthermore, $C \leftrightarrow A \to B$, $D \leftrightarrow A \to B$ and \ref{R2}, would imply $C \leftrightarrow B$ and $D \leftrightarrow B.$ In turn, $C \leftrightarrow B \to E$, $D \leftrightarrow B \to F$ and \ref{R2} would then imply  $C \leftrightarrow E$ and $D \leftrightarrow F$.  However, now, either $C \leftrightarrow E \circcirc F$ or $D \leftrightarrow F \circcirc E$ and \ref{R1} would imply either $E \to F$ or $F \to E$ which in both cases leads to a new unshielded collider (either $C \leftrightarrow E \leftarrow F$, or $D \leftrightarrow F \leftarrow E$). This is not allowed in any MAG represented by $\g$. 
\end{example}

We now present the most complicated new rule, which will be a revision of \ref{R4new} (Theorem \ref{thm:rule4newnew}).
 To do this, we first define an almost collider path and an almost discriminating path (Definitions \ref{def:almost-collider} and \ref{def:almost-discriminating}).

\begin{definition}[Almost collider path] 
\label{def:almost-collider} Let $\g = (\mathbf{V,E})$ be a 
partial mixed graph.
Let $p = \langle A=Q_0, Q_1, \dots Q_k \rangle, k \ge 2$ be a path in $\g$. Then $p$ is an almost collider path if
\begin{enumerate}[label = (\roman*)]
\item\label{starting-node} \begin{enumerate}[label=(\alph*)]
            \item $Q_1$ is a collider on $p$, or
            \item\label{first-node-b} $Q_{0} \bulletarrow Q_1 \circarrow Q_{2}$, and  $Q_{0} \bulletcirc Q_{2}$ are in $\g$, or
            \item\label{first-node-c} $Q_{0} \bulletcirc Q_1 \arrowbullet Q_{2}$ and {$Q_{0} \bulletarrow Q_{2}$} are in $\g$,
    \end{enumerate}
\item\label{inside-node} for $i \in \{2, \dots , k-2\}$
    \begin{enumerate}[label=(\alph*)]
    \item $Q_i$ is a collider on $p$, or
    \item\label{inside-node-b} $Q_{i-1} \bulletarrow Q_i \circarrow Q_{i+1}$, and $Q_{i-1} \arrowcirc Q_{i+2}$ are in $\g$, or
    \item\label{inside-node-c} $Q_{i-1} \arrowcirc Q_i \arrowbullet Q_{i+1}$ and $Q_{i-1} \circarrow Q_{i+1}$ are in $\g$,
\end{enumerate}  
\item\label{ending-node} \begin{enumerate}[label=(\alph*)]
            \item $Q_{k-1}$ is a collider on $p$, or
        \item\label{last-node-b} $Q_{k-2} \bulletarrow Q_{k-1} \circbullet Q_{k}$, and {$Q_{k-2} \arrowbullet Q_{k}$} are in $\g$, or     
        \item\label{last-node-c} $Q_{k-2} \arrowcirc Q_{k-1} \arrowbullet Q_{k}$ and $Q_{k-2} \circbullet Q_{k}$ are in $\g$.
    \end{enumerate}     
\end{enumerate}
\end{definition}

\begin{definition}[Almost discriminating path] 
\label{def:almost-discriminating} Let $\g = (\mathbf{V,E})$ be a 
partial mixed graph.
Let $p = \langle A=Q_0, Q_1, \dots Q_k, Q_{k+1} = B\rangle, k \ge 2$ 
be a path in $\g$. Then $p$ is an almost discriminating path for $Q_k$ if  
\begin{enumerate}[label = (\roman*)]
\item $A \notin \Adj(B, \g)$, and
\item for all $i \in \{1, \dots , k-1\}$, $Q_i \to B$ is in $\g$, and
\item $p(A, Q_{k})$ is an almost collider path. 
\end{enumerate}
\end{definition}

 Naturally, the definition above subsumes the definition of a discriminating path. This leads us to define a new orientation rule, which can be seen as a generalization of \ref{R4new}.

\begin{theorem} \label{thm:rule4newnew}  Let $\g = (\mathbf{V,E})$ be a 
partial mixed graph. 
\begin{enumerate}[label = R4]
    \item\label{R4}  If $\langle A=Q_0, Q_1, \dots Q_k, Q_{k+1} = B\rangle, k \geq 2$ is an almost discriminating path for   $Q_k$ in $\g$ and if $Q_k \circbullet B$ is in $\g$, then $Q_k \to B$ is present in all MAGs represented by $\g$. 
\end{enumerate}
\end{theorem}

 %We remark that \ref{R11} can be seen as a special case of \ref{R4}, but we feel this rule conflation would not be pedagogical, so we leave the two rules separate. 
\begin{figure}[!t]
\begin{subfigure}{.5\textwidth}
  \centering
	\begin{tikzpicture}[->,>=stealth',shorten >=1pt,auto,node distance=1.2cm,scale=.55,transform shape,font = {\huge\bfseries\sffamily}]
 \tikzstyle{every edge}=[draw,>=stealth',semithick]
	\tikzstyle{state}=[inner sep=5pt, minimum size=5pt]

       \node[state] (A) at (0,2) {$A$}; 
	\node[state] (B) at (4,2) {$B$}; 
	\node[state] (D) at (0,-2) {$D$}; 
	\node[state] (C) at (4,-2) {$C$}; 
	\node[state] (E) at (2, 0) {$E$};

	\draw[gray,<-o]  (B) edge (C);
	\draw[gray, o->] (C) edge (D);
	\draw[gray,<-o] (E) edge (A);
	\draw[gray,o-o] (E) edge (B);
	\draw[gray,<-o] (E) edge (C);
	\draw[gray,o-o] (E) edge (D);
	\draw[gray,o->] (A) edge (B);
	\draw[gray,o->] (A) edge (D);

  \end{tikzpicture}
  \caption{}
\end{subfigure}
 \hspace{-2cm}
    \vrule
    \vspace{.5cm}
\begin{subfigure}{.5\textwidth}
  \centering
	\begin{tikzpicture}[->,>=stealth',shorten >=1pt,auto,node distance=1.2cm,scale=.55,transform shape,font = {\huge\bfseries\sffamily}]
 \tikzstyle{every edge}=[draw,>=stealth',semithick]
	\tikzstyle{state}=[inner sep=5pt, minimum size=5pt]
	\node[state] (A) at (-8,2) {$A$}; 
	\node[state] (B) at (-4,2) {$B$}; 
	\node[state] (D) at (-8,-2) {$D$}; 
	\node[state] (C) at (-4,-2) {$C$}; 
	\node[state] (E) at (-6, 0) {$E$};

	\draw[gray,<-o]  (B) edge (C);
	\draw[gray,o->] (C) edge (D);
	\draw[gray,<-o] (E) edge (A);
	\draw[->] (E) edge (B);
	\draw[gray,<-o] (E) edge (C);
	\draw[gray,o-o] (E) edge (D);
	\draw[->] (A) edge (B);
	\draw[<->] (A) edge (D);
          
     \coordinate [label=above:$\Rightarrow$] (L3) at (-3,0);

	\node[state] (A) at (-2,2) {$A$}; 
	\node[state] (B) at (2,2) {$B$}; 
	\node[state] (D) at (-2,-2) {$D$}; 
	\node[state] (C) at (2,-2) {$C$}; 
	\node[state] (E) at (0, 0) {$E$};
 
	\draw[blue,<-]  (B) edge (C);
	\draw[gray,o->] (C) edge (D);
	\draw[gray,<-o] (E) edge (A);
	\draw[->] (E) edge (B);
	\draw[gray,<-o] (E) edge (C);
	\draw[gray,o-o] (E) edge (D);
	\draw[->] (A) edge (B);
	\draw[<->] (A) edge (D);

  \end{tikzpicture}
  \caption{}
\end{subfigure}
\begin{subfigure}{\columnwidth}
  \centering
	\begin{tikzpicture}[->,>=stealth',shorten >=1pt,auto,node distance=1.2cm,scale=.55,transform shape,font = {\huge\bfseries\sffamily}]
 \tikzstyle{every edge}=[draw,>=stealth',semithick]
	\tikzstyle{state}=[inner sep=5pt, minimum size=5pt]

        \node[state] (A) at (-6,2) {$A$}; 
	\node[state] (B) at (-2,2) {$B$}; 
	\node[state] (D) at (-6,-2) {$D$}; 
	\node[state] (C) at (-2,-2) {$C$}; 
	\node[state] (E) at (-4, 0) {$E$};

	\draw[purple,<->]  (B) edge (C);
	\draw[gray,o->] (C) edge (D);
	\draw[gray,<-o] (E) edge (A);
	\draw[->] (E) edge (B);
	\draw[gray,o->] (E) edge (C);
	\draw[gray,o-o] (E) edge (D);
	\draw[->] (A) edge (B);
	\draw[<->] (A) edge (D);
          
	\coordinate [label=above:$\Rightarrow$] (L3) at (-1,0);

	\node[state] (A) at (0,2) {$A$}; 
	\node[state] (B) at (4,2) {$B$}; 
	\node[state] (D) at (0,-2) {$D$}; 
	\node[state] (C) at (4,-2) {$C$}; 
	\node[state] (E) at (2, 0) {$E$};

    \draw[purple,<->]  (B) edge (C);
	\draw[purple,->] (C) edge (D);
	\draw[gray,<-o] (E) edge (A);
	\draw[->] (E) edge (B);
	\draw[purple,<->] (E) edge (C);
	\draw[gray,o-o] (E) edge (D);
	\draw[->] (A) edge (B);
	\draw[<->] (A) edge (D);
          
	\coordinate [label=above:$\Rightarrow$] (L4) at (5,0);

	\node[state] (A) at (6,2) {$A$}; 
	\node[state] (B) at (10,2) {$B$}; 
	\node[state] (D) at (6,-2) {$D$}; 
	\node[state] (C) at (10,-2) {$C$}; 
	\node[state] (E) at (8, 0) {$E$};

    \draw[purple,<->]  (B) edge (C);
	\draw[purple,->] (C) edge (D);
	\draw[purple,<->] (E) edge (A);
	\draw[->] (E) edge (B);
	\draw[purple,<->] (E) edge (C);
	\draw[purple,->] (E) edge (D);
	\draw[->] (A) edge (B);
	\draw[<->] (A) edge (D);

  \end{tikzpicture}
  \caption{}
\end{subfigure}
\caption{(a) Essential ancestral graph $\g$, (b) Representation or \ref{R4}, and (c) additional graphs used in Example \ref{ex:r4}.}
\label{fig:counterexample}
\end{figure}
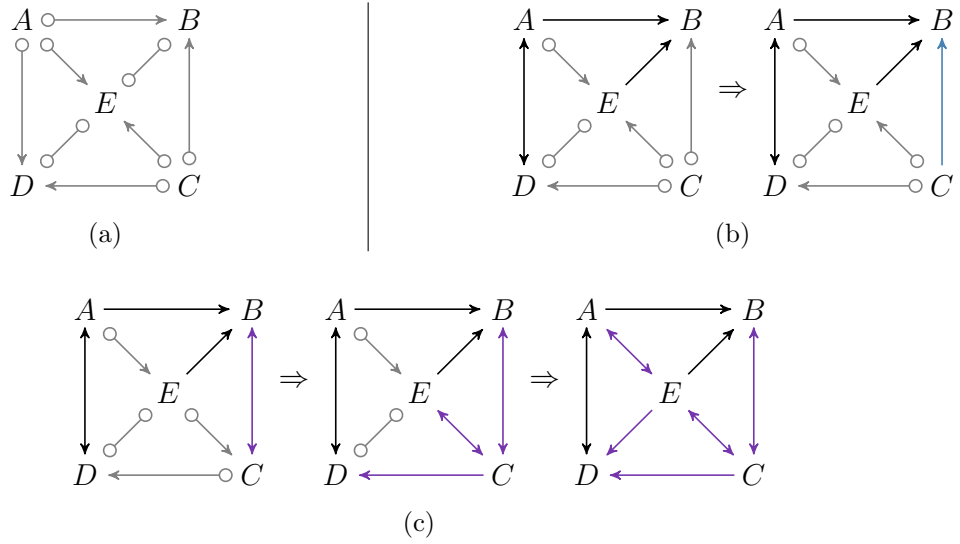

\begin{example}\label{ex:r4}
 Consider the essential ancestral graph $\g$ in Figure \ref{fig:counterexample}(a).  Suppose that we want to include expert knowledge $\mathcal{K}$ that $X_A$ is not a cause of $X_D$, $\mathcal{K} = \{A \arrowbullet D\}$. 
    Since $A \circarrow D$ is already in $\g$, adding our orientation knowledge results in $A \leftrightarrow D$, see graph $\g_{1}$ on the left-hand-side of Figure \ref{fig:counterexample}(b). Furthermore, since $D \notin \Adj(B, \g)$ and since $D \leftrightarrow A \circarrow B$ is in $\g_{1}$, \ref{R1} implies $A \to B$ is in $\g_{1}$. Furthermore,  the $\{D,A,B, E\}$  induced subgraph of $\g_{1}$ and  \ref{R11} 
    imply $E \to B$.

    However, orientations in $\g_{1}$ are still not closed under \ref{R4} due to path  $p = \langle D, A, E, C, B \rangle$, which is an almost discriminating path. To see this, consider that $D \notin \Adj(B, \g)$ and that $A \to B$, $E \to B$ are in $\g_{1}$. Furthermore, path $D \leftrightarrow A \circarrow E \arrowcirc C$ is an almost collider path in $\g_{1}$ due to the presence of the edge $D \circcirc E$.  Therefore, 
 \ref{R4} implies that $C \to B$ should be oriented in $\g_{1}$. We include this orientation to obtain a partial mixed graph $\g^{\prime}$ on the right-hand-side of Figure \ref{fig:counterexample}(b) which is {the} restricted essential ancestral graph.

    To explore why orienting $B \leftrightarrow C$ would lead to an issue, consider Figure \ref{fig:counterexample}(c). The left-hand-side graph in Figure \ref{fig:counterexample}(c) contains a graph derived from $\g_{1}$ by orienting {\color{purple} $B \leftrightarrow C$}. 
    The edge orientation {\color{purple} $B \leftrightarrow C$} now implies a few more orientations.
   For instance, {\color{purple}$C \leftrightarrow B$}, \ref{R2} and $E \to  {\color{purple} B \leftrightarrow C}$ imply {\color{purple}$E \leftrightarrow B$}. Furthermore, \ref{R1}, and {\color{purple} $B \leftrightarrow C \circarrow D$} imply {\color{purple}$C \to D$}. These two orientations are represented in the graph in the middle of Figure \ref{fig:counterexample}(c).
Next,  \ref{R11} implies {\color{purple} $E \to D$}. Lastly, \ref{R2} and ${\color{purple} E \to D }\leftrightarrow A$ imply {\color{purple} $E \leftrightarrow A$}. These two additional edge orientations are given in the mixed graph $\g^{*}$ on the right-hand side of Figure \ref{fig:counterexample}(c).

   Graph $\g^{*}$ is ancestral. However,  $\g^{*}$ contains path $q$ of the form $D \leftrightarrow {\color{purple}A \leftrightarrow E \leftrightarrow C \leftrightarrow B}$, and $D \notin \Adj(B, \g)$ meaning that $q$ is a new minimal collider path in $\g^{*}$ compared to $\g$. 
   Moreover, edges $A \to B$, $E \to B$ {\color{purple} $C \to D$} are in $\g^{*}$ implying that  $q$ is not only a new minimal collider path but also an inducing path in $\g^{*}$. Hence, $\g^{*}$ is not a MAG.
\end{example}

% \subsection{Additional Rules of \cite{wangpolynomial}}
\subsection{Rules of \cite{wangpolynomial}}
{
While our manuscript was under review, \cite{wangpolynomial} introduced two additional rules applicable to our setting. At the request of the reviewers, we reproduce these rules here and include them in our R package implementation (detailed in the next section). We label these rules as \ref{R14} and \ref{R18}, presenting them in Theorems \ref{thm:r14} and \ref{thm:r18}. To support \ref{R14}, we also include Definition \ref{def:priorto} from \cite{wangpolynomial}. This definition and both theorems have been slightly rephrased from \cite{wangpolynomial} to maintain consistent notation and simplify their structure. Nevertheless, Theorems \ref{thm:r14} and \ref{thm:r18} yield the same orientations as the original results in \cite{wangpolynomial}, following Propositions 1 and 2 of \cite{wangpolynomial}, Corollary \ref{cor:main-result-R13-eq}, and Lemma \ref{lemma:unshielded-poss-dir}.
%However, we note that these rules are not needed for the theoretical results of this manuscript this paper and are thus, not included in those.

% \begin{definition}[Unbridged path relative to $\mb{V'}$, \citealp{wangpolynomial}]\label{def:unbridged}
% Let $\mb{V'} \subseteq \mb{V}$ be a node set in a partial mixed graph $\g = (\mb{V,E})$. A path $p$, of the form $V_0 \leftrightarrow \dots \leftrightarrow V_k$, $k \ge 1$, is called an unbridged path relative to $\mb{V'}$ if there are nodes $C_1, C_2 \in \mb{V'}$ such that 
% \begin{enumerate}[label = (\roman*)]
%     \item $C_1 \arrowcirc V_0$ and $V_k \circarrow C_2$ are in $\g$,
%     \item $C_1 \leftrightarrow C_2$ is in $\g$, and
%     \item any two non-consecutive nodes on $r = \langle C_1, V_0\rangle \oplus p \oplus \langle V_k, C_2 \rangle$, other than $C_1, C_2$, are non-adjacent in $\g$.
% \end{enumerate}  % $C_1 \leftrightarrow C_2$ in $\g$.
% \end{definition}

\begin{definition}[\citealp{wangpolynomial}]\label{def:priorto}
Let $\g = (\mb{V,E})$ be a partial mixed graph and consider nodes $A$, $X$, and $B$ in $\g$. Also let $\mb{S}$ be the set of all nodes $S$, such that $S \leftrightarrow X$ is in $\g$. Then $A$ is said to be prior to $B$ relative to $X$ in $\g$ if
\begin{enumerate}[label = (\roman*)]
    \item  there is a set of nodes $\{F_0 = A, F_1, \dots , F_t = B\}, t \geq 1$, such that for all $i \in \{0, \dots, t-1\}, F_i \bulletcirc X$ is in $\g$, and 
    \item for each $i \in \{1, \dots, t-1\}$ one of the following holds:
 \begin{enumerate}
    \item  $F_i \to F_{i+1}$ in $\g$, or
    \item the unshielded possibly directed path $\langle X, F_i, \dots, M \rangle$ and  edge $M \to F_{i+1}$ are in $\g$, or
    \item there is an unshielded path of the form $C \arrowcirc V_1 \circcirc \dots \circcirc V_k \circarrow D$, $k >1$, where $C,D \in \mb{S} \cup \{F_{i+1}\}$ and  $V_1, \dots, V_i \notin \mb{S}\cup\{F_i, F_{i+1}\} = \emptyset$, and there exists an unshielded possibly directed path $\langle X, F_i, \dots, V_j \rangle$,  in $\g$,  for every $j \in \{1, \dots, k\}$.
\end{enumerate}   
\end{enumerate}
\end{definition}

\begin{theorem}%[c.f.\ R14 and Proposition 1 of \citealp{wangpolynomial}]
\label{thm:r14} Let $\g = (\mb{V,E})$ be a partial mixed graph. 
\begin{enumerate}[label= R14]
    \item\label{R14} Suppose an edge $A \circarrow B$ and path  $T_1 \bulletcirc X \circbullet T_2$ are in $\g$, such that both $T_1$ and $T_2$ are prior to $B$ relative to $X$ (Definition \ref{def:priorto}). Then  $A \leftrightarrow B$ is present in all MAGs represented by $\g$ if either 
    \begin{enumerate}
        \item $T_1$ is not adjacent to $T_2$, or 
        \item $T_1 \leftrightarrow T_2$ is in $\g$, and there is a path $t$ of the form $T_1 \arrowcirc V_1 \circcirc \dots \circcirc V_k \circarrow T_2$, $k >1$ in $\g$, such that each pair of non-consecutive nodes on $t$ (other than $T_1,T_2$) are not adjacent in $\g$ and such that $A  \circbullet T_i$ is in $\g$ for all $i \in \{1, \dots, k\}.$
    \end{enumerate}
    \end{enumerate}
\end{theorem}

\begin{theorem}%[c.f.\ R18 and Proposition 2 of \citealp{wangpolynomial}]
\label{thm:r18}
Let $\g = (\mb{V,E})$ be a partial mixed graph. 
\begin{enumerate}[label= R15]
    \item\label{R18} Suppose that $A \circarrow B \leftrightarrow D$ and two unshielded possibly directed paths $\langle D, T_1, \dots, A \rangle$ and $\langle D, T_2, \dots, A \rangle$ are in $\g$. Then $A \leftrightarrow B$ is present in all MAGs represented by $\g$ if either, 
    \begin{enumerate}
        \item $T_1$ is not adjacent to $T_2$ is in $\g$, or 
        \item $T_1 \leftrightarrow T_2$, and there is a path $t$ of the form $T_1 \arrowcirc V_1 \circcirc \dots \circcirc V_k \circarrow T_2$, $k >1$ in $\g$, such that each pair of non-consecutive nodes on $t$ (other than $T_1,T_2$) are not adjacent in $\g$ and such that $D  \circbullet T_i$ is in $\g$ for all $i \in \{1, \dots, k\}.$
    \end{enumerate}
\end{enumerate} 
\end{theorem}
}

\section{Incorporating Orientation Knowledge}
\label{sec:completeness}

We now introduce the \texttt{{addOrientationKnowledge}} algorithm (Algorithm \ref{alg:mpag}). Algorithm \ref{alg:mpag} takes as input a partial mixed graph $\g$, which could be an essential or a restricted essential ancestral graph, and a set of expert knowledge $\mathcal{K}.$ The algorithm proceeds to either create a partial mixed graph $\g^{\prime}$ or FAIL by adding   $\mathcal{K}$ and completing  \ref{R1}, \ref{R2}, \ref{R4}, \ref{R8}, \ref{R10}-{ \ref{R18}}. Algorithm \ref{alg:mpag} will fail if, at some point, it cannot add an element of $\mathcal{K}$ as orientation knowledge, that is, if an element of $\mathcal{K}$ is not admissible as per the following definition.  

 \begin{definition}[Admissible orientation]\label{def:admissible} 
 Let $\g = (\mathbf{V},\mathbf{E})$ be a partial mixed graph, and let $\langle \langle A, B \rangle \rangle $ be a piece of orientation knowledge. 
 Then $\langle \langle A, B \rangle \rangle $ is  admissible for $\g$ if {$A \in \Adj(B, \g)$}, %edge $\langle A, B \rangle$ is in $\g$, 
 and if one of the following hold
 \begin{enumerate}[label = (\roman*)]
    \item if $\langle \langle A, B \rangle \rangle$ is of the form $A \to B$, $\g$ contains  $A \circcirc B$, $A \circarrow B$ or $A \to B$, or
    \item if $\langle \langle A, B \rangle \rangle$ is of the form $A \leftarrow B$, $\g$ contains  $A \circcirc B$, $A \arrowcirc B$ or $A \leftarrow B$, or
     \item if $\langle \langle A, B \rangle \rangle$ is of the form $A \bulletarrow  B$, $\g$ contains $A \circcirc B$, $A \circarrow B$, {$A \leftrightarrow B$, or $A \to B$}
     \item  if $\langle \langle A, B \rangle \rangle $ is of the form $A \arrowbullet B$, $\g$ contains $A \circcirc B$, {$A \arrowcirc B$, $A \leftrightarrow B$, or $A \leftarrow B$.}
 \end{enumerate}
 \end{definition}

\begin{algorithm}[!t]
\caption{{addOrientationKnowledge}}
\label{alg:mpag}
\begin{algorithmic}[1]
    \Require Partial mixed graph $\g = (\mathbf{V,E})$, and  orientation knowledge set $\mathcal{K}$.
    \Ensure Partial mixed graph $\g^{\prime} = (\mathbf{V, E^{\prime}})$, or FAIL.
    \State Let $\g^{\prime} = \g$
    \For {piece of orientation knowledge $\langle \langle A, B \rangle \rangle \in \mathcal{K}$}
        \If {$\langle \langle A, B \rangle \rangle$ is admissible with $\g$}
            \State Orient  $\langle \langle A, B \rangle \rangle$ in $\g^{\prime}$
            \State Close orientations under \ref{R1}, \ref{R2}, \ref{R4}, \ref{R8}, \ref{R10}, \ref{R11}, \ref{R12}, \ref{R13},  {  \ref{R14} and \ref{R18}} in $\g^{\prime}$.
        \Else \ \Return $\mathbf{FAIL}$
        \EndIf
    \EndFor
    \State \Return  $\g^{\prime}$
\end{algorithmic}
\end{algorithm}

{ The \texttt{{addOrientationKnowledge}}$(\g,\mathcal{K})$ will fail if and only if one of the following holds: the input $\g$ does not represent any MAG (for instance, if $\g$ is not ancestral), or if the set of orientation knowledge $\mathcal{K}$ is not consistent for any MAG represented by $\g$.} For an example of the latter, consider that $\g$ is the essential ancestral graph in Figure \ref{fig:example_MEC_MAG}(c) and orientation knowledge is $\mathcal{K}_{1} = \{B \bulletarrow C, C \to D, D \to A\}$. Note that \texttt{{addOrientationKnowledge}}$(\g,\mathcal{K}_{1})$ will first  add $B \bulletarrow C$  to $\g$ and close the orientation rules to obtain graph $\g^{\prime}$ in Figure \ref{fig:example_MEC_MAG}(d). After that, $C \to D$ can be added without any additional change to $\g^{\prime}$, but the algorithm fails when it attempts to add the non-admissible orientation $D \to A$ to $\g^{\prime}$. 

 Proposition \ref{prop:alg2-sound} describes a scenario where Algorithm \ref{alg:mpag}  will not output a FAIL. Proposition \ref{prop:alg2-sound} holds directly by definition of consistent orientation knowledge and Corollary \ref{cor:old-rules-sound}, and Theorems \ref{thm:meek}, \ref{thm:rule12}, \ref{thm:rule13star}, \ref{thm:rule4newnew}, { \ref{thm:r14}, \ref{thm:r18}}.

\begin{proposition}
\label{prop:alg2-sound}
Let $\g = (\mb{V,E})$ be a restricted essential ancestral graph and $\mathcal{K}$ be a set of orientation knowledge edge marks consistent with $\g$. Then \texttt{{addOrientationKnowledge}($\g, \mathcal{K}$)} (Algorithm \ref{alg:mpag}) will not output FAIL.
\end{proposition}

One may be surprised that Algorithm \ref{alg:mpag} does not use  \ref{R3} and \ref{R9}. We show in Lemma \ref{lemma:no-contradiction-R9} that \ref{R3} and \ref{R9} are not needed when adding orientation knowledge to an essential or a restricted essential ancestral graph. Hence, we only recommend using   Algorithm \ref{alg:mpag} to add orientation knowledge to an essential or a restricted essential ancestral graph.

\begin{lemma}\label{lemma:no-contradiction-R9}  
 Let $\g= (\mb{V,E})$  be an essential ancestral graph or a restricted essential ancestral graph, and let $\mathcal{K}$ be a set of orientation knowledge edge marks consistent with $\g$. Let $\g^{\prime} = \texttt{{addOrientationKnowledge}}(\g, \mathcal{K})$. Then orientations in $\g^{\prime}$ are closed with respect to \ref{R3} and \ref{R9}.
\end{lemma}

\subsection{Properties of Partial Mixed Graphs with Sound Orientations}

Before considering the completeness of the new set of orientation rules, we reflect on properties that a partial mixed graph $\g^{\prime}$ must satisfy to have sound edge orientations. Hence, let $\g$ be an essential ancestral graph and let $\g^{\prime}$ be a graph on the same set of nodes and with the same adjacencies as $\g$ and such that every invariant edge mark in $\g$ is identical in $\g^{\prime}$.

For any MAG to be represented by a partial mixed graph $\g^{\prime}$, $\g^{\prime}$ must be ancestral and $\g^{\prime}$ cannot contain an inducing path. 
The following lemma then tells us that $\g^{\prime}$ will be ancestral as long as it does not contain directed or almost directed cycles of length $3$, that is, as long as it does not contain $V_1 \to V_2 \to V_3$ and an edge $V_1 \arrowbullet V_3$.

\begin{lemma}\label{lemma:nocycle3}
Let $\g = (\mathbf{V,E})$ be an essential ancestral graph and $\g^{\prime} = (\mathbf{V,E'})$ a partial mixed graph such that $\g$ and $\g^{\prime}$  have the same skeleton, and every invariant edge mark in $\g$ is identical in $\g^{\prime}$.
Furthermore, suppose that edge orientations in $\g^{\prime}$ are closed under \ref{R2}, \ref{R8}. If $\g^{\prime}$ is not ancestral, then there is a directed or almost directed cycle of length $3$ in $\g^{\prime}$. 
\end{lemma}

An ancestral mixed graph that contains no inducing paths is called maximal (see Corollary 4.4 of \citealp{richardson2002ancestral}). In order to define the maximal property for ancestral partial mixed graphs, we first expand the definition of inducing paths.

\begin{definition}[{ Possibly} inducing path] \label{def:poss-ind-path} Let  $\g = (\mathbf{V,E})$ be a partial mixed graph and $A, B \in \mb{V}$, $A \neq B$. A path $p = \langle A, Q_1, \dots, Q_k , B \rangle$, $k > 1$, is a { possibly } inducing path in $\g$ if $p$ is a collider path in $\g$,  $A \notin \Adj(B, \g)$, and $Q_i \in \PossAn(\{A,B\}, \g)$ for all $i \in \{1, \dots, k\}$.
\end{definition}

\begin{definition}[Maximal partial mixed graph] \label{def:max-pmg}  Let  $\g = (\mathbf{V,E})$ be an ancestral partial mixed graph. We say that $\g$ is maximal if $\g$ contains no { possibly } inducing paths.
\end{definition}

We now introduce two important results regarding the maximal property of ancestral partial mixed graphs. { Let $\g = (\mathbf{V,E})$ be an essential ancestral graph and  let $\g^{\prime} = (\mathbf{V,E'})$ be an ancestral partial mixed graph  such that $\g$ and $\g^{\prime}$ have the same skeleton, and every invariant edge mark in $\g$ is identical in $\g^{\prime}$.} Lemma \ref{lemma:pag-to-eg-to-mag} shows that as long as $\g^{\prime}$ is ancestral, and $\g^{\prime}$ and $\g$ contain the same minimal collider paths, $\g^{\prime}$ is maximal. Moreover, by Lemma \ref{lemma:no-new-mcps}, $\g^{\prime}$ and $\g$ contain the same minimal collider paths, as long as $\g^{\prime}$ does not contain new unshielded colliders or new colliders discriminated by a path compared to $\g$.

\begin{lemma} \label{lemma:pag-to-eg-to-mag}
Let $\g = (\mathbf{V,E})$ be an essential ancestral graph and   $\g^{\prime} = (\mathbf{V,E'})$ be an ancestral partial mixed graph, such that $\g$ and $\g^{\prime}$ have the same skeleton and minimal collider paths, and every invariant edge mark in $\g$ is identical in $\g^{\prime}$. Then $\g^{\prime}$ is maximal.
\end{lemma}

\begin{lemma}\label{lemma:no-new-mcps} 
Let $\g= (\mathbf{V, E})$ be an essential ancestral graph and $\g^{\prime} = (\mathbf{V, E'})$ be an ancestral partial mixed graph such that $\g$ and $\g^{\prime}$ have the same skeleton and such that every invariant edge mark in $\g$ is identical in $\g^{\prime}$. Furthermore, suppose that  orientations in $\g^{\prime}$ are closed under \ref{R1}, \ref{R2}, and \ref{R4}. Then every minimal collider path in $\g^{\prime}$ is also a minimal collider path in $\g$ if and only if:
\begin{enumerate}[label = (\roman*)]
\item\label{no-new-mcp:1} All unshielded colliders in $\g^{\prime}$ are also unshielded colliders in $\g$, and
\item\label{no-new-mcp:2} for every discriminating collider path $\langle A, Q_1, \dots, Q_k, B \rangle$, $k \ge 2$ in $\g^{\prime}$, $Q_{k-1} \bulletarrow Q_k \arrowbullet B$ is in $\g$.
\end{enumerate}
 \end{lemma}

\subsection{Completeness of Orientations Rules in Certain Scenarios}\label{section:partial-completeness}

We now prove that a graph output by Algorithm \ref{alg:mpag} will be sound and complete in specific scenarios. In Theorem \ref{thm:chordal-completeness}, we show that Algorithm \ref{alg:mpag} is sound and complete if the input essential ancestral graph $\g$ has no minimal collider paths. Theorem \ref{thm:chordal-completeness} can be seen as a generalization of Theorem 4 of \cite{meek1995causal}. We also note that our proof corrects an error in the proof given by \cite{meek1995causal} (see Example \ref{ex:lemma6-meek-issue} in Section \ref{sec:meek-analogue-main} for details). We include a proof sketch for Theorem \ref{thm:chordal-completeness} below, while the full proof is given in Section \ref{sec:meek-analogue-main}.

\begin{theorem}
\label{thm:chordal-completeness}
Suppose that $\g^{\prime}= (\mb{V,E'})$  is an ancestral and maximal partial mixed graph with no minimal collider paths, such that the {skeleton} of $\g^{\prime}$ is chordal (Definition \ref{def:chordal-graph}) and such that the edge orientations in {$\g^{\prime}$}  are {closed} under { \ref{R1}, \ref{R2}, \ref{R8}, \ref{R11}}. %\ref{R1}-\ref{R4}, \ref{R8}-\ref{R13}. 
{Then, the orientations in $\g^{\prime}$ are sound and complete. Specifically,} 
\begin{enumerate}[label=(\roman*)]
\item\label{claim1:chor-comp} If $A \circcirc B$ is in $\g^{\prime}$, then there are {at least} three MAGs $\g[M]_1$, $\g[M]_2$, and $\g[M]_3$ represented by $\g^{\prime}$ such that $A \to B$ is in $\g[M]_1$,   $A \leftarrow B$ is in $\g[M]_2$, and $A \leftrightarrow B$ is in $\g[M]_3$. 
\item  If $A \circarrow B$ is in $\g^{\prime}$, then there are {at least} two MAGs $\g[M]_1$ and $\g[M]_2$ represented by $\g^{\prime}$ such that $A \to B$ is in $\g[M]_1$, and  $A \leftrightarrow B$ is in $\g[M]_2$.
\end{enumerate}
\end{theorem}

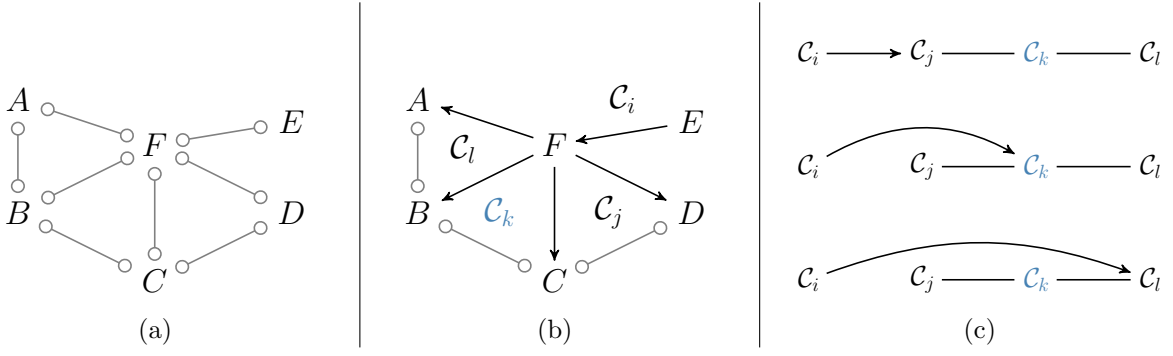
\begin{figure}
    \centering
    \begin{subfigure}{.31\textwidth}
          \tikzstyle{every edge}=[draw,>=stealth',semithick]
        \vspace{1cm}
        \centering
        \begin{tikzpicture}[->,>=stealth',shorten >=1pt,auto,node distance=1.2cm,scale=0.6,transform shape,font = {\huge\bfseries\sffamily}]
	\tikzstyle{state}=[inner sep=5pt, minimum size=5pt]

        \node[state] (A) at (-3,2.5) {$A$};
	\node[state] (B) at (-3,0) {$B$};
	\node[state] (C) at (0,-1.5) {$C$};
	\node[state] (F) at (0,1.5) {$F$};
	\node[state] (D) at (3,0) {$D$};
          \node[state] (E) at (3,2) {$E$};

	\draw[gray,o-o]  (C) edge (D);
	\draw[gray,o-o] (F) edge (D);
	\draw[gray,o-o] (F) edge (B);
	\draw[gray,o-o] (B) edge (C);
	\draw[gray,o-o] (C) edge (F);

        \draw[gray,o-o] (F) edge (A);
        \draw[gray,o-o] (B) edge (A);
 	\draw[gray,o-o] (F) edge (E);
  \end{tikzpicture}
\caption{}
\label{fig:jointree000} 
    \end{subfigure}
    \vrule
    \begin{subfigure}{.31\textwidth}
          \tikzstyle{every edge}=[draw,>=stealth',semithick]
        \vspace{1cm}
        \centering
        \begin{tikzpicture}[->,>=stealth',shorten >=1pt,auto,node distance=1.2cm,scale=0.6,transform shape,font = {\huge\bfseries\sffamily}]
	\tikzstyle{state}=[inner sep=5pt, minimum size=5pt]

        \node[state] (A) at (-3,2.5) {$A$};
	\node[state] (B) at (-3,0) {$B$};
	\node[state] (C) at (0,-1.5) {$C$};
	\node[state] (F) at (0,1.5) {$F$};
	\node[state] (D) at (3,0) {$D$};
          \node[state] (E) at (3,2) {$E$};

	\draw[gray,o-o]  (C) edge (D);
	\draw[->] (F) edge (D);
	\draw[->] (F) edge (B);
	\draw[gray,o-o] (B) edge (C);
	\draw[<-] (C) edge (F);

        \draw[->] (F) edge (A);
        \draw[gray,o-o] (B) edge (A);
 	\draw[<-] (F) edge (E);

        \node[state] (Cl) at (-2,1.4) {$\g[C]_l$};
	\node[state] (Ck) at (-1.2,0) {\color{blue} $\g[C]_k$};
	\node[state] (Cj) at (1.2,0) {$\g[C]_j$};
 	\node[state] (Ci) at (1.5,2.5) {$\g[C]_i$};
  \end{tikzpicture}
\caption{}
\label{fig:jointree111} 
    \end{subfigure}
    \vrule
    \begin{subfigure}{.35\textwidth}
        \centering
       \tikzstyle{every edge}=[draw,>=stealth',semithick]
	\begin{tikzpicture}[->,>=stealth',shorten >=1pt,auto,node distance=1.2cm,scale=0.5,transform shape,font = {\huge\bfseries\sffamily}]
	\tikzstyle{state}=[inner sep=5pt, minimum size=5pt]

	\node[state] (J1i) at (-3,2) {$\g[C]_i$};
	\node[state] (J1j) at (0,2) {$\g[C]_j$};
	\node[state] (J1k) at (3,2) {\color{blue}  $\g[C]_k$};
        \node[state] (J1l) at (6,2) {$\g[C]_l$};
       
        \draw[->]  (J1i) edge (J1j);
        \draw[-]  (J1j) edge (J1k);
        \draw[-]  (J1k) edge (J1l);

        \node[state] (J2i) at (-3,-1) {$\g[C]_i$};
	  \node[state] (J2j) at (0,-1) {$\g[C]_j$};
        \node[state] (J2k) at (3,-1) {\color{blue}  $\g[C]_k$};
        \node[state] (J2l) at (6,-1) {$\g[C]_l$};
       
        \draw[->, out = 30, in= 150]  (J2i) edge (J2k);
        \draw[-]  (J2j) edge (J2k);
        \draw[-]  (J2k) edge (J2l);
     
 	\node[state] (J3i) at (-3,-4) {$\g[C]_i$};
	\node[state] (J3j) at (0,-4) {$\g[C]_j$};
	\node[state] (J3k) at (3,-4) {\color{blue} $\g[C]_k$};
        \node[state] (J3l) at (6,-4) {$\g[C]_l$};
       
        \draw[->, out = 20, in= 160]  (J3i) edge (J3l);
        \draw[-]  (J3j) edge (J3k);
         \draw[-]  (J3k) edge (J3l);
	
     \end{tikzpicture}
    \caption{}
    \label{fig:jointree222}
    \end{subfigure}
    \caption{(a) Essential ancestral graph $\g$, (b) partial mixed graph $\g^{\prime}$ and (c) three partially directed join trees for $\g^{\prime}$ all explored in the proof sketch of Theorem \ref{thm:chordal-completeness} and Examples \ref{ex:l7p2} and \ref{ex:algo4-demo} in Section \ref{sec:meek-analogue-main}.}
    \label{fig:lemma7-issue-p2}
\end{figure}

\begin{figure}
\centering
 \begin{subfigure}{.4\textwidth}
 \centering
        \tikzstyle{every edge}=[draw,>=stealth',semithick]
	\begin{tikzpicture}[->,>=stealth',shorten >=1pt,auto,node distance=1.2cm,scale=0.5,transform shape,font = {\huge\bfseries\sffamily}]
	\tikzstyle{state}=[inner sep=5pt, minimum size=5pt]

	\node[state] (J1i) at (-3,2) {};
	\node[state] (J1j) at (0,2) {};
	\node[state] (J1k) at (3,2) {};
        \node[state] (J1l) at (6,2) {};
        
        \node[state] (J2i) at (-3,-1) {$\g[C]_i$};
	  \node[state] (J2j) at (0,-1) {$\g[C]_j$};
        \node[state] (J2k) at (3,-1) {\color{blue}  $\g[C]_k$};
        \node[state] (J2l) at (6,-1) {$\g[C]_l$};
       
        \draw[->, out = 30, in= 150]  (J2i) edge (J2k);
        \draw[<-, color=blue]  (J2j) edge (J2k);
        \draw[->, color=blue]  (J2k) edge (J2l);

       \node[state] (J3i) at (-3,-4) {};
	\node[state] (J3j) at (0,-4) {};
	\node[state] (J3k) at (3,-4) {};
        \node[state] (J3l) at (6,-4) {};
  \end{tikzpicture}
\caption{}
\label{fig:jointree-final1} 
    \end{subfigure}
    \vrule
    \begin{subfigure}{.5\textwidth}
    \centering
       \tikzstyle{every edge}=[draw,>=stealth',semithick]
	\begin{tikzpicture}[->,>=stealth',shorten >=1pt,auto,node distance=1.2cm,scale=0.6,transform shape,font = {\huge\bfseries\sffamily}]
	\tikzstyle{state}=[inner sep=5pt, minimum size=5pt]

        \node[state] (A) at (-3,2.5) {$A$};
	\node[state] (B) at (-3,0) {$B$};
	\node[state] (C) at (0,-1.5) {$C$};
	\node[state] (F) at (0,1.5) {$F$};
	\node[state] (D) at (3,0) {$D$};
          \node[state] (E) at (3,2) {$E$};

	\draw[->, color=blue]  (C) edge (D);
	\draw[->] (F) edge (D);
	\draw[->] (F) edge (B);
	\draw[gray,o-o] (B) edge (C);
	\draw[<-] (C) edge (F);

        \draw[->] (F) edge (A);
        \draw[->, color=blue] (B) edge (A);
 	\draw[<-] (F) edge (E);

        \node[state] (Cl) at (-2,1.4) {$\g[C]_l$};
	\node[state] (Ck) at (-1.2,0) {\color{blue}$\g[C]_k$};
	\node[state] (Cj) at (1.2,0) {$\g[C]_j$};
 	\node[state] (Ci) at (1.5,2.5) {$\g[C]_i$};
  \end{tikzpicture}
\caption{}
\label{fig:jointree-final2} 
    \end{subfigure}
    \caption{(a) Directed join tree $\mathcal{T}$ and (b) partial mixed graph $\g^{\prime}_{\pi}$ used in the proof sketch of Theorem \ref{thm:chordal-completeness} and Example \ref{ex:algo4-demo} in Section \ref{sec:meek-analogue-main}.}
    \label{fig:last-join-tree}
\end{figure}
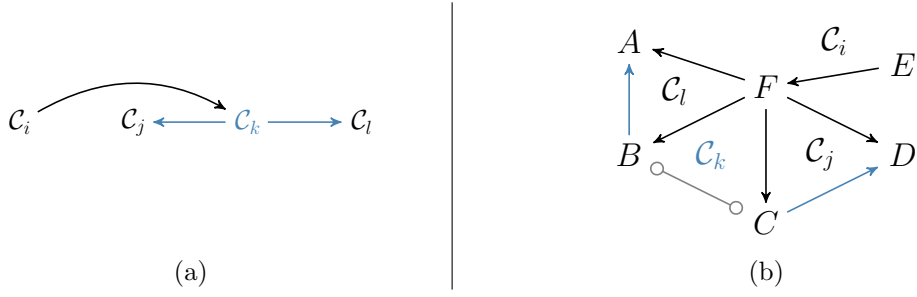

\begin{proofsketch}[Theorem \ref{thm:chordal-completeness}]
     Consider the essential ancestral graph $\g$ in Figure \ref{fig:lemma7-issue-p2}(a), as well as the partial mixed graph $\g^{\prime}$ in Figure  \ref{fig:lemma7-issue-p2}(b) which can be obtained as $\g^{\prime} = \texttt{{addOrientationKnowledge}}(\g, \{E \to F\})$. Note that $\g$ and $\g^{\prime}$ satisfy assumptions of Theorem \ref{thm:chordal-completeness}. 
    
    Suppose that we want to show that claim \ref{claim1:chor-comp} of Theorem \ref{thm:chordal-completeness} holds for the edge $B \circcirc C$ in $\g^{\prime}.$ (The proof sketch for an $\circarrow$ edge would be analogous.) Hence, we want to find three MAGs represented by $\g^{\prime}$ that contain $B \to C$, $B \leftarrow C$ and $B \leftrightarrow C$ respectively. 
    To do this, we will exploit the fact that essential ancestral graph $\g$ is equal to its circle component $\g_{C}$ and, as such, is a chordal graph (see Section \ref{sec:meek-analogue-main} for additional definitions).  Every chordal graph $\g$ can be represented by a meta graph $\g[T]$, where the nodes of $\g[T]$ are maximal cliques of $\g$ and $\g[T]$ is a tree graph which satisfies a running intersection property with respect to $\g$ (Definition \ref{def:join-tree}). Such a meta graph $\g[T]$ for $\g$ is called a \textit{junction tree} or \textit{join tree} for $\g$. 
    
    The maximal cliques of $\g$ in Figure \ref{fig:lemma7-issue-p2}(a) are $\mathcal{C}_{i} = \{E,F\}, \mathcal{C}_{j} = \{C,D,F\}, \mathcal{C}_{k} = \{B,C,F\}$, and $ \mathcal{C}_{l} = \{A,B,F\}$. Three partially directed join-trees representing $\g$ and $\g^{\prime}$ are given in Figure \ref{fig:lemma7-issue-p2}(c). An edge in Figure \ref{fig:lemma7-issue-p2}(c) is directed between two cliques $\mathcal{C}_{1}$ and $\mathcal{C}_2$ as $\mathcal{C}_1 \to \mathcal{C}_2$ if all edges between $\mathcal{C}_1 \cap \mathcal{C}_2$ and $\mathcal{C}_2 \setminus \mathcal{C}_1$ in $\g^{\prime}$ are out of $\mathcal{C}_1 \cap \mathcal{C}_2$ and into $\mathcal{C}_2 \setminus \mathcal{C}_1$ and there is at least one edge between $\mathcal{C}_1 \setminus \mathcal{C}_2$ and $\mathcal{C}_1 \cap \mathcal{C}_2$ that is into the latter in $\g^{\prime}$ (see Section \ref{sec:gen-join-tree-l4}).

    We will construct the desired MAGs by orienting a specific join tree in Figure \ref{fig:lemma7-issue-p2}(c) into a directed tree graph and incorporating these orientations into $\g^{\prime}$.
   Consider that edge $\langle B,C \rangle$ belongs to clique $\mathcal{C}_k$. We hence, choose the partially directed join tree $\mathcal{T}_{0}$ in the middle row of Figure \ref{fig:lemma7-issue-p2}(c) as this is the only partially directed join tree of $\g^{\prime}$  that is \textit{anchored} around $\mathcal{C}_k$ (Definition \ref{def:anchored}). By anchored, we mean that  $\PossAn(\mathcal{C}_k, \mathcal{T}_{0}) \equiv \An(\mathcal{C}_k, \mathcal{T}_{0})$ (Section \ref{sec:finding-right-jointree} shows how to construct a partially directed join tree for $\g$ and $\g^{\prime}$ anchored around a specific clique.)

    We orient $\mathcal{T}_0$ into a directed join tree $\mathcal{T}$ in Figure \ref{fig:last-join-tree}(a), where $\mathcal{T}$ has no unshielded colliders and no new edges into $\mathcal{C}_k$. The orientations in $\mathcal{T}$ can be applied to $\g^{\prime}$ to create partial mixed graph $\g^{\prime}_{\pi}$ in Figure \ref{fig:last-join-tree}(b). We show how to obtain such a directed tree and partial mixed graph in Section \ref{sec:orienting-join-tree-and-cliques}. 
    Now, applying any of the desired orientations to $\langle B,C \rangle$ in $\g^{\prime}_{\pi}$ will lead to one of the MAGs $\g[M]_1$, $\g[M]_2$, or $\g[M]_3$ described by \ref{claim1:chor-comp} of Theorem \ref{thm:chordal-completeness}, which gives us our desired result.
    After applying orientations from a directed join tree to $\g^{\prime}$, there may still be remaining variant edge marks on edges that do not lie between two cliques. In those cases, we show how to obtain MAGs $\g[M]_1$, $\g[M]_2$, or $\g[M]_3$ through generalizations of  the \cite{dor1992simple} algorithm in Lemmas \ref{lemma:existence-spo2} and \ref{lemma:existence-spo} in Section \ref{sec:clique}. 
\end{proofsketch}

Next, we prove the completeness of edge orientations in partial mixed graphs that allow for minimal collider paths but restrict expert knowledge  on $\circarrow$ edges within an essential ancestral graph. Namely, we show completeness if the expert knowledge or subsequent orientation rules application never orients such an edge as bidirected (Theorem \ref{thm:3} and Corollary \ref{cor:2}). We also show completeness in the case where expert knowledge (and subsequent orientation rules application) fully specifies all variant edge marks on $\circarrow$ edges within an essential ancestral graph (Theorem \ref{thm:chordal-subgraph-completeness} and Corollary \ref{cor:4}).
In the main text, we only give proof sketches for Theorems \ref{thm:3} and \ref{thm:chordal-subgraph-completeness}. Their proofs are in Section \ref{appendix:completeness1}.  Corollaries \ref{cor:2} and \ref{cor:4} follow directly from Theorems \ref{thm:3} and \ref{thm:chordal-subgraph-completeness} and the definition of consistent orientation knowledge (Definition \ref{def:consistentbg}).

\begin{theorem}
\label{thm:3}
Let $\g = (\mb{V,E})$ be an essential ancestral graph and $\g^{\prime} = (\mb{V,E'})$  be an ancestral partial mixed graph, such that $\g$ and $\g^{\prime}$ have the same skeleton,  the same set of minimal collider paths, and all invariant edge marks in $\g$ are identical in $\g^{\prime}$. Suppose also, that orientations in $\g^{\prime}$ are closed under { \ref{R1}, \ref{R2}, \ref{R4}, \ref{R8}, \ref{R11}} %\ref{R1}-\ref{R4}, \ref{R8}-\ref{R13}
and that every  $A \circarrow B$ in $\g$ corresponds to $A \to B$ or $A \leftrightarrow B$ in  $\g^{\prime}$. Then $\g^{\prime}$ is a restricted essential ancestral graph.
\end{theorem}

\begin{figure}
    \centering
    \begin{subfigure}{.31\textwidth}
          \tikzstyle{every edge}=[draw,>=stealth',semithick]
        \vspace{1cm}
        \centering
        \begin{tikzpicture}[->,>=stealth',shorten >=1pt,auto,node distance=1.2cm,scale=0.5,transform shape,font = {\huge\bfseries\sffamily}]
	\tikzstyle{state}=[inner sep=5pt, minimum size=5pt]

        \node[state] (A) at (-3,2.5) {$A$};
	\node[state] (B) at (-3,0) {$B$};
	\node[state] (C) at (0,-1.5) {$C$};
	\node[state] (F) at (0,1.5) {$F$};
	\node[state] (D) at (3,0) {$D$};
          \node[state] (E) at (3,2) {$E$};

    \node[state] (G) at (2,4) {$G$};  
     \node[state] (H) at (-.5,4) {$H$};

	\draw[gray,o-o]  (C) edge (D);
	\draw[gray,o-o] (F) edge (D);
	\draw[gray,o-o] (F) edge (B);
	\draw[gray,o-o] (B) edge (C);
	\draw[gray,o-o] (C) edge (F);

        \draw[gray,o-o] (F) edge (A);
        \draw[gray,o-o] (B) edge (A);
 	\draw[gray,o-o] (F) edge (E);

 	\draw[gray,o->] (H) edge (G);
 	\draw[gray,<-o] (G) edge (E);
   \end{tikzpicture}
\caption{}
    \end{subfigure}
    \vrule
    \begin{subfigure}{.31\textwidth}
          \tikzstyle{every edge}=[draw,>=stealth',semithick]
        \vspace{1cm}
        \centering
        \begin{tikzpicture}[->,>=stealth',shorten >=1pt,auto,node distance=1.2cm,scale=0.5,transform shape,font = {\huge\bfseries\sffamily}]
	\tikzstyle{state}=[inner sep=5pt, minimum size=5pt]

        \node[state] (A) at (-3,2.5) {$A$};
	\node[state] (B) at (-3,0) {$B$};
	\node[state] (C) at (0,-1.5) {$C$};
	\node[state] (F) at (0,1.5) {$F$};
	\node[state] (D) at (3,0) {$D$};
          \node[state] (E) at (3,2) {$E$};

    \node[state] (G) at (2,4) {$G$};  
     \node[state] (H) at (-.5,4) {$H$};

	\draw[gray,o-o]  (C) edge (D);
	\draw[->] (F) edge (D);
	\draw[->] (F) edge (B);
	\draw[gray,o-o] (B) edge (C);
	\draw[<-] (C) edge (F);

        \draw[->] (F) edge (A);
        \draw[gray,o-o] (B) edge (A);
 	\draw[<-] (F) edge (E);

 	\draw[->] (H) edge (G);
 	\draw[<->] (G) edge (E);
  
  \end{tikzpicture}
\caption{}
    \end{subfigure}
    \vrule
    \begin{subfigure}{.31\textwidth}
          \tikzstyle{every edge}=[draw,>=stealth',semithick]
        \vspace{1cm}
        \centering
        \begin{tikzpicture}[->,>=stealth',shorten >=1pt,auto,node distance=1.2cm,scale=0.5,transform shape,font = {\huge\bfseries\sffamily}]
	\tikzstyle{state}=[inner sep=5pt, minimum size=5pt]

        \node[state] (A) at (-3,2.5) {$A$};
	\node[state] (B) at (-3,0) {$B$};
	\node[state] (C) at (0,-1.5) {$C$};
	\node[state] (F) at (0,1.5) {$F$};
	\node[state] (D) at (3,0) {$D$};
          \node[state] (E) at (3,2) {$E$};

    \node[state] (G) at (2,4) {$G$};  
     \node[state] (H) at (-.5,4) {$H$};

	\draw[gray,o-o]  (C) edge (D);
	\draw[->] (F) edge (D);
	\draw[->] (F) edge (B);
	\draw[gray,o-o] (B) edge (C);
	\draw[<-] (C) edge (F);

        \draw[->] (F) edge (A);
        \draw[gray,o-o] (B) edge (A);
 	\draw[<-] (F) edge (E);

 	\draw[->] (H) edge (G);
 	\draw[gray,<-o] (G) edge (E);
  \end{tikzpicture}
\caption{}
    \end{subfigure}
    \caption{(a) Essential ancestral graph $\g$, (b) graph $\g_{1}$ used in the proof sketch of Theorem \ref{thm:3}, and (c)  graph $\g_{2}$ used in the proof sketch of Theorem \ref{thm:chordal-subgraph-completeness}.}
    \label{fig:thm2-3}
\end{figure}
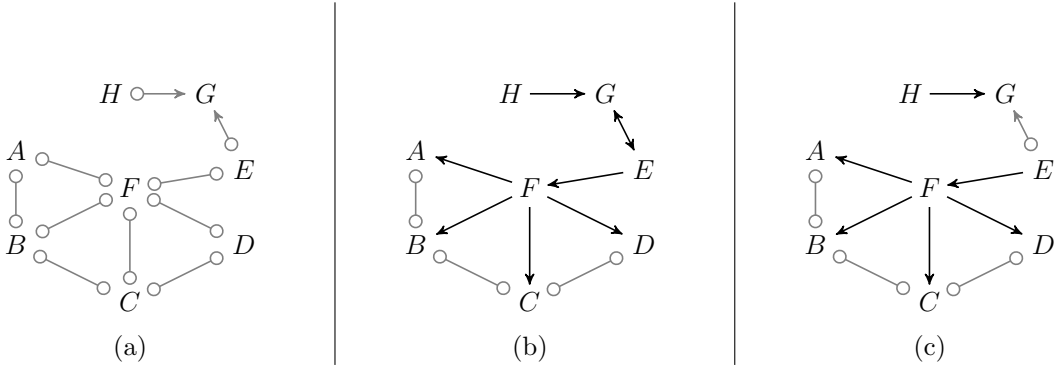

\begin{proofsketch}[Theorem \ref{thm:3}]
 Consider the essential ancestral graph $\g$ in Figure \ref{fig:thm2-3}(a), and the partial mixed graph $\g^{\prime}$ in Figure \ref{fig:thm2-3}(b) where $\g^{\prime} = \texttt{{addOrientationKnowledge}}(\g, \{H \to G, G \bulletarrow E \})$. Note that $\g$ and $\g^{\prime}$ satisfy assumptions of Theorem \ref{thm:3}. 
To show that $\g^{\prime}$ is a restricted essential ancestral graph, it is enough to show that for any of the edges $\langle A, B \rangle, \langle B, C \rangle, \langle C, D \rangle$ one can obtain a MAG represented by $\g^{\prime}$, where the edge of interest is oriented as $\to, \leftarrow$, or $\leftrightarrow.$ Let us consider edge $\langle B, C\rangle$ and show how to obtain MAGs represented by $\g^{\prime}$ that contain $B \to C$, $B \leftarrow C$ and $B \leftrightarrow C$. 

We rely on Theorem \ref{thm:chordal-completeness} to do this. Notice that the circle component of $\g$, $\g_{C}$, looks the same as the graph in Figure \ref{fig:lemma7-issue-p2}(a). Furthermore, the induced subgraph of $\g^{\prime}$ that corresponds to $\g_{C}$, called $\g^{\prime}_{C}$ exactly matches the graph in Figure \ref{fig:lemma7-issue-p2}(b). Hence, using the same reasoning as in the proof sketch of Theorem \ref{thm:chordal-completeness}, we can obtain a MAG $\g[M]$ represented by $\g^{\prime}_{C}$ that contains a desired orientation of $\langle B, C \rangle$. 
Then, it is enough to show that constructing a mixed graph by adding invariant orientations from $\g[M]$  to  $\g^{\prime}$ leads to a MAG $\g[M]^{\prime}$ represented by $\g^{\prime}$. For instance, per the proof sketch of Theorem \ref{thm:chordal-completeness}, consider a MAG $\g[M]$ that contains $B \to A$, $C \to D$, and $B \leftarrow C$. Adding these orientations to $\g^{\prime}$ to create $\g[M]^{\prime}$ clearly results in a MAG that has no new minimal collider paths compared to $\g^{\prime}$ (see also results in  Section \ref{section:supporting-thm3}). 
\end{proofsketch}

\begin{corollary}
\label{cor:2}
Let $\g= (\mb{V,E})$  be an essential ancestral graph and $\mathcal{K}$ be a set of orientation knowledge edge marks consistent with $\g$. Let $\g^{\prime} = \texttt{{addOrientationKnowledge}}(\g, \mathcal{K})$. 
If every edge of the form $A \circarrow B$ in $\g$ corresponds to $A \to B$ or $A \leftrightarrow B$ in $\g^{\prime}$, then $\g^{\prime}$ is {the} $\mathcal{K}$-restricted essential ancestral graph.
\end{corollary}

\begin{theorem}
\label{thm:chordal-subgraph-completeness}
Let $\g = (\mb{V,E})$ be an essential ancestral graph and $\g^{\prime} = (\mb{V,E'})$  be an ancestral partial mixed graph, such that $\g$ and $\g^{\prime}$ have the same skeleton,  the same set of minimal collider paths, and all invariant edge marks in $\g$ are identical in $\g^{\prime}$. Suppose furthermore,  that orientations in $\g^{\prime}$ are closed under { \ref{R1}, \ref{R2}, \ref{R4}, \ref{R8}, \ref{R11}.} % \ref{R1}-\ref{R4}, \ref{R8}-\ref{R13}. 
If there are no edges of the form $A \leftrightarrow B$ in $\g^{\prime}$ that correspond to $A \circarrow B$ in $\g$, then the following hold:
\begin{enumerate}[label=(\roman*)]
    \item\label{case3:chordal-sub-comp} For any edge $A \circarrow B$ in $\g^{\prime}$ such that $A \circarrow B$ is in $\g$, there is a MAG $\g[M]_1$ represented by $\g^{\prime}$ such that $A \to B$ is in $\g[M]_1$.
    \item\label{case1:chordal-sub-comp} For any edge $A \circcirc B$ in $\g^{\prime}$, there are three MAGs $\g[M]_1$, $\g[M]_2$ and $\g[M]_3$ represented by $\g^{\prime}$ such that $A \to B$ is in $\g[M]_1$,  $A \leftarrow B$ is in $\g[M]_2$, and $A \leftrightarrow B$ is in $\g[M]_3$.
    \item\label{case2:chordal-sub-comp} For any edge $A \circarrow B$ in $\g^{\prime}$ that corresponds to $A \circcirc B$ in $\g$, there are two MAGs $\g[M]_1$ and $\g[M]_2$ represented by $\g^{\prime}$ such that $A \to B$ is in $\g[M]_1$, and  $A \leftrightarrow B$  is in $\g[M]_2$.
\end{enumerate}
\end{theorem}

\begin{proofsketch}[Theorem \ref{thm:chordal-subgraph-completeness}]
   Consider the essential ancestral graph $\g$ in Figure \ref{fig:thm2-3}(a), as well as the partial mixed graph $\g^{\prime}$ in Figure \ref{fig:thm2-3}(c) which can be obtained as $\g^{\prime} = \texttt{{addOrientationKnowledge}}(\g, \{H \to G \})$. Then $\g$ and $\g^{\prime}$ satisfy assumptions of Theorem \ref{thm:chordal-subgraph-completeness}. 
To show that Theorem \ref{thm:chordal-subgraph-completeness} holds, it is enough to show that there is a MAG represented by $\g^{\prime}$ that contains $E \to G$ and such that a desired edge from the following set $\langle A, B \rangle, \langle B, C \rangle, \langle C, D \rangle$ is oriented as $\to, \leftarrow$, or $\leftrightarrow.$ 

We will rely on Theorem \ref{thm:3}. Namely, it is enough to show that we can orient all the remaining $\circarrow$ edges in $\g^{\prime}$ that also correspond to $\circarrow$ in $\g$ as $\to$ without incurring additional orientations or creating an almost directed cycle or a new minimal collider path. This is similar to a proof strategy used by \cite{zhang2008completeness} for essential ancestral graphs, except that we have additional orientation rules and already incorporated orientation knowledge to consider. We show that indeed holds in Theorem \ref{thm:adding-tails-rules-complete}. For $\g^{\prime}$ in Figure \ref{fig:thm2-3}(c), it is almost immediately apparent that orienting $E \to G$ does not incur any ancestral issues, new minimal collider paths, or edge orientations. Additionally, once $E \to G$ is oriented, our new partial mixed graph satisfies Theorem \ref{thm:3} and hence, the rest of the claim follows. 
\end{proofsketch}

\begin{corollary}
\label{cor:4}
Let $\g= (\mb{V,E})$  be an essential ancestral graph and $\mathcal{K}$ be a set of orientation knowledge consistent with $\g$.  Let $\g^{\prime} = \texttt{{addOrientationKnowledge}}(\g, \mathcal{K})$. 
 If there are no $A \leftrightarrow B$ edges in $\g^{\prime}$ that correspond to  $A \circarrow B$ in $\g$, then Theorem \ref{thm:chordal-subgraph-completeness} holds for $\g^{\prime}$.
\end{corollary}

\subsection{General Completeness of Orientation Rules}
\label{section:full-complete}

\begin{algorithm}
\caption{verifyCompleteness}
\label{alg:verify-completeness2}
\begin{algorithmic}[1]
    \Require Essential ancestral graph $\g$, orientation knowledge $\mathcal{K}$, and partial mixed graph $\g^{\prime}$, such that $\g^{\prime} = \texttt{{addOrientationKnowledge}}(\g, \mc{K})$
    \Ensure TRUE or FALSE.
    \State Let $\mc{A}_{\g^{\prime}}$ be the set of all $\circarrow$ edges in $\g$ that are still $\circarrow$ in $\g^{\prime}$
    \State Let $\g_{C}$ be the circle component of $\g$
    \State Let $\g^{\prime}_C$ be the induced subgraph of $\g^{\prime}$ that corresponds to $\g_{C}$
    \State Let $Invariant^{\prime}_{C}$ be the set of all invariant edge marks in $\g^{\prime}_C$
    \If {$\mc{A}_{\g^{\prime}} \neq \emptyset$} \label{line:inner-loop-start}
        \State Let $k$ be the length of  $\mc{A}_{\g^{\prime}}$ \label{line:startif}
        \State Let $\mc{O}_{1}$ be a list such that $\mc{O}_{1}[[i]]= (A \to B, A \leftrightarrow B)$, $\forall A \circarrow B \in \mc{A}_{\g^{\prime}}$, $i \in \{1, \dots, k\}$
        \State Initialize list $GCList = \emptyset$
        \State Initialize $count = 0$
        \For {$i \text{ in } 1 : k$}
            \For {$j \text{ in } 1 : 2$}
                \If {there exists $\g^{\prime \prime}$ that contains $\mc{O}_{1}[[i]][j]$ and all invariant orientations of $\g^{\prime}$ and if $\g^{\prime \prime}$ satisfies Theorem \ref{thm:3} relative to $\g$}
                        \State $count = count + 1$
                        \State Let $\g^{\prime \prime}_C$ be the induced subgraph of $\g^{\prime \prime}$ that corresponds to $\g_{C}$
                        \State Let $Invariant^{\prime \prime}_{C}$ be the set of all invariant edge marks in $\g^{\prime \prime}_C$
                        \State $GCList[[count]] = Invariant^{\prime \prime}_{C}$
                \Else \ \Return FALSE
                \EndIf
            \EndFor
        \EndFor
        \State Let $Invariant_{final} = \cap_{i =1}^{count} GCList[[i]]$
        \If {$Invariant_{final} \setminus Invariant^{\prime}_{C} \neq \emptyset$} \label{line:complex} 
            \State Let $r$ be the length of $Invariant_{final}  \setminus Invariant^{\prime}_{C}$
            \State Let $\mc{O}_{2}$ be the list of complementary orientations to $Invariant_{final} \setminus Invariant^{\prime}_{C}$
            \State Initialize $check = 1$
             \While {$check <= r$}
                \If {there exists $\g^{\prime \prime}$ that contains $\mc{O}_{2}[[check]]$ and all invariant orientations of $\g^{\prime}$ and if $\g^{\prime \prime}$ satisfies Theorem \ref{thm:3} relative to $\g$}
                \State $check = check + 1$
                \Else \ \Return FALSE
                \EndIf 
            \EndWhile 
        \EndIf \label{line:end-extra-check}
    \EndIf \label{line:inner-loop-end}
    \If {there is a directed or almost directed cycle of length 3 in $\g^{\prime}$ (Lemma \ref{lemma:nocycle3}), or there is a new unshielded collider or a new collider discriminated by a path in $\g^{\prime}$ compared to $\g$ (Lemma \ref{lemma:no-new-mcps})} \label{line:check-anc-and-max}
        \Return FALSE
    \EndIf 
    \State \Return  TRUE
\end{algorithmic}
\end{algorithm}

Unfortunately, our orientation rules are not necessarily complete in the general setting. %{  This is true even with the addition of orientation rules \ref{R14} and \ref{R18} of \cite{wangpolynomial}, see their discussion for more details.} %refer the reader to R14 and R18 of \cite{wangpolynomial}, for additional orientation rules which were discovered our work was under review. We have included these rules in Appendix... for completeness of this manuscript.. 
%The  additional orientation rules may be applicable when certain $\circarrow$ edge in the essential ancestral graph are oriented as $\leftrightarrow$ either by expert knowledge or by completion of another rule. 
In the absence of proof of general completeness,
we devise the \texttt{verifyCompleteness} algorithm for checking whether a partial mixed graph is a restricted essential ancestral graph.  The pseudocode of algorithm \texttt{verifyCompleteness}
is given in Algorithm \ref{alg:verify-completeness2}. 
Algorithm \ref{alg:verify-completeness2} relies on the results of Theorem \ref{thm:3} and Lemmas \ref{lemma:nocycle3} and \ref{lemma:no-new-mcps} to verify soundness and completeness of orientations in a partial mixed graph $\g^{\prime}$ obtained from an essential ancestral graph $\g$, orientation knowledge $\g[K]$, through $\g^{\prime} = \texttt{{addOrientationKnowledge}}(\g, \mc{K})$. %\footnote{We do not include R14 of \cite{wangpolynomial} in \texttt{{addOrientationKnowledge}} in order to ensure results within our manuscript are  self-contained.} 
The algorithm returns TRUE if $\g^{\prime}$ is {the} $\mc{K}$-restricted essential ancestral graph and FALSE otherwise. 

To explain the reasoning in more detail, let $\mc{A}_{\g^{\prime}}$, be the set of all  $A \circarrow B$ edges which are in both $\g$ and $\g^{\prime}$.
If $\mc{A}_{\g^{\prime}} = \emptyset$ or if we have reached Line \ref{line:inner-loop-end} of Algorithm \ref{alg:verify-completeness2}, we only need to check that $\g^{\prime}$ is ancestral and has the same minimal collider paths as $\g$, which is done in Line \ref{line:check-anc-and-max}. If this check is passed, we have that $\g^{\prime}$ satisfies Theorem \ref{thm:3} relative to $\g$, and so it is a restricted essential ancestral graph, and Algorithm \ref{alg:verify-completeness2} returns TRUE. Otherwise, Algorithm \ref{alg:verify-completeness2} returns FALSE. 

If $\mc{A}_{\g^{\prime}} \neq \emptyset$, we enter Line \ref{line:startif} of Algorithm \ref{alg:verify-completeness2}. Now it suffices to verify that for all  $A \circarrow B$ edges in  $\mc{A}_{\g^{\prime}} \neq \emptyset$, there are graphs $\g_{1}$ and $\g_{2}$ such that where $A \to B$ is in $\g_{1}$ and $A \leftrightarrow B$ is in $\g_{2}$ and such that  all invariant edgemarks in $\g^{\prime}$ are identical in $\g_{1}$ and $\g_{2}$,  $\g_{1}$ and $\g_{2}$ individually satisfy  Theorem \ref{thm:3} relative to $\g$. Note that for $\g_{1} $ and $\g_{2}$ to satisfy Theorem \ref{thm:3} relative to $\g$ all variant edge marks from $\mc{A}_{\g^{\prime}}$ must be invariant in these graphs.  This check is done between the lines \ref{line:inner-loop-start} and \ref{line:inner-loop-end}. Note that this means we do not necessarily need to construct graphs for every combination of edge orientations in $\mc{A}_{\g^{\prime}}$. Rather, if $|\mc{A}_{\g^{\prime}}| =k$, we need to construct $2k$ graphs between the lines \ref{line:inner-loop-start} and \ref{line:inner-loop-end}.

Importantly, in Line \ref{line:complex}, we also check that across the creation of these $2k$  graphs, we do not always encounter some other invariant edge mark in the former circle component of $\g$ that is labeled as a variant in $\g^{\prime}$. Having such an invariant edge mark across all $2k$ graphs does not immediately indicate an issue, as we did not check exhaustively over all combinations of edge orientations on $\circarrow$ edges from $\mc{A}_{\g^{\prime}}$. However,
 we need to perform an additional check, making sure that the complements of those orientations are viable, which we do between Lines \ref{line:complex} and \ref{line:end-extra-check}. By complementary orientation, we mean that if $C \arrowbullet D$ was always encountered in these $2k$ graphs, we check that $C \to D$ is viable, and if $C \to D$ was encountered across all graphs, we check that $C \arrowbullet D $ is a viable orientation. 

\subsection{Simulation Results} \label{sec:simulation}

\begin{figure}[!t]
    \centering
    \begin{subfigure}[t]{0.48\textwidth}
        \centering
        \includegraphics[height=2.5in]{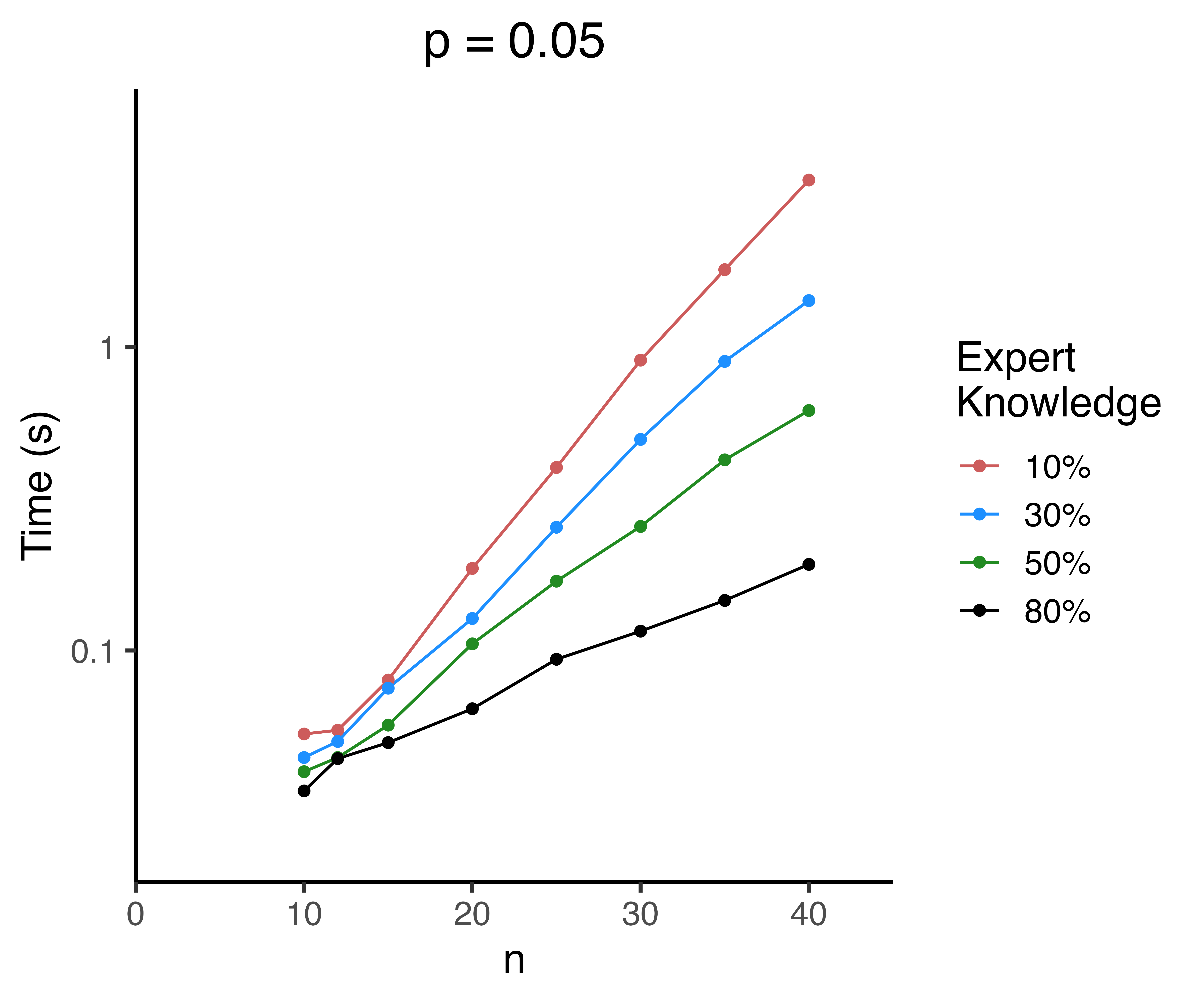}
    \caption{}
    \end{subfigure}
    ~
    \begin{subfigure}[t]{0.48\textwidth}
        \centering
        \includegraphics[height=2.5in]{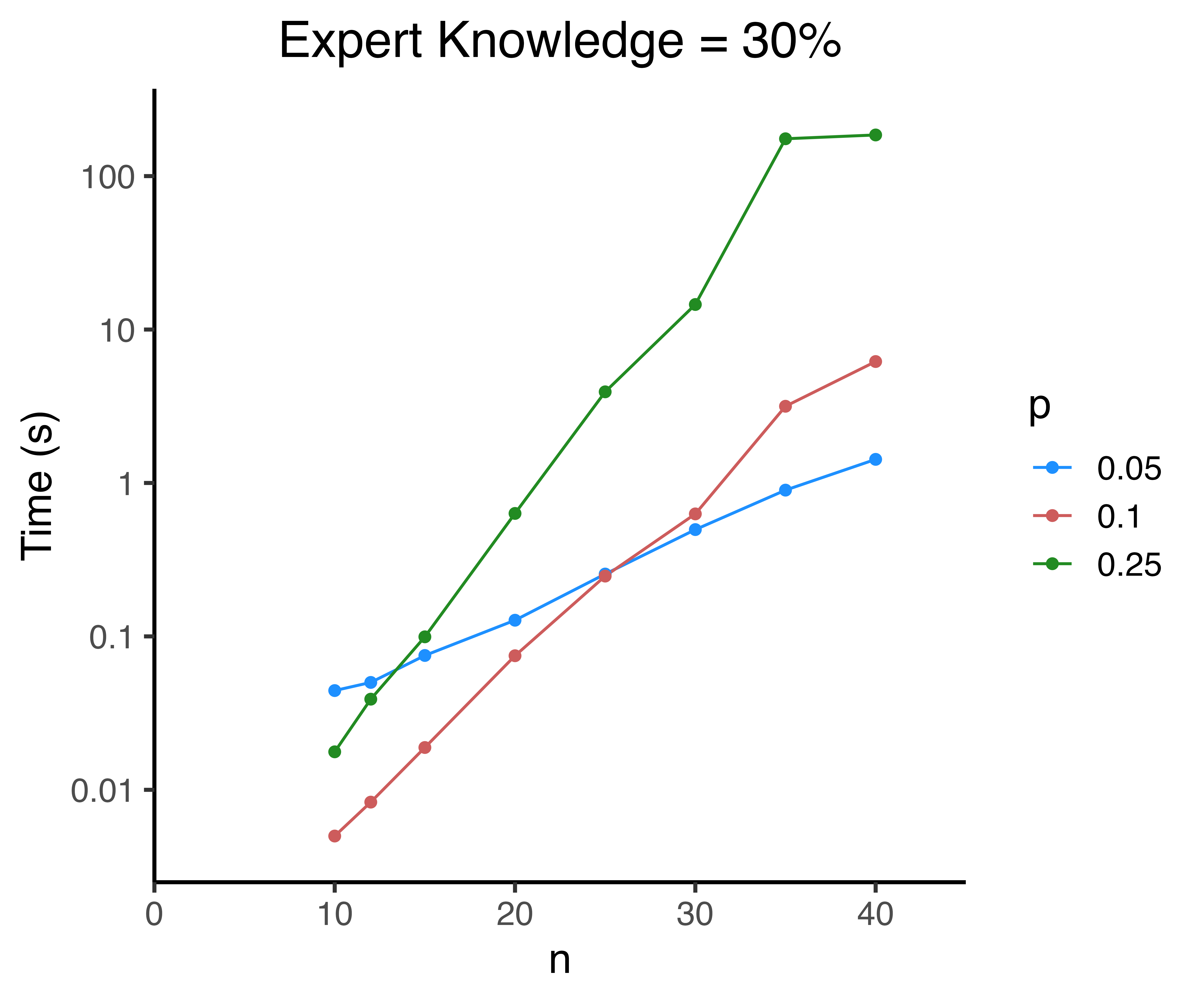}
        \caption{}
    \end{subfigure}
    \caption{(a) Average runtime of Algorithm \ref{alg:verify-completeness2} for various $n$ and percentage of revealed $\circ$ edge marks in $p = 0.05$ regime. 
    (b) Average runtime of Algorithm \ref{alg:verify-completeness2} for various $n$ and $p$, under a fixed percentage of revealed $\circ$ edge marks.}
    \label{fig:conjecture-runtime}
\end{figure}

We perform simulations to explore the runtime of Algorithm \ref{alg:verify-completeness2}. Our simulations used \texttt{R} v4.3.0 and \texttt{pcalg} v2.7-8 on a CPU with 4 cores and 30 GB RAM limit. Our implementation of Algorithms \ref{alg:mag2pag}-\ref{alg:verify-completeness2} is available through \texttt{R} package, \texttt{expertOrientR}, on GitHub (\url{https://github.com/AparaV/expertOrientR}). %{ Notably, we have also implemented R14 and R18 of \cite{wangpolynomial} as part of \texttt{{addOrientationKnowledge}} in our R package \texttt{expertOrientR} and have used these rules in the simulation below. However, we chose not to edit our manuscript pseudocode for {addOrientationKnowledge} in Section \ref{sec:bg-knowledge}, as this was not part of our original manuscript and results. Furthermore, the theoretical results included in the manuscript do not depend on these two orientation rules.}

We start by generating DAGs using the \texttt{randomDAG} function from the \texttt{pcalg} package with the Erd\H{o}s-R\'enyi $G(n, p)$ model, where $n$ is the number of nodes and $p$ is the probability of an edge existing between two nodes. We generate 1000 DAGs for each combination of $n \in \{10, 12, 15, 20, 25, 30, 35, 40\}$ and $p \in \{0.05, 0.1, 0.25\}$. For each generated DAG $\g[D]$, we randomly select $10\%$ of its source nodes to be designated as latent and construct the corresponding MAG $\g[M]$ on the observed nodes. The MAGs $\g[M]$ generated in this way contain, on average, 1-2 fewer nodes compared to the original DAGs, and the probability of an edge existing between two nodes in $\g[M]$ is on average $\{0.055, 0.11, 0.295\}$ for the corresponding DAG settings.  We also construct the essential ancestral graph $\g$ of $\g[M]$. For each $\g$ generated in this fashion, we choose  $k \%$ of $\circ$ edge marks, $k \in \{10, 30, 50, 80\}$, in $\g$ to reveal as orientation knowledge $\mathcal{K}$, using the true edge marks in $\g[M]$. If one of the edge marks we choose to reveal is a tail, we also reveal the arrowhead edge mark on the same edge in $\mathcal{K}$. We then obtain the partial mixed graph $\g^{\prime}$, as $\g^{\prime} = \texttt{{addOrientationKnowledge}}(\g, \mathcal{K})$. Now, for each combination of $\g$, $\mathcal{K}$, and $\g^{\prime}$ we run \texttt{verifyCompleteness}$(\g, \mathcal{K}, \g^{\prime})$ and record its runtime. In all of our simulations, \texttt{verifyCompleteness}$(\g, \mathcal{K}, \g^{\prime})$ has never returned FALSE.

We report the average runtime in seconds as a function of $n$, $p$, and orientation knowledge percentage in the two plots in Figure \ref{fig:conjecture-runtime}. In Figure \ref{fig:conjecture-runtime}(a), we can see the average runtime in seconds as a function of $n$ and orientation knowledge percentage for a fixed $p$, $p = 0.05$. As expected, the algorithm's runtime increases with graph size $n$, though we can notice that the runtime can be improved by revealing more orientation knowledge.
In Figure \ref{fig:conjecture-runtime}(b), we can see the average runtime in seconds as a function of $n$ and $p$ when about $30\%$ of circle edge marks are revealed by orientation knowledge. In this plot, it is clear that the starting DAG density has an enormous impact on algorithm runtime. This is because the size and density of a generated DAG affect the size and density of the associated MAG. In turn, the MAG influences the size of the Markov equivalence class of the essential ancestral graph. For dense and large graphs, this class can be substantial. As a result, verifying completeness becomes computationally challenging (see Figure 4 of \citealp{wang2024efficient} for a simulation investigating  sizes of the Markov equivalence classes of MAGs).

\begin{table}
\parbox{.43\linewidth}{
\centering
\begin{tabular}{c||c|c|c}
        \backslashbox[10mm]{$n$}{$p$} & \textbf{0.05} & \textbf{0.10} & \textbf{0.25} \\ 
        \hline
        \hline
        \textbf{10} & 0.42 (0) & 1.33 (0) & 4.76 (5) \\ 
        \textbf{12} & 0.74 (0) & 2.41 (2) & 6.74 (6) \\
        \textbf{15} & 1.47 (2) & 4.36 (4) & 9.13 (9) \\ 
        \textbf{20} & 3.49 (3) & 8.71 (8) & 12.50 (12) \\ 
        \textbf{25} & 6.43 (6) & 13.41 (13) & 13.82 (13) \\ 
        \textbf{30} & 10.26 (10) & 18.16 (18)  & 14.86 (14) \\ 
        \textbf{35} & 14.39 (14) & 22.83 (23)  & 15.69 (14) \\ 
        \textbf{40} & 19.11 (19) & 26.57 (26) & 15.94 (15) \\ 
    \end{tabular}
    \caption{Average (median) number of  $\circarrow$ edges in $\g$ for each $(n, p)$.}
    \label{tab:pd-size}
}
\quad\quad\quad
\parbox{.45\linewidth}{
\centering
 \begin{tabular}{c||c|c|c}
        \backslashbox[10mm]{$n$}{$p$} & \textbf{0.05} & \textbf{0.10} & \textbf{0.25} \\ 
        \hline
        \hline
        \textbf{10} & 3.67 (4) & 1.33 (6) & 11.29 (11) \\
        \textbf{12} & 5.37 (0) & 8.94 (8) & 14.28 (13) \\
        \textbf{15} & 8.03 (2) & 13.02 (13) & 16.68 (16) \\
        \textbf{20} & 13.65 (3) & 20.01 (20) &  20.31 (20) \\
        \textbf{25} & 19.69 (20) & 26.41 (26)  & 21.54 (21) \\
        \textbf{30} & 26.71 (26) & 32.29 (32)  & 23.00 (22) \\
        \textbf{35} & 33.43 (33) & 37.63 (37) & 23.72 (22) \\
        \textbf{40} & 40.32 (40) & 41.70 (42) & 24.03 (23) \\
    \end{tabular}
    \caption{Average (median) number of $\circ$ marks in $\g$ for each $(n, p)$.}
    \label{tab:circle-edgemarks-size}
}
\end{table}

The primary driver of the increase in runtime observed in Figure \ref{fig:conjecture-runtime} are $\circarrow$ edge in $\g$ which remain $\circarrow$ in $\g^{\prime}$. 
A secondary driver of longer runtime is the general number of $\circ$ edge marks, which must be considered when completing the orientation rules. 
We report the average (median) number of $\circarrow$ edges as well as the average (median) number of $\circ$ edge marks in $\g$ for each $(n,p)$ combination in Tables \ref{tab:pd-size} and \ref{tab:circle-edgemarks-size} respectively.

{
In Section \ref{subsection:real-data-sims}, we perform additional simulations comparing how often the true restricted essential graph agrees with the restricted essential graph learned from finite data.
}

\section{Discussion}
\label{sec:discussion}

We considered using expert knowledge of edge marks from the true MAG to restrict a Markov equivalence class. {We refer to this type of expert knowledge as \emph{orientation knowledge}.} Orientation knowledge is more general compared to tiered or local knowledge, when imposed on existing edges in the graph, in that it allows specifying bidirected edges, but also does not require all edge marks incident to a node to be specified \citep{andrews2020completeness, mooij2020joint,wang2022sound,wang2023estimating,wang2024new}. Our results bridge several characterizations of Markov equivalence (Section \ref{sec:eq-class}), and we provide several new graphical orientation rules for restricting such a class (Section \ref{sec:new-rules}). 
We construct an algorithm to add orientation knowledge into an essential ancestral graph (Algorithm \ref{alg:mpag}) and show that it is complete in specific settings (Section \ref{sec:completeness}) by generalizing results of \cite{meek1995causal} and \cite{zhang2008completeness}. Outside of these settings, we devise an algorithm (Algorithm \ref{alg:verify-completeness2}) to check whether the output of our Algorithm \ref{alg:mpag} is complete (Section \ref{section:full-complete}) and discuss its runtime (Section \ref{sec:simulation}).

Proving a general completeness result for a partial mixed graph $\g^{\prime}$ is challenging due to the existence of bidirected edges in $\g^{\prime}$ that correspond to $\circarrow$ edges in the essential ancestral graph.  One strategy employed by \cite{zhang2008completeness} to show the completeness of rules for constructing an essential ancestral graph involves considering the circle and non-circle components separately. We use a similar strategy for our results in Section \ref{sec:completeness}. However, this approach does not work in general. 
 For instance, consider the essential ancestral graph $\g$  in  Figure \ref{fig:disc}(a) and a partial mixed graph $\g^{\prime}$ constructed as $\g^{\prime} = \texttt{addBgKnowedge}(\g, \{B \bulletarrow D\})$ in Figure \ref{fig:disc}(b). Graph $\g^{\prime}$ is a restricted essential ancestral graph as  \texttt{verifyCompleteness}($\g,\langle \{B \bulletarrow D\}, \g^{\prime})$ returns TRUE.
It is impossible to orient all remaining $\circarrow$ edges in $\g^{\prime}$ as $\to$ without incurring a new unshielded collider $\langle A, D, C \rangle$. Furthermore, orienting either $A \to B$ or $C \to B$ in $\g^{\prime}$ leads to orientations on  $\langle A,D \rangle$ and $\langle D, C \rangle$ edges.

Another strategy for showing the completeness of orientation rules employed by \cite{meek1995causal} and Theorem \ref{thm:chordal-completeness} above relies on exploiting properties of chordal graphs. However, $\g^{\prime}$ will not generally be a chordal graph. See, for instance, $\g^{\prime}$ in Figure \ref{fig:r13}(a) and, in particular, the cycle $C \leftrightarrow D \arrowcirc F \circcirc E \circarrow C$ which is not chordal.

\begin{figure}
        \vspace{1cm}
   \centering
    \begin{subfigure}{.45\textwidth}
          \tikzstyle{every edge}=[draw,>=stealth',semithick]
        \vspace{1cm}
        \centering
        \begin{tikzpicture}[->,>=stealth',shorten >=1pt,auto,node distance=1.2cm,scale=0.7,transform shape,font = {\Large\bfseries\sffamily}]
	\tikzstyle{state}=[inner sep=5pt, minimum size=5pt]

	\node[state] (A) at (-2,0) {$A$};
	\node[state] (B) at (0,-1.5) {$B$};
	\node[state] (D) at (0,1.5) {$D$};
	\node[state] (C) at (2,0) {$C$};
      
	\draw[gray, <-o]  (B) edge (C);
	\draw[gray, o-o] (D) edge (C);
	\draw[gray, o-o] (D) edge (A);
	\draw[gray, o->] (A) edge (B);
	\draw[gray, o->] (D) edge (B);

  \end{tikzpicture}
\caption{}
\label{fig:disca} 
    \end{subfigure}
    \vrule
    \begin{subfigure}{.45\textwidth}
        \centering
       \tikzstyle{every edge}=[draw,>=stealth',semithick]
	 \begin{tikzpicture}[->,>=stealth',shorten >=1pt,auto,node distance=1.2cm,scale=0.7,transform shape,font = {\Large\bfseries\sffamily}]
	\tikzstyle{state}=[inner sep=5pt, minimum size=5pt]

	\node[state] (A) at (-2,0) {$A$};
	\node[state] (B) at (0,-1.5) {$B$};
	\node[state] (D) at (0,1.5) {$D$};
	\node[state] (C) at (2,0) {$C$};
  
	\draw[gray, <-o]  (B) edge (C);
	\draw[gray, o-o] (D) edge (C);
	\draw[gray, o-o] (D) edge (A);
	\draw[gray, o->] (A) edge (B);
	\draw[<->] (D) edge (B);

  \end{tikzpicture}
\caption{}
\label{fig:discb}
    \end{subfigure}
    \caption{(a) An essential ancestral graph $\g$, (b) a restricted essential ancestral graph $\g^{\prime}$.}
    \label{fig:disc}
\end{figure}

The general completeness problem remains open.
In our simulations, we never encounter a case where our orientation rules are incomplete; that is, we never observe a case where \texttt{verifyCompleteness} outputs a FALSE given an essential ancestral graph and consistent orientation knowledge.
However, \cite{wangpolynomial} discovered R14, applicable when assumptions of Theorem \ref{thm:chordal-subgraph-completeness} are violated (when bidirected edges exist in $\g^{\prime}$ that correspond to $\circarrow$ edges in $\g$), while our work was under review. This suggests that cases of incompleteness occur in graphs that are difficult to elicit through simulations. This is bolstered by the fact that the simplest examples of these new rules (our \ref{R4} and \ref{R13}, and \cite{wangpolynomial}'s R14) require dense graphical models that are challenging to generate.

We note that our paper does not cover the topics of causal effect identification or estimation given a restricted essential ancestral graph. Instead, we leave these questions open for future investigations. 
 We also believe that some of our results should help improve causal discovery and potentially, for proving consistency results for existing causal discovery algorithms \citep{triantafillou2016score,rantanen2021maximal,claassen2022greedy,hu2024fast}. The remaining open questions also include considerations of expert knowledge in the presence of selection bias.

\section*{Acknowledgements}
This material is based upon work supported by the National Science Foundation under Grant No. 2210210. We gratefully acknowledge Chris Meek, Thomas Richardson, and Tian-Zuo Wang for helpful discussions. We thank the reviewers for their careful feedback and suggestions.

%===============================%
%===============================%
%
%    Supplementary
%
%===============================%
%===============================%

\newpage
\appendix

\centerline{\LARGE{Supplement to:}}
\centerline{\LARGE{Towards Complete Causal Explanation with Expert Knowledge}}

\tableofcontents

\section{Additional Preliminaries and Existing Results} \label{sec:supp-prelim}

We denote sets of nodes in bold (for example $\mathbf{V}$), graphs in calligraphic font (for example $\g$) and nodes in a graph in uppercase letters (for example $V$).

\textbf{Paths.} A \textit{path} $p$ from $A$ to $B$ in~$\g$ is a sequence of distinct nodes $\langle A, \dots,B \rangle$ on which every pair of successive nodes are adjacent in $\g$. 
If $p = \langle V_1, V_2, \dots , V_k, \rangle, k \ge 2$, then $V_1$ and $V_k$ are \textit{endpoints} of $p$, and any other node $V_i, 1 <i<k,$ is a \textit{non-endpoint} node on $p$.
The \textit{length} of a path $p$, labeled $|p|$ equals the number of edges on $p$.
 A \textit{subsequence} of path $p$ is a sequence of nodes obtained by deleting some nodes from $p$ without changing the order of the remaining nodes.
 For a path $p = \langle V_1,V_2,\dots,V_m \rangle$, the \textit{subpath} from $V_i$ to $V_k$ ($1\le i\le k\le m)$ is the path $p(V_i,V_k) = \langle V_i,V_{i+1},\dots,V_{k}\rangle$. 
If $p = \langle V_1, V_2, \dots , V_k, \rangle, k \ge 2$, then with $-p$ we denote the path $\langle V_k, \dots , V_2, V_1 \rangle$. 
For two disjoint subsets $\mathbf{A}$ and $\mathbf{B}$ of $\mathbf{V}$, a path from $\mathbf{A}$ to $\mathbf{B}$ is a path from some $A \in \mathbf{A}$ to some $B \in \mathbf{B}$.
If $\g$ and $\g^*$ are two graphs with identical adjacencies and $p$ is a path in $\g$, then the \textit{corresponding path} $\pstar$ is the path in $\g^*$ constituted by the same sequence of nodes as $p$. 

\noindent\textbf{Concatenation of paths.}  We denote the concatenation of paths by $\oplus$, so that for example $p = p(V_1,V_{k}) \oplus p(V_{k},V_{m})$. In this paper, we only concatenate paths if the result of the concatenation is again a path. 

\begin{definition}[Chordal Graph]\label{def:chordal-graph}
Graph $\g = (\bV, \bE)$ is chordal if for every path $p =\langle V_1, V_2, \dots, V_k \rangle$, $k>3$ in $\g$ such that  edge $\langle V_1, V_k \rangle$ is also in $\g$, there is an edge $\langle V_i , V_j \rangle$, $1\le i < j \le k$ in $\g$, such that $j - i >1$.
\end{definition}

\subsection{Existing Results}

\begin{theorem}[Theorem 2.1 of \citealp{zhao2005markov}]\label{thm:21zhao}
Let $\g[M]_{1}$ and $\g[M]_{2}$ be two MAGs on the same set of nodes $\mathbf{V}$. Then $\g[M]_{1}$ and $\g[M]_{2}$ are Markov equivalent if and only if 
$\g[M]_{1}$ and $\g[M]_{2}$ have the same skeleton and the same minimal collider paths. 
\end{theorem}

\begin{lemma}[c.f.\ Lemmas 4.1, A.1, B.7, and B.8 of \citealp{zhang2008completeness}]
\label{lemma:circle-chordal}
Let $\g$ be an essential ancestral graph. Then, the circle component of $\g$ i.e., a subgraph of $\g$ containing only edges of type $\circcirc$ is a union of disconnected chordal graphs $\g_{C_1}, \dots , \g_{C_k}$, $k \ge 1$. Moreover, $\g_{C_i}$ for every $i \in \{1, \dots, k \}$ is an induced subgraph of $\g$.
\end{lemma}

\begin{lemma}[Lemmas B.4, B.5 and Corollary B.6 of \citealp{zhang2008completeness}] \label{lemma:zhang-pdpath-edge-not-into}
    Let $\g$ be an essential ancestral graph. If path $p = \langle V_1, \dots, V_k \rangle, k> 1$, does not contain any edge of the form $V_i \arrowbullet V_{i+1}, 1 \le i \le k-1$ and if there is an edge $\langle V_1, V_k \rangle$ in $\g$, then $V_1 \to V_k$, or $V_1 \circbullet V_k$ is in $\g$.
Furthermore, if $V_{k-1} \bulletarrow V_k$ is in $\g$, then $V_{1} \to V_k$, or $V_1 \circarrow V_k$ is in $\g$.
\end{lemma}

\begin{lemma}[Lemmas B.7 of \citealp{zhang2008completeness}] \label{lemma:zhang-circ-path}
    Let $\g$ be an essential ancestral graph. If path $p = \langle V_1, \dots, V_k \rangle, k> 1$, is of the form $V_1 \circcirc V_2 \circcirc \dots \circcirc V_k$ and there is an edge $\langle V_1, V_k \rangle$ in $\g$, then $V_1 \circcirc V_k$ is in $\g.$
\end{lemma}

\begin{lemma}[Lemmas B.8 of \citealp{zhang2008completeness}] \label{lemma:zhang-b8-circ-path}
    Let $\g$ be an essential ancestral graph. If path $p = \langle V_1, \dots, V_k \rangle, k> 3$, is an unshielded path of the form $V_1 \circcirc V_2 \circcirc \dots \circcirc V_k$ in $\g$, then there is no edge $\langle V_i, V_j \rangle$ in $\g$, where $1 \le i < j \le k.$
\end{lemma}

\begin{lemma}[Lemma A.1 of \citealp{zhang2008completeness}]\label{lemma:p1zhang}
Let $\g$ be an essential ancestral graph, and let $A, B,$ and $C$ be three distinct nodes in $\g.$ If $A \bulletarrow B \circbullet C$ is in $\g$, then $A \bulletarrow C$ is also in $\g$. Furthermore, if $A \to B$ is in $\g,$ then $A \to C$, or $A \circarrow C$ is in $\g.$
 \end{lemma}

 \begin{lemma}[Lemma 7.5 of \citealp{maathuis2015generalized}]
\label{lemma:maathuis-7_5}
Let $A$ and $B$ be two distinct nodes in an essential  graph $\g$. If  edge $A \arrowbullet B$ is in $\g$ then any path $p = \langle A = V_1, V_2, \dots, V_k = B\rangle, k>1$ from $A$ to $B$, must contain at least one edge of the form  $V_{i} \arrowbullet V_{i+1}$, $i \in \{1, \dots, k-1\}$. 
Conversely, if a path $q = \langle V_1, \dots, V_r \rangle, r >1$ does not contain any edge of the form $V_j \arrowbullet V_{j+1}$, $j \in \{1, \dots, r-1\},$ then $q$ is a possibly directed path from $V_1$ to $V_r$. 
\end{lemma}

\section{Auxiliary Results}

We first generalize a few important and well known essential ancestral graph properties to our general partial mixed graph setting.  Corollary \ref{cor:unshielded-gen} generalizes Lemma B.1 of \citet{zhang2008completeness}, Corollary \ref{cor:b2} generalizes Lemma B.2 of \citet{zhang2008completeness}, and Lemma \ref{lemma:prop1new} generalizes Lemma 1 of \citet{meek1995causal} and Lemma A.1 of \citet{zhang2008completeness} (given in Lemma \ref{lemma:p1zhang} above).

\begin{corollary}\label{cor:unshielded-gen} 
Let $\g = (\mathbf{V,E})$ be a partial mixed graph. 
Let $p = \langle V_1, \dots, V_k\rangle, k > 1$ be a  possibly directed path in $\g$. Then there is a subsequence of $p$ called $p^{\prime}$, $p^{\prime} = \langle V_1 = V^{\prime}_1, V^{\prime}_2, \dots, V^{\prime}_{\ell} = V_k \rangle $, $\ell >1$, such that $p^{\prime}$ is an unshielded possibly directed path. 
\end{corollary}

\begin{proofof}[Corollary \ref{cor:unshielded-gen}]
Observe that any subsequence of $p$ that is a path would necessarily be a possibly directed path. Hence, if $p$ is not unshielded, we can obtain $p^{\prime}$ through an iterative process of skipping over shielded nodes.  
\end{proofof}

\begin{corollary}\label{cor:b2}  Let $\g = (\mathbf{V,E})$ be a partial mixed graph such that the orientations in $\g$ are closed under rule $\ref{R1}$.
Let $p = \langle A, \dots, B \rangle$ be an unshielded possibly directed path in $\g$. Then:
\begin{enumerate}[label=(\roman*)]
    \item If there is a $\circarrow$ or $\to$ edge on $p$, then all edges after that edge on $p$ are of type $\to$
    \item If there is a $\circcirc$ edge on $p$, this edge occurs before a $\circarrow$ or $\to$ edge on $p$.
    \item There is at most one $\circarrow$ edge on $p$
\end{enumerate}
\end{corollary}
\begin{proofof}[Corollary \ref{cor:b2}] Follows from the fact that orientations in $\g$ are closed under \ref{R1} and the fact that $p$ is an unshielded possibly directed path. 
\end{proofof}

\begin{lemma}\label{lemma:prop1new}
 Let $\g = (\mathbf{V,E})$ be a partial mixed graph such that the orientations in $\g$ are closed under rules $\ref{R1}$ and $\ref{R2}$. For any three nodes $A, B, C \in \mathcal{G}$ such that $A \bulletarrow B \circbullet C$. Then there is an edge between $A$ and $C$ in $\g$ that is not of the form $A \leftarrow C$. Moreover, if $A \to B \circbullet C$ is in $\g$, then the edge between $A$ and $C$ is also not of the form $A \leftrightarrow C$.
\end{lemma}

\begin{proofof}[Lemma \ref{lemma:prop1new}]
Since orientations in $\g$ are closed under \ref{R1}, there must be an edge between $A$ and $C$. 
The edge between $A$ and $C$ cannot be of the form $A \leftarrow C$, since that would imply that orientations in $\g$ are not closed under \ref{R2}. Similarly, if the edge between $A$ and $B$ in $\g$ is $A \to B$, then due to \ref{R2}, $A \leftrightarrow C$ is also not in $\g$.
\end{proofof}

\section{Supplement to Section \ref{sec:eq-class}}

\begin{proofof}[Theorem \ref{thm:ali-conjecture}]
Let $C = Q_k$. To begin, we note that $p$ forms a collider path in $\g[M]$. If $p$ is a minimal collider path, then $\langle Q_{k-1}, C, B \rangle$ will be a collider in every $\G[M]^*$ that is Markov equivalent to $\G[M]$ (Theorem \ref{thm:21zhao}). 

Otherwise, $p$ is not a minimal collider path in $\g[M]$. Then there is a subsequence $p^{\prime} = \langle A = Q_{n_0}, Q_{n_1}, \dots, Q_{n_m}, C, B \rangle$ of $p$ in $\g[M]$, such that  $p^{\prime}$ is a minimal collider path in $\g[M]$ and if $m>0$, $\{ Q_{n_j}\}_{j=1}^{m} \subset \{ Q_i \}_{i=1}^{k-1}$. Note that $B$ must be in $p^{\prime}$ as $Q_i \to B$ for all $i$ by definition of discriminating path. There are two possibilities for $Q_{n_m}$ on $p^{\prime}$
\begin{enumerate}[label=(\roman*)]
    \item $Q_{n_m} = Q_{k-1}$: Then, by Theorem \ref{thm:21zhao}, $\langle  Q_{k-1}, C, B \rangle$ forms a collider in every MAG $\G[M]^*$ that is Markov equivalent to $\G[M]$.
    \item $n_m < k-1$: Then, we have $Q_{n_m} \bulletarrow C \arrowbullet B$ is in $\g[M]$. By Theorem \ref{thm:21zhao}, we have $Q_{n_m} \bulletarrow C \arrowbullet B$ is in every $\G[M]^*$ that is Markov equivalent to $\G[M]$. So, we now only need to show that $Q_{k-1} \bulletarrow C$ is in every $\G[M]^*$ that is Markov equivalent to $\G[M]$.
    
    For the sake of contradiction, assume that there is at least one MAG Markov equivalent to $\g[M]$ that does not contain  $Q_{k-1} \bulletarrow C$. Therefore, in the essential ancestral graph $\g$ of $\G[M]$, we have $Q_{k-1} \bulletcirc C \arrowbullet B$. Then by  Lemma \ref{lemma:p1zhang}, the edge between $Q_{k-1}$ and $B$ of the form $B \bulletarrow Q_{k-1}$. This is a contradiction to the assumption that $p$ is a discriminating path in $\G[M]$ as that implies that $Q_{k-1}$ is a parent of $B$ in $\g[M]$. Therefore, we must have $Q_{k-1} \bulletarrow C \arrowbullet B$ in the essential ancestral graph $\g$ and therefore, $\langle Q_{k-1}, C, B \rangle$ is a collider in every $\G[M]^*$ that is Markov equivalent to $\G[M]$.
\end{enumerate}
\end{proofof}

\begin{proofof}[Theorem \ref{thm:mag2pag}] 
We will use the following notation: if $p_{\g[M]}$ is a path in $\g[M]$, then $p_{\g}$ denotes the corresponding path in $\g$. 
Based on the construction of $\g$ by Algorithm \ref{alg:mag2pag}, we know that all minimal collider paths in $\g[M]$ are also in $\g$. Furthermore, any edge orientation done by Algorithm \ref{alg:mag2pag} should match the same edge orientation in $\g[M]$ as long as the use of  \ref{R4new} does not induce a different orientation in $\g$. 
Hence, for $\g$ to be an essential ancestral graph of $\g[M]$, it is  sufficient to show that all colliders discriminated by a path $p_{\g[M]}$ in $\g[M]$ are also colliders on $p_{\g}$ in $\g$ (see also Theorem \ref{thm:ali-conjecture}).

Hence, consider a discriminating path $p_{\g[M]} = \langle A = Q_0, Q_1, \dots, Q_k , B\rangle$, $k \ge 2$ in $\g[M]$ such that $Q_{k}$ is a collider on $p_{\g[M]}.$
If $p_{\g[M]}$ is a minimal collider path, then $Q_k$ is a collider on $p_{\g}$ and we are done. Hence, for the rest of the proof suppose that $p_{\g[M]}$ is not a minimal collider path, and let $p_{\g[M]}^{\prime}$ be a subsequence of $p_{\g[M]}$ that forms a minimal collider path in $\g[M]$.

Since $A \notin \Adj(B, \g[M])$ and since $Q_{i} \to B$ is in $\g[M]$ for all $i \in \{1, \dots , k-1\}$ it follows $\langle Q_k, B \rangle$ is on  $p_{\g[M]}^{\prime}$, that is,  $p_{\g[M]}^{\prime}$ is of the form $p_{\g[M]}^{\prime} = \langle A = Q_{n_0}, Q_{n_1}, \dots, Q_{n_{\ell}}, Q_k, B \rangle$,  $\ell \ge 0$. Let $p_{\g}^{\prime}$ be the corresponding minimal collider path in $\g$.  Hence, $Q_{k} \arrowbullet B$ is in $\g$, and we only need to show that $Q_{k-1} \bulletarrow Q_k$ is also in $\g$. Of course, this immediately holds if $Q_{n_{\ell}} = Q_{k-1}$, so for the rest of the proof consider the case where $Q_{k-1}$ is not on $p_{\g[M]}^{\prime}$.

Note that since, $Q_k \arrowbullet B$ is in $\g$ for an arbitrarily chosen discriminating collider path $\langle A = Q_0, Q_1, \dots, Q_k , B\rangle$ in $\g[M]$, we can conclude that all orientations  made by completing the \ref{R4new} in Algorithm \ref{alg:mag2pag} match the orientations on the corresponding edge in $\g[M]$.  Therefore, since we know that $Q_{k-1} \leftrightarrow Q_k$ is in $\g[M]$, we know that $Q_{k-1} \leftarrow Q_k$, or $Q_{k-1} \to Q_k$ cannot be in $\g$.
Furthermore, this implies that  $\g$ is an ancestral partial mixed graph that does not contain any inducing paths.

Now, to show that $Q_{k-1} \circarrow Q_k$ or $Q_{k-1} \leftrightarrow Q_k$ is in $\g$, we show that the other remaining option for this edge in $\g$: $Q_{k-1} \bulletcirc Q_k$ leads to a contradiction. Hence, suppose that  $Q_{k-1} \bulletcirc Q_k$ is in $\g$. Then since  $p_{\g}^{\prime} = \langle A = Q_{n_0}, Q_{n_1}, \dots, Q_{n_{\ell}}, Q_k, B \rangle$ is a minimal collider path in $\g$, such that $A \notin \Adj(B, \g)$ and $Q_{k-1}$ is not on $p_{\g}^{\prime}$, but $Q_k \circbullet Q_{k-1}$ is in $\g$, Lemma \ref{lemma:induction-result-prime} would imply that $B \bulletarrow Q_{k-1}$ is in $\g$. However, this now leads us to a contradiction with the fact that invariant edge marks in $\g$ match those in $\g[M]$, since we know that $B \leftarrow Q_{k-1}$ is in $\g[M]$.
 \end{proofof}

\begin{proofof}[Lemma \ref{lemma:MCP-format}] 
Note that $p$ is of the form $P_1 \bulletarrow P_2 \leftrightarrow \dots \leftrightarrow P_{k-1} \arrowbullet P_k$ and that $P_1 \notin \Adj(P_k, \g)$ by definition. Hence, if $k = 3$, the claim holds by definition.

For the rest of the proof suppose that $k>3$ and let $i \in \{2, \dots, k-1\}$. If $P_{i-1} \notin \Adj(P_{i+1}, \g)$, then we are in case \ref{caseMCP1:1} and we are done.
Otherwise, $P_{i-1} \in \Adj(P_{i+1}, \g)$, so by \ref{zhao3} of Lemma \ref{lemma:zhao-mcp}, we have that either $P_{i-1} \to P_{i+1}$ or $P_{i-1} \leftarrow P_{i+1}$ is in $\g.$

Assume without loss of generality that $P_{i-1} \to P_{i+1}$ is in $\g$. We will show that in this case, we end up having a discriminating collider path for $P_i$ of the form in \ref{caseMCP1:2}. If $P_{i-1} \leftarrow P_{i+1}$ was in $\g$, an analogous argument can be used to show the existence of a discriminating collider path for $P_i$ of the form in \ref{caseMCP1:3}.

Since $P_{i-1} \to P_{i+1}$ is in $\g$, by \ref{zhao1} of Lemma \ref{lemma:zhao-mcp}, we have that $i -1 \neq 1$, that is $i>2$. If $i =3$, we must have that  $P_{i-2} \notin \Adj(P_{i+1}, \g)$,  by \ref{zhao1} and \ref{zhao4} of Lemma \ref{lemma:zhao-mcp}, and in this case we immediately have that $p(P_{i-2}, P_{i+1})$ is a discriminating collider path of the form \ref{caseMCP1:2}.

Otherwise, $i >3$, and either $P_{i-2} \notin \Adj(P_{i+1}, \g)$, in which case we again have that $p(P_{i-2}, P_{i+1})$ is a discriminating collider path of the form \ref{caseMCP1:2}, or $P_{i-2} \to P_{i+1}$ by  \ref{zhao4} of  Lemma \ref{lemma:zhao-mcp}. Now, we can apply the above argument iteratively, since if $i = 4$, we have that $P_{i-3} \notin \Adj(P_{i+1}, \g)$ by \ref{zhao1} and \ref{zhao4} of  Lemma \ref{lemma:zhao-mcp}, and otherwise, we have that $P_{i-3} \to P_{i+1}$ is in $\g$ and we consider the presence of edge $\langle P_{i-4}, P_{i+1}\rangle$.
\end{proofof}

\subsection{Supporting Results}

Figure \ref{figproof1} includes the proof structure for Lemma \ref{lemma:induction-result-prime}.

\begin{lemma}
\label{lemma:induction-result-prime}
Let $\g = (\mb{V,E})$ be an {ancestral partial mixed graph that does not contain inducing paths and such that orientations in $\g$ are closed under \ref{R1}-\ref{R3}, \ref{R4new}}. Furthermore,  let $A$ and $B$ be distinct nodes in $\g$ such that $A \notin \Adj(B, \g)$. Suppose that there is a minimal collider path $p = \langle A =Q_{l_k},\dots, Q_{l_{1}}, Q,  Q_{r_1}, \dots, Q_{r_m}= B \rangle$, $k, m \geq 1$,   in $\g$ and a node $W$ not on $p$ such that $W \bulletcirc Q$ is in $\g$. Then the following hold:
\begin{enumerate}[label = (\roman*)]
    \item Either $A \bulletarrow W$ is in $\g$, or $k>1$ and there is an $i \in \{1, \dots , k-1\}$ such that $Q_{l_i} \leftrightarrow W$ is in $\g$.
    \item Either $B \bulletarrow W$ is in $\g$, or $m >1$ and there is an $j \in \{1, \dots , m-1\}$ such that $Q_{r_j} \leftrightarrow W$ is in $\g$.
\end{enumerate}
\end{lemma}

\begin{proofof}[Lemma \ref{lemma:induction-result-prime}] 
First note, that by Lemma \ref{lemma:helper3}, we have that $Q_{l_1} \bulletarrow W \arrowbullet Q_{r_1}$. The claim then follows by iterative application of Lemma \ref{lemma:induction-result-prime-helper}.
\end{proofof}

\begin{figure}[!t]
    \centering
      \begin{tikzpicture}[->,>=latex,shorten >=1pt,auto,node distance=0.8cm,scale=.8,transform shape,font = {\large\sffamily}]
  \tikzstyle{state}=[inner sep=2pt, minimum size=12pt]

  \node[state] (L1) at (-2,-2) {Lemma \ref{lemma:MCP-format}};
  \node[state] (L2) at (-3,-4) {Lemma \ref{lemma:zhao-mcp} };
  \node[state] (L3) at (0.5,-4) {Lemma \ref{lemma:helper1} };
  \node[state] (L4) at (-2,-6) {Lemma \ref{lemma:helper11}};
  \node[state] (L5) at (2,-6) {Lemma \ref{lemma:helper4}};
  \node[state] (L6) at (4,-4) {Lemma \ref{lemma:helper3} }; 
  \node[state] (L7) at (8,-4) {Lemma \ref{lemma:induction-result-prime-helper}};
  \node[state] (L8) at (8,-6) {\textbf{Lemma \ref{lemma:induction-result-prime}}};

\draw[->,arrows= {-Latex[width=5pt, length=5pt]}] (L2) edge (L3);

\draw[->,arrows= {-Latex[width=5pt, length=5pt]}] (L2) edge (L1);
\draw[->,arrows= {-Latex[width=5pt, length=5pt]}] (L1) edge (L3);
\draw[->,arrows= {-Latex[width=5pt, length=5pt]}] (L2) edge (L4);
\draw[->,arrows= {-Latex[width=5pt, length=5pt]}] (L2) edge (L5);
\draw[->, out=30,in=160,arrows= {-Latex[width=5pt, length=5pt]}] (L2) edge (L6);

\draw[->,arrows= {-Latex[width=5pt, length=5pt]}] (L3) edge (L4);
\draw[->,arrows= {-Latex[width=5pt, length=5pt]}] (L3) edge (L5);
\draw[->,arrows= {-Latex[width=5pt, length=5pt]}] (L3) edge (L6);

\draw[->,arrows= {-Latex[width=5pt, length=5pt]}] (L4) edge (L5);
\draw[->,arrows= {-Latex[width=5pt, length=5pt]}] (L5) edge (L6);

\draw[->,arrows= {-Latex[width=5pt, length=5pt]}] (L6) edge (L7);
\draw[->,arrows= {-Latex[width=5pt, length=5pt]}] (L6) edge (L8);

\draw[->,arrows= {-Latex[width=5pt, length=5pt]}] (L7) edge (L8);

\end{tikzpicture}
   \caption{Proof structure of Lemma \ref{lemma:induction-result-prime}}
    \label{figproof1}
\end{figure}

\begin{definition}[Distance to $\mathbf{Z}$; cf. \citealp{zhang2006causal}, \citealp{perkovic2018complete}]  \label{def:dist-to-z}
Let $\g = (\mb{V,E})$ be a partial mixed graph, $p$ a path in  $\g$ and $\mathbf{Z} \subset \mb{V}$. Suppose that every node on $p = \langle V_1, \dots V_k \rangle$ is in $\PossAn(\mathbf{Z}, \g)$. Then the distance to $\mathbf{Z}$ for each node $V_i$, $i \in \{1 , \dots , k\}$ on $p$ is the length of a shortest possibly causal path from  $V_i$ to $\mathbf{Z}$. The distance to $\mathbf{Z}$ for the entire path $p$ is equal to the sum of the distances to $\mathbf{Z}$ for each node on $p$.
\end{definition}

The following Lemma is similar to Lemma 2.1 of \citet{zhao2005markov}.

\begin{lemma}\label{lemma:zhao-mcp} 
Let $\g = (\mb{V,E})$ be an ancestral partial mixed graph and let $p$, be a  minimal collider path in $\g$, $p =  \langle A = Q_0, Q_1, \dots, Q_k, Q_{k+1} =B \rangle, k \ge 2$. Furthermore, suppose that the edge orientations in $\g$ are closed under \ref{R1}, \ref{R2}, \ref{R4new}. 
Then the following hold 
\begin{enumerate}[label= (\roman*)]
    \item\label{zhao1} If edge $\langle Q_i, A\rangle$ is in $\g$ for some $i \in \{2, \dots , k\}$, then this edge is of the form $Q_i \to A$.
    \item\label{zhao2}  If edge $\langle Q_i, B\rangle$ is in $\g$ for some $i \in \{1, \dots, k-1\}$, then it is of the form $Q_i \to B$.
    \item\label{zhao3}  If edge $\langle Q_i, Q_j \rangle$ is in $\g$ for some $i,j \in \{1, \dots, k-1\}, i < j-1$, then this edge is either $Q_i \to Q_j$ or $Q_j \to Q_i$. 
    \item\label{zhao4} If  $\langle Q_i, Q_{j} \rangle$ and $\langle Q_i, Q_{j+1} \rangle$ are edges in $\g$ for some $i, j \in \{1, \dots, k-1\}, i \neq j$, then these edges are either $Q_{j} \to Q_i \leftarrow Q_{j+1}$, or $Q_{j} \leftarrow Q_i \to Q_{j+1}$ in $\g$.
\end{enumerate}
\end{lemma}

\begin{proofof}[Lemma \ref{lemma:zhao-mcp}]
Note that since $p$ is a minimal collider path in $\g$, we have that $A \notin \Adj(B, \g).$ 

\begin{enumerate}
    \item[\ref{zhao1}, \ref{zhao2}]   We only prove the claim \ref{zhao1}, since the proof for claim \ref{zhao2} is symmetric. Note that the edge $\langle Q_i, A \rangle$ cannot be of the form $Q_i \arrowbullet A$, since in this case, $p$ is not a minimal collider path in $\g$. Hence, we only need to show that this edge is also not of the form $Q_i \circbullet A$.

Since $Q_{k+1} = B$ is not adjacent to $A$ in $\g$, there is at least one node on $p(Q_{i+1}, Q_{k+1})$ that is not adjacent to $A$. Let $Q_r, i < r \le k+1$ be the closest node to $Q_i$ on $p(Q_{i}, Q_{k+1})$ such that $Q_r \notin \Adj(A, \g)$. 
Then $Q_j \in \Adj(A, \g)$ for all $j \in \{i, \dots, r-1 \}$. Additionally, $Q_j \arrowbullet A$ is not in $\g$  for any $j \in \{i, \dots, r-1 \}$ as that would contradict that $p$ is a minimal collider path. If $Q_{r-1} \circbullet A$ was in $\g$, $Q_{r} \bulletarrow Q_{r-1} \circbullet A $ and $Q_r \notin \Adj(A, \g)$ would contradict Lemma \ref{lemma:prop1new}. Hence,  $A \leftarrow Q_{r-1}$ is in $\g$.

If $i = r-1 $ we are done. 
Otherwise, consider the path  $Q_{r} \bulletarrow Q_{r-1} \leftrightarrow Q_{r-2}$ and edge $Q_{r-1} \to A$ in $\g$. Since orientations in $\g$ are closed under \ref{R4new} and since $Q_{r-2} \in \Adj(A, \g)$, $A \leftarrow Q_{r-2}$ is in $\g$.  We can apply this same reasoning iteratively for all (if any) remaining $j \in \{i, \dots, r-2 \}$ to show that  $Q_j \to A$ is in $\g$.

\item[\ref{zhao3}]  Since $p$ is a minimal collider path in $\g$, it is clear that $Q_i \leftrightarrow Q_j$ is not in $\g$. Hence, we only need to show that  $Q_i \circbullet Q_j$ and $Q_i \bulletcirc Q_j$ are not in $\g$.  We will do this by contradiction.

Suppose first that $Q_i \circbullet Q_j$ is in $\g$. Since $i \ge 1$, $Q_{i-1} \bulletarrow Q_i$  is in $\g$. Hence, by Lemma \ref{lemma:prop1new},  $Q_{i-1} \to Q_j$, $Q_{i-1} \circbullet Q_j$ or $Q_{i-1} \arrowcirc Q_{j}$ is in $\g$. Then if $i = 1$, by \ref{zhao1} above, we immediately reach a contradiction.

If $Q_{i-1} \to Q_{j}$, or $Q_{i-1} \circbullet Q_j$ is in $\g$, then consider that $Q_{i-2} \bulletarrow Q_{i-1}$ is also in $\g$, and since orientations in $\g$ are closed under \ref{R1} and \ref{R4new} it follows that $\langle Q_{i-2}, Q_j \rangle$ must be in $\g$. Similarly, by the ancestral property of $\g$ and by Lemma \ref{lemma:prop1new}, $Q_{i-2} \leftarrow Q_j$ is not in $\g$. Hence, by \ref{zhao1} $A \neq Q_{i-2}$, that is $i > 2$ and $Q_{i-2} \arrowcirc Q_j$, $Q_{i-2} \to Q_j$, or $Q_{i-2} \circbullet Q_j$ is in $\g$.  

If $Q_{i-1} \arrowcirc Q_j$ is in $\g$, then by \ref{zhao2}, $Q_j \neq B$ and hence, $j < k+1$. Therefore, in this case, we can consider that $Q_{i-1} \arrowcirc Q_j \arrowbullet Q_{j+1}$ implies by Lemma \ref{lemma:prop1new} that edge $\langle Q_{i-1}, Q_{j+1} \rangle$ is in $\g$ and it is not of the form $Q_{i-1} \to Q_{j+1}.$ Hence, by \ref{zhao2}, $B \neq Q_{j+1}$, that is $j <  k$, and $Q_{i-1} \arrowcirc Q_{j+1}$, $Q_{i-1} \leftarrow Q_{j+1}$, or $Q_{i-1} \circbullet Q_{j+1}$ is in $\g$. 

Next, we can apply the same reasoning as above to conclude that $i > 3$, and or $j < k-1$, and so forth. Since $i<j$, we will eventually run into a contradiction. 

Analogously we can derive a contradiction when assuming that $Q_i \bulletcirc Q_j$ is in $\g$. Hence, $Q_i \to Q_j$, or $Q_i \leftarrow Q_j$ are in $\g$. 

\item[\ref{zhao4}]  This case follows from the fact that $\g$ is ancestral and cases \ref{zhao1}-\ref{zhao3} above.
\end{enumerate}
\end{proofof}

\begin{lemma}\label{lemma:helper1}
 Let $\g = (\mb{V,E})$ be an ancestral partial mixed graph that does not contain inducing paths. Furthermore, suppose that the edge orientations in $\g$ are closed under rules \ref{R1}, \ref{R2}, and \ref{R4new}, and let $p =  \langle A = Q_0, Q_1, \dots, Q_k, Q_{k+1} =B \rangle, k \ge 2$ be a minimal collider path in $\g$. 
Then the following hold 
\begin{enumerate}[label= (\roman*)]
    \item\label{helper11} For any subpath $p(Q_i, Q_{j}), 0 \le i < j-1 \le k$, there is at least one non-endpoint node $Q_l$, $l \in \{i+1, \dots , j-1\}$ such that $Q_l \notin \An(\{Q_i,Q_j\}, \g)$.
    \item\label{helper12} There is at least one unshielded triple on $p$.
    \item\label{helper13} Suppose that there is an edge $Q_i \to Q_j$, $i,j \in \{1, \dots ,k+1\}$, $i < j$ in $\g$. Then there is a node $Q_l$, $0 \le l <i$, such that $Q_l \notin \Adj(Q_j, \g)$ and $Q_{l_1} \to Q_{j}$ is in $\g$ for all $l_1 \in \{l+1, \dots , i \}$.
    \item\label{helper14} Suppose that there is an edge $Q_i \leftarrow Q_j$, $i,j \in \{0,1, \dots ,k\}$, $i < j$ in $\g$. Then there is a node $Q_r$, $j < r \le k+1$, such that $Q_r \notin \Adj(Q_i, \g)$ and $Q_{r_1} \to Q_{i}$ is in $\g$ for all $r_1 \in \{j, \dots , r-1 \}$.
\end{enumerate}
\end{lemma}

\begin{proofof}[Lemma \ref{lemma:helper1}]  Since $p$ is a minimal collider path in $\g$, $A \notin \Adj(B, \g)$.

\begin{enumerate}
    \item[\ref{helper11}]  Suppose for a contradiction that there is a subpath $p(Q_i, Q_{j})$, of $p$ such that  for all $l \in \{i+1, \dots , j-1\}$, $Q_l \in \An(\{Q_i,Q_j\}, \g)$. Since there are no inducing paths in $\g$, $Q_i \in \Adj(Q_j, \g)$. Then by  Lemma \ref{lemma:zhao-mcp}, either $Q_i \to Q_j$, or $Q_i \leftarrow Q_j$ is in $\g$. However both options,  $Q_i \to Q_j \bulletarrow Q_{j-1} \to \dots \to Q_i$ or $Q_j \to Q_i \bulletarrow Q_{i+1} \to \dots \to Q_j$ contradict that $\g$ is an ancestral graph. 
    
    \item[\ref{helper12}] Suppose for a contradiction that every consecutive triple on $p$ is shielded. Then by Lemma \ref{lemma:zhao-mcp} it follows that $Q_0 \leftarrow Q_2$ and $Q_{k-1} \to Q_{k+1}$ is in $\g$. If $k = 2$, we then immediately reach a contradiction with  \ref{helper11} above.
    
Otherwise, suppose $k>2$.
 Since $Q_1 \leftrightarrow Q_2 \leftrightarrow Q_3$ is a shielded triple, it follows that $Q_1 \leftarrow Q_3$ or $Q_1 \to Q_3$ is in $\g$ (Lemma \ref{lemma:zhao-mcp}). However, since  $Q_0 \leftarrow Q_2$ is also in $\g$, by \ref{helper11} above, 
 we  conclude that $Q_1 \leftarrow Q_3$ must be in $\g$.  In fact, we can apply this argument iteratively to the remaining consecutive triples on $p$, until we reach $p(Q_{k-2},Q_{k+1})$ contradicting \ref{helper11} above. 
 \item[\ref{helper13},\ref{helper14}] Both of these cases follow from Lemma \ref{lemma:MCP-format}.
 \end{enumerate}
\end{proofof}

\begin{lemma}\label{lemma:helper11} 
 Let $\g = (\mb{V,E})$ be an ancestral partial mixed graph that does not contain inducing paths. Furthermore, suppose that the edge orientations in $\g$ are closed under rules \ref{R1} - \ref{R3}, and \ref{R4new}.  Suppose that there is a minimal collider path $p = \langle A =Q_{l_k},\dots, Q_{l_{1}}, Q =  Q_{r_0},  Q_{r_1}, \dots, Q_{r_m}= B \rangle$, $ m \geq 1$, $k>1$ in $\g$, and  a node $W$ that is not on $p$ such that the following are in $\g$:
\begin{enumerate}[label = (\alph*)]
    \item $W \circcirc Q$, and
    \item $W \circbullet Q_{r_1}$, and
    \item $W \arrowcirc Q_{l_i}$, or $W \leftarrow Q_{l_i}$, for $i \in \{1, \dots , k_1 \}$, $1\le k_1 \le k-2$ and
    \item $W \bulletcirc Q_{l_{k_1 +1}}$.
\end{enumerate}
Then 
\begin{enumerate}[label = (\roman*)]
     \item $Q_{l_{k_1 +2}} \in \Adj(W, \g)$, and
     \item $W \circbullet Q_{l_{k_1 +2}}$ is not in $\g$.
\end{enumerate}
\end{lemma}

\begin{proofof}[Lemma \ref{lemma:helper11}] Since $Q_{l_{k_1+2}} \bulletarrow Q_{l_{k_1 +1}} \circbullet W$ is in $\g$,  Lemma \ref{lemma:prop1new} implies that $Q_{l_{k_1 +2}} \in \Adj(W, \g)$. So it only remains to show that  $W \circbullet Q_{l_{k_1 +2}}$ is not in $\g$.
Suppose for a contradiction that $W \circbullet Q_{l_{k_1 +2}}$ is in $\g$. Below we obtain a contradiction if \ref{k1bigger1} $k_1 > 1$, and \ref{k1equal1} $k_1 = 1$.

\begin{enumerate}[label = (\arabic*)]
    \item\label{k1bigger1} Suppose first that $k_1 > 1$. If $W \circcirc Q_{l_{k_1 +1}}$ is in $\g$, then $Q_{l_1} \bulletarrow W \circcirc Q_{l_{k_1+1}}$ together with Lemmas \ref{lemma:prop1new} and \ref{lemma:zhao-mcp} implies that  $Q_{l_1} \to Q_{l_{k_1 +1}}$ is in $\g$. By the same reasoning $Q_{l_i} \bulletarrow W \circcirc Q$ implies that $Q_{l_i} \to Q$ is in $\g$, for $i \in \{2, \dots , k_1\}$. However now, $p(Q_{l_{k_1 +1}}, Q)$ contradicts \ref{helper11} of Lemma \ref{lemma:helper1}.
     
    Otherwise,  $W \arrowcirc Q_{l_{k_1 +1}}$ is in $\g$. But in this case,
    $Q_{l_1} \bulletarrow W \circbullet Q_{l_{k_1+2}}$
    together with Lemmas \ref{lemma:prop1new} and \ref{lemma:zhao-mcp} implies that $Q_{l_1} \to Q_{l_{k_1 +2}}$ is in $\g$. By the same reasoning  $Q_{l_i} \bulletarrow W \circcirc Q$ implies that $Q_{l_i} \to Q$ is in $\g$, for $i \in \{2, \dots , k_1 +1\}$. However now, $p(Q_{l_{k_1 +2}}, Q)$ contradicts \ref{helper11} of Lemma \ref{lemma:helper1}.  

    \item\label{k1equal1} Next, consider the case when $k_1 = 1$.  Since  having $Q \leftrightarrow Q_{l_1} \to W$ and $W \circcirc Q$ in $\g$  would contradict that orientations in $\g$ are closed under \ref{R2}, we must have that  $Q_{l_1} \circarrow W$ is in $\g$. Moreover, since $Q_{l_1} \circarrow W \circbullet Q_{r_1}$ and $Q_{l_1} \circarrow W \circbullet Q_{l_3}$ are in $\g$, Lemmas \ref{lemma:prop1new} and \ref{lemma:zhao-mcp} imply that $Q_{l_1} \to Q_{r_1}$ and $Q_{l_1} \to Q_{l_3}$ are in $\g$. 

    If $W \arrowcirc Q_{l_2}$ is in $\g$, then since  $Q_{l_2} \circarrow W \circcirc Q$ is in $\g$, Lemmas \ref{lemma:prop1new} and \ref{lemma:zhao-mcp} would lead us to conclude that $Q_{l_2} \to Q$ is in $\g$, making $p(Q_{l_3}, Q)$ contradict \ref{helper11} of Lemma \ref{lemma:helper1}.
    Alternatively, if $Q_{l_2} \circcirc W$ is in $\g$, then
    $Q_{l_{2}} \circcirc W \circcirc Q$, $Q_{l_{2}} \bulletarrow Q_{l_1} \leftrightarrow Q$, and  $Q_{l_1} \circbullet W$, together with \ref{R3} and Lemma \ref{lemma:zhao-mcp},
     would imply that either $Q_{l_2} \to Q$, or $Q_{l_2} \leftarrow Q$ are in $\g$. Having both $Q_{l_2} \to Q$ and $Q_{l_1} \to Q_{l_3}$ in $\g$, would make $p(Q_{l_3}, Q)$ contradict \ref{helper11} of Lemma \ref{lemma:helper1}. Alternatively, having both $Q_{l_2} \leftarrow Q$ and $Q_{l_1} \to Q_{r_1}$ in $\g$, would make $p(Q_{l_2}, Q_{r_1})$ contradict \ref{helper11} of Lemma \ref{lemma:helper1}.
\end{enumerate}
\end{proofof}

\begin{lemma}\label{lemma:helper4} 
Let $\g = (\mb{V,E})$ be an {ancestral partial mixed graph that does not contain inducing paths and such that orientations in $\g$ are closed under \ref{R1}-\ref{R3}, \ref{R4new}}.  Suppose that there is a minimal collider path $p = \langle A =Q_{l_k},\dots, Q_{l_{1}}, Q =  Q_{r_0},  Q_{r_1}, \dots, Q_{r_m}= B \rangle$, $ m \geq 1$, $k>1$, in $\g$, and  a node $W$ not on $p$ such that
\begin{enumerate}[label = (\alph*)]
    \item $W \bulletcirc Q$, and
    \item $W \circbullet Q_{r_1}$, and
    \item $W \bulletcirc Q_{l_i}$, or $W \leftarrow Q_{l_i}$, for $i \in \{1, \dots , k_1\}$, $k_1<k$ and
    \item $Q_{l_i} \to Q_{r_1}$ are in $\g$, for $i \in \{1, \dots , k_1\}$, $k_1<k$.
\end{enumerate}
Then 
\begin{enumerate}[label = (\roman*)]
     \item $\langle Q_{l_{k_1 +1}}, W \rangle$ is in $\g$, but not of the form $W \to Q_{l_{k_1 +1}} $, and
     \item $Q_{l_{k_1 +1}} \to Q_{r_1}$ is in $\g$.
\end{enumerate}
\end{lemma}

\begin{proofof}[Lemma \ref{lemma:helper4}]
This proof is split into three cases depending on the forms of edges $\langle Q_{l_{k_1}},W \rangle$ and $\langle W,Q \rangle$:  \ref{case:a} $Q_{l_{k_1}} \to W$ is in $\g$, \ref{case:b}  $Q_{l_{k_1}} \circbullet W$ and  $W \arrowcirc Q$ are in $\g$, and \ref{case:c}  $Q_{l_{k_1}} \circbullet W$ and  $W \circcirc Q$ are in $\g$.
\begin{enumerate}[label = (\alph*)]
 \item\label{case:a} In this case we assume that $Q_{l_{k_1}} \to W$ is in $\g$. Let $i_1 \in \{1, \dots , k_1\}$ be the largest index such that $Q_{l_{i_1}} \circbullet W$ is in $\g$. If such an index does not exist then let $i_1 = 0$ and $Q_{l_{0}} = Q$ since  $Q_{l_0} \circbullet W$ is in $\g$. 
 
Since $k>k_1,$ we now have that  $Q_{l_{k_1 +1}} \bulletarrow Q_{l_{k_1}} \leftrightarrow \dots \leftrightarrow Q_{l_{i_1}}  \circbullet W$ is in $\g$. Furthermore,  $Q_{l_{i'}} \to W$ is in $\g$ for all $i' \in \{i_1 +1, \dots , k_1\}$. Hence, since orientations in $\g$ are closed under \ref{R4new},  $Q_{l_{k_1 +1}} \in \Adj(W, \g)$. Furthermore,  $Q_{l_{k_1 +1}} \leftarrow W$ is not in $\g$ since $\g$ is ancestral. In fact, since orientations in $\g$ are closed under \ref{R2}, $Q_{l_{k_1 +1}} \bulletarrow W$ is in $\g$. 
Now, $Q_{l_{k_1 +1}} \bulletarrow W \circbullet Q_{r_1}$ implies that  $Q_{l_{k_1 +1}} \to Q_{r_1}$ is in $\g$, by Lemmas \ref{lemma:prop1new} and \ref{lemma:zhao-mcp}.
 
    \item\label{case:b} In this case we assume that  $Q_{l_{k_1}} \circbullet W$ and  $W \arrowcirc Q$ are in $\g.$ 
Since $k>k_1$, and $Q_{l_{k_1 +1}} \bulletarrow Q_{l_{k_1}} \circbullet W$ is  in $\g$, Lemma \ref{lemma:prop1new} implies that $Q_{l_{k_1 +1}} \in \Adj(W,\g)$ and that  $Q_{l_{k_1 +1}} \leftarrow W$ is not in $\g.$ 

Note also that $Q_{l_{k_1 +1}} \bulletcirc W$ is not possible, since $Q_{l_{k_1 +1}} \bulletcirc W \arrowcirc Q$ would by Lemmas \ref{lemma:prop1new} and \ref{lemma:zhao-mcp} imply that $Q_{l_{k_1 +1}} \leftarrow Q$ thus, together with $Q_{l_i} \to Q_{r_1}$ for all $i \in \{1, \dots , k_1\}$, making $p(Q_{l_{k_1 +1}}, Q_{r_1})$ contradict \ref{helper11} of Lemma \ref{lemma:helper1}. 
Hence, $Q_{l_{k_1 +1}} \bulletarrow W \circbullet Q_{r_1}$ is in $\g$ implying that $Q_{l_{k_1 +1}} \to Q_{r_1}$ is also in $\g$ by Lemmas \ref{lemma:prop1new} and \ref{lemma:zhao-mcp}.

 \item\label{case:c} n this case we assume that   $Q_{l_{k_1}} \circbullet W$ and $W \circcirc Q$ are in $\g.$ Let $Q_{l_0} = Q$.  As in the above cases, note that since $k>k_1$, $Q_{l_{k_1 +1 }}\bulletarrow Q_{l_{k_1 }} \circbullet W$ is  in $\g$. Therefore, Lemma \ref{lemma:prop1new} implies that $Q_{l_{k_1 +1 }} \in \Adj(W, \g)$  and $Q_{l_{k_1 +1 }} \leftarrow W$ is not in $\g.$  If $Q_{l_{k_1 +1 }} \bulletarrow W$ is in $\g,$ we can use exactly the same argument as in \ref{case:b}  to show that $Q_{l_{k_1 +1 }} \to Q_{r_1}$ is in $\g.$ 
 
{ Otherwise, $Q_{l_{k_1 +1 }} \bulletcirc W$  is in $\g$. Suppose first that $k_1 = 1$. Then $Q_{l_{2}} \bulletcirc W \circcirc Q$, $Q_{l_{2}} \bulletarrow Q_{l_1} \leftrightarrow Q$, and  $Q_{l_1} \circbullet W$, together with \ref{R3} and Lemma \ref{lemma:zhao-mcp}, imply that $Q_{l_2} \to Q$, or $Q_{l_2} \leftarrow Q$ is in $\g$. Since $Q_{l_2} \leftarrow Q$ together with $Q_{l_1} \to Q_{r_1}$ would imply that $p(Q_{l_2}, Q_{r_1})$ contradicts \ref{helper11} of Lemma \ref{lemma:helper1}, it must be that $Q_{l_2} \to Q$ is in $\g$. We can now apply \ref{R3} and Lemma \ref{lemma:zhao-mcp}  to  $Q_{l_{2}} \bulletcirc W \circbullet Q_{r_1}$, $Q_{l_{2}} \to  Q \arrowbullet Q_{r_1}$, and  $Q \circcirc W$ to conclude that $Q_{l_2} \to Q_{r_1}$ must be in $\g.$ 

Next, suppose that $k_1 >1$.
Note, that if there is any edge  $Q_{l_{i_1}} \circarrow W$, or $Q_{l_{i_1}} \to W$ in $\g$, for $ i_1 \in \{1, \dots, k_1-1\}$, we can construct a contradiction with Lemma \ref{lemma:helper11}. Hence, all edges $\langle Q_{l_{i_1}}, W \rangle, i_1 \in \{0, \dots, k_1-1\}$ must be of the form $ Q_{l_{i_1}} \circcirc W$ in $\g$.} 

Note that  $Q_{l_{k_1 +1 }} \bulletcirc W \circcirc Q_{l_{k_1 -1}}$ and $Q_{l_{k_1}} \circbullet W$ with $Q_{l_{k_1 +1 }} \bulletarrow Q_{l_{k_1 }} \leftrightarrow Q_{l_{k_1-1}}$ and \ref{R3} imply that $Q_{l_{k_1 +1 }} \in \Adj(Q_{l_{k_1-1}},\g)$. Due to Lemmas  \ref{lemma:zhao-mcp}, and \ref{lemma:helper1}, this edge must be of the form $Q_{l_{k_1 +1 }} \to Q_{l_{k_1-1}}$.

Then $Q_{l_{k_1 +1 }} \bulletcirc W \circcirc Q_{l_{k_1 -2}}$, $Q_{l_{k_1}} \circcirc W$, and $Q_{l_{k_1 +1 }} \to Q_{l_{k_1-1 }} \leftrightarrow Q_{l_{k_1-2}}$ are in $\g$. Hence, by \ref{R3},  Lemma \ref{lemma:zhao-mcp}, and Lemma \ref{lemma:helper1},  $Q_{l_{k_1 +1 }} \to Q_{l_{k_1-2}}$ is in $\g$. Since $ Q_{l_{i_1}} \circcirc W$ for all $i_1 \in \{0, \dots , k_1\}$, we can keep iterating the above procedure until we get that $Q_{l_{k_1 +1 }} \to Q$ is in $\g$. The conclusion that $Q_{l_{k_1 +1 }} \to Q_{r_1}$ is in $\g$, then follows from the above paragraph.  
\end{enumerate}
\end{proofof}

\begin{lemma}\label{lemma:helper3} 
Let $\g = (\mb{V,E})$ be an {ancestral partial mixed graph that does not contain inducing paths and such that orientations in $\g$ are closed under \ref{R1}-\ref{R3}, \ref{R4new}}.  Suppose that there is a minimal collider path $p = \langle A =Q_{l_k},\dots, Q_{l_{1}}, Q =  Q_{r_0},  Q_{r_1}, \dots, Q_{r_m}= B \rangle$, $k, m, \geq 1$, $m +k \ge 2$ in $\g$, and  a node $W$ that is not on $p$ such that  $W \bulletcirc Q$ is in $\g$. Then edges $\langle  Q_{l_1}, W \rangle$, and $\langle  W, Q_{r_1} \rangle$ are in $\g$. Furthermore, both of these edges are into $W$.
\end{lemma}

\begin{proofof}[Lemma \ref{lemma:helper3}]
First note that edges  $\langle  Q_{l_1}, W \rangle$, and $\langle  W, Q_{r_1} \rangle$ are in $\g$ by Lemma \ref{lemma:prop1new}. Furthermore, by the same lemma, neither $W \to Q_{l_1}$,  nor $W \to Q_{r_1}$ is in $\g$. Hence, we have the following options for the triple $\langle Q_{l_1}, W, Q_{r_1} \rangle$, $Q_{l_1} \bulletarrow W \arrowbullet Q_{r_1}$, $Q_{l_1} \bulletarrow W \circbullet Q_{r_1}$, $Q_{l_1} \bulletcirc  W \arrowbullet Q_{r_1}$, $Q_{l_1} \bulletcirc W \circbullet Q_{r_1}$. For the remainder of the proof, our goal is to rule out the latter three options.

Note that if $k =1$, we can rule out that $Q_{l_1} \bulletarrow W \circbullet  Q_{r_1}$ is in $\g$, since in this case Lemma \ref{lemma:prop1new}, would imply that $Q_{l_1} \bulletarrow Q_{r_1}$ is in $\g$, but since $Q_{l_1} = A$ that would contradict that $p$ is a minimal collider path. 
Similarly, if $m =1$, we can rule out that  $Q_{l_1} \bulletcirc W \arrowbullet  Q_{r_1}$ is in $\g$, by a symmetric argument. 
Furthermore, if $k = m = 1$, we can also rule out that  $Q_{l_1} \bulletcirc W \circbullet Q_{r_1}$, since that in combination with $Q \circbullet W$, and $Q_{l_1} \bulletarrow Q \arrowbullet Q_{r_1}$ and $Q_{l_1} \notin \Adj(Q_{r_1}, \g)$ contradicts that orientations in $\g$ are closed under \ref{R3}. Hence, if $k = m = 1$, we are done.

For the rest of the proof, suppose that  either $k>1$ or $m >1$ and for contradiction suppose that one of the following is in $\g$: $Q_{l_1} \bulletarrow W \circbullet Q_{r_1}$, $Q_{l_1} \bulletcirc  W \arrowbullet Q_{r_1}$, or $Q_{l_1} \bulletcirc W \circbullet Q_{r_1}$.
Note also that if $Q_{l_1} \bulletarrow W \circbullet Q_{r_1}$ is in $\g$, then $Q_{l_1} \to Q_{r_1}$ is in $\g$, by Lemmas \ref{lemma:prop1new} and \ref{lemma:zhao-mcp}, so either $k >1$ or we have reached a contradiction with $p$ being a minimal collider path.  Similarly if $Q_{l_1} \bulletcirc W \arrowbullet Q_{r_1}$ is in $\g$ then $Q_{l_1} \leftarrow Q_{r_1}$ is in $\g$, by Lemmas \ref{lemma:prop1new} and \ref{lemma:zhao-mcp}, so either $m >1$ or we have reached a contradiction with $p$ being a minimal collider path.
Lastly, if $Q_{l_1} \bulletcirc W \circbullet Q_{r_1}$ is in $\g$, then since $W \bulletcirc Q$ and $Q_{l_1} \bulletarrow Q \arrowbullet Q_{r_1}$ are also in $\g$ and since orientations in $\g$ are closed under \ref{R3},  $Q_{l_1} \in \Adj(Q_{r_1}, \g)$ is in $\g$. 
By Lemma \ref{lemma:zhao-mcp}, $Q_{l_1} \to Q_{r_1}$, or $Q_{l_1} \leftarrow Q_{r_1}$ is in $\g$. Note that if  $Q_{l_1} \to Q_{r_1}$ then either $k >1$, or we have reached a contradiction with $p$ being a minimal collider path, and similarly, if $Q_{l_1} \leftarrow Q_{r_1}$ then either $m >1$, or we have reached a contradiction with $p$ being a minimal collider path.
Therefore, the following combinations remain to be discussed:
\begin{enumerate}[label=(\alph*)]
    \item $k>1$, and $W \circbullet Q_{r_1}$ and $Q_{l_1} \to Q_{r_1}$ are in $\g$, or
    \item $m>1$, and $W \circbullet Q_{l_1}$ and $Q_{l_1} \leftarrow Q_{r_1}$ are in $\g$.
\end{enumerate}
The proof for the above cases is symmetric, so without loss of generality we will assume that $k>1$,  $W \circbullet Q_{r_1}$ and $Q_{l_1} \to Q_{r_1}$ are in $\g$ and show that assumption leads to a contradiction.
We will show a contradiction under the following assumptions: \ref{case148}  there is no $i \in \{1, \dots , k\}$ such that $Q_{l_i} \arrowbullet W$ is in $\g$, and \ref{case248} there exists an $i \in \{1, \dots , k\}$ such that $Q_{l_i} \arrowbullet W$ is in $\g$.
\begin{enumerate}[label=(\arabic*)]
    \item\label{case148}  There is no $i \in \{1, \dots , k\}$ such that $Q_{l_i} \arrowbullet W$ is in $\g$. In this case, $Q_{l_1} \circbullet W$, or $Q_{l_1} \to W$ is in $\g$ and by assumption $Q_{l_1} \to Q_{r_1}$ and $W \circbullet Q_{r_1}$ are also in $\g$. We can now use Lemma \ref{lemma:helper4} iteratively to show that $Q_{l_{i}} \circbullet W$, or $Q_{l_{i}} \to W$ is in $\g$, for all $i \in \{1, \dots , k\}$. Additionally, by the same lemma, we will also have that $Q_{l_{i}} \to Q_{r_1}$, for all $i \in \{1, \dots , k\}$. Since $Q_{l_{k}} = A$, we now reach a contradiction with Lemma \ref{lemma:zhao-mcp}.
    
    \item\label{case248} There is an $i \in \{1, \dots , k\}$ such that $Q_{l_i} \arrowbullet W$ is in $\g$, and $Q_{l_{i_1}}$ is the closest such node to $Q$ on $p(A, Q)$.  In this case, $Q_{l_{i_1}} \arrowbullet W$ is in $\g$ and $Q_{l_{i}} \circbullet W$ or $Q_{l_{i}} \to W$ is in $\g$, for all $i \in \{1, \dots, i_1-1 \}$. Furthermore, by Lemma \ref{lemma:helper4}, $Q_{l_{i}} \to Q_{r_1}$ is in $\g$, for all $i \in \{1, \dots , i_1 \}$. Since $Q_{l_{i_1}} \leftrightarrow Q_{l_{i_1-1}} \to W$, or $Q_{l_{i_1}} \leftrightarrow Q_{l_{i_1-1}} \circbullet W$, by the ancestral property of $\g$ and Lemma \ref{lemma:prop1new}, $Q_{l_{i_1}} \leftarrow W$ is not in $\g$. Hence, $Q_{l_{i_1}} \arrowbullet W$ is either $Q_{l_{i_1}} \leftrightarrow W$ or $Q_{l_{i_1}} \arrowcirc W$.  

    Now, since $Q_{l_{i_1}} \to Q_{r_1}$ is in $\g$, either $i_1 = k$ and we have reached a contradiction with Lemma \ref{lemma:zhao-mcp}, or by Lemma \ref{lemma:helper1}, there is a node $Q_{l_{i_2}}$ on $p(A, Q_{l_{i_1}})$ such that $Q_{l_{i_2}} \notin \Adj(Q_{r_1}, \g)$, and $Q_{l_i} \to Q_{r_{1}},$ for all $i \in \{i_1, \dots , i_2 - 1\}$. 
    But in this case, we also have the path $p(Q_{l_{i_2}}, Q_{l_{i_1}}) \oplus \langle Q_{l_{i_1}}, W \rangle \oplus \langle W, Q_{r_1} \rangle$ which  contradicts that orientations in $\g$ are closed under \ref{R4new}.
\end{enumerate}
\end{proofof}

\begin{lemma}
\label{lemma:induction-result-prime-helper} 
Let $\g = (\mb{V,E})$ be an {ancestral partial mixed graph that does not contain inducing paths and such that orientations in $\g$ are closed under \ref{R1}-\ref{R3}, \ref{R4new}}. Suppose that there is a minimal collider path $p = \langle A=Q_{l_k},\dots, Q_{l_{1}}, Q,  Q_{r_1}, \dots, Q_{r_m}= B \rangle$, $m \geq 1$, $k >1$,  in $\g$ and a node $W$ not on $p$ such that 
\begin{enumerate}[label = (\alph*)]
   \item  $W \bulletcirc Q$ is in $\g$, and
    \item $Q_{l_i} \circbullet W$ or $Q_{l_i} \to W$  is in $\g$ for $i \in \{1, \dots , k_1\}$, $k_1  < k$.
\end{enumerate}
Then $Q_{l_{k_1 +1}} \bulletarrow W$ is in $\g$.
\end{lemma}

\begin{proofof}[Lemma \ref{lemma:induction-result-prime-helper}] Suppose first that  $Q_{l_{k_1}} \circarrow W$ is in $\g$. Then directly by Lemma \ref{lemma:helper3}, $Q_{l_{k_1 +1}} \bulletarrow W$ is in $\g$.
Hence, for the remainder suppose that $Q_{l_{k_1}} \to W$ and let $Q_{l_0} \equiv Q$.
Let $i_1 \in \{0, \dots , k_1 -1\}$ be such that $Q_{l_{i_1}}$ is the closest node to $Q_{l_{k_1}}$ on $p(Q_{l_{k_1}}, Q)$ such that $Q_{l_{i_1}} \circbullet W$ is in $\g$. Now, $Q_{l_i} \to W$ for all $i \in \{i_1 +1, \dots , k_1\}$ and $Q_{l_{k_1 +1}} \bulletarrow Q_{l_{k_1}} \leftrightarrow \dots \leftrightarrow Q$ is in $\g$, so since orientations in $\g$ are closed under \ref{R4new}, it follows that $Q_{l_{k_1 +1}} \in \Adj(W, \g).$ Since $\g$ is ancestral and $Q_{l_{k_1 +1}} \bulletarrow Q_{l_{k_1}} \to W$ is in $\g$, $Q_{l_{k_1 +1}} \leftarrow W$ is not in $\g$. Additionally, since orientations in $\g$ are closed under \ref{R2},  $Q_{l_{k_1 +1}} \bulletcirc W$ is also not in $\g$. Hence,  $Q_{l_{k_1 +1}} \bulletarrow W$ is in $\g$.
\end{proofof}

\section{Supplement to Section \ref{sec:new-rules}}
\label{appendix:new-rules}

\begin{proofof}[Theorem \ref{thm:meek}]
 We prove the theorem by contradiction while considering different possibilities for the orientation of the $A \bulletbullet B$ edge. Hence, suppose for a contradiction that there is a MAG $\g[M]$ represented by $\g$ that contains $A \arrowbullet D$ and \ref{R11-case-i} $A \arrowbullet B$, or \ref{R11-case-ii} $A \to B$.

\begin{enumerate}[label=(\roman*)]
\item\label{R11-case-i} We immediately have the contradiction in this case, as $D \bulletarrow A \arrowbullet B$ is an unshielded collider in $\g[M]$ that is not in $\g$. Hence, $\g[M]$ cannot be represented by $\g$.

\item\label{R11-case-ii} We assume that  $A \to B$ and $A \arrowbullet D$ are in $\g[M]$. Then $D \to A$ cannot be in $\g[M]$, as $C \to D \to A \to B \bulletarrow C$ is either a directed or an almost directed cycle. Hence, $D \leftrightarrow A$ is in $\g[M]$. Furthermore, using similar reasoning, $C \to D \leftrightarrow A \to B$ implies that $B \leftrightarrow C$ is in $\g[M]$, and $C \to D \leftrightarrow A$ implies that $A \leftrightarrow C$. But this gives us an inducing path $D \leftrightarrow A \leftrightarrow C \leftrightarrow B$ in $\g[M]$, which is a contradiction.
\end{enumerate}
\end{proofof}

\begin{proofof}[Theorem \ref{thm:rule12}]
Suppose for a contradiction that there exists a MAG $\g[M]$ represented by $\g$ such that $V_1 \to V_2$ is in $\g[M]$ and let $p = \langle V_1, \dots , V_i \rangle$, and $q = \langle V_i, V_{i+1}, V_1 \rangle.$.

Since $\g[M]$ contains only those unshielded colliders already present in $\g$ and since $p$ is an unshielded possibly directed path in $\g$, we will have that the path corresponding to $p$ in $\g[M]$ is of the form $V_1 \to V_2 \to \dots \to V_i$. Hence, the paths corresponding to $p$ and $q$ in $\g[M]$  an almost directed cycle, which is a contradiction with $\g[M]$ being an ancestral graph.
\end{proofof}

\begin{proofof}[Theorem \ref{thm:rule13star}] Suppose for a contradiction that there is a MAG $\g[M]$ represented by $\g$ that contains $A \to B$.
Since $\g[M]$ does not  contain new unshielded colliders compared to $\g$, the paths corresponding to $\langle A, B, \dots , V_i \rangle$, must be of the form $A \to B \to \dots \to V_i$ in $\g[M]$ for all $i \in \{1, \dots, k\}$. 
Furthermore, since $\g[M]$ is ancestral, the path $C \leftrightarrow A \to \dots \to \dots V_1$ in $\g[M]$ implies that the edge $C \arrowcirc V_1$ in $\g$ is oriented as $C \leftrightarrow V_1$ in $\g[M]$. Similarly, path $D \leftrightarrow A \to \dots \to V_k$ in $\g[M]$ implies $D \leftrightarrow V_k$ is in $\g[M]$. However, now, any orientation of the remaining edges on unshielded path $\langle C, V_1, \dots, V_k , D \rangle$ implies a presence of a new unshielded collider in $\g[M]$ compared to $\g$, which is a contradiction. 
\end{proofof}

\begin{proofof}[Theorem \ref{thm:rule4newnew}]  For the sake of contradiction, assume that there is a MAG $\g[M]$ represented by $\g$ that contains $Q_k \arrowbullet B$. Let $p_{\g[M]}$ be the path in $\g[M]$ corresponding to $p$ in $\g$. 
Note that $p$ is not a collider path. Moreover, there cannot be a subsequence of $p$ that forms a collider path in $\g$ since that would require an edge of the form $Q_j \arrowbullet B$, $j \in \{0, \dots, k\},$ and by choice of $p$ there is no such edge in $\g$. 

We will derive the contradiction by proving that there is a subsequence of $p_{\g[M]}$ that forms a collider path from $A$ to $B$ in $\g[M].$ Hence, there is also a subsequence of $p_{\g[M]}$ that forms a minimal collider path from $A$ to $B$, which ultimately gives us the contradiction with $\g[M]$ being represented by $\g$ by Definition \ref{def:represent}.

 Note that since $Q_k \arrowbullet B \leftarrow Q_{k-1}$ is in $\g[M]$, and since $\g[M]$ is ancestral it follows that $Q_k \arrowbullet Q_{k-1}$ is in $\g[M]$, that is $Q_k$ is a collider on $p_{\g[M]}$.
 If the remaining non-endpoint nodes on $p_{\g[M]}$ are colliders, then the contradiction is immediate. 
Otherwise, there is at least one non-endpoint node on $p_{\g[M]}$ that is a non-collider. Let $ \{Q_{k_1}, \dots , Q_{k_m} \} $, $m \ge 1$ and $1 \le k_i < k_j \le k -1 $, $ 1 \le i < j \le m$, be the non-colliders on $p_{\g[M]}$. We will show how to ``skip over'' one or two of these non-colliders and construct a subsequence of $p_{\g[M]}$ called $p_{\g[M]}^{1}$ that has one fewer non-collider, or a subsequence of $p_{\g[M]}$ called $p_{\g[M]}^{2}$ that has two fewer non-colliders. This process can then be applied again on the obtained subsequence until we reach a subsequence of $p_{\g[M]}$ called $p_{\g[M]}^{m}$ that is a collider path, thereby deriving the contradiction.

Hence, let $i= k_j$. Since $Q_i$ is a non-collider on $p_{\g[M]}$, $Q_i$ satisfies \ref{starting-node}\ref{first-node-b}, \ref{starting-node}\ref{first-node-c}, \ref{inside-node}\ref{inside-node-b}, \ref{inside-node}\ref{inside-node-c}, \ref{ending-node}\ref{last-node-b}, or \ref{ending-node}\ref{last-node-c} of Definition \ref{def:almost-collider} on $p$. 
We now discuss each of these cases and show how to construct $p_{\g[M]}^{1}$.

\begin{enumerate}
    \item[\ref{starting-node}\ref{first-node-b}] $Q_0 \bulletarrow Q_1 \circarrow Q_2$ and $Q_0 \bulletcirc Q_2$ is in $\g$.
Since $Q_1$ is a non-collider on $p_{\g[M]}$, $Q_0 \bulletarrow Q_1 \to Q_2$ is in $\g[M]$. Additionally, since $\g[M]$ is an ancestral graph, the edge between $Q_0$ and $Q_2$ is $Q_0 \bulletarrow Q_2$. Hence, let $p_{\g[M]}^{1} = \langle Q_{0}, Q_{2} \rangle \oplus p_{\g[M]}(Q_{2}, B)$.

If $Q_2$ is a collider on both $p_{\g[M]}$ and  $p_{\g[M]}^{1}$, then $p_{\g[M]}^{1}$ has one fewer non-collider.
If however, $Q_2$ is a non-collider on $p_{\g[M]}$, then $Q_2 \to Q_3$ is on $p_{\g[M]}$ as well. Therefore, $Q_1 \circarrow Q_2 \circarrow Q_3$ is on $p$. And by choice of $p$, $Q_1 \arrowcirc Q_3$ would need to be in $\g$. Then $Q_1 \to Q_2 \to Q_3$ and $Q_1 \arrowbullet Q_3$ would imply that $\g[M]$ is not ancestral, which is a contradiction.

\item[\ref{starting-node}\ref{first-node-c}] $Q_0 \bulletcirc Q_1 \arrowbullet Q_2$ and {$Q_0 \bulletarrow Q_2$}  is in $\g$. Since $Q_1$ is a non-collider on $p_{\g[M]}$, $Q_0 \leftarrow Q_1 \arrowbullet Q_2$ is in $\g[M]$. Additionally, since $\g[M]$ is an ancestral graph, the edges between $Q_0$ and $Q_2$, and $Q_1$ and $Q_2$ must be  $Q_0 \leftrightarrow Q_2$, $Q_1 \leftrightarrow Q_2$. 
Now, if $Q_2$ is a collider on $p_{\g[M]}$, let as above $p_{\g[M]}^{1} = \langle Q_{0}, Q_{2} \rangle \oplus p_{\g[M]}(Q_{2}, B)$ and we are done. 

Otherwise, $Q_2$ a non-collider on $p_{\g[M]}$, meaning that  $Q_0 \leftarrow Q_1 \leftrightarrow Q_2 \to Q_3$ is in $\g[M]$. Consider what this implies in $\g$, we know that $Q_0 \bulletcirc Q_1 \arrowbullet Q_2$ is in $\g$ and we know that $Q_2 \to Q_3$ is in $\g[M]$. By properties of $p$ as an almost discriminating path, $Q_2 \circarrow Q_3$ must be in $\g$. This furthermore implies that $Q_1 \leftrightarrow Q_2 \circarrow Q_3$, and $Q_1 \arrowcirc Q_3$ is in $\g$.
Hence, since $Q_0 \leftarrow Q_1 \leftrightarrow Q_2 \to Q_3$ is in $\g[M]$, for $\g[M]$ to be ancestral, $Q_1 \leftrightarrow Q_3$ is also in $\g[M]$. 

Therefore, we have that $Q_0 \leftrightarrow Q_2 \leftrightarrow Q_1 \leftrightarrow Q_3$, and $Q_2 \to Q_3$, $Q_1 \to Q_0$ are in $\g[M]$. Now, since $\g[M]$ is a maximal graph, edge $\langle Q_0, Q_3 \rangle$ is in $\g[M]$. Furthermore, for $\g[M]$ to be ancestral, it must be of the form $Q_0 \leftrightarrow Q_3$. 

{{Now, there are two possibilities---either $Q_3 \arrowbullet Q_4$ is on $p$, or $Q_3 \circbullet Q_4$ and $Q_2 \arrowcirc Q_4$ are on $p$. In the first case, $Q_3$ is already a collider on $p$. In the second case, since we also have that $Q_2 \to Q_3$, for $\g[M]$ to be ancestral it must be that $Q_3 \arrowbullet Q_4 $ is in $\g[M]$. Therefore,  $Q_3$ is  collider on $p_{\g[M]}$ regardless of its status on $p$. 
Hence, let $p_{\g[M]}^{2} = \langle Q_{0}, Q_{3} \rangle \oplus p_{\g[M]}(Q_{3}, B)$. Then $p_{\g[M]}^{2}$ has two  fewer non-colliders than  $p_{\g[M]}$.}}

\item[\ref{inside-node}\ref{inside-node-b}] $Q_{i-1} \bulletarrow Q_i \circarrow Q_{i+1}$,  and $Q_{i-1} \arrowcirc Q_{i+1}$ are in $\g$ and $i \in \{2, \dots , k-2\}$. Since $Q_i$ is a non-collider on $p_{\g[M]}$, $Q_{i-1} \bulletarrow Q_i \to Q_{i+1}$ is in $\g[M]$. Additionally, since $Q_{i-1} \bulletarrow Q_i \to Q_{i+1}$, $\g[M]$ is an ancestral graph, and $Q_{i-1} \arrowcirc Q_{i+1}$ is in $\g$, the edges between $Q_{i-1}$ and $Q_{i+1}$ and $Q_{i-1}$ and $Q_i$ are $Q_{i-1} \leftrightarrow Q_{i+1}$, $Q_{i-1} \leftrightarrow Q_i$. 

Now, we know that $Q_{i-1} \leftrightarrow Q_i \to Q_{i+1}$ and $Q_{i-1} \leftrightarrow Q_{i+1}$ are in $\g[M]$. First we show that $Q_{i+1}$ is a collider on $p_{\g[M]}$. Note that $Q_{i+1}$ is either already a collider on $p$, or $Q_{i} \circarrow Q_{i+1} \circbullet Q_{i+2}$ and $Q_{i} \arrowcirc Q_{i+2}$ are in $\g$. In the latter case,  since $Q_{i} \to Q_{i+1}$ is in $\g[M]$ and since $\g[M]$ is ancestral, {$Q_{i+1} \arrowbullet Q_{i+2}$ is in $\g[M]$.} Hence, $Q_{i+1}$ is a collider on $p_{\g[M]}.$

Note that $Q_{i-1} \leftrightarrow Q_i$ is on $p_{\g[M]}$, so if  $Q_{i-1}$ is also a collider on $p_{\g[M]}$, let $p_{\g[M]}^{1} = p_{\g[M]}^{1}(A,Q_{i-1}) \oplus \langle Q_{i-1}, Q_{i+1} \rangle \oplus p_{\g[M]}(Q_{i+1}, B)$ and we are done. 

Otherwise, $Q_{i-1}$ is a non-collider on $p_{\g[M]}$, so since $Q_{i-1} \leftrightarrow Q_i$ is in $\g[M]$, it follows that $Q_{i-2} \bulletarrow Q_{i-1}$ cannot on $p$. Since $p$ is an almost discriminating path it must be that  $Q_{i-2} \bulletcirc Q_{i-1} \leftrightarrow Q_{i}$ and $Q_{i-2} \circarrow Q_{i}$ are in $\g$.
Then for $Q_{i-1}$ to be a non-collider on $p_{\g[M]}$, we have that $Q_{i-1} \leftarrow Q_{i-1} \leftrightarrow Q_i$ in $\g[M]$, and since $\g[M]$ is ancestral, and $Q_{i-2} \circarrow Q_{i} $ is in $\g$, $Q_{i-2} \leftrightarrow Q_{i}$ is in $\g[M]$. 

Consider that now we know that $Q_{i-2} \leftrightarrow Q_{i} \leftrightarrow Q_{i-1} \leftrightarrow Q_{i+1}$, $Q_{i} \to Q_{i+1}$ and $Q_{i-1} \to Q_{i-2}$ are in $\g[M]$. Hence, since $\g[M]$ is maximal $\langle Q_{i-2}, Q_{i+1} \rangle$ must also be in $\g[M]$. Furthermore, since $\g[M]$ is ancestral this edge between $Q_{i-2}$ and $Q_{i+1}$ is of the form $Q_{i-2} \leftrightarrow Q_{i+1}$.

If $i =2$, let  $p_{\g[M]}^{2} =  \langle Q_{i-2}, Q_{i+1} \rangle \oplus p_{\g[M]}(Q_{i+1}, B)$ and we are done. 
Otherwise, $i >2$, so edge $Q_{i-2} \bulletcirc Q_{i-1}$ is of the form $Q_{i-2} \arrowcirc Q_{i-1}$ on $p$. Furthermore, then either $Q_{i-2}$ is a collider on $p$, or $Q_{i-3} \bulletcirc Q_{i-2} \arrowcirc Q_{i-1}$ and $Q_{i-3} \circarrow Q_{i-1}$ is in $\g$. In the latter case,  since $\g[M]$ is an ancestral graph and since $Q_{i-2} \leftarrow Q_{i-1}$ is in $\g[M],$ $Q_{i-3} \leftrightarrow Q_{i-2}$ and $Q_{i-3} \leftrightarrow Q_{i-1}$ are also in $\g[M]$.  Hence, under both options, we have that $Q_{i-2}$ is a collider on $p_{\g[M]}$. Hence, $p_{\g[M]}^{2} = p_{\g[M]}(A,Q_{i-2}) \oplus \langle Q_{i-2}, Q_{i+1} \rangle \oplus p_{\g[M]}(Q_{i+1}, B)$ is a subsequence of $p_{\g[M]}$ with two fewer non-colliders.

\item[\ref{inside-node}\ref{inside-node-c}] $Q_{i-1} \arrowcirc Q_i \arrowbullet Q_{i+1}$, and $Q_{i-1} \circarrow Q_{i+1}$ are in $\g$ and $i \in \{2, \dots , k-2\}$. This case is exactly symmetric to the case \ref{inside-node}\ref{inside-node-b}. {Using a symmetric argument we can conclude that $Q_{i-1}$ is always a collider on $p_{\g[M]}$. Additionally, if $Q_{i+1}$ is not a collider on $p_{\g[M]}$, then $Q_{i+2}$ will be a collider on $p_{\g[M]}$. So we either show that $Q_{i-1} \leftrightarrow Q_{i+1}$ is in $\g[M]$ and construct the path $p_{\g[M]}^{1} = p_{\g[M]}(A,Q_{i-1}) \oplus \langle Q_{i-1}, Q_{i+1} \rangle \oplus p_{\g[M]}(Q_{i+1}, B)$ with one fewer non-collider compared to $p_{\g[M]}$, or show that $Q_{i-1} \leftrightarrow Q_{i+2}$ is in $\g[M]$ and construct the path $p_{\g[M]}^{2} = p_{\g[M]}(A,Q_{i-1}) \oplus \langle Q_{i-1}, Q_{i+2} \rangle \oplus p_{\g[M]}(Q_{i+2},B)$ with two fewer non-colliders.}

 \item[\ref{ending-node}\ref{last-node-b}] $Q_{k-2} \bulletarrow Q_{k-1} \circarrow Q_{k}$, and {$Q_{k-2} \arrowbullet Q_{k}$} is in $\g$. This case is symmetric to \ref{starting-node}\ref{first-node-c} and holds by an analogous argument. 

 \item[\ref{ending-node}\ref{last-node-c}] that is $Q_{k-2} \arrowcirc Q_{k-1} \arrowbullet Q_{k}$, and $Q_{k-2} \circbullet Q_{k}$ is in $\g$. This case is symmetric to \ref{starting-node}\ref{first-node-b} and holds by an analogous argument. 
\end{enumerate}
\end{proofof}

\subsection{Results Related \cite{wang2024new} and \cite{wangpolynomial}} \label{supp:r13}

\subsubsection{\ref{R13} and \cite{wang2024new}}
In this section, we show that our phrasing of \ref{R13} leads to equivalent orientations as the phrasing of the rule originally given by \cite{wang2024new}. We first state \cite{wang2024new}'s version of the rule in Theorem \ref{thm:wang2024} below (labeled \ref{WangR13}), which used their concept of an unbridged path, also defined below. We note that \cite{wang2024new} refer to the rule presented in Theorem \ref{thm:wang2024} as R12. 

Corollary \ref{cor:main-result-R13-eq} shows that \ref{WangR13} and \ref{R13} will lead to the same edge mark orientations when executed in combination with the remaining orientation rules.  The proof of Corollary \ref{cor:main-result-R13-eq} relies on  Corollary \ref{cor:1wang} which is based on \cite{wang2024new} proof of Theorem \ref{thm:wang2024} and Lemma \ref{lemma:equivalenceR13andR13wang} below. %The proof of Lemma \ref{lemma:equivalenceR13andR13wang} relies on a few supporting results given subsequently. We include a sketch of how the supporting results come together to prove Lemma \ref{lemma:equivalenceR13andR13wang} in Figure \ref{fig:eqR13}.

\begin{definition}[Unbridged path relative to $\mb{V}^{\prime}$, Definition 1 of  \citealp{wang2024new}]\label{def:unbridged-o} Let $\g = (\mb{V,E})$ be a partial mixed graph and $\mb{V}^{\prime} \subset \mb{V}$. If there is an unshielded path $p = \langle V_1, \dots, V_k \rangle$, $k >1$ of the form $V_1 \circcirc V_2 \circcirc \dots \circcirc V_k$ in $\g$,  $\{V_1, \dots, V_k \} \cap \mb{V}^{\prime} = \emptyset$ and such that $\mathcal{F}_{1} \setminus \mathcal{F}_2 \neq \emptyset,$ and $\mathcal{F}_{k} \setminus \mathcal{F}_{k-1} \neq \emptyset,$ where $\mathcal{F}_{i} = \{V \in \mb{V}^{\prime} : V \bulletcirc V_i \text{, or } V \bulletarrow V_i \text{ is in $\g$}\},$ then $p$ is called an unbridged path relative to $\mb{V}^{\prime}.$
\end{definition}

\begin{theorem}[Theorem 1 of \citealp{wang2024new}]\label{thm:wang2024}
Let $\g = (\mb{V,E})$ be a partial mixed graph.
\begin{enumerate}[label = Wang-R13, leftmargin = 2cm]
    \item\label{WangR13}  Suppose edge $A \circbullet B$ is in $\g$ and let $\mb{S}_A = \{V \in \mb{V} : V \bulletarrow A \text{ is in 
 } \g \} \cup \{A\}$. If there is an unbridged path $\langle V_1, \dots, V_k \rangle, k >1,$  relative to $\mb{S}_A$ in $\g$ such that for every $i \in \{1, \dots, k\}$ there is an unshielded path $p_i = \langle W_{i_1} = A, W_{i_2}= B, \dots, W_{i_m}= V_i \rangle, {{m} \geq 3}$  with no edge  $W_{i_j} \arrowbullet W_{i_{j+1}}$, $j \in \{1, \dots, m-1\}$ on $p_i$, then $A \arrowbullet B$ is in every MAG represented by $\g$. 
\end{enumerate}
\end{theorem}

\begin{corollary}[c.f. Proof of Theorem 1 of \citealp{wang2024new}]\label{cor:1wang} Let $\g = (\mathbf{V,E})$ be an essential ancestral graph and   $\g^{\prime} = (\mathbf{V,E'})$ be an ancestral partial mixed graph  such that $\g$ and $\g^{\prime}$ have the same skeleton, the same set of minimal collider paths, and every invariant edge mark in $\g$ is identical in $\g^{\prime}$. Suppose furthermore that edge orientations in $\g^{\prime}$ are closed under  \ref{R1}-\ref{R4}, \ref{R8}-\ref{R12}. Let $A \circbullet B$ be an edge in $\g^{\prime}$ and let $\mb{S}_A = \{V \in \mb{V} : V \bulletarrow A \text{ is in 
 } \g^{\prime} \} \cup \{A\}$. Suppose that  there is also an unbridged path $\langle V_1, \dots, V_k \rangle, k >1,$  relative to $\mb{S}_A$ in $\g^{\prime}$ such that for every $i \in \{1, \dots, k\}$ there is an unshielded path  $p_i = \langle W_{i_1} = A, W_{i_2}= B, \dots, W_{i_m}= V_i \rangle, {{m} \geq 3}$  with no edge  $W_{i_j} \arrowbullet W_{i_{j+1}}$, $j \in \{1, \dots, m-1\}$ on $p_i$.
 Then there are nodes 
 $C_1, C_2 \in \mb{S}_A$ such that  $C_1 \in \mathcal{F}_1 \setminus \mathcal{F}_2$, and $C_1 \notin \Adj(V_2, \g^{\prime})$, and  $C_2 \in \mathcal{F}_k \setminus \mathcal{F}_{k-1}$, and $C_2 \notin \Adj(V_{k-1}, \g^{\prime})$.
\end{corollary}

\begin{corollary}\label{cor:main-result-R13-eq}
  Let $\g = (\mathbf{V,E})$ be an essential ancestral graph and   $\g^{\prime} = (\mathbf{V,E'})$ be an ancestral partial mixed graph  such that $\g$ and $\g^{\prime}$ have the same skeleton, the same set of minimal collider paths, and every invariant edge mark in $\g$ is identical in $\g^{\prime}$. Suppose furthermore that edge orientations in $\g^{\prime}$ are closed under  \ref{R1}, \ref{R2}, \ref{R4}. \ref{R8},  \ref{R10}-\ref{R12}.  If there is an edge $A \circbullet B$ in $\g^{\prime}$ such that $A \arrowbullet B$ would be implied by \ref{WangR13}, then $A \arrowbullet B$ would also be implied by \ref{R13}.
\end{corollary}

\begin{proofof}[Corollary \ref{cor:main-result-R13-eq}] 
  By Lemma \ref{lemma:r9-notneeded}, orientations in $\g^{\prime}$ are closed under \ref{R9}. Furthermore, since $\g^{\prime}$ and $\g$ contain the same minimal collider paths and since orientations in $\g$ are closed under \ref{R3} (Corollary \ref{cor:old-rules-sound}), the orientations in $\g^{\prime}$ are also closed under \ref{R3}. The Corollary then  holds by Lemma \ref{lemma:equivalenceR13andR13wang} and Corollary \ref{cor:1wang}.
\end{proofof}

\begin{lemma}\label{lemma:equivalenceR13andR13wang}
 Let $\g = (\mathbf{V,E})$ be an essential ancestral graph and   $\g^{\prime} = (\mathbf{V,E'})$ be an ancestral partial mixed graph  such that $\g$ and $\g^{\prime}$ have the same skeleton, the same set of minimal collider paths, and every invariant edge mark in $\g$ is identical in $\g^{\prime}$. Suppose furthermore that edge orientations in $\g^{\prime}$ are closed under  \ref{R1}, \ref{R2}, \ref{R8}, \ref{R11}, \ref{R12}.  Let $A \circbullet B$ be an edge in $\g^{\prime}$ and let $\mb{S}_A = \{V \in \mb{V} : V \bulletarrow A \text{ is in 
 } \g^{\prime} \} \cup \{A\}$. Suppose that  there is also an unbridged path $\langle V_1, \dots, V_k \rangle, k >1,$  relative to $\mb{S}_A$ in $\g^{\prime}$ such that for every $i \in \{1, \dots, k\}$ there is an unshielded path  $p_i = \langle W_{i_1} = A, W_{i_2}= B, \dots, W_{i_m}= V_i \rangle,{{m} \geq 3}$  with no edge  $W_{i_j} \arrowbullet W_{i_{j+1}}$, $j \in \{1, \dots, m-1\}$ on $p_i$.
Let $C_1, C_2  \in \mb{S}_A$ be nodes such that  $C_1 \in \mathcal{F}_1$, and $C_1 \notin \Adj(V_2, \g^{\prime})$, and  $C_2 \in \mathcal{F}_k$, and $C_2 \notin \Adj(V_{k-1}, \g^{\prime})$.  Then the following hold:
\begin{enumerate}[label = (\roman*)]
    \item\label{case1:equivalenceR13} $C_1 \bulletcirc V_1$ and $C_2 \bulletcirc V_k$ is in $\g^{\prime}$.
    \item\label{case2:equivalenceR13} For every $i \in \{1, \dots, k\}$, $p_i$ is a possibly directed path from $A$ to $V_i$.
    \item\label{case3:equivalenceR13} $A \notin \cup_{i=1}^{k} \Adj(\{V_1,V_k\}, \g^{\prime})$ which also implies that $A \notin \{C_1, C_2\}$.
    \item\label{case4:equivalenceR13} $C_1 \bulletcirc A \circbullet C_2$ is in $\g$.
    \item\label{case5:equivalenceR13} $V_1 \circarrow C_1 \leftrightarrow C_2 \arrowcirc V_k$ is in $\g$.
    \item\label{case6:equivalenceR13} $A \circarrow C_1 \arrowcirc V_1$ and $A \circarrow C_2 \arrowcirc V_k$ are in $\g$.  
\end{enumerate}
\end{lemma}

\begin{proofof}[Lemma \ref{lemma:equivalenceR13andR13wang}]
\begin{enumerate}
    \item[\ref{case1:equivalenceR13}] Note that since $C_1 \in \mathcal{F}_1$ it follows that $C_1 \bulletcirc V_1$ or $C_1 \bulletarrow V_1$ is in $\g^{\prime}$. However, since $V_1 \circcirc V_2$ and since $C_1 \notin \Adj(V_2, \g^{\prime})$, it follows that $C_1 \bulletarrow V_1$ cannot be in $\g^{\prime}$ (otherwise, orientations are not closed under  \ref{R1}). Therefore, $C_1 \bulletcirc V_1$ is in $\g^{\prime}$. 
    We can obtain that $C_2 \bulletcirc V_k$ is in $\g^{\prime}$ using analogous reasoning.
     \item[\ref{case2:equivalenceR13}] Follows by Lemma \ref{lemma:unshielded-poss-dir}.  
    %Since every $p_i$ is an unshielded path and orientations in $\g^{\prime}$ are closed under \ref{R1} it follows that if there is an arrowhead at $W_{i_{j+1}}$ on any edge $W_{i_j} \bulletarrow W_{i_{j+1}}$, then $p_i (W_{i_{j+1}}, W_{i_m})$ must be a directed path. Furthermore, Lemma \ref{lemma:new-circle-path} implies  $p_i(W_{i_1}, W_{i_{j+1}})$ is an unshielded possibly directed path. Now, lastly, we have by Lemma \ref{lem:concatenation} that $p(W_{i_1}, W_{i_{j+1}}) \oplus p_i(W_{i_{j+1}}, W_{i_m})$ is an unshielded possibly directed path. 
    \item[\ref{case3:equivalenceR13}] We will only prove that $A \notin \Adj(V_1, \g^{\prime})$ by contradiction. The proof of  $A \notin \Adj(V_k \g^{\prime})$ would be exactly symmetric. Hence, suppose for a contradiction that $A \in \Adj(V_1, \g^{\prime})$. 
    
    Note that since $p_1$ is a possibly directed unshielded path (by \ref{case2:equivalenceR13} above), we have that $A \circcirc V_1$, $A \circarrow V_1$ or $A \to V_1$ is in $\g^{\prime}$. We first show that having $A \circcirc V_1$ in $\g$ already leads to a contradiction. 
         {Since $p_1= \langle W_{1_1} = A, B, \dots, W_{1_m} =V_1 \rangle$ is a possibly directed unshielded path from $A$ to $V_1$ and $\langle A, V_1 \rangle$ is in $\g$, it must be that $m \geq 4$ i.e., $p_1$ must contain at least four nodes.}
         Moreover, if $A \circcirc V_1$ is in $\g$, then Lemma \ref{lemma:zhang-pdpath-edge-not-into} implies that $p_1$ must be a circle path in $\g$. Together, $p_1$ and $A \circcirc V_1$ contradict Lemma \ref{lemma:zhang-b8-circ-path}. 
         Therefore, $A \circcirc V_1$ is not in $\g$, and so $A \circcirc V_1$ is also not in $\g^{\prime}$.
        
        Note that by Definition \ref{def:unbridged-o} and Theorem \ref{thm:wang2024}, it is technically possible to have $A \equiv C_1$, but in this case we cannot have $A \circarrow V_1$, or $A \to V_1$, by \ref{case2:equivalenceR13}. Lastly, consider that $A \neq C_1$ and that $A \circarrow V_1$ or $A \to V_1$ is in $\g^{\prime}$ and also that the corresponding edge in $\g$ is $A \circarrow V_1$ or $A \to V_1$ .

        Note that since $V_1 \circbullet C_1 \bulletarrow A$ is in $\g^{\prime}$, having $A \to V_1$ in $\g^{\prime}$ (or in $\g$) would contradict that orientations in $\g^{\prime}$ are clsoed under \ref{R2}. Therefore, the only remaining option is that $A \circarrow V_1$ is in $\g^{\prime}$ and $\g$.

        Then since $A \circarrow V_1 \circcirc V_2$ is in $\g$, we have that $A \bulletarrow V_2$ is also in $\g$ by Lemma \ref{lemma:p1zhang}. 
        Note also that we know that $C_1 \notin \Adj(V_2, \g^{\prime})$ and that $C_1 \bulletarrow A$ is in $\g^{\prime}$. Therefore, either $C_1 \bulletarrow A \leftrightarrow V_2$ is already an unshielded collider in $\g$, or $C_1 \bulletarrow A \to V_2$ is in $\g^{\prime}$.
       { In the former case, we now obtain a contradiction with $A \circarrow V_1$ being in $\g^{\prime}$ and $\g^{\prime}$  having the same minimal collider paths as $\g$. This is because $\g^{\prime}$ also contains $C_1 \bulletcirc V_1 \circcirc V_2$ and unshielded collider $C_1 \bulletarrow A \leftrightarrow V_2$ and $C_1 \notin \Adj(V_2, \g^{\prime})$. Since $\g$ would need to contain the same unshielded collider and since orientations in $\g$ are closed under \ref{R3} (Corollary \ref{cor:old-rules-sound}), we obtain the contradiction.}
       In the latter case, we obtain a contradiction with $V_1 \circcirc V_2$ being in $\g^{\prime}$ and orientations in $\g^{\prime}$ being closed under \ref{R11}, since now we have that $C_1 \bulletarrow A \to V_2$, $C_1 \bulletcirc V_1 \circcirc V_2$, $A \circarrow V_1$, and $C_1 \notin \Adj(V_2, \g^{\prime})$. 
        This concludes deriving the contradiction in the case $A \in \Adj(V_1, \g^{\prime})$.
       
   \item[\ref{case4:equivalenceR13}]   We will prove that $A \circbullet C_1$ must be in $\g$ and a symmetric argument can be used to show $A \circbullet C_2$ is in $\g$.
   Note that $A \arrowbullet C_1$ is in $\g^{\prime}$, so that $A \circbullet C_1$ or $A \arrowbullet C_1$ must be in $\g$. We will assume that $A \arrowbullet C_1$ is in $\g^{\prime}$ and show that leads to a contradiction.

   Since $A \circbullet B$ is in $\g^{\prime}$, and $p_1 = \langle W_{1_1} = A, B, \dots, V_1 = W_{1_m} \rangle$, $m >2$ is an unshielded possibly directed path in $\g^{\prime}$, we also have that $A \circbullet B$ is in $\g$ and that the corresponding path in $\g$, $\langle W_{1_1} = A, B, \dots, V_1 =W_{1_m}\rangle$ is also unshielded and possibly directed. Note that $\langle W_{1_1} = A, B, \dots, V_1 =W_{1_m}\rangle$ is either a circle path in $\g^{\prime}$, or there is an arrowhead at some $W_{1_l}, l\ge 2$, on edge $\langle W_{1_{l-1}}, W_{1_{l}} \rangle$ after which $W_{1_l} \to \dots \to W_{1_m},$ by \ref{R1}. 
   Therefore, by iterative application of Lemma \ref{lemma:p1zhang}, we have that $C_1 \bulletarrow B$ is in $\g$, and also that $C_1 \bulletarrow W_{1_j}$ is also in $\g^{\prime}$, for every $1 \le j \le l$ (in the case where $\langle W_{1_1} = A, B, \dots, V_1 =W_{1_m}\rangle$ is a circle path $C_1 \bulletarrow W_{1_j}$ for all $1 \le j \le m$). 

   Then $\langle W_{1_1} = A, B, \dots, V_1 =W_{1_m}\rangle$ cannot be a circle path otherwise $C_1 \bulletarrow V_1$ would be in $\g$ and contradict case \ref{case1:equivalenceR13} above. Hence, there must be an $l < m$ such that $C_1 \bulletarrow W_{1_l} \to W_{1_{l+1}} \to \dots \to V_1$ is in $\g$ and also in $\g^{\prime}$. However, this path together with $C_1 \bulletcirc V_1$ in $\g^{\prime}$ now { contradicts Corollary \ref{cor:possible-cycle}}. 

 \item[\ref{case5:equivalenceR13}] By case \ref{case4:equivalenceR13} we have that $C_1 \bulletcirc A \circbullet C_2$ is in $\g$ and by $C_1 \in \mathcal{F}_1$, $C_2 \in \mathcal{F}_2$, we know that $C_1 \bulletarrow A \arrowbullet C_2$ is in $\g^{\prime}$. Since $\g^{\prime}$ does not contain any new unshielded colliders compared to $\g$, it must be that $C_1 \in \Adj(C_2, \g).$  We also know that  $V_1 \circbullet C_1 \bulletbullet C_2 \bulletcirc V_k$ is in $\g^{\prime}$ (and $\g$). Below, we first show by contradiction that $V_1 \circcirc C_1 \bulletbullet C_2 \circcirc V_k$ is not in $\g$. { We subsequently argue depending on the size of $k$ that this implies that $V_1 \circarrow C_1 \leftrightarrow C_2 \arrowcirc V_k$ must be in $\g$ by invoking Lemma \ref{lemma:r9-notneeded}.}

 Suppose for a contradiction that $V_1 \circcirc C_1 \bulletbullet C_2 \circcirc V_k$ is in $\g$. The presence of the path $C_1 \circcirc V_1 \circcirc \dots \circcirc V_k \circcirc C_2$ in $\g$ implies, by {Lemma \ref{lemma:zhang-circ-path}}, that edge $\langle C_1, C_2 \rangle$ must be of the form $C_1 \circcirc C_2$ in $\g$. However, note that path $C_1 \circcirc V_1 \circcirc \dots \circcirc V_k \circcirc C_2$ is unshielded in $\g$ and that it contains at least four nodes. The presence of edge $C_1 \circcirc C_2$ then contradicts Lemma \ref{lemma:zhang-b8-circ-path}.
Hence, we conclude that $V_1 \circarrow C_1$, or $V_k \circarrow C_2$ is in $\g$.

Suppose without loss of generality that $V_1 \circarrow C_1$ is in $\g$.
Suppose additionally, that $V_k \in \Adj(C_1, \g)$.
{Note that, here, we consider $k > 2$. Otherwise, we have a contradiction with $C_1 \notin \Adj(V_2, \g)$.}
Let $V_t$, $t \in \{2, \dots, k-1 \}$ be a node with the largest index $t$ such that $V_t \notin \Adj(C_1, \g)$. This node surely exists, since $V_2 \notin \Adj(C_1, \g)$. Then consider the edge $\langle V_{t+1}, C_1 \rangle$. Since $V_t \circcirc V_{t+1}$ and $\langle V_{t+1}, C_1 \rangle$ is in $\g$ and since $V_t \notin \Adj(C_1, \g)$ it follows by Lemma \ref{lemma:p1zhang} that $\langle V_{t+1}, C_1 \rangle$ is of one of these forms in $\g$, $V_{t+1} \to C_1$ or $V_{t+1} \circarrow C_1$. { In either case, we have that by concatenating $V_1 \circcirc V_2 \circcirc \dots \circcirc V_{t} \circcirc V_{t+1}$ and $\langle V_{t+1}, C_1 \rangle$ we obtain an unshielded possibly directed path from $V_1$ to $C_1$ in $\g$ (Lemma \ref{lemma:maathuis-7_5}) such that $V_2 \notin \Adj(C_1, \g)$. Since additionally, we have that $V_1 \circarrow C_1$ is in $\g$, we obtain a contradiction with Lemma \ref{lemma:r9-notneeded}.}

Otherwise, $C_1 \notin \Adj(V_k, \g)$ (in this case it is possible that $k =2$), in which case $\langle C_1 , C_2, V_k \rangle$ is of one of the following forms in $\g$ (Lemma \ref{lemma:p1zhang}), $C_1 \bulletarrow C_2 \arrowcirc V_k$,  $C_1 \leftarrow C_2 \bulletcirc V_k$, $C_1 \bulletcirc C_2 \circcirc V_k$.{ In the latter two cases, similarly to above, we obtain a contradiction with Lemma \ref{lemma:r9-notneeded}, because $V_2 \notin \Adj(C_1, \g)$ and by concatenating $V_1 \circcirc \dots \circcirc V_k$ and $\langle V_k , C_2, C_1 \rangle$ we obtain an unshielded possibly directed path from $V_1$ to $C_1$ (Lemma \ref{lemma:maathuis-7_5}) in addition to edge $V_1 \circarrow C_1$. }

We now have that $V_1 \circarrow C_1 \bulletarrow C_2 \arrowcirc V_k$ is in $\g$, so in order to show that $V_1 \circarrow C_1 \leftrightarrow C_2 \arrowcirc V_k$ is in $\g$ we only need to rule out that $V_1 \circarrow C_1 \circarrow C_2 \arrowcirc V_k$ and $V_1 \circarrow C_1 \to C_2 \arrowcirc V_k$ is in $\g$.

Suppose for a contradiction that either $V_1 \circarrow C_1 \circarrow C_2 \arrowcirc V_k$ and $V_1 \circarrow C_1 \to C_2 \arrowcirc V_k$ is in $\g$. In the former case we have that $V_1 \in \Adj(C_2, \g)$ (Lemma \ref{lemma:p1zhang}), while in the latter case its possible that $V_1 \notin \Adj(C_2, \g)$. In either case let $V_s$, $s \in \{1, \dots k-1 \}$ be the node with the smallest index that is not adjacent to $C_2$ in $\g$. Such a node surely exists since $V_{k-1} \notin \Adj(C_2, \g)$.
Then since $V_{s-1} \circcirc V_{s}$  and $\langle V_{s-1} , C_2 \rangle$ are in $\g$ and $V_s \notin \Adj(C_2, \g)$, Lemma \ref{lemma:p1zhang} lets us conclude that $V_{s-1} \circbullet C_2$ or $V_{s-1} \to C_2$ is in $\g$. { Therefore, similarly to above we now have that $V_k \circarrow C_2$ is in $\g$ and $V_{k-1} \notin \Adj(C_2, \g)$ and concatenating $V_k \circcirc \dots \circcirc V_{s} \circcirc V_{s-1}$ and $\langle V_{s-1}, C_2 \rangle$ ({Lemma \ref{lemma:maathuis-7_5}}) yields an unshielded possibly directed path from $V_k$ to $C_2$ in $\g$ which contradicts Lemma \ref{lemma:r9-notneeded}. }

\item[\ref{case6:equivalenceR13}] We have shown in case \ref{case5:equivalenceR13} that $V_1 \circarrow C_1$ and $V_k \circarrow C_2$ are in $\g$ and thus, these are also in $\g^{\prime}$ by \ref{case1:equivalenceR13}. Similarly, we have shown in case \ref{case4:equivalenceR13} that $C_1 \bulletcirc A$ and $C_2 \bulletcirc A$ are in $\g$, and by case  \ref{case3:equivalenceR13} we have that $A \notin \Adj(V_1, \g)$
 and $A \notin \Adj(V_k, \g)$. Therefore, Lemma \ref{lemma:p1zhang} leads us to conclude that $A \circarrow C_1 \arrowcirc V_1$ is an unshielded collider in $\g$ and so is  $A \circarrow C_1 \arrowcirc V_k$.
 \end{enumerate} 
\end{proofof}

\subsubsection{On Possibly Directed and Unshielded paths}

This subsection contains a few results on possibly directed paths, unshielded paths, and path concatenations. The main result of this section are Lemmas \ref{lemma:unshielded-poss-dir} and \ref{lem:concatenation}, and the relevant proof structure is given in Figure \ref{fig:unshielded-poss-dir}.

\begin{figure}[!t]
    \centering
      \begin{tikzpicture}[[->,>=stealth', auto,node distance=0.8cm,scale=.8,transform shape,font = {\large\sffamily}]
  \tikzstyle{state}=[inner sep=2pt, minimum size=12pt]

    \node[state] (E5) at (4, 0) {\textbf{Lemma \ref{lemma:unshielded-poss-dir}}};
    \node[state] (E6) at (0, 0) {Lemma \ref{lem:concatenation}};
    \node[state] (E7) at (-4, 0) {Lemma \ref{lemma:concat-edge}};
    \node[state] (E8) at (-4, 2) {Lemma \ref{lemma:not-allowed-paths}};
    \node[state] (E9) at (4, 2) {Lemma \ref{lemma:new-circle-path}};
    \node[state] (E10) at (-8, 1.3) {Corollary \ref{cor:possible-cycle}};
    \node[state] (E12) at (-8, 0) {Lemma \ref{lemma:possible-cycle}};
    
    \node[state] (E11) at (0, 3) {Lemma \ref{lemma:r9-notneeded}};

    \draw[->,arrows= {-Latex[width=5pt, length=5pt]}] (E6) edge (E5);
    \draw[->,arrows= {-Latex[width=5pt, length=5pt]}] (E10) edge (E7);
    
    \draw[->,arrows= {-Latex[width=5pt, length=5pt]}] (E7) edge (E6);

   \draw[->,arrows= {-Latex[width=5pt, length=5pt]}] (E8) edge (E7);
    \draw[->,arrows= {-Latex[width=5pt, length=5pt]}] (E10) edge (E8);
    \draw[->,arrows= {-Latex[width=5pt, length=5pt]}] (E9) edge (E8);
    \draw[->,arrows= {-Latex[width=5pt, length=5pt]}] (E9) edge (E5);

    \draw[->,arrows= {-Latex[width=5pt, length=5pt]}] (E11) edge (E9);
    \draw[->,arrows= {-Latex[width=5pt, length=5pt]}] (E11) edge (E8);
    \draw[->,arrows= {-Latex[width=5pt, length=5pt]}] (E12) edge (E10);
\end{tikzpicture}
   \caption{Proof structure of Lemma \ref{lemma:unshielded-poss-dir}.}
    \label{fig:unshielded-poss-dir}
\end{figure}

\begin{lemma}[Possibly Directed Status of an Unshielded Path]\label{lemma:unshielded-poss-dir}
  Let $\g = (\mathbf{V,E})$ be an essential ancestral graph and   $\g^{\prime} = (\mathbf{V,E'})$ be an ancestral partial mixed graph  such that $\g$ and $\g^{\prime}$ have the same skeleton, the same set of minimal collider paths, and every invariant edge mark in $\g$ is identical in $\g^{\prime}$. Suppose furthermore that edge orientations in $\g^{\prime}$ are closed under  
  \ref{R1}, \ref{R2}, \ref{R8}, \ref{R11}, \ref{R12}.  
  If $p = \langle P_1, \dots , P_k \rangle $, $k \ge 1$ is an unshielded path in $\g^{\prime}$ such that there is no edge of the form $P_i \arrowbullet P_{i+1}$, then  $p$ is a possibly directed path in $\g^{\prime}$. 
\end{lemma}

\begin{proofof}[Lemma \ref{lemma:unshielded-poss-dir}]
    Since $p$ is an unshielded path and orientations in $\g^{\prime}$ are closed under \ref{R1} it follows that if there is an arrowhead at $P_{i}$ on any edge $P_i \bulletarrow P_{i+1}$, then $p(P_{i+1}, P_k)$. must be a directed path. Furthermore, Lemma \ref{lemma:new-circle-path} implies  $p(P_1, P_{i+1})$ is an unshielded possibly directed path. Now, lastly, we have by Lemma \ref{lem:concatenation} that $p(P_1,P_{i+1}) \oplus p(P_{i+1}, P_k)$ is an unshielded possibly directed path. 
\end{proofof}

\begin{lemma}[Possibly Directed Path Concatenation]\label{lem:concatenation}
Let $\g = (\mathbf{V,E})$ be an essential ancestral graph and   $\g^{\prime} = (\mathbf{V,E'})$ be an ancestral partial mixed graph  such that $\g$ and $\g^{\prime}$ have the same skeleton, the same set of minimal collider paths, and every invariant edge mark in $\g$ is identical in $\g^{\prime}$. Suppose furthermore that edge orientations in $\g^{\prime}$ are closed under  
  \ref{R1}, \ref{R2}, \ref{R8}, \ref{R11}, \ref{R12}.   If $p = \langle P_1, \dots , P_k \rangle $, $k \ge 1$ is an unshielded possibly directed path in $\g^{\prime}$ and $q = \langle P_k , \dots, P_{k+r} \rangle, r \ge 1$ is a directed path in $\g^{\prime}$, then  $p \oplus q$ is a possibly directed path in $\g^{\prime}$. 
\end{lemma}

\begin{proofof}[Lemma \ref{lem:concatenation}] 
Follows by iterative applications of Lemma \ref{lemma:concat-edge} and Corollary \ref{cor:unshielded-gen}.
\end{proofof}

\begin{lemma}[Towards Possibly Directed Path Concatenation]\label{lemma:concat-edge}
Let $\g = (\mathbf{V,E})$ be an essential ancestral graph and   $\g^{\prime} = (\mathbf{V,E'})$ be an ancestral partial mixed graph  such that $\g$ and $\g^{\prime}$ have the same skeleton, the same set of minimal collider paths, and every invariant edge mark in $\g$ is identical in $\g^{\prime}$. Suppose furthermore that edge orientations in $\g^{\prime}$ are closed under  
  \ref{R1}, \ref{R2},  \ref{R8},  \ref{R11}, \ref{R12}.  If $p = \langle P_1, \dots , P_k \rangle $, $k \ge 1$ is an unshielded possibly directed path in $\g^{\prime}$ and if $P_k \to P_{k+1}$ is in $\g^{\prime}$, then $p \oplus \langle P_k , P_{k+1} \rangle$ is a possibly directed path in $\g^{\prime}$. 
\end{lemma}

\begin{proofof}[Lemma \ref{lemma:concat-edge}]
It is enough to show that $P_{i} \arrowbullet P_{k+1}$, $i \in \{1, \dots, k-1 \rangle$ is not in $\g^{\prime}$. 

This claim holds for $i = k-1$ since otherwise, $P_{k-1} \arrowbullet P_{k+1} \leftarrow P_k$ and the fact that $\g^{\prime}$ is ancestral and that  orientations are closed under \ref{R2} would imply that $P_{k-1} \arrowbullet P_k$ is on $p$. And this fact would in turn contradict that $p$ is possibly directed from $P_1$ to $P_k$. 

Hence, suppose for a contradiction that $P_{i} \arrowbullet P_{k+1}$ is in $\g^{\prime}$ for some $i \in \{1, \dots , k-2\}$, and let $P_j$ be the closest such node to $P_k$ on $p$. Furthermore, note that $P_j \arrowcirc P_{k+1}$ is not in $\g^{\prime}$, as $P_j \arrowcirc P_{k+1} \leftarrow P_k$ and the fact that orientations in $\g^{\prime}$ are closed under \ref{R1} implies that $P_k$ and $P_j$ are adjacent. Further, \ref{R2} implies that $P_j \arrowbullet P_k$ is in $\g^{\prime}$ thus contradicting the assumption that $p$ is a possibly directed path in $\g^{\prime}$. 

Then $P_j \leftarrow P_{k+1}$ or $P_{j} \leftrightarrow P_{k+1}$ is in $\g^{\prime}$.
Since $\g^{\prime}$ is an ancestral graph, we can also conclude that $p(P_j, P_k)$ is not a directed path from $P_j$ to $P_k$. Furthermore, since orientations in $\g^{\prime}$ are closed under \ref{R1} and since $p$ is an unshielded path, this also implies that $P_1 \circcirc P_2 \circcirc \dots \circcirc P_j \circbullet P_{j+1}$ is in $\g^{\prime}$ by Corollary \ref{cor:b2}.

Now, let $P_l$ be the closest node to $P_1$ on $p$ such that $P_l \to \dots \to P_k$ is in $\g^{\prime}$ and if no such node is in $p$, then let $P_l \equiv P_k$. 
Consider paths $P_j \circcirc \dots \circcirc P_l$ and $q= p(P_l, P_k) \oplus \langle P_k, P_{k+1}, P_j \rangle$, where by construction, $q$ is of one of the following forms $P_l \to \dots \to P_{k+1} \to P_{j}$, or $P_l \to \dots \to P_{k+1} \leftrightarrow P_{j}$. If $l >j+1,$ these two paths contradict Lemma \ref{lemma:not-allowed-paths}. If $l =j+1,$ then since orientation in $\g^{\prime}$ are closed under \ref{R1}, $P_{k+1} \in \Adj(P_{l}, \g^{\prime})$ and furthermore, { Corollary \ref{cor:possible-cycle} implies that $P_{l} \to P_{k+1}$ is in $\g^{\prime}$.} But closure under \ref{R2} implies that $P_j \arrowbullet P_l$ contradicting our assumption that $p$ is a possibly directed path.
\end{proofof}

\begin{lemma} \label{lemma:not-allowed-paths}
Let $\g = (\mathbf{V,E})$ be an essential ancestral graph and   $\g^{\prime} = (\mathbf{V,E'})$ be an ancestral partial mixed graph  such that $\g$ and $\g^{\prime}$ have the same skeleton, the same set of minimal collider paths, and every invariant edge mark in $\g$ is identical in $\g^{\prime}$. Suppose furthermore that edge orientations in $\g^{\prime}$ are closed under   \ref{R1}, \ref{R2},  \ref{R8},  \ref{R11}, \ref{R12}.  Then there are no two paths $p =  \langle V_1, \dots, V_i \rangle, i>1$ and $q = \langle V_i, \dots, V_k, V_1 \rangle $, $k>i$ in $\g^{\prime}$ such that $p$ and $q$ have the same endpoint nodes and are of the following forms:
\begin{enumerate}[label = (\arabic*)]
    \item \label{bla1} $p$ is an unshielded path  of the form $V_1 \circcirc V_2  \dots \circcirc V_{i-1} \circbullet V_i$, and 
    \item \label{bla} $q$ is one of the following forms 
    \begin{enumerate}[label = (\roman*)]
        \item \label{case:0} $V_i \to \dots \to V_k \to V_1$, or 
        \item \label{case:1} $V_i \to \dots \to V_{k} \bulletarrow V_1$, or 
        \item \label{case:2} $V_i \bulletarrow V_{i+1} \to \dots \to V_{k} \to V_1$, or 
        \item \label{case:3} $V_i \to \dots \to V_j \bulletarrow V_{j+1} \to \dots \to V_{k} \to V_1$, $k > j >i$. 
    \end{enumerate}
\end{enumerate}
\end{lemma}

\begin{proofof}[Lemma \ref{lemma:not-allowed-paths}]
Suppose for a contradiction that there are two paths with the same endpoints that are of the forms as discussed in \ref{bla1} and \ref{bla} in $\g^{\prime}$. Choose among all such pairs in $\g^{\prime}$ the paths $p$ and $q$ with endpoints $V_1$ and $V_i$ such that for any other pair of paths $p'$ and $q'$ with endpoints $V_1^{'}$ and $V_{i}^{'}$ and such that $p'$ is of the form \ref{bla1}, and $q'$ is of the form \ref{bla}, the following hold:
either $|p| < |p'|$ and $|q| \le |q'|$, or $|p| = |p'|$ and $|q| \le |q'|.$

By choice of $p$ and $q,$ there cannot be any subsequence of $q$ that forms a path in $\g^{\prime}$, that is of one of the forms: \ref{bla}\ref{case:0} - \ref{bla}\ref{case:3}. In conjunction with { Corollary \ref{cor:possible-cycle}, we then have that there cannot be any edge between any two non-consecutive nodes on $q$.} Hence, $q$ is an unshielded path. 
This further implies that $V_i \notin \Adj(V_1, \g^{\prime}),$ and hence, $i > 2$ on $p$.

Next, consider path $p$. By assumption $p$ is an unshielded path and above we concluded that $|p|>1$. Additionally, by Lemma \ref{lemma:new-circle-path}, there is no edge of the form $V_l \arrowbullet V_r$, $1 \le l < r \le i$ in $\g^{\prime}$. By the same reasoning, there is also no edge of the form $V_l \bulletarrow V_r$, for $1 \le l < r \le i-1$. Furthermore, by choice of $p$ and $q$ there also cannot be an edge $V_l \to V_i$, or $V_l \circarrow V_i$, $1 \le l < i$ in $\g^{\prime}$. Lastly, by choice of $p$ there also cannot be an edge of the form  $V_l \circcirc V_r$, $1 \le l < r \le i$ in $\g^{\prime}$. Hence, not only is $p$ unshielded, but similarly to $q$, there is no edge between any two non-consecutive nodes on $p.$

Revisiting the fact that $q$ is unshielded, together with the assumption that orientations in $\g^{\prime}$ are closed under \ref{R1}, the $\bulletarrow$ edge on $q$ must be either $\to$, or (if $q$ starts with $\circarrow$ as in case \ref{bla}\ref{case:2}, then $p$ must end with $\circcirc$ and we just redefine $p$ to include the $\circarrow$ edge). 
We now break the rest of the proof up into cases depending on the form of $q$.

\begin{enumerate}[label=(\roman*), leftmargin = 1.5cm]
\item[\ref{case:0}] 
Since $V_k \to V_1 \circcirc V_2$ is in $\g^{\prime}$, and since orientations in $\g^{\prime}$ are closed under \ref{R1}, $V_k \in \Adj(V_2, \g^{\prime})$. The edge $\langle V_k, V_2 \rangle$ cannot be of the form $V_k \bulletarrow  V_2$ as that contradicts the choice of $q$ (this would fall under case \ref{case:1}). It also cannot be of the form   $V_k \arrowbullet V_2$ as that contradicts that orientations are closed under \ref{R2}.  

Hence,  $V_k \circcirc V_2$ must be in $\g^{\prime}$.
Since we now have that $V_{k-1} \to V_k \circcirc V_2$ is in $\g^{\prime}$, and since orientations in $\g^{\prime}$ are closed under \ref{R1},  $\langle V_{k-1} , V_2 \rangle$ is in $\g^{\prime}$. By the same reasoning as above, we now have that $V_{k-1}  \circcirc V_{2}$ must be in $\g^{\prime}$. However, now $\g^{\prime}$ contains the unshielded triples $V_{k-1} \to V_k \to V_1$  and $V_{k-1} \circcirc V_2 \circcirc V_1$ and edge $V_{k} \circcirc V_2$ which contradicts that orientations in $\g^{\prime}$ are closed under  \ref{R11}. 

\item[\ref{case:1}] 
Since we already discussed the case when $q$ is a directed path in \ref{case:0}, we will assume that $V_k \leftrightarrow V_1$ is in $\g^{\prime}$.  Furthermore, since orientations in $\g^{\prime}$ are   closed under \ref{R12}, we know that $|q|>2$, that is, $k > i+1$.

Since $V_k \leftrightarrow V_1 \circcirc V_2$ is in $\g^{\prime}$, and since orientations in $\g^{\prime}$ are closed under \ref{R1}, $V_k \in \Adj(V_2, \g^{\prime})$. Note, furthermore, that the edge $\langle V_k, V_2 \rangle$ cannot be of the form $V_k \bulletarrow  V_2$ as that contradicts the choice of $q$. Additionally,  $V_k \leftarrow V_2$ contradicts that orientations are closed under \ref{R2}, since in this case $V_2 \to V_k \bulletarrow V_1$ and $V_1 \circcirc V_2$ would be in $\g^{\prime}$. 

Thus, $V_k \arrowcirc V_2$ or $V_k \circcirc V_2$ are in $\g^{\prime}$.
Let us first consider the case where $V_k \circcirc V_2$ in $\g^{\prime}$.
Now have that $V_{k-1} \to V_k \circcirc V_2$ is in $\g^{\prime}$, so that  since orientations in $\g^{\prime}$ are closed under \ref{R1},  $\langle V_{k-1} , V_2 \rangle$ is in $\g^{\prime}$. Note that $V_{k-1} \arrowbullet V_{2}$ contradicts that orientations are closed under \ref{R2}, and $V_{k-1} \bulletarrow V_{2}$ contradicts the choice of $q$. Hence $V_{k-1} \circcirc V_2$ is in $\g^{\prime}$. { But now, consider the unshielded collider $V_{k-1} \to V_k \leftrightarrow V_1$, $V_{k-1} \circcirc V_2 \circcirc V_1$ and $V_k \circcirc V_2$. By assumption $V_{k-1} \bulletarrow V_k \arrowbullet V_1$, $V_{k-1} \circcirc V_2 \circcirc V_1$ and $V_k \circcirc V_2$ are also in $\g$, which contradicts that orientations in $\g$ are closed under \ref{R3} (Corollary \ref{cor:old-rules-sound}).}

Hence, it is left to consider the case when $V_k \arrowcirc V_2$ is in $\g^{\prime}$.
Consider that $p(V_2, V_i)$ is a possibly directed path in $\g^{\prime}$ and that $q(V_i, V_k)$ is a directed path in $\g^{\prime}$ and moreover, that there cannot be any edge $V_l \arrowbullet V_r$, $2 \le l < r \le k$ in $\g^{\prime}$ as that contradicts either that $\g^{\prime}$ is ancestral, or the choice of $p$ and $q.$ Hence $t = p(V_2, V_i) \oplus q(V_i, V_k)$ is a possibly directed path in $\g^{\prime}$.

Note that if there is any edge $\langle V_l, V_r \rangle$, $2 \le l < i < r \le k$ in $\g^{\prime}$, by choice of $p$ and $q$, this edge cannot be of the form $V_l \to V_r$, or $V_l \arrowbullet V_r$.
Hence, any such edge must be of the form $V_l \circbullet V_r$.

Furthermore, consider any edge $\langle V_l, V_k \rangle$ $2 \le l < i$. Then since $V_1 \leftrightarrow V_k$ is in $\g^{\prime}$ and $V_1 \notin \Adj(V_l, \g^{\prime})$ and since orientations in $\g^{\prime}$ are closed under \ref{R1}, we can conclude that the $\bullet$ on edge $V_l \circbullet V_k$ must be an arrowhead, that is $V_l \circarrow V_k$.  Now, let $V_s, s \in \{2, \dots, i-1\}$ be the closest node to $V_i$ on $t$ such that $V_s \circarrow V_k$ is in $\g^{\prime}$.

Consider again that any edge  $\langle V_l, V_r \rangle$, $2 \le l < i < r \le k$ in $\g^{\prime}$ must be of the form $V_l \circbullet V_r$.
Since orientations in $\g^{\prime}$ are closed under \ref{R12} and $V_{k-1} \to V_{k} \leftrightarrow V_1$ is in $\g^{\prime}$ and $V_{1} \notin \Adj(V_{k-1}, \g^{\prime})$  this implies that we cannot have an edge $\langle V_2, V_{k-1} \rangle $ in $\g^{\prime}$. Moreover, if $i >3$, then  since $p(V_1, V_3)$ is an unshielded path of the form $V_1 \circcirc V_2 \circcirc V_3$ and since $V_1, V_2 \notin \Adj(V_{k-1}, \g^{\prime})$, we also cannot have an edge $\langle V_3, V_{k-1} \rangle$ in $\g^{\prime}$. We can apply the same reasoning to conclude that $V_2, \dots , V_{i-1} \notin \Adj(V_{k-1}, \g^{\prime})$.
Hence, also $V_s, V_{s+1}, \dots, V_{i-1} \notin \Adj(V_{k-1}, \g^{\prime})$.

Now we have that $V_s \circarrow V_k$ is in $\g^{\prime}$, $V_s, \dots V_{i-1} \notin \Adj(V_{k-1}, \g^{\prime})$, and $V_{s+1}, \dots, V_i \notin \Adj(V_k, \g^{\prime})$. Additionally, $t(V_s, V_k)$ is of the form, $V_s \circcirc \dots \circcirc V_{i-1} \circbullet V_i \to \dots \to V_{k-1} \to V_k$ and is a possibly directed path. Now we can  choose nodes $V_a$ and $V_b$ such that $V_a$ is on $t(V_s, V_i)$, $ a \neq s$, and $V_b$ is on $t(V_i, V_k)$, $b \notin \{k-1,k\}$, and $V_a \circbullet V_b$ is in $\g^{\prime}$ ($a = i - 1$ and $b = i$ is a valid choice, so such pairs $a, b$ exist). Moreover, we can choose $V_a, V_b$ such that   $w =t(V_s, V_a) \oplus \langle V_a, V_b \rangle \oplus t(V_b, V_k)$ is an unshielded possibly directed path in $\g^{\prime}$. { Then note that $V_s \circarrow V_k$ is also in $\g^{\prime}$ and that $V_{s+1} \notin \Adj(V_k, \g^{\prime})$ by choice of $s$, which contradicts Lemma \ref{lemma:r9-notneeded}.}

\item[\ref{case:2}, \ref{case:3}] Since $V_k \to V_1 \circcirc V_2$ is in $\g^{\prime}$ and since orientations in $\g^{\prime}$ are closed under \ref{R1}, $V_k \in \Adj(V_2, \g^{\prime})$. As in the proof of case \ref{case:0}, the edge $\langle V_k, V_2 \rangle$  must be of the form $V_k \circbullet V_2$ is in $\g^{\prime}$. Now, $V_{k-1} \bulletarrow V_k \circbullet V_2$ and orientations in $\g^{\prime}$ being closed under \ref{R1} imply that edge $\langle V_{k-1}, V_2 \rangle$ is in $\g^{\prime}$. Furthermore, as $q$ is unshielded, we know that $V_{k-1} \notin \Adj(V_1, \g^{\prime})$. Putting it all together, we now have that $V_{2} \bulletbullet V_{k-1} \bulletarrow V_k \to V_1$, $V_1 \circcirc V_2 \bulletcirc V_k $, and $V_{k-1} \notin \Adj(V_1, \g^{\prime})$,  which contradicts that orientations in $\g^{\prime}$ are closed under \ref{R11}.
\end{enumerate}
\end{proofof}

\begin{lemma}
\label{lemma:new-circle-path}
Let $\g = (\mathbf{V,E})$ be an essential ancestral graph and   $\g^{\prime} = (\mathbf{V,E'})$ be an ancestral partial mixed graph  such that $\g$ and $\g^{\prime}$ have the same skeleton, the same set of minimal collider paths, and every invariant edge mark in $\g$ is identical in $\g^{\prime}$. Suppose furthermore that edge orientations in $\g^{\prime}$ are closed under  \ref{R2},  \ref{R12}.
  Suppose furthermore that there is an unshielded path $q = \langle V_1, V_2, \dots, V_k \rangle, k\ge3$ in $\g^{\prime}$ of the form $V_1 \circcirc V_2 \circcirc \dots V_{k-1} \circbullet V_k$. Then there is no edge  $V_1 \arrowbullet V_k$ in $\g^{\prime}$.
\end{lemma}

\begin{proofof}[Lemma \ref{lemma:new-circle-path}]
If $k = 3$, then since $q$ is unshielded, $V_1 \notin \Adj(V_3, \g^{\prime})$. For the rest of the proof, suppose that $k>3$ and let $q^{*}$ be the path in $\g$ that corresponds to $q$ in $\g^{\prime}$. Additionally, suppose for a contradiction that $V_1 \arrowbullet V_k$ is in $\g^{\prime}$.

Consider the case where  $V_{k-1} \circcirc V_k$ is in $\g$. 
 Since  $q^{*}$ is of the form $V_1 \circcirc \dots V_{k-1} \circcirc V_{k} $, $k > 3$ in $\g$, the edge $\langle V_1, V_k \rangle$ is of the form $V_1 \circcirc V_k$ in $\g$ by Lemma \ref{lemma:zhang-circ-path}. But now this contradicts Lemma \ref{lemma:zhang-b8-circ-path} in $\g$. 
 Since $\g$ and $\g^{\prime}$ have the same skeleton we reach a contradiction. 

For the rest of the proof, we consider the case where $V_{k-1} \circarrow V_k$ is in $\g$, and therefore also in $\g^{\prime}$. By Lemma \ref{lemma:maathuis-7_5}, path $q^{*}$ is an unshielded possibly directed path from $V_1$ to $V_k$ in $\g$. Further, it also ends with an arrowhead pointing to $V_k$. Hence, Lemma \ref{lemma:zhang-pdpath-edge-not-into} implies that edge $\langle V_1, V_k \rangle$ in $\g$ is of the form  $V_1 \circarrow V_k$, or $V_1 \to V_k$. Since $V_1 \arrowbullet V_k$ is supposed to be in $\g^{\prime}$, we now conclude that $V_1 \circarrow V_k$ must be in $\g,$ which further implies that $V_1 \leftrightarrow V_k$ is in $\g^{\prime}$.

{ Furthermore, since $q^{*}$ is an unshielded possibly directed path from $V_1$ to $V_k$ in $\g$  (Lemma \ref{lemma:maathuis-7_5}), and $k>3,$ and since $V_1 \circarrow V_k$ is in $\g$ by Lemma \ref{lemma:r9-notneeded}, it follows that $\langle V_2, V_k \rangle$ is in $\g$. If $k = 4,$ we now reach a contradiction with $q$ being an unshielded path.} Otherwise, $k>4$, and by Lemma \ref{lemma:zhang-pdpath-edge-not-into} edge $\langle V_2, V_k \rangle$  is of the form  $V_2 \circarrow V_k$, or $V_2 \to V_k$ in $\g$. Note that, $V_2 \to V_k$ cannot be in $\g$, otherwise $V_2 \to V_k \leftrightarrow V_1 \circcirc V_2$ is in $\g^{\prime}$ which contradicts that orientations in $\g^{\prime}$ are closed under \ref{R2}. Hence, $V_2 \circarrow V_k$ is in $\g$.

{ Now, similarly to above, consider that $q^{*}(V_2, V_k)$ is an unshielded possibly directed path from $V_2$ to $V_k$ in $\g^{\prime}$, and that $k>4$, and that $V_2 \circarrow V_k$ is in $\g$. Hence, it follows by Lemma \ref{lemma:r9-notneeded} that $\langle V_3, V_k \rangle$ is in $\g$.} If $k = 5,$ we now reach a contradiction with $q$ being an unshielded path. Otherwise, $k>5$, and by Lemma \ref{lemma:zhang-pdpath-edge-not-into} edge $\langle V_3, V_k \rangle$  is of the form  $V_3 \circarrow V_k$, or $V_3 \to V_k$ in $\g$. Note that, $V_3 \to V_k$ cannot be in $\g$, otherwise $V_3 \to V_k \leftrightarrow V_1$ is in $\g^{\prime}$ and $V_1 \circcirc V_2 \circcirc V_3$ is an unshielded path in $\g^{\prime}$ which contradicts  that orientations in $\g^{\prime}$ are closed under \ref{R12}. Hence, $V_3 \circarrow V_k$ is in $\g$.

Note that the above argument can be repeated to conclude that $V_4 \circarrow V_k, \dots, V_{k-2} \circarrow V_k$ are all in $\g$, which ultimately leads to a contradiction with the assumption that $q$ is an unshielded path.
\end{proofof}

 \begin{corollary} \label{cor:possible-cycle} 
Let $\g = (\mathbf{V,E})$ be an essential ancestral graph and   $\g^{\prime} = (\mathbf{V,E'})$ be an ancestral partial mixed graph  such that $\g$ and $\g^{\prime}$ have the same skeleton, the same set of minimal collider paths, and every invariant edge mark in $\g$ is identical in $\g^{\prime}$. Suppose furthermore that edge orientations in $\g^{\prime}$ are closed under \ref{R1}, \ref{R2}, \ref{R8}. Suppose that there is a path $p = \langle P_1 , \dots , P_k \rangle,$ $k\ge 3$ and edge $\langle P_1, P_k \rangle $ in $\g^{\prime}$. Then the following hold
\begin{enumerate}[label = (\roman*)]
    \item\label{case01:poss-cycle} If $p$ is a directed path from $P_1$ to $P_k$, then $P_1 \to P_k$ is in  $\g^{\prime}$.
    \item\label{case02:poss-cycle} If $P_i \to P_{i+1}$ for all $i \in \{1, \dots, k-1\} \setminus \{j\}, 1  \le j \le  k-1$ and  $P_j \bulletarrow P_{j+1}$, then $P_1 \bulletarrow P_k$ is in $\g^{\prime}$.
\end{enumerate}
\end{corollary}

\begin{proofof}[Corollary \ref{cor:possible-cycle}]
  Follows by Lemmas \ref{lemma:possible-cycle} and \ref{lemma:r9-notneeded}.  
\end{proofof}

 \begin{lemma}[Maintaining the Ancestral Property] \label{lemma:possible-cycle} 
Suppose that $\g = (\mb{V,E})$ is an ancestral partial mixed graph with orientations closed under  \ref{R1}, \ref{R2}, \ref{R8}, \ref{R9}. Suppose that there is a path $p = \langle P_1 , \dots , P_k \rangle,$ $k\ge 3$ and edge $\langle P_1, P_k \rangle $ in $\g$. Then the following hold
\begin{enumerate}[label = (\roman*)]
    \item\label{case1:poss-cycle} If $p$ is a directed path from $P_1$ to $P_k$, then $P_1 \to P_k$ is in  $\g$.
    \item\label{case2:poss-cycle} If $P_i \to P_{i+1}$ for all $i \in \{1, \dots, k-1\} \setminus \{j\}, 1  \le j \le  k-1$ and  $P_j \bulletarrow P_{j+1}$, then $P_1 \bulletarrow P_k$ is in $\g$.
\end{enumerate}
\end{lemma}

\begin{proofof}[Lemma \ref{lemma:possible-cycle}]
We prove the two statements by induction on the length of $p$. For the base case of the induction $k = 3$, and we have that both cases \ref{case1:poss-cycle} and \ref{case2:poss-cycle} hold because 
   $\g$ is an ancestral partial mixed graph and because orientations in $\g$ are closed under \ref{R2} and \ref{R8}.
Next, we show the induction step in each of the two cases.  
\begin{enumerate}
    \item[\ref{case1:poss-cycle}] Suppose that claim \ref{case1:poss-cycle} holds for all paths $p'$ of length $n \leq k$, where $k \geq 3$.
Let $p$ be a directed  path with $k+1$ nodes,   $p = \langle P_1, \dots,  P_{k+1}\rangle$ such that the edge $\langle P_1, P_{k+1} \rangle$ is also in $\g$. Let $p' = \langle P_1 = Q_1, \dots,  Q_m = P_{k+1}\rangle, m >1$ be a shortest subsequence of $p$ that forms a directed path from $P_1$ to $P_{k+1}$ in $\g$. If $m \le k$, then $P_1 \to P_{k+1}$ is in $\g$ by the induction assumption.
Otherwise $m >k$, that is $m = k+1$ and $p' \equiv p$, meaning that  $p$ is an unshielded path in $\g$.
Since $\g$ is ancestral, this edge cannot be $P_1 \leftarrow P_{k+1}$ or $P_1 \leftrightarrow P_{k+1}$.
Below we argue by contradiction that edge $\langle P_1, P_{k+1} \rangle$ cannot be $P_1 \bulletcirc P_{k+1}$ or $P_1 \circarrow P_{k+1}$ in $\g.$

Suppose for a contradiction that $P_1 \bulletcirc P_{k+1}$ is in $\g$. Since $P_1 \bulletcirc P_{k+1} \leftarrow P_{k}$ is in $\g$, and since orientations in $\g$ are closed under \ref{R1} it follows that $P_{k} \in \Adj(P_1, \g)$. Hence, by the induction assumption, $P_1 \to P_{k}$ is in $\g$. But this further implies that $P_1 \to P_{k} \to P_{k+1}$  which is a subsequence of $p$ that is a directed path is in $\g$, and that contradicts that $p' \equiv p$.

Next, suppose for a contradiction that $P_1 \circarrow P_{k+1}$ is in $\g$. { Note that since $P_1 \to \dots \to P_k \to P_{k+1}$ is an unshielded directed path in the ancestral graph $\g$ and since edge mark orientations in $\g$ are   closed under \ref{R9}, it follows that   $P_2 \in \Adj(P_{k+1}, \g)$. Since $P_2 \in \Adj(P_{k+1}, \g)$ and $P_2 \to \dots \to P_{k+1}$ is in $\g$, by the induction assumption, $P_2 \to P_{k+1}$ is in $\g$.} But now $P_1 \to P_2 \to P_{k+1}$ is a subsequence of $p$ that is a directed path is in $\g$. This contradicts that $p' \equiv p$. 
\item[\ref{case2:poss-cycle}]  Suppose that claim \ref{case2:poss-cycle} holds for all paths $p'$ of length $n \leq k$, where $k \geq 3$.
 Let $p$ be a  path with $k+1$ nodes,   $p = \langle P_1, \dots,  P_{k+1}\rangle$ such that $P_j \bulletarrow P_{j+1}$, for some  $j \in \{1, \dots, k\}$, but $P_i \to P_{i+1}$ for all $i \in \{1, \dots, k \} \setminus \{j\}$ and also such that the edge $\langle P_1, P_{k+1} \rangle$ is in $\g$. 

Since $\g$ is ancestral, the edge $\langle P_1, P_{k+1} \rangle$ cannot be of the form $P_1 \leftarrow P_{k+1}$. Hence, for the claim \ref{case2:poss-cycle}, it is enough to show that this edge is also not of the form  $P_1 \bulletcirc P_{k+1}$ in $\g$.

 Suppose for a contradiction that $P_1 \bulletcirc P_{k+1}$ is in $\g$. Since $P_1 \bulletcirc P_{k+1} \arrowbullet P_{k}$ is in $\g$ and since orientations in $\g$ are closed under \ref{R1} it follows that $P_{k} \in \Adj(P_1, \g)$. If $p(P_1, P_{k})$ is a directed path, then by \ref{case1:poss-cycle} above, we have that $P_1 \to P_{k}$ is in $\g$. But then $P_1 \to P_{k} \bulletarrow P_{k+1}$ together with $P_1 \bulletcirc P_{k+1}$ contradicts that orientations in $\g$ are closed under \ref{R2}. Otherwise, $p(P_1, P_{k})$ contains either $\circarrow$ or a $\leftrightarrow$ edge, so by the induction step $P_1 \bulletarrow P_{k}$ is in $\g$.
 However, in this case $P_1 \bulletarrow P_{k} \to P_{k+1}$ and $P_1 \bulletcirc P_{k+1}$ are in $\g$, which contradicts that orientations in $\g$ are closed under \ref{R2}.
\end{enumerate}
\end{proofof}

\begin{lemma}\label{lemma:r9-notneeded}
Let $\g = (\mathbf{V,E})$ be an essential ancestral graph and   $\g^{\prime} = (\mathbf{V,E'})$ be an ancestral partial mixed graph  such that $\g$ and $\g^{\prime}$ have the same skeleton, the same set of minimal collider paths, and every invariant edge mark in $\g$ is identical in $\g^{\prime}$.
Suppose furthermore that the edge $A \circarrow C$ is in $\g^{\prime}$ and that there is an unshielded possibly directed  path $p$, from $A$ to $C$, $p= \langle A = P_1, P_2, \dots, P_k = C \rangle$, $k>3$ in $\g^{\prime}$. Then $P_2 \in \Adj(C, \g)$.
\end{lemma}

\begin{proofof}[Lemma \ref{lemma:r9-notneeded}]
    Let $p^{*}$ be the path in $\g$ that corresponds to $p$ in $\g^{\prime}$. 
    Since $\g$ and $\g^{\prime}$ have same skeleton, $p^{*}$ is an unshielded path in $\g$.
Furthermore, since $\g^{\prime}$ has additional edge orientations compared to $\g$, any possibly directed path in $\g^{\prime}$ corresponds to a possibly directed path in $\g.$
Therefore, $p^{*}$ is a possibly directed unshielded path in $\g$. 

Suppose first that $A \circarrow C$ is in $\g$. Then $P_2 \in \Adj(C, \g) \equiv \Adj(C, \g^{\prime})$ because otherwise, orientations in $\g$ are not   closed under \ref{R9}. 
Next, suppose $A \circcirc C$ is in $\g$. Then Lemma \ref{lemma:zhang-pdpath-edge-not-into} and Corollary \ref{cor:b2} together imply that $p^{*}$ is an unshielded path of the form $A \circcirc P_2 \circcirc \dots \circcirc C$  in $\g$. Furthermore, since by assumption $|p^{*}| \ge 3$  and since $A \circcirc C$ we obtain a contradiction with Lemma Lemma \ref{lemma:zhang-b8-circ-path}.
\end{proofof}

\section{Supplement to Section \ref{sec:completeness}}
\label{appendix:incorporating-bg-knowledge}

\begin{proofof}[Lemma \ref{lemma:no-contradiction-R9}]
Completeness of orientations with respect to \ref{R9} follows from Lemma \ref{lemma:r9-notneeded}.
Completeness of orientations with respect to \ref{R3} follows from the fact that we are adding consistent orientation knowledge to $\g$, which means we never elicit a new unshielded collider in $\g^{'}$. For \ref{R3} to be invoked a new unshielded collider would be needed.
\end{proofof}

\begin{proofof}[Lemma \ref{lemma:nocycle3}]
Let $p_{\g} = \langle P_1 , P_2 , \dots, P_k \rangle$ be a path in $\g^{\prime}$ that makes up a shortest directed or an almost directed cycle with edge $\langle P_1, P_k \rangle$. If $k =3$, we are done. 

Hence, suppose for a contradiction that $k>3 $ and let $p_{\g}$ be the path in $\g$ that  corresponds to path $p_{\g^{\prime}}$ in $\g$. 
Note that $p_{\g^{\prime}}$ must be an unshielded path since due to the completion of orientations in $\g^{\prime}$ under \ref{R2} and \ref{R8}, any shield  $\langle P_i, P_{i+2} \rangle$ would imply the existence of a shorter directed or almost directed cycle in $\g^{\prime}$. Therefore, $p_{\g}$ is an unshielded path of length $k>3$. Hence, it cannot be a circle path (Lemma \ref{lemma:zhang-b8-circ-path}).

By Corollary \ref{cor:b2}, it follows that $P_{k-1} \bulletarrow P_{k}$ is in $\g$. 
Using the same reasoning as in the previous paragraph, we can also conclude that  $P_2 \notin \Adj(P_k , \g)$ and that $P_1 \notin \Adj(P_{k-1}, \g)$.
Since orientations in $\g$ are closed under \ref{R9}, it therefore follows that we cannot have $P_1 \circarrow P_k$ in $\g$.

Hence $P_1 \leftarrow P_k$, or $P_1  \arrowcirc P_k$, or $P_1 \circcirc P_k$ is in $\g$. Since $P_1 \notin \Adj(P_{k-1}, \g)$,   and since $\g$ is ancestral, Lemma \ref{lemma:maathuis-7_5} implies that  $P_1 \to P_2 \dots \to P_k$ cannot be in $\g$.
Hence $P_1 \circbullet P_2$ is in $\g$. 

But now $P_1 \circbullet P_2$, $P_2 \notin \Adj(P_{k},\g)$,  and Lemma \ref{lemma:p1zhang}, imply that $P_1 \arrowbullet P_k$ is not in $\g$. Hence, $P_1 \circcirc P_k$ is in $\g$.

But now  $P_1 \circcirc P_k$ and the path $p_{\g}$ from $P_1$ to $P_k$ that does not contain $P_i \arrowbullet P_{i+1}$, $i \in \{1, \dots, k-1\}$ and ends with $P_{k-1}\bulletarrow P_k$ contradict Lemma \ref{lemma:zhang-pdpath-edge-not-into}.
\end{proofof}

\begin{proofof}[Lemma \ref{lemma:pag-to-eg-to-mag}]
Suppose for a contradiction that $\g^{\prime}$ is not maximal, that is, there is a { possibly} inducing path in $\g^{\prime}$. Then there is also a minimal collider path that is a { possibly} inducing path in $\g^{\prime}$.  The corresponding path in $\g$ must then also be a minimal collider path and a { possibly} inducing path. 

Among all shortest { possibly} inducing paths that are minimal collider paths in $\g$ choose a path that has the shortest distance to its endpoints (Definition \ref{def:dist-to-z}). Let this path be $p = \langle A, Q_1, \dots Q_k, B \rangle, k>1$. Then $Q_i \in \PossAn(\{A,B\}, \g)$ for all $i \in \{1, \dots, k\}$ and there is at least one $i \in \{1, \dots, k\}$ such that $Q_i \notin \An(\{A,B\}, \g)$ (otherwise $p$ is an inducing path). Note that $\g$ cannot contain inducing paths as it is an essential ancestral graph and there is at least one MAG it represents \citep{zhang2008completeness}.

 Let $Q_j$, $j \in \{1, \dots, k\}$, be the closest node to $A$ on $p$, such that $Q_j \notin \An(\{A,B\}, \g)$ and suppose without loss of generality that  $Q_j \in \PossAn(B,\g)$. Hence, let $q = \langle Q_j= Q_{j,1}, Q_{j,2}, \dots , Q_{j, k_j} = B \rangle, k_1 \ge 2$ be a shortest possibly directed path from $Q_j$ to $B$ in $\g$. By Corollary \ref{cor:unshielded-gen}, $q$ is then an unshielded possibly directed path. Hence, by  {Lemma \ref{lemma:zhao-mcp}} (ii) on path $p$,  $k_j >2$ (otherwise,  $Q_j \in \An(B,\g)$). Furthermore, by Corollary \ref{cor:b2}, $q$ must start with edge $Q_{j} \circbullet Q_{j,2}$ in $\g$.

Now, by Lemma \ref{lemma:induction-result-prime}, either $B \bulletarrow Q_{j,2}$ or there is some $Q_{j_+}$, $j_+ \in \{j+1, \dots, k\}$, such that $Q_{j_+} \leftrightarrow Q_{j,2}$. We cannot have $B \bulletarrow Q_{j,2}$ as that contradicts $q$ being an unshielded possible directed path from $Q_j$ to $B$. Similarly, either $A \bulletarrow Q_{j, 2}$ or there is some $Q_{j_-}$, $j_- \in \{1, \dots, j-1\}$, such that $Q_{j_-} \leftrightarrow Q_{j,2}$. In the former case, consider the path $p_1 = \langle A, Q_{j,2}, Q_{j_+} \rangle \oplus p(Q_{j_+}, B)$ and in the latter case, consider the path $p_2 = p(A, Q_{j_-}) \oplus \langle Q_{j_-}, Q_{j,2}, Q_{j_+} \rangle \oplus p(Q_{j_+}, B)$. Either way, we have now obtained either a shorter minimal collider path than $p$ that is a { possibly } inducing path in $\g$, or one that is of the same length but with a shorter distance to its endpoints, which is a contradiction.
\end{proofof}

 \begin{proofof}[Lemma \ref{lemma:no-new-mcps}]
We consider both directions below. 
   \begin{itemize}
       \item[$\Leftarrow$:] If conditions \ref{no-new-mcp:1} and \ref{no-new-mcp:2}  are satisfied, then Lemma \ref{lemma:MCP-format} and the fact that \ref{R4} subsumes \ref{R4new} immediately allow us to conclude that $\g$ and $\g^{\prime}$ have identical minimal collider paths. 
    \item[$\Rightarrow$:]  Suppose that every minimal collider path $p_{\g^{\prime}} = \langle V_1, \dots, V_k \rangle, k >1$ in $\g^{\prime}$ corresponds to a minimal collider path $p_{\g} = \langle V_1, \dots, V_k \rangle, k >1$ in $\g$. We need to show that this implies \ref{no-new-mcp:1} and \ref{no-new-mcp:2} hold. 
  
   Since every unshielded collider is a minimal collider path, the unshielded colliders in $\g$ and $\g^{\prime}$ must be identical. Hence, \ref{no-new-mcp:1} holds.   Furthermore, every discriminating collider path $q_{\g^{\prime}} = \langle A, Q_1, \dots, Q_m, B \rangle, m \ge 2$ in $\g^{\prime}$ that is also a minimal collider path in $\g^{\prime}$, will definitely satisfy \ref{no-new-mcp:2} in $\g$. 
   
   Lastly, suppose that $q_{\g^{\prime}} =  \langle A, Q_1, \dots, Q_m, B \rangle, m \ge 2$ is a discriminating collider path in $\g^{\prime}$, but not a minimal collider path in $\g^{\prime}.$ Then there must be a subsequence $q_{\g^{\prime}}^{\prime}$ of $q_{\g^{\prime}}$ that is a minimal collider path in $\g^{\prime}$. Furthermore, note that since $Q_i \to B$ is in $\g^{\prime},$ for all $i \in \{1, \dots, m -1 \}$, and $A \notin \Adj(B, \g^{\prime})$ the subsequence of $q_{\g^{\prime}}^{\prime}$ that forms a minimal collider path in $\g^{\prime}$ must contain $Q_m \arrowbullet B$.
  Therefore, $Q_m \arrowbullet B$ is in $\g$.

  Now, note that $Q_{m-1} \leftarrow Q_{m}$ cannot be in $\g$ since we know that all invariant orientations in $\g$ are also in $\g^{\prime}$ and we also know that $Q_{m-1} \bulletarrow Q_m$ is in $\g^{\prime}$. Therefore, either $Q_{m-1} \bulletcirc Q_{m}$ $Q_{m-1} \bulletarrow Q_{m}$ is in $\g$. We can now rule out that $Q_{m-1} \bulletcirc Q_m$ is in $\g$ as $Q_{m-1} \bulletcirc Q_m \arrowbullet B$ and Lemma \ref{lemma:p1zhang} would imply that $Q_{m-1} \arrowbullet B$ is in $\g$ and therefore in $\g^{\prime}$ contradicting our assumption that $\g^{\prime}$ contains $Q_{m-1} \to B$. 
  Therefore, $Q_{m-1} \bulletarrow Q_m \arrowbullet B$ is in $\g$.  
    \end{itemize}  
 \end{proofof}

\subsection{Theorem \ref{thm:3}}
\label{appendix:completeness2}

\begin{figure}[!t]
    \centering
      \begin{tikzpicture}[[->,>=stealth', auto,node distance=0.8cm,scale=.8,transform shape,font = {\large\sffamily}]
  \tikzstyle{state}=[inner sep=2pt, minimum size=12pt]

    \node[state] (T65) at (0, 0) {\textbf{Theorem \ref{thm:3}}};
    \node[state] (F3) at (0, 2) {Lemma \ref{lemma:merging-graphs-helper-case1}};
    \node[state] (L68) at (4, 2) {Lemma \ref{lemma:nocycle3}};
    \node[state] (L411) at (-4, 0) {Lemma \ref{lemma:pag-to-eg-to-mag}};
    \node[state] (F4) at (-4, 2) {Lemma \ref{lemma:merging-graphs-helper33}};
    \node[state] (G12) at (4, 0) {Theorem \ref{thm:chordal-completeness}};

 \draw[->,arrows= {-Latex[width=5pt, length=5pt]}] (L411) edge (T65);
 \draw[->,arrows= {-Latex[width=5pt, length=5pt]}] (G12) edge (T65);
 \draw[->,arrows= {-Latex[width=5pt, length=5pt]}] (L68) edge (T65);
 \draw[->,arrows= {-Latex[width=5pt, length=5pt]}] (F4) edge (T65);
 \draw[->,arrows= {-Latex[width=5pt, length=5pt]}] (F3) edge (T65);
\end{tikzpicture}
   \caption{Proof structure of Theorem \ref{thm:3}}
    \label{fig:thm-3}
\end{figure}

Figure \ref{fig:thm-3} displays how the supporting results come together to prove Theorem \ref{thm:3}.

\begin{proofof}[Theorem \ref{thm:3}]
Consider the following procedure.
First, we identify the circle component of $\g = (\mathbf{V}, \mathbf{E})$. This is the subgraph of $\g$ containing only $\circcirc$ edges, $\mathbf{E}_C$. Call this $\g_C = (\mathbf{V}, \mathbf{E}_C)$. Consider the same edges present in $\g^{\prime} = (\mathbf{V}, \mathbf{E}^{\prime})$, which might potentially have different edge mark orientations, $\mathbf{E}_C^{\prime}$. 
Note that by Lemma \ref{lemma:circle-chordal}, $\g_C = (\mathbf{V}, \mathbf{E}_C)$ is a collection of undirected connected chordal components $\g_{C_1}, \dots \g_{C_k}$, $k \ge 1$, each of which is an induced subgraph of $\g$.
We will refer to the corresponding induced subgraphs of $\g^{\prime}$ as $\g_{C_1}^{\prime}, \dots, \g_{C_k}^{\prime}$. Theorem \ref{thm:chordal-completeness} tells us that each individual induced subgraph  $\g_{C_i}^{\prime}$, $i \ge 1$ of $\g_C^{\prime}$ is a restricted essential ancestral graph. That is, each  $\g_{C_i}^{\prime}$  can be oriented into a MAG $\g[M]_i$ with no minimal collider paths, and with the desired edge orientation of a particular edge $\langle A,B \rangle$.

Now, suppose we construct a new directed mixed graph $\g[M] = (\mathbf{V}, \mathbf{E}_{\g[M]})$ obtained by taking the union of all invariant edge marks in $\g^{\prime}$ and $\g[M]_i$ for all $i \in \{1, \dots, k\}$. We will now show that $\g[M]$ is a MAG represented by $\g^{\prime}$. That is $\g[M]$ is an ancestral graph with the same minimal collider paths as $\g^{\prime}$ (Lemma \ref{lemma:pag-to-eg-to-mag}).
In particular, it suffices to show that there are no directed cycles or almost directed cycles in $\g[M]$ that contain some edges from $\g[M]_i$, $i \in \{1, \dots, k \}$ and some edges from $\g^{\prime}$ that are not in any $\g[M]_i$,  $i \in \{1, \dots, k \}$, and also that there are no minimal collider paths in $\g[M]$ that are made up of edges from $\g[M]_i$, $i \in \{1, \dots, k \}$ and edges outside of  $\g[M]_i$, $i \in \{1, \dots, k \}$ that are in $\g^{\prime}$.

First, we show that $\g[M]$ is ancestral. 
By Lemma \ref{lemma:nocycle3}, it is enough to show that there are no directed cycles or almost directed cycles of length 3 in $\g[M]$.  For sake of contradiction, we will suppose that there is a triple $A \to B \to C$ and edge $A \arrowbullet C$ in $\g[M]$. Furthermore, since $\g^{\prime}$ and $\g[M]_i$, for all $i$ are ancestral, and since $\g_{C_i}^{\prime}$  are induced subgraphs of $\g_{C}^{\prime}$ (Lemma \ref{lemma:circle-chordal})  $\forall i$, exactly two of the nodes $A,B,C$ are in $\g_{C_j}^{\prime}$ for some $j \ge 1$. We consider the options below:

\begin{enumerate}[label=(\alph*)] 
\item Suppose that $A,C$ are in $\g_{C_j}^{\prime},$ and $B \notin \g_{C}$. Note again that  $A \to B \to C$ and $A \arrowbullet C$ is in $\g[M]$. Furthermore, since $A,C \in \g_{C_j}^{\prime}$ and $B \notin \g_{C}$, we have that $A \circcirc C$ is in $\g$, and also that $B \to C$ or $B \circarrow C$ is in $\g$. Now Lemma \ref{lemma:p1zhang}, implies that $B \bulletarrow A$ must have been in $\g$, which leads us to a contradiction.
\item Suppose that $A,B$ are in $\g_{C_j}^{\prime},$ and $C \notin \g_{C}$. Again, consider that $A \to B \to C$ and $A \arrowbullet C$ are in $\g[M]$. Therefore, similarly to above, we have that $A \circcirc B$ is in $\g$ and since $C \notin \g_{C}$, $A \arrowbullet C$ is in $\g$. Hence, we obtain a contradiction with Lemma \ref{lemma:p1zhang}  as in the previous case.
\item Suppose that $B,C$ are in $\g_{C_j}^{\prime},$ and $A \notin \g_{C}$. Now again $A \to B \to C$ and $A \arrowbullet C$ are in $\g[M]$. Now, $C \to A \to B$ or $C \leftrightarrow A \to B$ are in $\g^{\prime}$. So since edge mark orientations in $\g^{\prime}$ are closed under \ref{R2}, the edge $\langle C, B \rangle$ must have an arrowhead at $B$ in $\g^{\prime}$. But, this contradicts that $B \to C$ is in $\g[M]$. 
\end{enumerate}

Therefore, $\g[M]$ is ancestral.
It remains to prove that $\g[M]$ has the same minimal collider paths as $\g^{\prime}$.
Suppose for a contradiction, there is a minimal collider path $p_{\g[M]} = \langle V_1, \dots, V_r \rangle$, $r \ge 3$ in $\g[M]$ such that the corresponding path $p_{\g^{\prime}}$ in $\g^{\prime}$ is not a collider path. Furthermore, we will choose the shortest such path $p_{\g[M]}$ and denote the corresponding paths (same sequences of nodes) in $\g^{\prime}$ as $p_{\g^{\prime}}$ and in $\g$ as $p_{\g}$.

Since there are no minimal collider paths in $\g[M]_i$, $i \in \{1 \dots, k\}$, and since a node in $\g[M]_i$ is not in $\g[M]_j$, for $i \neq j$, $i,j \in \{1, \dots, k\}$, we know that at least one edge on $p_{\g[M]}$ is in $\g^{\prime}$, but not in $\g^{\prime}_{C}$. Since $\g^{\prime}$ contains exactly the same minimal collider paths as $\g$, there is also at least one edge mark on $p_{\g[M]}$ that is in $\g[M]_i$, $i \in \{1 \dots, k\}$, but not in $\g^{\prime}$. 

Note first that $p_{\g[M]}$ cannot be an unshielded collider itself, and that $p_{\g[M]}$ cannot contain an unshielded collider that is not on $p_{\g^{\prime}}$. This is because none of the $\g[M]_i$, $i \in \{1 \dots, k\}$, graphs contain unshielded colliders, and $\g^{\prime}$ itself does not contain unshielded collider that are not already in $\g$. 
Furthermore, we cannot have a path $\langle A, B, C \rangle$ in $\g[M]$, where $\langle A, B \rangle$ is in $\g[M]_i$, and $\langle B, C \rangle$ is in $\g[M]_j$, where $i, j \in \{1, \dots, k\}$, and $i \neq j$ (due to Lemma \ref{lemma:circle-chordal}).
Furthermore, we know that $\g^{\prime}$ does not contain any unshielded collider  $A \bulletarrow B \arrowbullet C$, where $\langle A,B \rangle$ is in $\g^{\prime}_{C}$, and $\langle B,C \rangle$ is in $\g^{\prime}$ but not in $\g^{\prime}_C$, or vice versa (based on Lemma \ref{lemma:p1zhang} the fact that $\g^{\prime}$ does not contain new unshielded colliders compared to $\g$) and also that  $\g^{\prime}$ also cannot contain $A \bulletarrow B \circbullet C$, where $A \notin \Adj(C, \g^{\prime})$, due to orientations in $\g^{\prime}$ being closed under \ref{R1}.

Hence, any consecutive triple of nodes on $p_{\g[M]}$ is either shielded, or the corresponding triple is already an unshielded collider on $p_{\g}$. In particular, any triple $\langle V_l, V_{l+1}, V_{l+2} \rangle$, $l \in \{1, \dots, r-2\}$ on $p_{\g[M]}$ such that  $\langle V_l, V_{l+1} \rangle$ is in $\g^{\prime}$, but not in $\g^{\prime}_C$ and $\langle V_{l+1}, V_{l+2} \rangle$ is in $\g^{\prime}_{C_i}$ for some $i \in \{1, \dots, k \}$ is shielded. 

Since $p_{\g^{\prime}}$ is not a collider path we now consider the following options for choosing a triple on $p_{\g^{\prime}}$ which will be used to derive our desired contradiction. 
\begin{enumerate}[label = (\alph*)]
    \item\label{case11:nomcp-triple} Choose a triple $\langle V_l, V_{l+1}, V_{l+2} \rangle$, { with the smallest index} $l \in \{1, \dots, r-2\}$ on $p_{\g^{\prime}}$ that is of one of the following forms in $\g^{\prime}$:
    \begin{enumerate}[label = (a\arabic*)]
        \item\label{a1-case-noMCP} $V_l \bulletarrow V_{l+1} \circbullet V_{l+2}$ such that $V_l \in \Adj(V_{l+2}, \g^{\prime})$ and 
    such that $\langle V_l, V_{l+1} \rangle$ is in $\g^{\prime}$, but not in $\g^{\prime}_C$ and $\langle V_{l+1}, V_{l+2} \rangle$ is in $\g^{\prime}_{C_i}$ for some $i \in \{1, \dots, k \}$, or
        \item\label{a2-case-noMCP} $V_l \bulletarrow V_{l+1} \arrowbullet V_{l+2}$ such that $V_l \bulletarrow V_{l+2}$ is also in $\g^{\prime}$, and such that $\langle V_l, V_{l+1} \rangle$ is in $\g^{\prime}$, but not in $\g^{\prime}_C$ and $\langle V_{l+1}, V_{l+2} \rangle$ is in $\g^{\prime}_{C_i}$ for some $i \in \{1, \dots, k \}$, or
        \item\label{a3-case-noMCP} $V_l \bulletarrow V_{l+1} \arrowbullet V_{l+2}$ such that $V_l \bulletarrow V_{l+2}$ is also in $\g^{\prime}$, and such that $\langle V_l, V_{l+1} \rangle$ is in $\g^{\prime}_{C_i}$ for some $i \in \{1, \dots, k \}$  and $\langle V_{l+1}, V_{l+2} \rangle$ is in $\g^{\prime}$, but not in $\g^{\prime}_C$.
    \end{enumerate}
    \item\label{case22:nomcp-triple} Choose a triple $\langle V_l, V_{l+1}, V_{l+2} \rangle$, {with the largest index} $l \in \{1, \dots, r-2\}$ on $p_{\g^{\prime}}$, that is of one of the following forms in $\g^{\prime}$:
    \begin{enumerate}[label = (b\arabic*)]
        \item\label{b1-case-noMCP} $V_l \bulletcirc V_{l+1} \arrowbullet V_{l+2}$ in $\g^{\prime}$  such that $V_l \in \Adj(V_{l+2}, \g^{\prime})$ and such that $\langle V_l, V_{l+1} \rangle$ is in $\g^{\prime}_{C_i}$  for some $i \in \{1, \dots, k \}$, and $\langle V_{l+1}, V_{l+2} \rangle$ is in $\g^{\prime}$  but not in $\g^{\prime}_C$, or
         \item\label{b2-case-noMCP} $V_l \bulletarrow V_{l+1} \arrowbullet V_{l+2}$ such that $V_l \arrowbullet V_{l+2}$ is also in $\g^{\prime}$, and such that $\langle V_l, V_{l+1} \rangle$ is in $\g^{\prime}_{C_i}$ for some $i \in \{1, \dots, k \}$  and $\langle V_{l+1}, V_{l+2} \rangle$ is in $\g^{\prime}$, but not in $\g^{\prime}_C$, or
        \item\label{b3-case-noMCP} $V_l \bulletarrow V_{l+1} \arrowbullet V_{l+2}$ such that $V_l \arrowbullet V_{l+2}$ is also in $\g^{\prime}$, and such that $\langle V_l, V_{l+1} \rangle$ is in $\g^{\prime}$, but not in $\g^{\prime}_C$ and $\langle V_{l+1}, V_{l+2} \rangle$ is in $\g^{\prime}_{C_i}$ for some $i \in \{1, \dots, k \}$.
    \end{enumerate}
\end{enumerate}

Note that cases \ref{case11:nomcp-triple} and \ref{case22:nomcp-triple} cover all options for the form of the triple $\langle V_l, V_{l+1}, V_{l+2} \rangle$ on $p_{\g^{\prime}},$ so we are assured that one of the above options will exist on $p_{\g^{\prime}}.$ Also, note that case \ref{case22:nomcp-triple} is symmetric to case \ref{case11:nomcp-triple}, and the proof will be using exactly the same arguments Hence, without loss of generality, we only derive a contradiction for cases  \ref{case11:nomcp-triple}.

\begin{enumerate}
    \item[\ref{case11:nomcp-triple}] We discuss all three possible forms of the triple  $\langle V_{l}, V_{l+1}, V_{l+2} \rangle$ below and derive a contradiction in each case. 
\begin{enumerate}[label = (a\arabic*), leftmargin=\parindent,align=left,labelwidth=\parindent]
  \item[\ref{a1-case-noMCP}  or \ref{a2-case-noMCP}]  In this case we assume that either:
  \begin{itemize}
      \item  $V_{l} \bulletarrow V_{l+1} \circbullet V_{l+2}$ is in $\g^{\prime}$ and  $V_l \in \Adj(V_{l+2}, \g^{\prime})$ and    moreover, $\langle V_l, V_{l+1} \rangle$ is in $\g^{\prime}$, but not in $\g^{\prime}_C$ and $\langle V_{l+1}, V_{l+2} \rangle$ is in $\g^{\prime}_{C_i}$ for some $i \in \{1, \dots, k \}$.
      \item Or that $V_{l} \bulletarrow V_{l+1} \arrowbullet V_{l+2}$  and  $V_l \bulletarrow V_{l+2}$ are in $\g^{\prime}$,  and 
    moreover, $\langle V_l, V_{l+1} \rangle$ is in $\g^{\prime}$, but not in $\g^{\prime}_C$ and $\langle V_{l+1}, V_{l+2} \rangle$ is in $\g^{\prime}_{C_i}$ for some $i \in \{1, \dots, k \}$.
  \end{itemize}
    Hence, consider the form of edge $\langle V_{l}, V_{l+1} \rangle$ in $\g$. If this edge is of the form $V_l \to V_{l+1}$, $V_l \arrowcirc V_{l+1},$ or $V_l \leftrightarrow V_{l+1}$ in $\g$, then Lemma \ref{lemma:merging-graphs-helper-case1} tells us that the form of the edge $\langle V_l, V_{l+2} \rangle$ in $\g^{\prime}$ and $\g[M]$ would allow us  construct a shorter minimal collider path than $p_{\g[M]}$ by skipping over $V_{l+1}$, which leads us to a contradiction. 
    
    Next, we consider the case where $\langle V_l, V_{l+1} \rangle$ is of the form $V_l \circarrow V_{l+1}$ in $\g$. Then Lemma \ref{lemma:merging-graphs-helper-case1} implies that $V_l \to V_{l+2}$ or $V_l \leftrightarrow V_{l+2}$ is in $\g^{\prime}$ and $\g[M]$. In the latter case, we again get a contradiction with $p_{\g[M]}$ being a minimal collider path, as we could replace $\langle V_l, V_{l+1}, V_{l+2} \rangle$ with $\langle V_l , V_{l+2} \rangle$. Similarly, we get the same contradiction if $\langle V_l, V_{l+1} \rangle$ is the first edge on $p_{\g^{\prime}}$ and $p_{\g[M]}$, regardless of the form of the $\langle V_l, V_{l+2} \rangle$ edge in $\g^{\prime}$ and $\g[M]$.  

    Hence, suppose that $V_l \to V_{l+2}$ is in $\g^{\prime}$ and $\g[M]$ and that $l>1$, meaning that $V_l \leftrightarrow V_{l+1}$ is in $\g^{\prime}$ and $\g[M]$ (corresponding to $V_l \circarrow V_{l+1}$ in $\g$). 
    Next, note that if $p_{\g^{\prime}}(V_1, V_{l+1})$ is of the form $V_1 \bulletarrow V_2 \leftrightarrow \dots \leftrightarrow V_{l+1}$, case \ref{merging-h3:case3} of  Lemma \ref{lemma:merging-graphs-helper-case1} would imply that we can choose a subsequence of $p_{\g[M]}$ as a shorter minimal collider path, which is a contradiction. Otherwise, there is at least one edge $\langle V_{j}, V_{j+1} \rangle$, $1 \le j < l$ on $p_{\g^{\prime}}(V_1, V_{l+1})$ that corresponds to $V_j \circcirc V_{j+1}$ in $\g$, and also by case \ref{merging-h3:case3} of  Lemma \ref{lemma:merging-graphs-helper-case1},   there are edges $V_i \to V_{l+2}$ in $\g^{\prime}$ for every $j +1< i \le l$, and also that $V_{j+1} \circarrow V_{l+2}$ or $V_{j+1} \to V_{l+2}$ is in $\g^{\prime}$. Let $\langle V_j, V_{j+1} \rangle$, $1 \le j <l$  be indeed such an edge on $p_{\g^{\prime}}(V_1, V_{l+1})$ chosen so that the index $j$ is the largest possible.

    Now, consider the triple $\langle V_j, V_{j+1}, V_{j+2} \rangle$ in $\g^{\prime}$. By choice of our original triple $\langle V_{l}, V_{l+1}, V_{l+2} \rangle$, we can conclude that the triple $\langle V_{j}, V_{j+1}, V_{j+2} \rangle$ must be of one of the forms in \ref{case22:nomcp-triple}, and more precisely, either of the form described in case \ref{b1-case-noMCP} or case \ref{b2-case-noMCP}. 

    In either case, we have that either $V_{j} \leftrightarrow V_{j+2}$  or $V_{j} \leftarrow V_{j+2}$ is in $\g^{\prime}$ by Lemma \ref{lemma:merging-graphs-helper-case1}. If $V_j \leftrightarrow V_{j+2}$ is in $\g^{\prime}$,  we obtain our desired contradiction by constructing a shorter collider path $p_{\g[M]}(V_1, V_j) \oplus \langle V_j, V_{j+2} \rangle \oplus p_{\g[M]}(V_{j+2}, V_r)$.
       If $V_{j} \leftarrow V_{j+2}$ is in $\g^{\prime}$, then we must be in  case \ref{merging-h3:case3} of  Lemma \ref{lemma:merging-graphs-helper-case1}, so that $V_{j} \leftarrow V_{s}$, or $V_{j} \leftrightarrow V_s$, $j+2 \le s \le l$ and either $V_{j} \leftarrow V_{l+1}$, $V_{j} \leftrightarrow V_{l+1},$ or $V_{j} \arrowcirc V_{l+1}$ is in $\g^{\prime}.$ If any of the mentioned edges is of the form $\leftrightarrow$ in $\g^{\prime}$, we obtain a contradiction. Otherwise, we consider the edges between the following nodes in $\g$: $V_{j}, V_{j+1}, V_{l+1}, V_{l+2}$.

        We know that $V_{j} \circcirc V_{j+1}$ and $V_{l+1} \circcirc V_{l+2}$ is in $\g$. We also know that $V_{l+1} \to V_{j}$ or $V_{l+1} \circarrow V_{j}$ are in $\g^{\prime}$ and that similarly $V_{j+1} \to V_{l+2}$ or $V_{j+1} \circarrow V_{l+2}$ is in $\g^{\prime}$. 

        If $V_{l+1} \to V_j \circcirc V_{j+1}$ or $V_{l+1} \to V_{j} \circcirc V_{j+1}$ is in $\g$, then Lemma \ref{lemma:p1zhang}  and completeness of \ref{R2} in $\g$ 
        imply that $V_{l+1} \to V_{j+1}$ or $V_{l+1} \circarrow V_{j+1}$ is in $\g$. Similarly, if $V_{j+1} \to V_{l+2} \circcirc V_{l+1}$ or $V_{j+1} \to V_{l+2} \circcirc V_{l+1}$ are in $\g$, then Lemmas \ref{lemma:p1zhang} and completeness of \ref{R2} in $\g$  
        imply that $V_{j+1} \to V_{l+1}$ or $V_{j+1} \circarrow V_{l+1}$ is in $\g$.
        Both of these cannot be true at the same time, so at least one of the edges $\langle V_{l+1}, V_j \rangle$ or $\langle V_{j+1}, V_{l+2} \rangle$ are of the form $\circcirc$ in $\g.$

        Furthermore, if $V_{l+2} \circcirc V_{l+1} \circcirc V_j \circcirc V_{j+1}$ is in $\g$, then the edge $\langle V_{l+2}, V_{j +1} \rangle$ must also be of the form $\circcirc$ in $\g$ (Lemma \ref{lemma:zhang-circ-path}). Analogously, if $V_{j} \circcirc V_{j+1} \circcirc V_{l+2} \circcirc V_{l+1},$ is in $\g$, we conclude that $V_{j} \circcirc V_{l+1}$ is in $\g$ as well.

        Hence, now we have an undirected cycle of length $4$ in $\g$. Then by the chordal property of the circle component of $\g$ (Lemma \ref{lemma:circle-chordal}), either $V_{j} \circcirc V_{l+2}$ or $V_{j+1} \circcirc V_{l+1}$ is in $\g$. Let us assume without loss of generality that $V_j \circcirc V_{l+2}$ is in $\g$, and consider the form of this edge in $\g[M]$.
        If $V_{j} \to V_{l+2}$ is in $\g[M]$, then this edge together with $V_{l+2} \bulletarrow V_{l+1} \to V_{j}$ contradicts that $\g[M]$ is ancestral. If $V_{j} \leftarrow V_{l+2}$ is in $\g[M]$, then this edge together with $V_{j} \bulletarrow V_{j+1} \to V_{l+2}$ contradicts that $\g[M]$ is ancestral.
        Hence, the only option is for $V_j \leftrightarrow V_{l+2}$ to be in $\g[M]$, in which case $p_{\g[M]}(V_1, V_j) \oplus \langle V_j, V_{l+2} \rangle \oplus p_{\g[M]}(V_{l+2}, V_r)$ is a subsequence of $p_{\g[M]}$ in $\g[M]$ that forms a shorter collider path, which is a contradiction. 
  
\item[\ref{a3-case-noMCP}]  $V_l \bulletarrow V_{l+1} \arrowbullet V_{l+2}$ such that $V_l \bulletarrow V_{l+2}$ is also in $\g^{\prime}$, and $\langle V_l, V_{l+1} \rangle$ is in $\g^{\prime}_{C_i}$ for some $i \in \{1, \dots, k \}$  and $\langle V_{l+1}, V_{l+2} \rangle$ is in $\g^{\prime}$, but not in $\g^{\prime}_C$. By Lemma \ref{lemma:merging-graphs-helper33}, we have that either $V_l \leftrightarrow V_{l+2}$ or $V_l \to V_{l+2}$ is in $\g^{\prime}$. In the former case, we again get a contradiction with $p_{\g[M]}$ being a minimal collider path, as we could replace $\langle V_l, V_{l+1}, V_{l+2} \rangle$ with $\langle V_l , V_{l+2} \rangle$. Similarly, we get the same contradiction if $\langle V_l, V_{l+1} \rangle$ is the first edge on $p_{\g^{\prime}}$ and $p_{\g[M]}$, regardless of the form of the $\langle V_l, V_{l+2} \rangle$ edge in $\g^{\prime}$ and $\g[M]$.  

    Hence, suppose that $V_l \to V_{l+2}$ is in $\g^{\prime}$ and $\g[M]$ and that $l>1$, meaning that $V_l \leftrightarrow V_{l+1}$ is in $\g^{\prime}$ and $\g[M]$ (corresponding to $V_l \circcirc V_{l+1}$ in $\g$). 
    Suppose first that $V_{l-1}$ is also in $\g^{\prime}_{C_i}.$
    
    Since $V_{l+2} \bulletarrow V_{l+1}$ is in $\g^{\prime}$, if $V_{l+2} \notin \Adj(V_{l-1}, \g)$, we have that $V_{l+1} \to V_{l-1}$ is in $\g^{\prime}$ by \ref{R1}, and therefore, $\langle V_{l+2}, V_{l+1}, V_{l}, V_{l-1} \rangle$  would be a minimal discriminating collider path for $V_{l}$ that is in $\g^{\prime}$ but not in $\g$, therefore giving us our contradiction. 
    
    Otherwise, $V_{l+2} \in \Adj(V_{l-1}, \g)$. In this case consider again the edge $\langle V_{l-1}, V_{l+1} \rangle $ in $\g[M]$. If $V_{l-1} \leftrightarrow V_{l+1}$ is in $\g[M]$ we obtain a contradiction with our choice of path. If $V_{l-1} \leftarrow V_{l+1},$ then due to ancestrality of $\g[M]$, we have that $V_{l-1} \leftrightarrow V_{l+2}$ is in $\g[M]$, then again there is a subsequence of $p_{\g[M]}$ that forms a collider path in $\g[M]$ which also gives us a contradiction.
    
    Otherwise, $V_{l+2} \in \Adj(V_{l-1}, \g)$ and  $V_{l-1} \to V_{l+1}$  is in $\g[M]$. Let $j$ be chosen as the smallest  index on $p_{\g[M]}(V_1, V_{l-1})$ such that $V_j, \dots, V_{l}, V_{l+1}$ are all in $\g^{\prime}_{C_i}$.   Then all of the nodes in $V_j, \dots , V_{l+1}$ must be in the same clique since we do not create any minimal collider paths in $\g[M]_i$. Furthermore, if any edge $V_d \leftrightarrow V_s$, $j \le d < d+1 < s \le l+1$ is in $\g[M]$, we can choose a subsequence of $p_{\g[M]}$ that is a shorter collider path. Moreover, since $\g[M]_i$ and $\g[M]$ are ancestral, it follows that either $V_d \to V_s$ for all pairs $j \le d < d+1 < s \le l+1$ or $V_d \to V_s$.  If $j = 1$, we now have that $\langle V_1, V_{l+1} \rangle \oplus p_{\g[M]}(V_{l+1}, V_r)$ is a collider path, which is a contradiction.

    Otherwise, $j \neq 1$, and consider the triple $\langle V_{j-1}, V_j, V_{j+1} \rangle$ in $\g^{\prime}.$ Note that $\langle V_{j-1}, V_j \rangle$ cannot be in $\g^{\prime}_{C}$, otherwise it would be in $\g^{\prime}_{C_i}$ (Lemma \ref{lemma:circle-chordal}). Hence, $\langle V_{j-1}, V_{j} \rangle$ is in $\g^{\prime}$ but not in $\g^{\prime}_{C}$, and $\langle V_{j}, V_{j+1} \rangle$ is in $\g^{\prime}_{C_i}.$
      By choice of our original triple $\langle V_{l}, V_{l+1}, V_{l+2} \rangle$, we can conclude that the triple $\langle V_{j-1}, V_{j}, V_{j+1} \rangle$ must be of the form in case \ref{b3-case-noMCP}, that is $V_{j-1} \arrowbullet V_{j+1}$ is in $\g[M]$.

      If $V_{j-1} \leftrightarrow V_{j+1}$ is in $\g[M]$, then of course, $p_{\g[M]}(V_1, V_{j-1}) \oplus \langle V_{j-1}, V_{j+1} \rangle \oplus p_{\g[M]}(V_{j+1}, V_r)$ is a subsequence of $p_{\g[M]}$ that forms a minimal collider path and give us our contradiction. 
      
      Otherwise, $V_{j-1} \leftarrow V_{j+1}$ is in $\g[M]$ and we focus on the subpath $\langle V_{j-1}, V_j, V_{j+1}, V_{j+2} \rangle$. 
      Since $V_{j} \to V_{j+2}$ is in $\g[M]$ by assumption, we have that if edge $\langle V_{j-1}, V_{j+2} \rangle$ is in $\g$, then due to the ancestral property of $\g[M]$, $V_{j-1} \leftrightarrow V_{j+2}$ is in $\g[M]$ and then similarly to above, $p_{\g[M]}(V_1, V_{j-1}) \oplus \langle V_{j-1}, V_{j+2} \rangle \oplus p_{\g[M]}(V_{j+2}, V_r)$ is a subsequence of $p_{\g[M]}$ that forms a minimal collider path and give us our contradiction.  If however $V_{j-1} \notin \Adj(V_{j+2}, \g)$, then consider that  $\langle V_{j-1}, V_j, V_{j+1}, V_{j+2} \rangle$ is an inducing path and a minimal collider path in $\g[M]$. Since  $\langle V_{j-1}, V_j, V_{j+1}, V_{j+2} \rangle$ is an inducing path in $\g[M]$, this path cannot be collider path in $\g^{\prime}$ (otherwise, it would be a possibly inducing path and contradict Lemma \ref{lemma:pag-to-eg-to-mag}). Furthermore, since   $1 \le j\le l-1 < r$,  $\langle V_{j-1}, V_j, V_{j+1}, V_{j+2} \rangle$ is shorter than $p_{\g[M]}$, so our choice of a minimal collider path is incorrect and we obtain our contradiction.  
      \end{enumerate}
\end{enumerate}
\end{proofof}

\subsubsection{Supporting Results for Theorem \ref{thm:3}} \label{section:supporting-thm3}

\begin{lemma}\label{lemma:merging-graphs-helper-case1}  
Let $\g^{\prime} = (\mathbf{V,E'})$ be an ancestral partial mixed graph and $\g= (\mathbf{V,E})$ be an essential ancestral graph such that $\g$ and $\g^{\prime}$ have the same skeleton,  the same set of minimal collider paths, and all invariant edge marks in $\g$ exist and are identical in $\g^{\prime}$. 
 Suppose furthermore, that all $A\circarrow B$ edges in $\g$ correspond to $A \to B$ or $A \leftrightarrow B$ edges in $\g^{\prime}$ and that orientations in $\g^{\prime}$ are closed under {\ref{R1}, \ref{R2}, \ref{R4}}. %, \ref{R8}-\ref{R13}.
Suppose that $E \bulletarrow C$ is in $\g^{\prime}$, where this edge is of one of the following forms in $\g$:  $E \arrowcirc C$, $E \to C$, $E \circarrow C$, or $E \leftrightarrow C$. Furthermore, suppose that there is an edge $\langle C, D \rangle$ in $\g^{\prime}$ that corresponds to $C \circcirc D$ in $\g$, and also suppose that edge $\langle E, D \rangle$ is in $\g^{\prime}$.
\begin{enumerate}[label = (\arabic*)]
    \item\label{option1-merging-graphs-case1} If the form of the edge $\langle C, D \rangle$ is $C \circbullet D$ in $\g^{\prime}$, or
    \item\label{option2-merging-graphs-case1} if the form of the edge $\langle C, D \rangle$ is $C \arrowbullet D$ in $\g^{\prime}$, while the form of the edge $\langle E, D \rangle$ is $E \bulletarrow D$ in $\g^{\prime}$,
\end{enumerate}
then the following hold:
\begin{enumerate}[label = (\roman*)]
\item\label{merging-h3:case0} If $E \to C$ is in $\g$, then $E \to C$ is in $\g^{\prime}$ and also $E \to D$ or $E \leftrightarrow D$ is in $\g^{\prime}$.
\item\label{merging-h3:case1} If $E \arrowcirc C$  is in $\g$, then $E \leftrightarrow C$ is in $\g^{\prime}$ and also $E \leftrightarrow D$ is in $\g^{\prime}$.
    \item\label{merging-h3:case2} If $E \leftrightarrow C$  is in $\g$, then $E \leftrightarrow C$ is in $\g^{\prime}$ and also  $E \leftrightarrow D$ is in $\g^{\prime}$.
\item\label{merging-h3:case3} If  $E \circarrow C$ is in $\g$, then either 
\begin{itemize}
    \item $E \to C$ and $E \to D$ are in $\g^{\prime}$, or
    \item $E \to C$ and $E \leftrightarrow D$ are in $\g^{\prime}$, or
    \item $E \leftrightarrow C$ and $E \leftrightarrow D$ are in $\g^{\prime}$, or 
    \item $E \to C$ and $E \leftrightarrow D$ are in $\g^{\prime}$. Furthermore, in this setting, we have that 
  \begin{enumerate}
    \item\label{merging-h3:case2b} for every $P_1$ in $\g^{\prime}$   such that $P_1 \bulletarrow E$ is in $\g^{\prime}, $ $P_1 \bulletarrow D$ is in $\g^{\prime}$, and
    \item\label{merging-h3:case2c} for every $P_1 \bulletarrow P_2 \leftrightarrow \dots \leftrightarrow P_k$, $P_k \equiv E$, $k> 1$
    either there is an $i \in \{1, \dots, k\}$ such that $P_i \leftrightarrow D$ and $P_j \to D$, for all $j \in \{i+1 , \dots k\}$ or $P_1 \bulletarrow D$ and $P_i \to D$ for all $i \in \{2, \dots, k\}$ is in $\g^{\prime}$.
    \end{enumerate}
\end{itemize}
\end{enumerate}
 \end{lemma}

 \begin{proofof}[Lemma \ref{lemma:merging-graphs-helper-case1}] 
\begin{enumerate}
    \item[\ref{merging-h3:case0}] Since $E \to C \circcirc D$ is in $\g$, Lemma \ref{lemma:p1zhang} implies that $E \to D$ or $E \circarrow D$ are in $\g$. Since all $\circarrow$ edges in $\g$ correspond to $\to$ or $\leftrightarrow$ edges in $\g^{\prime}$, we know that $E \to D$, or $E \leftrightarrow D$ is in $\g^{\prime}$. 

\item[\ref{merging-h3:case1}] Since $E \arrowcirc C \circcirc D$ is in $\g$, and $E \in \Adj(D, \g)$, we have by Lemmas \ref{lemma:p1zhang}
and the fact orientations in $\g$ are closed under \ref{R2},  that $E \arrowcirc D$ or $E \leftarrow D$ is in $\g$. Then $E \leftrightarrow D$ or $E \leftarrow D$ is in $\g^{\prime}$.

In case \ref{option2-merging-graphs-case1}, we then immediately have that $E \leftrightarrow D$ is in $\g^{\prime}$.
Now, in case \ref{option1-merging-graphs-case1}, $E \leftrightarrow  C \circbullet D$ in $\g^{\prime}$, and the fact that orientations in $\g^{\prime}$ are closed with respect to \ref{R2} would imply that $E \leftarrow D$ cannot be in $\g$. Hence, $E \arrowcirc D$ is in $\g$ and therefore, $E \leftrightarrow D$ is in $\g^{\prime}$.

\item[\ref{merging-h3:case2}]  If $E \leftrightarrow C$ is in $\g$, then since $C \circcirc D$ is in $\g$,  Lemma \ref{lemma:p1zhang} and completeness of \ref{R2} in $\g$ imply that $E \leftrightarrow D$ is also in $\g$. Hence, $E \leftrightarrow C$ and $E \leftrightarrow D$ are also in $\g^{\prime}$.

\item[\ref{merging-h3:case3}] If $E \circarrow C$ is in $\g$, then since $C \circcirc D$ is in $\g$, Lemma \ref{lemma:p1zhang} 
implies $E \circarrow  D$ or $E \to D$ is in $\g$. Then we have the combination of cases as listed above. In particular, if   $E \leftrightarrow C$ and $E \to D$ are in $\g^{\prime}$, we also have that cases \ref{merging-h3:case2b} and \ref{merging-h3:case2c} hold because $\g^{\prime}$ is ancestral and that $\g^{\prime}$ has the same minimal collider paths as $\g$. Note that  Lemma \ref{lemma:no-new-mcps} unshielded collider in $\g^{\prime}$ is also an unshielded collider in $\g$ and every collider discriminated by a path in $\g^{\prime}$ must be a collider on the corresponding path in $\g$. Since we know that $C \circbullet D$ is in $\g$, we know that the paths of the form $P_i \bulletarrow P_{i+1} \leftrightarrow \dots \leftrightarrow P_k \leftrightarrow E \leftrightarrow C \arrowbullet D$, $i \in \{1, \dots, k\}$ in $\g^{\prime}$ cannot be discriminating paths, hence $P_i \in \Adj(D, \g^{\prime})$. The rest of the argument follows by using completeness of orientation rules \ref{R1}, \ref{R2}, and \ref{R4} in $\g^{\prime}$.
\end{enumerate}
  \end{proofof}

\begin{lemma}\label{lemma:merging-graphs-helper33}  
Let $\g^{\prime} = (\mathbf{V,E'})$ be an ancestral partial mixed graph and $\g = (\mathbf{V,E})$ be an essential ancestral graph such that $\g$ and $\g^{\prime}$ have the same skeleton,  the same set of minimal collider paths, and all invariant edge marks in $\g$ exist and are identical in $\g^{\prime}$. 
Suppose furthermore, that all $A\circarrow B$ edges in $\g$ correspond to $A \to B$ or $A \leftrightarrow B$ edges in $\g^{\prime}$ and that orientations in $\g^{\prime}$ are closed under {\ref{R1}, \ref{R2}, \ref{R4}.} %, \ref{R8}-\ref{R13}.
Suppose that $C \arrowbullet D$ is an edge in $\g^{\prime}$ that corresponds to $C \circcirc D$ in $\g$, and also that $E \bulletarrow C$ is in $\g^{\prime}$, where this edge is of one of the following forms in $\g$:  $E \arrowcirc C$, $E \to C$, $E \circarrow C$, or $E \leftrightarrow C$, then there is an edge $\langle E, D \rangle$ in $\g^{\prime}$ and suppose that this edge is of the form $E \arrowbullet D$ is in $\g^{\prime}.$ 
Then
\begin{enumerate}[label = (\roman*)]
\item\label{merging-h33:case0} If $E \to C$ is in $\g$, then $E \to C$ is in $\g^{\prime}$ and $E \leftrightarrow D$ is in $\g^{\prime}$.
    \item\label{merging-h33:case2} If $E \leftrightarrow C$  is in $\g$, then $E \leftrightarrow C$ is in $\g^{\prime}$ and also  $E \leftrightarrow D$ is in $\g^{\prime}$. 
\item\label{merging-h33:case3} If  $E \circarrow C$ is in $\g$, then either $E \to C$ or $E \leftrightarrow C$ is in $\g^{\prime}$ and $E \leftrightarrow D$ is in $\g^{\prime}$.
\item\label{merging-h33:case1} If $E \arrowcirc C$  is in $\g$, then $E \leftrightarrow C$ is in $\g^{\prime}$ and $E \leftrightarrow D$ or $E \leftarrow D$ is in $\g^{\prime}$.
\end{enumerate}
 \end{lemma}

  \begin{proofof}[Lemma \ref{lemma:merging-graphs-helper33}]
\begin{enumerate}
    \item[\ref{merging-h33:case0}] Since $E \to C \circcirc D$ is in $\g$, Lemma \ref{lemma:p1zhang} implies that $E \to D$ or $E \circarrow D$ are in $\g$. Since all $\circarrow$ edges in $\g$ correspond to $\to$ or $\leftrightarrow$ edges in $\g^{\prime}$, we know that $E \to D$, or $E \leftrightarrow D$ is in $\g^{\prime}$. By assumption, we know $E \arrowbullet D$ is in $\g^{\prime}$, and hence, $E \leftrightarrow D$ must be in $\g^{\prime}$.

\item[\ref{merging-h33:case2}]  If $E \leftrightarrow C$ is in $\g$, then since $C \circcirc D$ is in $\g$,  Lemma \ref{lemma:p1zhang} and completeness of \ref{R2} in $\g$, imply that $E \leftrightarrow D$ is also in $\g$. Hence, $E \leftrightarrow C$ and $E \leftrightarrow D$ are also in $\g^{\prime}$.

\item[\ref{merging-h33:case3}] If $E \circarrow C$ is in $\g$, then since $C \circcirc D$ is in $\g$, Lemma \ref{lemma:p1zhang} 
implies $E \circarrow  D$ or $E \to D$ is in $\g$. 
Since we know, that $E \arrowbullet D$ is in $\g^{\prime}$, it must be that $E \circarrow D$ is in $\g$ and  $E \leftrightarrow D$ is in $\g^{\prime}$.

\item[\ref{merging-h33:case1}]  Since $E \arrowcirc C \circcirc D$ is in $\g[{P}]$, and $E \in \Adj(D, \g)$, we have by Lemmas \ref{lemma:p1zhang} 
and the fact that orientations under \ref{R2} are closed in $\g$ that $E \arrowcirc D$ or $E \leftarrow D$ is in $\g$. Since we assume that $E \arrowbullet D$ is in $\g^{\prime}$, this implies that  $E \leftrightarrow D$ or $E \leftarrow D$   is in $\g^{\prime}$.
\end{enumerate}
  \end{proofof}

\subsection{Theorem \ref{thm:chordal-subgraph-completeness}}
\label{appendix:completeness1}

\begin{proofof}[Theorem \ref{thm:chordal-subgraph-completeness}]
Consider constructing the graph $\g^{\prime \prime}$ by replacing all edges $\langle S, T \rangle$  in $\g^{\prime}$ that are of the form $S \circarrow T$ in both $\g^{\prime}$ and $\g$ with $S \to T$.
By Theorem \ref{thm:adding-tails-rules-complete}, 
$\g^{\prime \prime}$ is ancestral, has the same minimal collider paths as $\g^{\prime}$, and edge mark orientations in $\g^{\prime \prime}$  are  closed under { \ref{R1}, \ref{R2}, \ref{R4}, \ref{R8}, \ref{R11}} % and \ref{R8}-\ref{R13}. 
The proof is now complete as $\g$ and $\g^{\prime \prime}$ satisfy Theorem \ref{thm:3}.
\end{proofof}

\begin{theorem}
\label{thm:adding-tails-rules-complete} 
Let $\g^{\prime} = (\mathbf{V,E})$ be an ancestral partial mixed graph and $\g$ be an essential ancestral graph such that $\g$ and $\g^{\prime}$ have the same skeleton,  the same set of minimal collider paths, and all invariant edge marks in $\g$ exist and are identical in $\g^{\prime}$. Suppose furthermore, that every edge $A \circarrow B$ in $\g$ corresponds either to $A \to B$ or to $A \circarrow B$ in $\g^{\prime}$ and that edge mark orientations in $\g^{\prime}$ are closed under { \ref{R1}, \ref{R2}, \ref{R4}, \ref{R8}, \ref{R11}.} %\ref{R1}-\ref{R4}, \ref{R8}-\ref{R13}. 
Let $\g^{\prime \prime}$ be identical to $\g^{\prime}$ except all $A\circarrow B$ edges in $\g$ correspond to $A \to B$ edges in $\g^{\prime \prime}$. Then edge mark orientations in $\g^{\prime\prime}$ are closed under \ref{R1}-\ref{R4}, \ref{R8}-\ref{R13} and $\g^{\prime \prime}$ is ancestral and has the same minimal collider paths as $\g^{\prime}$.
\end{theorem}

\begin{proofof}[Theorem \ref{thm:adding-tails-rules-complete}]
We first consider showing that edge mark orientations in $\g^{\prime \prime}$ are closed under { \ref{R1}, \ref{R2}, \ref{R4}, \ref{R8}, \ref{R11}.} %\ref{R1}-\ref{R4} and \ref{R8}-\ref{R13}.
It is enough to consider each orientation rule and show that the antecedent for any rule will not occur in $\g^{\prime \prime}$ directly. A lot of the arguments below will use the fact that $\g^{\prime \prime}$ does not contain any new arrowhead edge marks compared to $\g^{\prime}$ and that edge mark orientations in $\g^{\prime}$ are already  closed under { \ref{R1}, \ref{R2}, \ref{R4}, \ref{R8}, \ref{R11}.} % \ref{R1}-\ref{R4}, \ref{R8}-\ref{R13}.
%First, note that completeness of edge marks under \ref{R3} and \ref{R9} follows immediately by Lemma \ref{lemma:no-contradiction-R9}. 
\begin{enumerate}
\item[\ref{R1}] The antecedent of \ref{R1} requires a triple $A \bulletarrow B \circbullet C$ to exist in  $\g^{\prime \prime}$, and $A \notin \Adj(C, \g^{\prime \prime})$. We know this type of triple cannot exist in $\g^{\prime \prime}$ because we do not introduce any arrowhead edge marks in $\g^{\prime \prime}$ compared to $\g^{\prime}$, and edge mark orientations in $\g^{\prime}$ are closed  under \ref{R1}-\ref{R4}, \ref{R8}-\ref{R13}.

\item[\ref{R2}] Having the antecedent of \ref{R2} in $\g^{\prime \prime}$ but not in $\g^{\prime}$ would require that there is a triple $A,B,C$ in $\g$ such that
\begin{itemize}
    \item $A \bulletcirc C$ is in $\g^{\prime \prime}$, $\g^{\prime}$, and in $\g$, and
    \item $A \bulletarrow B \to C$ or $A\to B \bulletarrow C$  is in $\g^{\prime \prime}$, but
    \item  $A \bulletarrow B \circarrow C$ or $A \circarrow B \bulletarrow C$ is in $\g^{\prime}$, and by assumption
    \item $A \bulletarrow B \circarrow C$, or $A \circcirc B \circarrow C$ or $A \circarrow B \bulletarrow C$, or $A \circarrow B \circcirc C$ is in $\g$.
\end{itemize}
Note that if either $A \bulletarrow B \circarrow C$, or $A \circarrow B \circcirc C$ are in $\g$, then $A \bulletcirc C$ cannot be in $\g$ by Lemma \ref{lemma:p1zhang}. 
Similarly, having either $A \circcirc B \circarrow C$ or $A \circarrow B \bulletarrow C$ in $\g$, together with edge $A \bulletcirc C$ would imply a contradiction with Lemma \ref{lemma:p1zhang}, as Lemma \ref{lemma:p1zhang} would insist on an arrowhead at $A$ on edge $\langle A,B \rangle$.

\item [\ref{R4}] The antecedent of \ref{R4} would require the presence of:
\begin{itemize}
    \item an almost discriminating path $p = \langle A, Q_1, \dots, Q_k, Q_{k+1} = B, \rangle$ for $Q_k$ in $\g^{\prime \prime}$, $A \notin \Adj(B, \g)$, with
    \item $Q_k \circbullet B$ also being in $\g^{\prime \prime}$. 
    \item Then $p(A,Q_k)$ is then an almost collider path in $\g^{\prime \prime}$, and by inspecting the definition of an almost collider path (Definition \ref{def:almost-collider}), it is clear that
    \item $\langle A, Q_1, \dots, Q_k \rangle$ must also be an almost collider path  in $\g^{\prime}$.
\end{itemize}
However, since $\langle A, Q_1, \dots, Q_k,B \rangle$ is not an almost discriminating path in $\g^{\prime}$ (otherwise, $Q_k \to B$ would be in $\g^{\prime}$), at least one of the edges $\langle Q_i, B \rangle$ is of the form  $Q_i \circbullet B$, $i \in \{1, \dots, k-1\}$  in $\g^{\prime}$. Note that since all edges $\langle Q_i, B \rangle$, $i \in \{1, \dots, k-1\}$ are of the form $Q_i \to B$ in $\g^{\prime \prime}$, the form of all of these edges in $\g^{\prime}$ is either $\to$ or $\circarrow$.
Let $Q_j \circarrow B$, $j \in \{1, \dots, k-1\}$  be an edge in $\g^{\prime}$, chosen such that there is no edge of that form with a smaller index than $j$. 

If $j = 1$, then we know that $A \bulletarrow Q_1$ cannot be in $\g^{\prime}$, otherwise, edge mark orientations in $\g^{\prime}$ would not be  closed under \ref{R1}. Examining the definition of an almost collider path, we now know that $A \bulletcirc Q_1 \arrowbullet Q_2$ and {$A \bulletarrow Q_2$} are in $\g^{\prime}$, 
Furthermore,  {$A \bulletarrow Q_2$}  implies $Q_2 \to Y$ is in $\g^{\prime}$ by  \ref{R1}. Now consider the relationships between nodes $A,Q_1,Q_2$ and $B$ in $\g^{\prime}$:
\begin{itemize}
    \item $A \bulletarrow Q_1 \arrowbullet Q_2 \to B$ is in $\g^{\prime}$ and so are
    \item $A \bulletarrow Q_2$, and
    \item $Q_1 \circarrow B$, and in addition,
    \item $A \notin \Adj(B, \g^{\prime})$.
\end{itemize}
Now, the above implies that edge mark orientations in $\g^{\prime}$ are not  closed under  \ref{R11}, which is a contradiction. 

Next, suppose that $j>1$ and $Q_j \circarrow B$ is in $\g^{\prime}$. Now, by definition of the almost discriminating path we have that $\langle A,Q_1, \dots, Q_j \rangle \oplus \langle Q_j, B \rangle$  is an almost discriminating path for $Q_j$ in $\g^{\prime}$. However, this now implies that edge mark orientations in $\g^{\prime}$ are not  closed under  \ref{R4}, which is a contradiction.

\item [\ref{R8}] Having the antecedent of \ref{R8} in $\g^{\prime \prime}$ but not in $\g^{\prime}$ would require that there is a triple $A,B,C$ in $\g$ such that
\begin{itemize}
    \item $A \circarrow C$ is in $\g^{\prime \prime}$, and
    \item $A \to B \to C$  is in $\g^{\prime \prime}$, but
    \item  $A \circarrow B \to C$, or $A \to B \circarrow  C$, or $A \circarrow B \circarrow C$ is in $\g^{\prime}$.
\end{itemize}
Also, note that since  $A \circarrow C$ is in $\g^{\prime \prime}$, it must be that $A \circcirc C$ is in $\g$. Lemma \ref{lemma:p1zhang} then implies that $B \bulletarrow C$ cannot be in $\g$. Meaning that the only option among the above listed is to have $A \circcirc C$, $A \circarrow B \circcirc C$ in $\g$. However, this option also contradicts Lemma \ref{lemma:p1zhang}.

% \item [\ref{R10}] For the antecedent of \ref{R10} to exist in  $\g^{\prime \prime}$, by Lemma \ref{lemma:r10-needs-ucc}, we must have the following:
% \begin{itemize}
%     \item $B \to C \leftarrow D$, $A \circarrow C$,
%     $M_{11} \bulletarrow C \arrowbullet M_{21}$, are in $\g^{\prime \prime}$ and $M_{11} \notin \Adj(M_{21}, \g)$, and
%     \item $A \circbullet M_{11},$ or $A \to M_{11}$, and $A \circbullet M_{21},$ or $A \to M_{21}$, and are in $\g^{\prime \prime}$.
%     \item Then  $M_{11} \bulletarrow C \arrowbullet M_{21}$, is also in $\g$ and in $\g^{\prime}$ since we do not introduce new unshielded colliders into $\g^{\prime}$ or into $\g^{\prime \prime}$, and 
%    \item  similarly $A \circbullet M_{11},$ or $A \to M_{11}$, and $A \circbullet M_{21},$ or $A \to M_{21}$, and are in $\g^{\prime}$ and $\g$.
%     \item   also, by construction of $\g^{\prime \prime}$, it must be that $A \circcirc C$ is in $\g$.
% \end{itemize}
% Now, focus on the triple $A, C,$ and $M_{11}$ in $\g$. We know that $M_{11} \bulletarrow C \circcirc A$ is in $\g$ and since $M_{11} \in \Adj(A, \g)$, Lemma \ref{lemma:p1zhang} implies that $M_{11} \bulletarrow A$ is in $\g$ as well. But that contradicts that $A \circbullet M_{11}$, or $A \to M_{11}$ is in $\g$.

\item[\ref{R11}] For the antecedent of \ref{R11} consider the left panel of Figure \ref{fig:R11compressed}. To have this graph as an induced subgraph of $\g^{\prime \prime}$, but not of $\g^{\prime}$, edge $C \to D$ must have been $C \circarrow D$ in $\g^{\prime}$. However, this would contradict that edge mark orientations in  $\g^{\prime}$ are under \ref{R1}.

\end{enumerate}

Next we show that $\g^{\prime \prime}$ is ancestral and has the same minimal collider paths as $\g^{\prime}$. The latter follows immediately since we do not introduce any arrowheads in $\g^{\prime \prime}$, or remove edges compared to $\g^{\prime}$.
Suppose for a contradiction that there is a directed or almost directed cycle in $\g^{\prime \prime}$. By Lemma \ref{lemma:nocycle3}, there is also one such cycle of length $3$ in $\g^{\prime \prime}$. Let $A \to B \to C$, $A \arrowbullet C$ be one such cycle in $\g^{\prime \prime}$. Since $\g^{\prime}$ is ancestral, we know that the corresponding edges in $\g^{\prime}$  are in one of the following categories:
\begin{enumerate}[label=(\alph*)]
\item\label{case3:ancest-gii} $A \circarrow B \to C$ and $A \arrowbullet C$.
\item\label{case32:ancest-gii} $A \to B \circarrow C$ and  $A \arrowbullet C$.
\item\label{case2:ancest-gii} $A \circarrow B \circarrow C$ and $A \leftarrow C$.
\item\label{case1:ancest-gii} $A \circarrow B \circarrow C$ and $A \arrowcirc C$.
\item\label{case4:ancest-gii} $A \circarrow B \circarrow C$ and $A \leftrightarrow C$.
\end{enumerate}
Note that cases \ref{case3:ancest-gii}-\ref{case2:ancest-gii} contradict that edge mark orientations in $\g^{\prime}$ are  closed under  \ref{R2} and \ref{R8}.
The edges in case \ref{case1:ancest-gii}-\ref{case4:ancest-gii}  must be of that same form in $\g$. However, \ref{case1:ancest-gii} contradicts  Lemma \ref{lemma:maathuis-7_5} and \ref{case4:ancest-gii} contradicts Lemma \ref{lemma:p1zhang}.
\end{proofof}

{

\subsection{Additional Simulations}
\label{subsection:real-data-sims}

A key assumption in our work is that the orientation knowledge we add is consistent with at least the essential ancestral graph, if not the true underlying MAG. In practice, we can learn the essential ancestral graph only from finite samples, and therefore its correctness is not always correct. Therefore, the orientation knowledge may not be consistent with the learned essential graph even if it is consistent with the true MAG. In these scenarios, Algorithm \ref{alg:mpag} will fail.

We perform a basic simulation to quantify how often {the} true orientation knowledge is deemed to be inconsistent with the learned essential graph. And when it is consistent, we quantify how often the true restricted essential graph is recovered, and how many tail- and arrow-oriented edgemarks agree between the true and learned restricted essential graphs.

We generate a DAG with R's \texttt{pcalg::randomDAG} with $n$ nodes and probability, $p = 0.2$. We choose $L = 3$ root or confounder nodes as latent nodes to generate the MAG $\mc{M}$, and the corresponding essential graph, $\mc{G}$. We pick $k\% = 20\%$ of non-identified edgemarks in the essential graph as our orientation knowledge, $\mc{K}$. We learn the $\mc{K}$-restricted essential graph, $\mc{G}_{\mc{K}}$ using Algorithm \ref{alg:mpag}. We also generate $m = 1000$ data points using the DAG with \texttt{pcalg::rmvDAG}, remove the variables corresponding to the latent nodes. We estimate the essential graph using \texttt{pcalg::fci}, $\widehat{\mc{G}}$. And if $\mc{K}$ is consistent with $\widehat{\mc{G}}$, we also estimate $\widehat{\mc{G}}_{\mc{K}}$ with Algorithm \ref{alg:mpag}. We repeat this for $T = 100$ simulations. In scenarios where we can learn $\widehat{\mc{G}}_{\mc{K}}$, we also count how many arrowheads and tails agree with $\mc{G}_{\mc{K}}$. The results are summarized in Table \ref{tab:real-data-sims}.

\begin{table}[!tbh]
\centering
% \resizebox{\textwidth}{!}{%
\begin{tabular}{c||c|c|c|c|c|c}
 $n$ & \shortstack{\textbf{Conflict} \\ \textbf{rate}} & \shortstack{\textbf{Exact} \\ \textbf{match rate}} & \shortstack{\textbf{Arrowhead} \\ \textbf{correct rate}} & \shortstack{\textbf{Tail} \\ \textbf{correct rate}} & \shortstack{\textbf{Circle} \\ \textbf{correct rate}} & \shortstack{\textbf{Average} \\ \textbf{total edges}} \\
 \hline
 \hline
10 & 0.13 & 0.51 & 0.91 & 0.90 & 0.36 & 13.24 \\
11 & 0.15 & 0.40 & 0.86 & 0.90 & 0.32 & 17.62 \\
12 & 0.14 & 0.26 & 0.81 & 0.84 & 0.26 & 23.16 \\
13 & 0.15 & 0.13 & 0.71 & 0.73 & 0.22 & 27.22 \\
14 & 0.23 & 0.14 & 0.70 & 0.65 & 0.17 & 33.01 \\
15 & 0.24 & 0.08 & 0.65 & 0.66 & 0.14 & 39.53 \\
16 & 0.24 & 0.01 & 0.56 & 0.54 & 0.11 & 46.61 \\
17 & 0.30 & 0 & 0.49 & 0.49 & 0.09 & 55.66 \\
18 & 0.22 & 0.01 & 0.46 & 0.42 & 0.08 & 62.13 \\
19 & 0.25 & 0 & 0.39 & 0.34 & 0.07 & 72.27 \\
20 & 0.23 & 0 & 0.33 & 0.31 & 0.05 & 82.16
\end{tabular}
% }
\caption{Agreement between true restricted essential graph and restricted essential graph learned from finite data. Conflict rate is average number of graphs where the orientation knowledge is inconsistent with the learned essential graph. Exact match rate is the average number of graphs, when the background knowledge was consistent, where the true and learned restricted graphs identical.}
\label{tab:real-data-sims}
\end{table}

}

\section{Completeness of Edge Mark Orientations in Ancestral Partial Mixed Graphs with no Minimal Collider Paths}
\label{appendix:meek-analogue}

Consider a partial mixed graph $\g^{\prime}$ obtained as  output of Algorithm \ref{alg:mpag}.  We examine edge orientations of $\g^{\prime}_{C}$, which is the induced subgraph of $\g^{\prime}$ that corresponds to the circle component of the essential ancestral graph $\g$. We show that edge orientations within these types graphs are complete using an argument similar to \citet{meek1995causal}.  

Since the skeleton of  such a graph $\g^{\prime}_{C}$ is chordal \citep{zhang2008completeness}, we can construct a join trees on its maximal cliques (see Chapter 3.2 of \citealp{lauritzen1996graphical}, and Theorem \ref{thm:beeri-chordal-join-tree} below). Similar to \citet{meek1995causal}, we define a total ordering of maximal cliques in a join tree and show that this ordering induces a partial ordering of nodes in $\g^{\prime}_{C}$ that is consistent with prior edge mark orientations, maintains the ancestral property and does not introduce any minimal collider paths. Then, we show how to select two MAGs represented by $\g^{\prime}_{C}$ as extensions of these orderings with the required orientations of an edge in question.

Our main result is presented in Theorem \ref{thm:chordal-completeness} in Section \ref{sec:meek-analogue-main}. A map of how all results in this section are used to prove Theorem \ref{thm:chordal-completeness} is given in Figure \ref{figproof-chordal} of Section \ref{sec:meek-analogue-main}. Throughout this section we also include examples for intermediate results and algorithms, concluding with Example \ref{ex:algo4-demo}, which demonstrates the constructive process for obtaining the MAGs described in Theorem \ref{thm:chordal-completeness}.

In Table \ref{tab:results-comparing-meek} below, we make explicit the connections between our results and that of \citet{meek1995causal}. The second column provides locations or specific references to results in this manuscript that are somewhat analogous to those of \citet{meek1995causal}. In our proofs, we identify an important gap in Lemma 6 of \citet{meek1995causal}. Namely, Lemma 6 of \cite{meek1995causal} cannot hold as stated, which we illustrate in Example \ref{ex:lemma6-meek-issue} and the text following it. Since Lemmas 7, 8, and Theorem 4 of \cite{meek1995causal} rely on Lemma 6, their proofs also do not go through. Our Lemmas \ref{lemma:order-cliques-1} and \ref{lemma:rule-1-join-tree} present weaker versions of Lemmas 6 and 7 of \cite{meek1995causal} and we devise a different strategy for using their results that allows us to prove Theorem \ref{thm:chordal-completeness}. Since our setting is more general, our proof of Theorem \ref{thm:chordal-completeness} will also serve as a proof of  Theorem 4 of \citet{meek1995causal}.

\begin{table}[!tbh]
    \centering
    \begin{tabular}{c|c|c|c}
        \textbf{\citet{meek1995causal}} & \textbf{Our Results} & \textbf{Examples} & \textbf{Location} \\
        \hline
        \hline
      Lemma 6 &    Lemma \ref{lemma:order-cliques-1}  & Example \ref{ex:lemma6-meek-issue} & Section \ref{sec:gen-join-tree-l4} \\
       Lemma 7 &  Lemma \ref{lemma:rule-1-join-tree} & Examples \ref{ex:l7p11}-\ref{ex:l7p13} & Section \ref{sec:finding-right-jointree} \\ 
        Lemmas 4 and 8 & Algorithm \ref{alg:orient-join-tree} and 
 Lemma \ref{lem:correct-directed-tree} &---& Section \ref{sec:orienting-join-tree-and-cliques} \\
    \end{tabular}
    \caption{Locating Analogous Results to \citet{meek1995causal}.}
    \label{tab:results-comparing-meek}
\end{table}

\subsection{Section Specific Preliminaries}

\begin{definition}[Partial Order]\label{def:po}
Consider a set of elements $\mathbf{V}$. A relation $\leq$, between the elements of $\mathbf{V}$ is called a partial order if and only if for every $A,B, C \in \mb{V}$
\begin{enumerate}[label=(\roman*)]
    \item\label{def:po1} reflexive: $A \leq A$,
    \item\label{def:po2} antisymmetric: if $A \leq B$ and $B \leq A$, then $A = B$, and
    \item\label{def:po3} transitive: if $A \leq B$ and $A \leq C$, then $A \leq C$.
\end{enumerate}
\end{definition}

\begin{remark}
    If a pairwise relation $\pi$ on a set of elements of $\mb{V}$ is a partial ordering, then for elements $A,B \in \mb{V}$ such that $\pi(A,B)$ holds, we will also write $A \le_{\pi} B$. Note also, that not every two elements of $\mb{V}$ need to be comparable to have a partial ordering on $\mb{V}$. For distinct elements $A$ and $B$ in $\mb{V}$, if $A \not\le B$ and $B \not\le A$, then we say that $A$ and $B$ are incomparable and we denote this by $A \incomp B$ or, equivalently, $B \incomp A$ \citep{trotter1992combinatorics}.
\end{remark}

\begin{definition}[Extending Orders]\label{def:extends}
A partial order $\pi_1$ is an extension of a partial order $\pi_2$ if and only if $A \le_{\pi_2} B$ implies $ A \le_{\pi_1} B$.
\end{definition}

\begin{definition}[Compatible Order]\label{def:comp-order}
Let $\g = (\mb{V,E})$ be a partial mixed graph.  A partial order $\pi$ over $\mb{V}$  is compatible with $\g$ if and and only if for any pair of nodes $A$ and $B$ in $\g$
\begin{itemize}
    \item if $A \to B$ is in $\g$, then $A \le_{\pi} B$,
    \item if  $A \arrowbullet B$ is in $\g$, then $A \not \le_{\pi} B$.
\end{itemize}
\end{definition}

 \begin{definition}[Induced Orientation]\label{def:induced-p-orientation}  Let $\g = (\mb{V,E})$ be a partially directed mixed graph and let $\le_{\alpha}$ be a partial order on $\mb{V}$ that is compatible with $\g$. Then  $\le_{\alpha}$  induces a partial orientation as follows:
\begin{itemize}
    \item if $A \circbullet B$ is in $\g$ and $A \le_{\alpha} B$, or $\alpha(A,B)$, then orient $A \to B$.
\end{itemize}
The graph resulting from applying the above procedure is called $\g_{\alpha}$.
\end{definition}

\begin{lemma}\label{lemma:ancestral-prop-po} 
 Let $\g = (\mb{V,E})$ be a partially directed ancestral mixed graph. Let $\pi$ be a relation on the nodes of $\g$ induced by the ancestral relationships. That is $\pi(A, B)$ if and only if $A \in \An(B, \g)$. 
 Then $\pi$ is a partial ordering of $\mb{V}$ that is compatible with $\g$. 
\end{lemma}

\begin{proofof}[Lemma \ref{lemma:ancestral-prop-po}]
    By definition, every node in $\g$ is an ancestor of itself, hence $\pi$ is a reflexive relationship. To show that $\pi$ is antisymmetric note that $\g$ is ancestral, so if $A \in \An(B, \g)$, that is $\pi(A,B)$ and $B \in \An(A, \g)$, that is $\pi(B,A)$ holds, we must have $A \equiv B$.
   The transitive property also holds by definition. Therefore, $\pi$ is a partial ordering that is naturally compatible with $\g.$
\end{proofof}

\begin{definition}[Tree Graph]\label{def:tree}
A graph $\mathcal{T} = (\bV, \bE)$ is a tree if for any pair of nodes $A, B \in \bV$, there is exactly one path $p = \langle A = V_1, \dots, B = V_k \rangle$ in $\mathcal{T}$.
\end{definition}

\begin{definition}[Join Tree Graph]\label{def:join-tree}
Let $\g = (\bV, \bE)$ be a graph.  A join tree graph $\g[T] = (\mathbf{C}, \bE^{\prime})$ for $\g$ is an undirected tree graph whose nodes $\mathbf{C}$ are  a partition of $\bV$ with the following properties:
\begin{enumerate}[label=(\roman*)]
    \item for set of nodes $\mathbf{A} \subseteq \mb{V}$ that forms a maximal clique in $\g$,  $\mathbf{A} \equiv \mathcal{C}_{i}$, for some $\mathcal{C}_{i} \in \mathbf{C}$, and 
    \item (running intersection) for each pair $\mathcal{C}_{i}, \mathcal{C}_{j} \in \mathbf{C}$ such that $A \in (\mathcal{C}_{i} \cap \mathcal{C}_{j}) \subseteq \bV$,
    each node $\mathcal{C}_{k}$ on the unique path between  $\mathcal{C}_{i}$  and $\mathcal{C}_{j}$ in $\g[T]$ also contains $A$.
\end{enumerate}
\end{definition}

\begin{remark}
    Join trees are sometimes also called junction trees or chordal trees, due to the fact that only chordal graphs have a join tree.
We state the original result of \citet{beeri1983desirability} in Lemma \ref{thm:beeri-chordal-join-tree}. We refer the reader to \citet{jensen1994optimal} and \citet{lauritzen1996graphical} for a modern treatment of join trees and how to construct them. 
\end{remark}

\begin{lemma}[Theorem 3.4 of \citet{beeri1983desirability}]
\label{thm:beeri-chordal-join-tree}
A graph $\g = (\bV, \bE)$ has a join tree if and only if $\g$ is chordal.
\end{lemma}

\textbf{$\Lambda_{ij}$ notation.} Let $\g = (\bV, \bE)$ be a graph with a chordal skeleton.  For maximal cliques $\g[C]_i, \g[C]_j \subseteq \mb{V}$, we will use $\Lambda_{ij}$ to denote their intersection, that is, $\Lambda_{ij} = \g[C]_i \cap \g[C]_j$.

\begin{definition}[$\gamma$-relation] \label{def:gamma-order} 
  Let $\g = (\bV, \bE)$ be an ancestral partial mixed graph such that the skeleton of $\g$ is chordal and $\g$ contains no minimal collider paths. Let $\g[T] = (\mathbf{C}, \bE^{\prime})$ be an undirected join tree graph for $\g$. 
  Let $\g[C]_i, \g[C]_j \in \mathbf{C},$ and $\Lambda_{ij} = \g[C]_i \cap \g[C]_j$. 
    We define a relation $\gamma$ on the nodes of $\g[T]$ as follows: $\gamma(\g[C]_i, \g[C]_j)$ if and only if
\begin{enumerate}[label=(\roman*)]
\item $\Lambda_{ij} \neq \varnothing$, 
\item for all $B \in \Lambda_{ij}$ and $C \in \g[C]_j \setminus \Lambda_{ij}$, $B \to C$ is in $\g$, and
\item { there exist nodes $A \in \g[C]_i \setminus \Lambda_{ij}$ and  $B \in \Lambda_{ij}$ such that $A \bulletarrow B$ is in $\g$. } 
\end{enumerate}
\end{definition}

\begin{definition}[Partially Directed Join Tree]\label{def:partially-directed-join-tree}
     Let $\g = (\bV, \bE)$ be an ancestral partial mixed graph such that the skeleton of $\g$ is chordal and $\g$ contains no minimal collider paths. Let $\g[T] = (\mathbf{C}, \bE^{\prime})$ be an undirected join tree graph for $\g$ and let $\gamma$ be a relation on the nodes of $\g[T]$ defined in Definition \ref{def:gamma-order}. 
     We define a partially directed join tree graph $\g[T]_{\gamma} = (\mathbf{C}, \bE^{''})$ as follows:
     \begin{enumerate}[label = (\roman*)]
         \item The skeleton of $\g[T]_{\gamma}$ is identical to the skeleton of $\g[T]$.
         \item Edge $\langle \g[C]_i, \g[C]_j \rangle$ in $\g[T]$ corresponds to:
         \begin{itemize}
             \item $\g[C]_i \to \g[C]_j$ in $\g[T]_{\gamma}$ if $\gamma(\g[C]_i, \g[C]_j)$,
             \item $\g[C]_i \leftarrow \g[C]_j$ in $\g[T]_{\gamma}$ if $\gamma(\g[C]_j, \g[C]_i)$, and 
             \item $\g[C]_i - \g[C]_j$ in $\g[T]_{\gamma}$ if neither $\gamma(\g[C]_i, \g[C]_j)$ nor $\gamma(\g[C]_j, \g[C]_i)$.
         \end{itemize}
     \end{enumerate}
\end{definition}

\begin{remark}\label{red:arborescence}
    Note that the partially or fully directed trees we consider are not always arborescences in the graph theory sense. Meaning that our definition of a partially directed tree allows for more than one root node.
\end{remark}

\begin{definition}[Anchored Tree]\label{def:anchored}
Let $\g[T] = (\mathbf{C}, \bE^{\prime})$ be a partially directed tree graph and let $\g[C]_0 \in \mathbf{C}$. We say that $\g[T]$ is anchored around $\g[C]_0$ if $\PossAn(\g[C]_0, \g[T]) = \An(\g[C]_0, \g[T])$.
\end{definition}

\begin{definition}[Join Tree Induced Edge Orientations]
\label{def:jointree-inducedorder} 
 Let $\g = (\bV, \bE)$ be an ancestral partial mixed graph such that the skeleton of $\g$ is chordal and $\g$ contains no minimal collider paths.  Let $\g[T] = (\mathbf{C}, \bE^{\prime})$ be a partially directed join tree graph for $\g$ (Definition \ref{def:partially-directed-join-tree}) and suppose that $\pi_{\g[T]}$ is a partial ordering compatible with $\g[T]$, such that $\g[T]_{\pi_{\g[T]}}$ is a directed graph with no  colliders. Then, $\pi_{\g[T]}$ induces orientations  on the nodes of $\g$ using the following rule: 
\begin{enumerate}[label=(\roman*)]
    \item if $\pi_{\g[T]}(\g[C]_i, \g[C]_j)$, 
    then for all  $B \in \g[C]_i \cap \g[C]_j$ and $C \in \g[C]_j \setminus \g[C]_i$, orient $B  \to C$.
\end{enumerate}
The graph obtained as a result of this operation is called $\g_{\pi}$.
\end{definition}

\subsection{Main Result}\label{sec:meek-analogue-main}

\begin{figure}[!t]
    \centering
      \begin{tikzpicture}[[->,>=stealth', auto,node distance=0.8cm,scale=.8,transform shape,font = {\large\sffamily}]
  \tikzstyle{state}=[inner sep=2pt, minimum size=12pt]

 \node[state] (LB20) at (2,-2) {Lemma \ref{lemma:join-tree-single-operation}};
   \node[state] (LB21) at (-2,-2) {Lemma \ref{lemma:transform-join-tree-helper-terminates}};
 \node[state] (LB14) at (8,-2) {Lemma \ref{lemma:order-cliques-1}};

\node[state] (LB16) at (5,-4) {Lemma \ref{lemma:rule-1-join-tree}};
\node[state] (LB22) at (1.5,-4) {Lemma \ref{lemma:transform-join-tree-helper}};
 \node[state] (LB27) at (-2.5,-4) {Lemma \ref{lemma:join-tree-single-operation2}};

   \node[state] (LB28) at (-2,-6) {Lemma \ref{lemma:alg-transform-tree-terminates}};
    \node[state] (CB29) at (2,-6) {Corollary \ref{cor:algo-transformtree1-result}};
    \node[state] (LB30) at (6,-6) {Lemma \ref{lem:correct-directed-tree}};
  \node[state] (T13) at (10,-6) {\textbf{Theorem \ref{thm:chordal-completeness}}}; 

    \node[state] (LB32) at (9,-4) {Lemma \ref{lemma:existence-spo2}}; 
    \node[state] (LB33) at (11,-4) {Lemma \ref{lemma:existence-spo}}; 
  
 \draw[->,arrows= {-Latex[width=5pt, length=5pt]}] (LB32) edge (T13);
 \draw[->,arrows= {-Latex[width=5pt, length=5pt]}] (LB33) edge (T13);

\draw[->,arrows= {-Latex[width=5pt, length=5pt]}] (LB30) edge (T13);
\draw[->,arrows= {-Latex[width=5pt, length=5pt]}] (CB29) edge (LB30);
\draw[->,arrows= {-Latex[width=5pt, length=5pt]}](LB28) edge (CB29);

\draw[->,arrows= {-Latex[width=5pt, length=5pt]}] (LB27) edge (LB28);
\draw[->,arrows= {-Latex[width=5pt, length=5pt]}] (LB27) edge (CB29);

\draw[->,arrows= {-Latex[width=5pt, length=5pt]}] (LB21) edge (CB29);
\draw[->,arrows= {-Latex[width=5pt, length=5pt]}] (LB21) edge (LB22);

\draw[->,arrows= {-Latex[width=5pt, length=5pt]}] (LB22) edge (CB29);
\draw[->,arrows= {-Latex[width=5pt, length=5pt]}] (LB22) edge (LB27);

\draw[->,arrows= {-Latex[width=5pt, length=5pt]}] (LB20) edge (LB21);
\draw[->,arrows= {-Latex[width=5pt, length=5pt]}] (LB20) edge (LB22);

\draw[->,arrows= {-Latex[width=5pt, length=5pt]}] (LB16) edge (LB22);
\draw[->,arrows= {-Latex[width=5pt, length=5pt]}] (LB14) edge (LB16);
\draw[->,arrows= {-Latex[width=5pt, length=5pt]}] (LB14) edge (LB30);

\draw[->, out= 180, in =180,arrows= {-Latex[width=5pt, length=5pt]}] (LB21) edge (LB28);
\end{tikzpicture}
   \caption{Proof structure of Theorem \ref{thm:chordal-completeness}}
    \label{figproof-chordal}
\end{figure}

\begin{proofof}[Theorem \ref{thm:chordal-completeness}] 
Let $\mathcal{T}_{0}$ be a partially directed join tree of $\g^{\prime}$ (Definition \ref{def:partially-directed-join-tree}).
There is at least one clique $\g[C]_0$ such that $A, B \in \g[C]_0$.  We will first transform the partially directed join tree $\mathcal{T}_0$ of $\g^{\prime}$ into another partially directed join tree  of $\g^{\prime}$  called $\g[T]_1$ using the \texttt{transformTree} algorithm (Algorithm \ref{alg:join-tree-transformation} in Section \ref{sec:finding-right-jointree}), that is, $\g[T]_1 = \texttt{transformTree}(\g[T]_0, \g[C]_0)$. By Corollary \ref{cor:algo-transformtree1-result}, the partially directed join tree $\g[T]_1$ is anchored around clique $\g[C]_0$ (Definition \ref{def:anchored}), meaning that $\PossAn(\g[C]_0, \g[T]_1) = \An(\g[C]_0, \g[T]_1)$. Furthermore,  there are no unshielded colliders in $\g[T]_1$, or paths of the form $\g[C]_i \to \g[C]_j - \dots  -  \g[C]_k \leftarrow \g[C]_l$. 
Note that unshielded colliders, or paths of the form  $\g[C]_i \to \g[C]_j - \dots  -  \g[C]_k \leftarrow \g[C]_l$ can occur in an arbitrarily chosen partially directed join tree as per Figures \ref{fig:lemma7-issue-p1-1}-\ref{fig:lemma7-issue-p1-3} (see also the associated examples for more details). 
To construct join tree $\g[T]_1$, Algorithm \ref{alg:join-tree-transformation} relies on a few supporting algorithms (Algorithms \ref{alg:join-tree-transformation-helper}, \ref{alg:paths-for-join-tree-transformation}) and results in Sections \ref{sec:gen-join-tree-l4} and \ref{sec:finding-right-jointree}.

Next, we close the orientations in $\g[T]_1$ using algorithm \texttt{orientTree} (Algorithm \ref{alg:orient-join-tree} in Section \ref{sec:orienting-join-tree-and-cliques}), to construct a directed join tree $\g[T]$, that is, $\g[T] = \texttt{orientTree}(\g[T]_1, \g[C]_0)$. Let  $\pi_{\g[T]}$ be a partial order compatible with $\g[T]$. 
By case \ref{correct-directed-tree:case2} of Lemma \ref{lem:correct-directed-tree}, $\pi_{\g[T]}$ induces edge orientations that are compatible with $\g^{\prime}$ through the process described in Definition \ref{def:jointree-inducedorder}. 

Therefore, let $\g^{\prime}_{\pi}$ be the graph obtained from applying $\pi_{\g[T]}$  to $\g^{\prime}$ as in Definition \ref{def:jointree-inducedorder}. Then $\langle A, B \rangle$ is of the same form in $\g^{\prime}$ and $\g^{\prime}_{\pi}$ (\ref{correct-directed-tree:case4} of Lemma \ref{lem:correct-directed-tree}).
Furthermore, by case \ref{correct-directed-tree:case3} of Lemma \ref{lem:correct-directed-tree}, $\g^{\prime}_{\pi}$ is an ancestral partial mixed graph with no minimal collider paths and edge orientations closed under \ref{R2}, and \ref{R8}. 
Additionally, any ancestral directed mixed graph $\g[M]$ that is represented by $\g^{\prime}_{\pi}$ will be a MAG represented by $\g^{\prime}$.

Observe that all edges in $\g^{\prime}_{\pi}$ that are between two cliques are invariant. All variant edges ($\circarrow$ or $\circcirc$) are only present inside of cliques of $\g^{\prime}_{\pi}$. Therefore, to construct a MAG $\g[M]$ represented by $\g^{\prime}$ with the desired orientation of $\langle A, B \rangle$ edge, we now only need to orient $\g^{\prime}_{\pi}$ into an ancestral directed mixed graph. For this, it is enough to ensure that no directed or almost directed cycle is created within the maximal cliques of $\g^{\prime}_{\pi}$ when orienting it into $\g[M]$. To do this, we rely on Lemmas \ref{lemma:existence-spo2} and \ref{lemma:existence-spo} in Section \ref{sec:clique}, which give us two alternate procedures for orienting partially oriented cliques in $\g^{\prime}_{\pi}$ with desired edge marks on $\langle A, B \rangle$. 
\end{proofof}

\subsection{General Partially Directed Join Tree Properties}\label{sec:gen-join-tree-l4}

\begin{lemma}
\label{lemma:order-cliques-1} 
Let $\g$ be an ancestral partial mixed graph with a chordal skeleton such that $\g$ has no minimal collider paths such that the orientations in $\g$ are closed under \ref{R1} and \ref{R11}. Let $\mathcal{T}$ be a join tree for $\g$ and $\gamma$ a relation as defined in Definition \ref{def:gamma-order}. 
{ Let $\mc{C}_{i}$ and $\mc{C}_j$ be adjacent in $\g[T]$, and suppose that there is an}
unshielded triple $\langle A, B, C \rangle$ such that $A \bulletarrow B$ in $\g$,
and $A, B \in \g[C]_i$, $B, C \in \g[C]_j$, $A \notin \g[C]_j$, $C \notin \g[C]_i$. Then $\gamma(\g[C]_i, \g[C]_j)$.
\end{lemma}

\begin{figure}[!t]
    \centering
    \begin{subfigure}{.45\textwidth}
          \tikzstyle{every edge}=[draw,>=stealth',semithick]
        \vspace{1cm}
        \centering
        \begin{tikzpicture}[->,>=stealth',shorten >=1pt,auto,node distance=1.2cm,scale=0.7,transform shape,font = {\Large\bfseries\sffamily}]
	\tikzstyle{state}=[inner sep=5pt, minimum size=5pt]

	\node[state] (A) at (-3,0) {$A$};
	\node[state] (B) at (0,-1.5) {$B$};
	\node[state] (D) at (0,1.5) {$D$};
	\node[state] (C) at (3,0) {$C$};
	\draw[blue,->]  (B) edge (C);
	\draw[blue,->] (D) edge (C);
	\draw[*-*] (D) edge (A);
	\draw[*->] (A) edge (B);
	\draw[*-*] (B) edge (D);

	\node[state] (Ci) at (-4.5,0) {$\g[C]_i$};
	\node[state] (Cj) at (4.5,0) {$\g[C]_j$};

        \draw[black, thick] (-1.4,0) circle (2.5);
        \draw[black, thick] (1.4,0) circle (2.5);
  \end{tikzpicture}
\caption{}
\label{fig:meek-rule4-lemma} 
    \end{subfigure}
    \vrule
    \begin{subfigure}{.45\textwidth}
        \centering
       \tikzstyle{every edge}=[draw,>=stealth',semithick]
	\begin{tikzpicture}[->,>=stealth',shorten >=1pt,auto,node distance=1.2cm,scale=0.7,transform shape,font = {\Large\bfseries\sffamily}]
	\tikzstyle{state}=[inner sep=5pt, minimum size=5pt]

	\node[state] (A) at (-3,0) {$A$};
	\node[state] (B) at (0,-1.5) {$B$};
	\node[state] (D) at (0,1.5) {$D$};
        \node[state] (E) at (3,1.5) {$E$};
	\node[state] (C) at (3,-.5) {$C$};

	\draw[blue,->]  (B) edge (C);
	\draw[blue,->] (D) edge (C);
	\draw[*-*] (D) edge (A);
	\draw[*->] (A) edge (B);
	\draw[*-*] (B) edge (D);
 	\draw[*-*] (B) edge (E);
	\draw[*-*] (A) edge (E);
	\draw[*-*] (C) edge (E);
 	\draw[*-*] (D) edge (E);
  
	\node[state] (Ci) at (-4.5,0) {$\g[C]_i$};
        \node[state] (Cj) at (4.5,0) {$\g[C]_j$};
    
        \draw[black, thick] (-1.4,0) circle (2.5);
        \draw[black, thick] (1.4,0) circle (2.5);

  \end{tikzpicture}
\caption{}
\label{fig:meek-rule4-lemma2}
    \end{subfigure}
    \caption{Used in proof of Lemma \ref{lemma:order-cliques-1}.}
    \label{fig:meek-lemma6-proof-graph}
\end{figure}
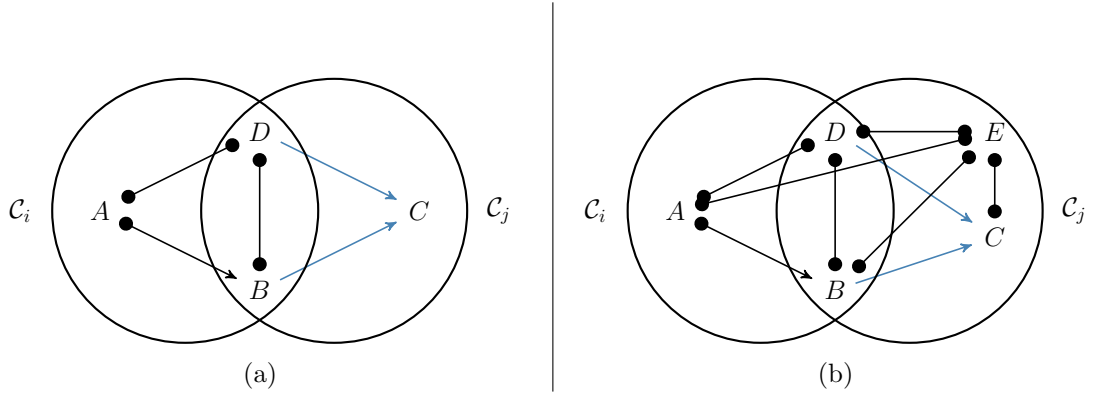

\begin{proofof}[Lemma \ref{lemma:order-cliques-1}]
Since $A \bulletarrow B$ is in $\g$ and since $\Lambda_{ij} \neq \varnothing$, for $\gamma(\mc{C}_{i},\mc{C}_{j})$ it is enough to show that for all $D \in \Lambda_{ij}$, $E \in \mc{C}_{j} \setminus \mc{C}_{i}$, $D \to E$ is in $\g$. There are three cases:

\begin{enumerate}[label=(\roman*)]
\item If $D \equiv B$, then for $E \equiv C$, or $E \notin \Adj(A, \g)$, $\langle A, D, E \rangle$, forms an unshielded triple in $\g$. Since $\g$ does not contain unshielded colliders,  by  \ref{R1}, we conclude that 
$D \to E$ is in $\g$.

\item For $D \not \equiv B$ but $E \equiv C$, we have that $\langle B,D \rangle$ edge is in $\g$ since $B,D \in \Lambda_{ij}$. Since $\g$ does not contain unshielded colliders or longer minimal collider paths, we conclude by \ref{R11} (see Figure \ref{fig:meek-rule4-lemma}) that $D \to C$, that is $D \to E$ is in $\g$.

\item For $D \not \equiv B$ and $E \not \equiv C$, we know that, $\langle B,D \rangle$ edge is in $\g$ since $B,D \in \Lambda_{ij}$ and also that $\langle D,C \rangle$, $\langle E,C \rangle$  are in $\g$, since $B,C,D,E \in \mc{C}_{j}.$
If $A \notin \Adj(E, \g)$, then as in the cases above, by \ref{R1} $B \to E$ is in $\g$ and by \ref{R11}, $D \to E$ is also in $\g$ and we are done.

Otherwise, $A \in \Adj(E, \g)$ as in Figure \ref{fig:meek-rule4-lemma2}.
However, this case is not possible. For sake of contradiction assume that this is possible.
Note that $A,B,D,E$ form a clique in $\g$, but since $E \notin \mc{C}_i$, there must be another maximal clique in $\g$, $\mc{C}_k$ that is a node in $\g[T]$, such that $A, B,D, E \in \mc{C}_{k}.$
Furthermore, $C \notin \mathcal{C}_{k}$, because $A \notin \Adj(C, \g)$.

There cannot be a path from $\mc{C}_{k}$ to $\mc{C}_{j}$ in $\g[T]$ that contains $\mc{C}_{i}$ as that violates the running intersection property ($\mc{C}_{k} \cap \mc{C}_{j} \subseteq \{B,D,E\} \not\subseteq \mc{C}_i$ as $E \notin \mc{C}_i$). 

Similarly, there is no path from $\mc{C}_{i}$ to $\mc{C}_{k}$ in $\g[T]$ that contains $\mc{C}_j$ as that also violates the running intersection property ($\mc{C}_{i} \cap \mc{C}_{k} \subseteq \{A,B,D\} \not \subseteq \mc{C}_j$ since $A \notin \mc{C}_j$).

And since we assume that $\mc{C}_{i}$ and $\mc{C}_{j}$ are adjacent in $\g[T]$ there cannot be a path from $\mc{C}_{i}$ to $\mc{C}_{j}$ that contains $\mc{C}_{k}$. Thus, we have a contradiction to $A \in \Adj(E, \g)$.
\end{enumerate}
Therefore, we have shown that $\gamma(\g[C]_i, \g[C]_j)$.
\end{proofof}

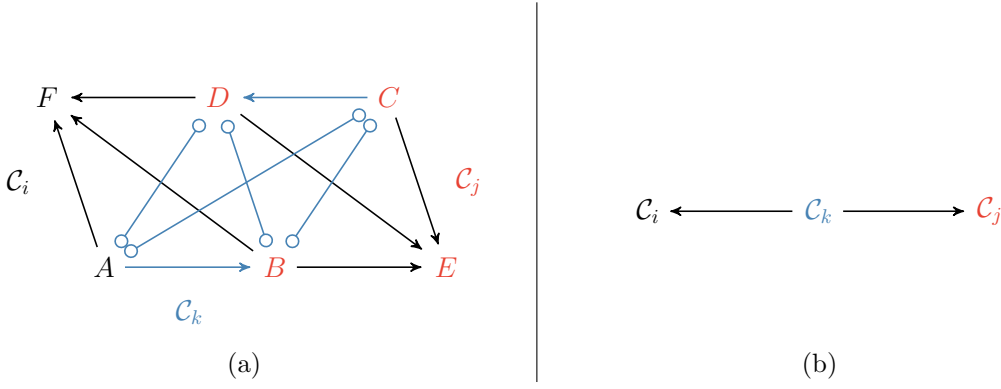
\begin{figure}[!t]
    \centering
    \begin{subfigure}{.45\textwidth}
          \tikzstyle{every edge}=[draw,>=stealth',semithick]
        \vspace{1cm}
        \centering
        \begin{tikzpicture}[->,>=stealth',shorten >=1pt,auto,node distance=1.2cm,scale=0.75,transform shape,font = {\Large\bfseries\sffamily}]
	\tikzstyle{state}=[inner sep=5pt, minimum size=5pt]

	\node[state] (A) at (-3,0) {$A$};
	\node[state, color = red] (B) at (0,0) {$B$};
        \node[state, color = red] (E) at (3,0) {$E$};
        \node[state] (F) at (-4,3) {$F$};
	\node[state, color = red] (D) at (-1,3) {$D$};
	\node[state, color = red] (C) at (2,3) {$C$};
         
	\draw[->, color = blue] (A) edge (B);
 	\draw[->, color = blue] (C) edge (D);
	\draw[->]  (B) edge (E);
 	\draw[->]  (C) edge (E);
   	\draw[->]  (D) edge (E);
	\draw[->] (D) edge (F);
	\draw[->] (B) edge (F);
 	\draw[->] (A) edge (F);

	\draw[o-o, color = blue] (A) edge (D);
        \draw[o-o, color = blue] (B) edge (D);
	\draw[o-o, color = blue] (A) edge (C);
        \draw[o-o, color = blue] (B) edge (C);

	\node[state] (Ci) at (-4.5,1.5) {$\g[C]_i$};
	\node[state, color = blue] (Ck) at (-1.5,-.8) {$\g[C]_k$};
 	\node[state, color = red] (Cj) at (3.4,1.5) {$\g[C]_j$};
  \end{tikzpicture}
\caption{}
\label{fig:graphglemma6} 
    \end{subfigure}
    \vrule
    \begin{subfigure}{.45\textwidth}
        \centering
       \tikzstyle{every edge}=[draw,>=stealth',semithick]
	\begin{tikzpicture}[->,>=stealth',shorten >=1pt,auto,node distance=1.2cm,scale=0.75,transform shape,font = {\Large\bfseries\sffamily}]
	\tikzstyle{state}=[inner sep=5pt, minimum size=5pt]

	\node[state] (J1i) at (-3,0) {$\g[C]_i$};
	\node[state, color = blue] (J1j) at (0,0) {$\g[C]_k$};
	\node[state, color = red] (J1k) at (3,0) {$\g[C]_j$};
       
        \draw[<-]  (J1i) edge (J1j);
        \draw[->]  (J1j) edge (J1k);

	\node[state] (J1l) at (0,-2) {};
  \end{tikzpicture}
\caption{}
\label{fig:graphlemma6-join-trees}
    \end{subfigure}
    \caption{\ref{fig:graphglemma6} partially mixed graph $\g$, \ref{fig:graphlemma6-join-trees} partially directed join tree $\g[T]$ of $\g$. These graphs are used in Example \ref{ex:lemma6-meek-issue}.}
    \label{fig:meek-lemma6-issue}
\end{figure}

\begin{example}\label{ex:lemma6-meek-issue}
   The condition of $\g[C]_i$ and $\g[C]_j$ being adjacent in the join tree is necessary for Lemma \ref{lemma:order-cliques-1} to hold. As an example illustrating this, consider the graphs in Figure \ref{fig:meek-lemma6-issue}. A partially directed ancestral mixed graph $\g = (\mb{V,E})$ in Figure \ref{fig:graphglemma6} has orientations that are    closed under  { \ref{R1}, \ref{R2}, \ref{R8}, \ref{R11}} 
  % \ref{R1}-\ref{R4}, \ref{R8}-\ref{R13} 
  and a chordal skeleton. In fact, the essential ancestral graph of $\g$ is fully undirected. 
   
   Three maximal cliques make up $\mb{V}$. These are $\g[C]_i = \{A,B,D,F\}, \g[C]_k = \{A,B,C,D\}$, and $\g[C]_j = \{B,C,D,E\}$. A partially directed join tree of $\g$, called $\g[T]$ is given in Figure \ref{fig:graphlemma6-join-trees}.
    In fact, $\g[T]$ is the only valid join tree of $\g$, since $\g[C]_k$ is a separator for $\g[C]_i$ and $\g[C]_j$.
    
    Now, note that $\g[C]_i$ and $\g[C]_j$ are not adjacent in $\g[T]$, but otherwise, satisfy conditions of Lemma \ref{lemma:order-cliques-1}. Notably also, $\Lambda_{ij} = \{B,D\}, \g[C]_i \setminus \Lambda_{ij} = \{A,F\}$, and $\g[C]_j \setminus \Lambda_{ij} = \{C,E\}$. However, looking at $\g$, we can conclude that $\neg \gamma(\g[C]_i, \g[C]_j)$ because $D \leftarrow C$ is in $\g$, and also $\neg \gamma(\g[C]_j, \g[C]_i)$ because $A \to B$ is in $\g$. Hence, this adjacency condition is necessary for Lemma \ref{lemma:order-cliques-1} to hold.
\end{example}

Lemma 6 of \citet{meek1995causal}, which is the analogous result to our Lemma \ref{lemma:order-cliques-1} does not require $\g[C]_i$ and $\g[C]_j$ being adjacent in the join tree. As shown in Example \ref{ex:lemma6-meek-issue}, this condition is necessary. In the following results, we show how to transform any join tree into another join tree that satisfies this adjacency condition. Thus, we provide a correct proof of Theorem 4 of \citet{meek1995causal}.

Based on the result of Lemma \ref{lemma:order-cliques-1}, one may assume that paths $\g[C]_1 \to \g[C]_2 - \g[C]_3$, or $\g[C]_1 \to \g[C]_2 \leftarrow \g[C]_3$ cannot occur in some partially directed join tree $\g[T]$. 
We consider this in Lemma \ref{lemma:rule-1-join-tree}, and show that contrary to the above intuition, the general join tree properties do not preclude such paths from existing. Subsequently, in Examples \ref{ex:l7p11}-\ref{ex:l7p13} and, later, in Example \ref{ex:l7p2}, we showcase a few partially directed join trees where such paths do occur.

We follow up Examples \ref{ex:l7p11}-\ref{ex:l7p13} with a result (Lemma \ref{lemma:join-tree-single-operation}) that shows how to move within the partially directed join tree space to a different partially directed join tree of $\g$ where some of these paths do not occur. Algorithm \ref{alg:join-tree-transformation-helper} operationalizes this result, and we show in Lemma \ref{lemma:transform-join-tree-helper} that the result of applying  Algorithm \ref{alg:join-tree-transformation-helper} is a partially directed join tree with our desired properties. Moreover, case \ref{case4:transformedtree1} of  Lemma \ref{lemma:transform-join-tree-helper} shows that the partially directed join tree resulting from the application of Algorithm \ref{alg:join-tree-transformation-helper} does not contain colliders. We demonstrate the Algorithm \ref{alg:join-tree-transformation-helper} in Examples \ref{ex:l7p111}-\ref{ex:l7p113}.

\begin{lemma}
\label{lemma:rule-1-join-tree}
Let $\g$ be an ancestral partial mixed graph with a chordal skeleton such that $\g$ has no minimal collider paths such that the orientations in $\g$ are closed under \ref{R1} and \ref{R11}. Let $\g[T]$ be a partially directed join tree for $\g$ as defined in Definition \ref{def:partially-directed-join-tree}. 

Consider any two  nodes $\g[C]_i$ and $\g[C]_j$ adjacent in $\g[T]$,  such that $\gamma(\g[C]_i, \g[C]_j)$. If there is node $\g[C]_k$ in $\g[T]$ that is distinct from $\g[C]_i$ and such that $\g[C]_j$ and $\g[C]_k$ are adjacent in $\g[T]$, then one of the following holds:
\begin{enumerate}[label = (\roman*)]
    \item\label{r1-lemma:case1} $\gamma(\g[C]_j, \g[C]_k)$, or
    \item\label{r1-lemma:case2} $\neg \gamma(\g[C]_j, \g[C]_k)$ and $\Lambda_{ik} = \Lambda_{jk} \subseteq \Lambda_{ij}$, or
    \item\label{r1-lemma:case3} $\neg \gamma(\g[C]_j, \g[C]_k)$ and $\Lambda_{ik} = \Lambda_{ij} \subset \Lambda_{jk}$. In this case,  $\gamma(\g[C]_i,\g[C]_k)$ holds. 
\end{enumerate}
\end{lemma}

\begin{proofof}[Lemma \ref{lemma:rule-1-join-tree}]
Let $\Lambda_{ij} = \g[C]_i \cap \g[C]_j$, and $\Lambda_{jk} = \g[C]_j \cap \g[C]_k$. By assumption, $\Lambda_{ij} \neq \varnothing \neq \Lambda_{jk}.$  Furthermore, by definition of $\gamma$, for all $B \in \Lambda_{ij}$ and $C \in \g[C]_j \setminus \Lambda_{ij}$, $B \to C$ and there is at least one $A \in \g[C]_i \setminus \Lambda_{ij}$ and $B \in \Lambda_{ij},$ such that $A \bulletarrow B$ is in $\g$. Note that $\g[C]_i$ and $\g[C]_k$ are not adjacent in $\g[T]$, because $\g[T]$ is a tree.
We also know that $\g[C]_{i} \cap \g[C]_k = \Lambda_{ik} \subseteq \g[C]_j$ by the running intersection property.
Consider the following possibilities:

\begin{enumerate}[label=(\alph*)]
    \item\label{jointtreelemma:case1}   $(\g[C]_j \setminus \Lambda_{ij}) \cap \Lambda_{jk} = \varnothing$.
    \item\label{jointtreelemma:case2} $(\g[C]_j \setminus \Lambda_{ij}) \cap \Lambda_{jk} \neq \varnothing$ and $(\g[C]_j \setminus \Lambda_{jk}) \cap \Lambda_{ij} \neq \varnothing$. 
    \item\label{jointtreelemma:case3}  $(\g[C]_j \setminus \Lambda_{ij}) \cap \Lambda_{jk} \neq \varnothing$ and $(\g[C]_j \setminus \Lambda_{jk}) \cap \Lambda_{ij} = \varnothing$.
\end{enumerate}

Cases \ref{jointtreelemma:case1}-\ref{jointtreelemma:case3} are mutually disjoint by construction and exhaust all possibilities for the relationship between $\g[C]_i, \g[C]_j,$ and $\g[C]_k$. We will show that they correspond to certain cases of Lemma \ref{lemma:rule-1-join-tree}.
In the proof below,  we make use of the following three set identities:
\begin{align}
    \text{For any two sets } \mc{X}, \mc{Y}\text{ such that } \mc{Y}\subseteq \mc{X}, & \text{ then } \mc{Y} = \mc{Y} \cap \mc{X}.  \label{jointreelemma:set-identity-1} \\
    \text{For any three sets } \mc{X}, \mc{Y}, \mc{Z} \text{ such that } \mc{Y} \subset \mc{Z}, & \text{ then } (\mc{X} \setminus \mc{Z}) \subset \mc{X} \setminus \mc{Y}.  \label{jointreelemma:set-identity-2} \\
    \text{For any three sets } \mc{X}, \mc{Y}, \mc{Z} \text{ such that } \mc{Y}, \mc{Z} \subseteq \mc{X}, & \text{ then } (\mc{X} \setminus \mc{Y}) \cap \mc{Z} = \varnothing \iff \mc{Z} \subseteq \mc{Y}.  \label{jointreelemma:set-identity-3}
\end{align}

\begin{enumerate}
    \item[\ref{jointtreelemma:case1}]
By identity \eqref{jointreelemma:set-identity-3} on $(\mc{C}_j, \Lambda_{ij}, \Lambda_{jk})$, we have $\Lambda_{jk} \subseteq \Lambda_{ij}$.
This, along with identity \eqref{jointreelemma:set-identity-1}, allows us to write
\begin{align*}
    \Lambda_{jk} = (\Lambda_{jk} \cap \g[C]_k) \subseteq (\Lambda_{ij} \cap \g[C]_k) = \Lambda_{ik}.
\end{align*}
Running intersection ($\Lambda_{ik} \subseteq \g[C]_j$) tells us $\Lambda_{ik} = (\Lambda_{ik} \cap \g[C]_k) \subseteq (\g[C]_j \cap \g[C]_k) = \Lambda_{jk}$.

Hence, we have that $\Lambda_{ik} = \Lambda_{jk} \subseteq \Lambda_{ij}.$
Therefore, if $\gamma(\g[C]_j, \g[C]_k)$ we are in case \ref{r1-lemma:case1} and otherwise, we are in case \ref{r1-lemma:case2}.  

\item[\ref{jointtreelemma:case2}] There is a node $A \in (\g[C]_j \setminus \Lambda_{jk}) \cap \Lambda_{ij}$ and also a node  $B \in ( \g[C]_j \setminus \Lambda_{ij}) \cap \Lambda_{jk}$ and for any such pair of nodes $(A, B)$,  $A \to B$ is in $\g$ by assumption that $\gamma(\g[C]_i, \g[C]_j)$ holds. Now, Lemma \ref{lemma:order-cliques-1} tells us that $\gamma(\g[C]_j, \g[C]_k)$. Hence, we are in case \ref{r1-lemma:case1}.

\item[\ref{jointtreelemma:case3}]  By identity \eqref{jointreelemma:set-identity-3} on $(\mc{C}_j, \Lambda_{jk}, \Lambda_{ij})$, we have $\Lambda_{ij} \subseteq \Lambda_{jk}$.

This, along with identity \eqref{jointreelemma:set-identity-1}, allows us to write
\begin{align*}
    \Lambda_{ij} = (\Lambda_{ij} \cap \g[C]_i) \subseteq (\Lambda_{jk} \cap \g[C]_i) = (\g[C]_j \cap \g[C]_k \cap \g[C]_i) = (\g[C]_{j} \cap \Lambda_{ik}) = \Lambda_{ik},
\end{align*}
where we used the running intersection property ($\Lambda_{ik} \subseteq \g[C]_j$) and identity \eqref{jointreelemma:set-identity-1} the last step.
Running intersection also tells us $ \Lambda_{ik} = (\Lambda_{ik} \cap \g[C]_i) \subseteq \Lambda_{ij}$.

Hence, $\Lambda_{ij} = \Lambda_{ik} \subseteq \Lambda_{jk}$.

Additionally, by identity \eqref{jointreelemma:set-identity-3} on $(\mc{C}_j, \Lambda_{ij}, \Lambda_{jk})$, we have $\Lambda_{jk} \not \subseteq \Lambda_{ij}$.
Therefore, $\Lambda_{ij} = \Lambda_{ik} \subset \Lambda_{jk}$.

To show that we are now either in case \ref{r1-lemma:case1} or \ref{r1-lemma:case3}, we will prove that  $\gamma(\g[C]_i, \g[C]_k)$ holds.
 Let $A, B$ be nodes such that $A \in \g[C]_{i} \setminus \g[C]_j$, $B \in \Lambda_{ij}$, and $A \bulletarrow B$ is in $\g$. 
Since $\Lambda_{ik} \subset \g[C]_j$, identity \eqref{jointreelemma:set-identity-2} says $\g[C]_{i} \setminus \g[C]_j \setminus \g[C]_{i} \setminus \Lambda_{ik}$. Therefore, $A \in \g[C]_{i} \setminus \g[C]_k$. Further, since $\Lambda_{ij} = \Lambda_{ik}$, $B \in \Lambda_{ik}$.

Furthermore, note that for any $C \in \g[C]_k \setminus \g[C]_i$, $A \notin \Adj(C, \g)$.
For sake of contradiction, assume that there is
some $C \in \g[C]_k \setminus \g[C]_i$, such that $A \in \Adj(C, \g)$, then there also must be a maximal clique $\g[C]_r$ in $\g$, such that $A, B, C \in \g[C]_r$. However, we know that $\langle \g[C]_i, \g[C]_j, \g[C]_k \rangle$ is in $\g[T]$ meaning that either (a) every path from $\g[C]_r$ to $\g[C]_i$ contains $\g[C]_j$, or (b) every path from $\g[C]_r$ to $\g[C]_k$ contains $\langle \g[C]_i, \g[C]_j, \g[C]_k \rangle$. Now the contradiction follows from the running intersection property since we have that $\g[C]_r \cap \g[C]_i \supseteq \{A\} \not \subseteq \g[C]_j$ and $\g[C]_r \cap \g[C]_k \supseteq \{C\} \not\subseteq \g[C]_i$.

Since every node in $C \in \g[C]_k \setminus \g[C]_i$ is not adjacent to $A$, $B \to C$ is in $\g$, and for every other node $D \in \Lambda_{ik}$, $\langle B, D \rangle$ is in $\g$ and $D \to C$ is in $\g$ using the fact that orientations in $\g$ are   closed under  \ref{R1} and \ref{R11}. Therefore, $\gamma(\g[C]_i, \g[C]_k)$ holds. 
\end{enumerate}

\end{proofof}

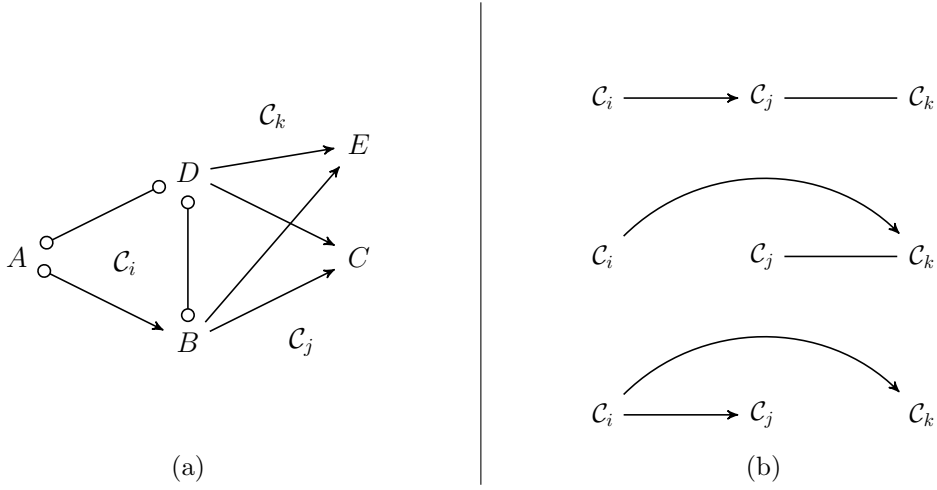
\begin{figure}[!t]
    \centering
    \begin{subfigure}{.45\textwidth}
          \tikzstyle{every edge}=[draw,>=stealth',semithick]
        \vspace{1cm}
        \centering
        \begin{tikzpicture}[->,>=stealth',shorten >=1pt,auto,node distance=1.2cm,scale=0.75,transform shape,font = {\Large\bfseries\sffamily}]
	\tikzstyle{state}=[inner sep=5pt, minimum size=5pt]

	\node[state] (A) at (-3,0) {$A$};
	\node[state] (B) at (0,-1.5) {$B$};
	\node[state] (D) at (0,1.5) {$D$};
	\node[state] (C) at (3,0) {$C$};
        \node[state] (E) at (3,2) {$E$};

	\draw[->]  (B) edge (C);
	\draw[->] (D) edge (C);
	\draw[o-o] (D) edge (A);
	\draw[o->] (A) edge (B);
	\draw[o-o] (B) edge (D);

 	\draw[->] (D) edge (E);
        \draw[->] (B) edge (E);

	\node[state] (Ci) at (-1.1,-.1) {$\g[C]_i$};
	\node[state] (Cj) at (2,-1.5) {$\g[C]_j$};
    \node[state] (Ck) at (1.5,2.5) {$\g[C]_k$};

    \node[state] (empty1) at (0,3) {};
    \node[state] (empty1) at (0,-3) {};
  \end{tikzpicture}
\caption{}
\label{fig:jointree11} 
    \end{subfigure}
    \vrule
    \begin{subfigure}{.45\textwidth}
        \centering
       \tikzstyle{every edge}=[draw,>=stealth',semithick]
	\begin{tikzpicture}[->,>=stealth',shorten >=1pt,auto,node distance=1.2cm,scale=0.7,transform shape,font = {\Large\bfseries\sffamily}]
	\tikzstyle{state}=[inner sep=5pt, minimum size=5pt]

	\node[state] (J1i) at (-3,3) {$\g[C]_i$};
	\node[state] (J1j) at (0,3) {$\g[C]_j$};
	\node[state] (J1k) at (3,3) {$\g[C]_k$};
       
        \draw[->]  (J1i) edge (J1j);
        \draw[-]  (J1j) edge (J1k);

	\node[state] (J2i) at (-3,0) {$\g[C]_i$};
	\node[state] (J2j) at (0,0) {$\g[C]_j$};
	\node[state] (J2k) at (3,0) {$\g[C]_k$};
    
        \draw[-]  (J2j) edge (J2k);
        \draw[->] (J2i) edge [out=45,in=135] (J2k);

    \node[state] (J2i) at (-3,-3) {$\g[C]_i$};
	\node[state] (J2j) at (0,-3) {$\g[C]_j$};
	\node[state] (J2k) at (3,-3) {$\g[C]_k$};
       
        \draw[->]  (J2i) edge (J2j);
        \draw[->] (J2i) edge [out=45,in=135] (J2k);
    
  \end{tikzpicture}
\caption{}
\label{fig:jointree22}
    \end{subfigure}
    \caption{\ref{fig:jointree11} Partially mixed graph $\g$, \ref{fig:jointree22} Three partially directed join trees for $\g$. These graphs are explored in Examples \ref{ex:l7p11} and \ref{ex:l7p111}. }
    \label{fig:lemma7-issue-p1-1}
\end{figure}

\begin{example}\label{ex:l7p11}
   A chordal and ancestral partial mixed graph $\g = (\mb{V,E})$ in Figure \ref{fig:jointree11} has orientations that are   closed under  { \ref{R1}, \ref{R2}, \ref{R8}, \ref{R11}}.  %\ref{R1}-\ref{R4}, \ref{R8}-\ref{R13}. 
   In fact, the essential ancestral graph of $\g$ is fully undirected. 
   
   Three maximal cliques make up $\mb{V}$. These are $\g[C]_i = \{A,B,D\}, \g[C]_j = \{B,C,D\}$, and $\g[C]_k = \{B,D,E\}$. Three different partially directed join trees for $\g$ are given in Figure \ref{fig:jointree22}. From top to bottom, these join trees are  $\g[T]_1$, $\g[T]_2$ and $\g[T]_3$. 
   As can be seen from the figure, orientations in these join trees are not necessarily   closed under  \ref{R1}. Based on $\g$, we have that $\gamma(\g[C]_{i}, \g[C]_j)$ and $\gamma(\g[C]_i, \g[C]_k)$. However, neither $\gamma(\g[C]_{j}, \g[C]_k)$, nor $\gamma(\g[C]_k, \g[C]_j)$ hold. 

\end{example}

\begin{example}\label{ex:l7p12}
     A chordal and ancestral partial mixed graph $\g = (\mb{V,E})$ in Figure \ref{fig:jointree13} has orientations that are   closed under { \ref{R1}, \ref{R2}, \ref{R8}, \ref{R11}}. %  \ref{R1}-\ref{R4}, \ref{R8}-\ref{R13}. 
     In fact, the essential ancestral graph of $\g$ is fully undirected. 
   
   Three maximal cliques make up $\mb{V}$. These are $\g[C]_i = \{A,B,D\}, \g[C]_j = \{B,C,D\}$, and $\g[C]_k = \{D,E\}$. Two partially directed join trees for $\g$ are given in Figure \ref{fig:jointree23}. From top to bottom, these join trees are  $\g[T]_1$, $\g[T]_2$. 
   As can be seen from the figure, orientations in $\g[T]_1$ are not   closed under  \ref{R1}.
  Based on $\g$, we have that $\gamma(\g[C]_{i}, \g[C]_j)$ but that is the only valid $\gamma$-relation on the maximal cliques of $\g$.
\end{example}

\begin{figure}[!t]
    \centering
    \begin{subfigure}{.45\textwidth}
          \tikzstyle{every edge}=[draw,>=stealth',semithick]
        \vspace{1cm}
        \centering
        \begin{tikzpicture}[->,>=stealth',shorten >=1pt,auto,node distance=1.2cm,scale=0.75,transform shape,font = {\Large\bfseries\sffamily}]
	\tikzstyle{state}=[inner sep=5pt, minimum size=5pt]

	\node[state] (A) at (-3,0) {$A$};
	\node[state] (B) at (0,-1.5) {$B$};
	\node[state] (D) at (0,1.5) {$D$};
	\node[state] (C) at (3,0) {$C$};
          \node[state] (E) at (3,2) {$E$};

	\draw[->]  (B) edge (C);
	\draw[->] (D) edge (C);
	\draw[o-o] (D) edge (A);
	\draw[->] (A) edge (B);
	\draw[o-o] (B) edge (D);

 	\draw[o-o] (D) edge (E);

	\node[state] (Ci) at (-1.1,-.1) {$\g[C]_i$};
	\node[state] (Cj) at (1.1,-.1) {$\g[C]_j$};
 	\node[state] (Ck) at (1.5,2.5) {$\g[C]_k$};
  \end{tikzpicture}
\caption{}
\label{fig:jointree13} 
    \end{subfigure}
    \vrule
    \begin{subfigure}{.45\textwidth}
        \centering
       \tikzstyle{every edge}=[draw,>=stealth',semithick]
	\begin{tikzpicture}[->,>=stealth',shorten >=1pt,auto,node distance=1.2cm,scale=0.7,transform shape,font = {\Large\bfseries\sffamily}]
	\tikzstyle{state}=[inner sep=5pt, minimum size=5pt]

	\node[state] (J1i) at (-3,3) {$\g[C]_i$};
	\node[state] (J1j) at (0,3) {$\g[C]_j$};
	\node[state] (J1k) at (3,3) {$\g[C]_k$};
       
        \draw[->]  (J1i) edge (J1j);
        \draw[-]  (J1j) edge (J1k);

	\node[state] (J2i) at (-3,0) {$\g[C]_i$};
	\node[state] (J2j) at (0,0) {$\g[C]_j$};
	\node[state] (J2k) at (3,0) {$\g[C]_k$};

        \draw[->]  (J2i) edge (J2j);
        \draw[-] (J2i) edge [out=45,in=135] (J2k);
  \end{tikzpicture}
\caption{}
\label{fig:jointree23}
    \end{subfigure}
    \caption{\ref{fig:jointree13} Partially mixed graph $\g$, \ref{fig:jointree23} Two partially directed join trees for $\g$. These graphs are explored in Examples \ref{ex:l7p12} and \ref{ex:l7p112}.}
    \label{fig:lemma7-issue-p1-2}
\end{figure}

\begin{figure}[!t]
    \centering
    \begin{subfigure}{.4\textwidth}
          \tikzstyle{every edge}=[draw,>=stealth',semithick]
        \vspace{1cm}
        \centering
        \begin{tikzpicture}[->,>=stealth',shorten >=1pt,auto,node distance=1.2cm,scale=0.75,transform shape,font = {\Large\bfseries\sffamily}]
	\tikzstyle{state}=[inner sep=5pt, minimum size=5pt]

	\node[state] (A) at (-3,0) {$A$};
	\node[state] (B) at (0,-1.5) {$B$};
	\node[state] (D) at (0,1.5) {$D$};
	\node[state] (C) at (3,0) {$C$};
        \node[state] (E) at (3,2) {$E$};
        \node[state] (F) at (0,3.2) {$F$};  

	\draw[->]  (B) edge (C);
	\draw[->] (D) edge (C);
	\draw[->] (D) edge (A);
	\draw[->] (A) edge (B);
	\draw[<-] (B) edge (D);
 	\draw[<-o] (D) edge (E);
  	\draw[o-o] (F) edge (E);

	\node[state] (Ci) at (-1.1,-.1) {$\g[C]_i$};
	\node[state] (Cj) at (1.1,-.1) {$\g[C]_j$};
 	\node[state] (Ck) at (1.5,2.1) {$\g[C]_k$};
   	\node[state] (Cl) at (1.7,3) {$\g[C]_l$};
  \end{tikzpicture}
\caption{}
\label{fig:jointree1} 
    \end{subfigure}
    \vrule
    \begin{subfigure}{.47\textwidth}
        \centering
       \tikzstyle{every edge}=[draw,>=stealth',semithick]
	\begin{tikzpicture}[->,>=stealth',shorten >=1pt,auto,node distance=1.2cm,scale=0.7,transform shape,font = {\Large\bfseries\sffamily}]
	\tikzstyle{state}=[inner sep=5pt, minimum size=5pt]

	\node[state] (J1i) at (-3,3) {$\g[C]_i$};
	\node[state] (J1j) at (0,3) {$\g[C]_j$};
	\node[state] (J1k) at (3,3) {$\g[C]_k$};
    \node[state] (J1l) at (6,3) {$\g[C]_l$};
       
        \draw[->]  (J1i) edge (J1j);
        \draw[<-]  (J1j) edge (J1k);
        \draw[-]  (J1l) edge (J1k);

	\node[state] (J2i) at (-3,0) {$\g[C]_i$};
	\node[state] (J2j) at (0,0) {$\g[C]_j$};
	\node[state] (J2k) at (3,0) {$\g[C]_k$};
    \node[state] (J2l) at (6,0) {$\g[C]_l$};
        
        \draw[->]  (J2i) edge (J2j);
         \draw[-]  (J2l) edge (J2k);
        \draw[->]  (J2k) edge [out=135,in=45] (J2i);

        \node[state] (J1r) at (3,-1) {}; 
  \end{tikzpicture}
\caption{}
\label{fig:jointree2}
    \end{subfigure}
    \caption{\ref{fig:jointree1} Partially mixed graph $\g$, \ref{fig:jointree2} Two partially directed join trees for $\g$. These graphs are explored in Examples \ref{ex:l7p13} and \ref{ex:l7p113}.}
    \label{fig:lemma7-issue-p1-3}
\end{figure}

\begin{example}\label{ex:l7p13}
  A chordal and ancestral partial mixed graph $\g = (\mb{V,E})$ in Figure \ref{fig:jointree1} has orientations that are   closed under { \ref{R1}, \ref{R2}, \ref{R8}, \ref{R11}}. %  \ref{R1}-\ref{R4}, \ref{R8}-\ref{R13}. 
  In fact, the essential ancestral graph of $\g$ is fully undirected. 
   
   Four maximal cliques make up $\mb{V}$. These are $\g[C]_i = \{A,B,D\}, \g[C]_j = \{B,C,D\}$, $\g[C]_k = \{D,E\}$, and $\g[C]_l = \{E,F\}$. Two partially directed join trees for $\g$ are given in Figure \ref{fig:jointree2}. From top to bottom, these join trees are  $\g[T]_1$, $\g[T]_2$. 
   As can be seen from the figure, $\g[T]_1$ contains an unshielded collider.
  Based on $\g$, the only valid $\gamma$-relations on the maximal cliques of $\g$ are $\gamma(\g[C]_{i}, \g[C]_j)$, $\gamma(\g[C]_{k}, \g[C]_i)$, and $\gamma(\g[C]_{k}, \g[C]_j)$.

\end{example}

\subsection{Finding the Appropriate Partially Directed Join Tree}\label{sec:finding-right-jointree}

\begin{lemma}
\label{lemma:join-tree-single-operation}
Let $\g = (\mathbf{V}, \mathbf{E})$ be an ancestral partial mixed graph with a chordal skeleton and such that $\g$ does not contain minimal collider paths. Let $\mathcal{T}_0 = (\mathbf{C}, \mathbf{E}_0)$ be a partially directed join tree for $\g$ (Definition \ref{def:partially-directed-join-tree}). Consider a triple $\langle \g[C]_1, \g[C]_2, \g[C]_3 \rangle$ in $\g[T]_0$ such that $\Lambda_{13} = \Lambda_{23} \subseteq \Lambda_{12}$. Suppose that $\gamma(\g[C]_1, \g[C]_2)$ holds, but not $\gamma(\g[C]_2, \g[C]_3)$. Then, the graph $\g[T]$ obtained  from $\g[T]_0$ by removing edge $\langle \g[C]_2, \g[C]_3 \rangle$ and adding edge 
\begin{itemize}
    \item $\g[C]_1 \leftarrow \g[C]_3$, if $\gamma(\g[C]_3, \g[C]_1)$, or
    \item $\g[C]_1  \to \g[C]_3$, if $\gamma(\g[C]_1, \g[C]_3)$, or
    \item $\g[C]_1  - \g[C]_3$, if neither $\gamma(\g[C]_1, \g[C]_3)$, nor $\gamma(\g[C]_3, \g[C]_1)$,
\end{itemize}
is still a partially directed join tree for $\g$.
\end{lemma}

\begin{proofof}[Lemma \ref{lemma:join-tree-single-operation}]
It is easy to see that $\g[T]$ is a tree: we replace edge $\langle \g[C]_2, \g[C]_3 \rangle$ with edge  $\langle \g[C]_1 , \g[C]_3 \rangle$ in $\g[T]_0$, and since $\g[T]_0$ is a tree, in doing so we do not create any cycles in the skeleton of $\g[T]$.
 
The nodes of $\g[T]$ are still maximal cliques of $\g$, and the orientations of edges in $\g[T]$ still follow the $\gamma$ relation by construction. So to show that $\g[T]$ is a join tree for $\g$, we need to show that the running intersection property still holds. 

Specifically, consider two maximal cliques $\g[C]_i, \g[C]_j$ in $\g$ such that $\Lambda_{ij} \neq \varnothing$. Suppose the unique path between $\g[C]_i$ and $\g[C]_j$ in $\g[T]_0$ is $p$. If $p$ does not contain edge $\langle \g[C]_2, \g[C]_3 \rangle$ then, $p$ also exists in $\g[T]$ and the running intersection holds for this path because $\g[T]_0$ is a join tree.

Suppose that $p$ contains  the subpath $\langle \g[C]_1, \g[C]_2, \g[C]_3 \rangle$  (with $\g[C]_1$ or $\g[C]_3$ possibly being the endpoints). Then, in $\g[T]$, the unique path between $\g[C]_i$ and $\g[C]_j$ is $q = p(\g[C]_i, \g[C]_1) \oplus \langle \g[C]_1, \g[C]_3 \rangle \oplus p(\g[C]_3, \g[C]_j)$. Since $\Lambda_{ij} \subseteq \g[C]_1$ and $\Lambda_{ij}\subseteq \g[C]_3$ holds already in $\g[T]_0$, the running intersection property is also satisfied in $\g[T]$. 
A symmetric argument can be made when $p$ contains the subpath $\langle \g[C]_3, \g[C]_2, \g[C]_1 \rangle$.

Next, suppose that $p$ contains the edge $\langle \g[C]_2, \g[C]_3 \rangle$ but does not contain node $\g[C]_1$.  Then, in $\g[T]$, the unique path between $\g[C]_i$ and $\g[C]_j$ is $q = p(\g[C]_i, \g[C]_2) \oplus \langle \g[C]_2, \g[C]_1, \g[C]_3 \rangle \oplus p(\g[C]_3, \g[C]_j)$.
Then, in the new tree $\g[T]$, the path must contain node $\g[C]_1$.  It is sufficient to show that $\Lambda_{ij} \subseteq \g[C]_1$.
Since  $\Lambda_{ij} \subseteq \g[C]_2$ and $\Lambda_{ij} \subseteq \g[C]_3$, we have $\Lambda_{ij} \subseteq \Lambda_{23}$. This implies $\Lambda_{ij} \subseteq \Lambda_{12}$ by assumption. As $\Lambda_{12} \subseteq \g[C]_1$, we have $\Lambda_{ij} \subseteq \g[C]_1$. Therefore, the running intersection property still holds.
A symmetric argument can be made when $p$ contains the edge $\langle \g[C]_3, \g[C]_2\rangle$ but not the node $\g[C]_1$.
\end{proofof}

\begin{algorithm}[!t]
\caption{transformTreeHelper}
\label{alg:join-tree-transformation-helper}
\begin{algorithmic}[1]
    \Require  Partially directed join tree $\g[T] = (\mathbf{C}, \mathbf{E})$ for an ancestral partial mixed graph $\g$ with a chordal skeleton, with edge orientations closed under{ \ref{R1}, \ref{R2}, \ref{R8}, \ref{R11}}  and such that $\g$ is without minimal collider paths.
    \Ensure Another join tree $\g[T]^{\prime} = (\mathbf{C}, \mathbf{E}^{\prime})$ for $\g$.
    
    \State $\g[T]^{\prime} \gets \g[T]$
    \State $\mathcal{Q} \gets \{ \langle \g[C]_i, \g[C]_j, \g[C]_k \rangle \mid \langle \g[C]_i, \g[C]_j \rangle,  \langle \g[C]_j, \g[C]_k \rangle \in \mathbf{E} \}$ \Comment{Set of triples yet to be verified} \label{line:tree-transform-queue}
    
    \While {$\mathcal{Q} \neq \varnothing$}
        \State $\langle \g[C]_i, \g[C]_j, \g[C]_k \rangle \gets \mathcal{Q}_1$ \Comment{Remove the first triple from $\mc{Q}$}
        \State $\mc{Q} \gets \mc{Q} \setminus \mathcal{Q}_1$
        \If {$\gamma(\g[C]_i, \g[C]_j)$ and $\neg \gamma(\g[C]_j, \g[C]_k)$} \label{line:tree-transform-condition}
            \State $\Lambda_{ij} \gets \g[C]_i \cap \g[C]_j$
            \State $\Lambda_{jk} \gets \g[C]_j \cap \g[C]_k$
            \State $\Lambda_{ik} \gets \g[C]_i \cap \g[C]_k$ 
            \If {$\Lambda_{ik} = \Lambda_{jk} \subseteq \Lambda_{ij}$}  \label{line:lemma-call}
                \State $\mathcal{A} \gets \{ \langle \g[C]_u, \g[C]_v, \g[C]_w \rangle \mid \langle \g[C]_u, \g[C]_v \rangle,  \langle \g[C]_v, \g[C]_w \rangle \in \mathbf{E}^{\prime} \}$
                \State $\mathbf{E}^{\prime} \gets ( \mathbf{E}^{\prime} \cup \langle \g[C]_i, \g[C]_k \rangle ) \setminus \langle \g[C]_j, \g[C]_k \rangle$ \Comment{Transform as in Lemma \ref{lemma:join-tree-single-operation}} \label{line:tree-transform-step}
                \State $\mathcal{B} \gets \{ \langle \g[C]_u, \g[C]_v, \g[C]_w \rangle \mid \langle \g[C]_u, \g[C]_v \rangle,  \langle \g[C]_v, \g[C]_w \rangle \in \mathbf{E}^{\prime} \}$
                \State $\mathcal{Q} \gets \left( \mathcal{Q} \setminus (\mathcal{A} \setminus \mathcal{B}) \right) \cup (\mathcal{B} \setminus \mathcal{A})$ \Comment{Update $\mc{Q}$ with triples present \emph{only} in $\mc{B}$} \label{line:tree-transform-update-queue}

            \EndIf
        \EndIf
    \EndWhile
    
    \State \Return  $\g[T]^{\prime}$
\end{algorithmic}
\end{algorithm}

Algorithm \ref{alg:join-tree-transformation-helper} presents a procedure leverages Lemma \ref{lemma:join-tree-single-operation} to remove triples $\langle \mc{C}_i, \mc{C}_j, \mc{C}_k \rangle$ such that $\Lambda_{ik} = \Lambda_{jk} \subseteq \Lambda_{ij}$ from the join tree $\g[T]$. The key idea for this algorithm is that we make an exhaustive list of triples in the join tree, $\mc{Q}$ (line \ref{line:tree-transform-queue}). Then, we go through every triple and check whether it meets the antecedent of Lemma \ref{lemma:join-tree-single-operation} (line \ref{line:tree-transform-condition}). If it does, then we operate on the tree as Lemma \ref{lemma:join-tree-single-operation} suggests (line \ref{line:tree-transform-step}). This results in a new tree where the set of triples have changed. Therefore, we update the set of triples, $\mc{Q}$ (line \ref{line:tree-transform-update-queue}). When we update $\mc{Q}$, we remove any triples present in the tree before the operation and add only the newly formed triples. This ensures that a triple present before the operation that we've already verified in line \ref{line:tree-transform-condition} does not get added back. We show that Algorithm \ref{alg:join-tree-transformation-helper} terminates in Lemma \ref{lemma:transform-join-tree-helper-terminates} and prove some important properties of its output in Lemma \ref{lemma:transform-join-tree-helper}.

\begin{lemma}
\label{lemma:transform-join-tree-helper-terminates}
Let $\g = (\mathbf{V}, \mathbf{E})$ be an ancestral partial mixed graph with a chordal skeleton such that $\g$ has no minimal collider paths. Let $\mathcal{T}_0$ be any partially directed join tree for $\g$ (Definition \ref{def:partially-directed-join-tree})  and $\gamma$ a relation as defined in Definition \ref{def:gamma-order}.
Then Algorithm \ref{alg:join-tree-transformation-helper} terminates with input $\g[T]_0$.
\end{lemma}

\begin{proofof}[Lemma \ref{lemma:transform-join-tree-helper-terminates}]

For sake of contradiction, suppose that Algorithm \ref{alg:join-tree-transformation-helper} does not terminate. Observe that there are only a finite number of possible triples in $\g[T]$, $\abs{\mathbf{C}} \times (\abs{\mathbf{C}} - 1) \times (\abs{\mathbf{C}} - 2)$. As Algorithm \ref{alg:join-tree-transformation-helper} does not terminate, it must be that Line \ref{line:tree-transform-condition} encounters some triple $\langle \g[C]_i, \g[C]_j, \g[C]_k \rangle$ again after previously operating on it according to Lemma \ref{lemma:join-tree-single-operation}.

The first time we encounter this triple, we operate as in Lemma \ref{lemma:join-tree-single-operation} to construct a new triple $\langle \g[C]_k, \g[C]_i, \g[C]_j \rangle$. In order to have encountered the triple $\langle \g[C]_i, \g[C]_j, \g[C]_k \rangle$ again, there must be another triple $\langle \g[C]_j, \g[C]_{j_2}, \g[C]_k \rangle$  (or $\langle \g[C]_k, \g[C]_{j_2}, \g[C]_j \rangle$), in a tree $\g[T]_1$, that gets operated on it as in Lemma \ref{lemma:join-tree-single-operation} to construct the triple $\langle \g[C]_k, \g[C]_j, \g[C]_{j2} \rangle$ (or $\langle \g[C]_{j}, \g[C]_k, \g[C]_{j2} \rangle$).

However, this must mean that there is an undirected cycle in the skeleton of $\g[T]_1$ made up by $p = \langle \g[C]_j, \g[C]_{j2}, \g[C]_k \rangle$ and $q = \langle \g[C]_k, \dots, \g[C]_i, \g[C]_j \rangle$. Here $q$ must contain the edge $\langle \g[C]_i, \g[C]_j \rangle$ in $\g[T]_1$. Further, $q(\g[C]_k, \g[C]_i)$ is either  the edge $\langle \g[C]_k, \g[C]_i \rangle$ that was obtained from operating on $\langle \g[C]_i, \g[C]_j, \g[C]_k \rangle$ the first time and is still present in $\g[T]_1$, or $\langle \g[C]_k, \g[C]_i \rangle$ was removed by some prior application of  Lemma \ref{lemma:join-tree-single-operation} in which case a longer path $q(\g[C]_k, \g[C]_i) = \langle \g[C]_k, \dots, \g[C]_i \rangle$ is present in $\g[T]_1$. Such a cycle with $p$ and $q$, of course, is a contradiction with $\g[T]_0$ being a tree, or the result of Lemma \ref{lemma:join-tree-single-operation}.
\end{proofof}

\begin{lemma}
\label{lemma:transform-join-tree-helper}
Let $\g = (\mathbf{V}, \mathbf{E})$ be an ancestral partial mixed graph with a chordal skeleton such that $\g$ has no minimal collider paths and such that orientations in $\g$ are closed under \ref{R1} and \ref{R11}. Let $\mathcal{T}_0 = (\mb{C,E_0})$ be any partially directed join tree for $\g$ (Definition \ref{def:partially-directed-join-tree})  and $\gamma$ a relation as defined in Definition \ref{def:gamma-order}.
Let $\g[T]  = (\mb{C,E})$ be the output of Algorithm \ref{alg:join-tree-transformation-helper} i.e., $\g[T] = \texttt{transformTreeHelper}(\g[T]_0)$. Then
\begin{enumerate}[label=(\roman*)]
    \item\label{case1:transformedtree1} $\g[T]$ is also a join tree for $\g$, and
    \item\label{case2:transformedtree1} for any pair of cliques, if $\gamma(\g[C]_i, \g[C]_j)$ in $\g[T]_0$, then $\gamma(\g[C]_i, \g[C]_j)$ in $\g[T]$ as well, and
    \item\label{case3:transformedtree1} for any path $\langle \g[C]_i, \g[C]_j, \g[C]_k \rangle$ in $\g[T]$ such that $\gamma(\g[C]_i, \g[C]_j)$ but not $\gamma(\g[C]_j, \g[C]_k)$, then $\Lambda_{ik} = \Lambda_{ij} \subset \Lambda_{jk}$ and  $\gamma(\g[C]_i,\g[C]_k)$ holds.
    \item\label{case4:transformedtree1} $\g[T]$ does not contain any path of the form $\g[C]_i \to \g[C]_j \leftarrow \g[C]_k$ for any $\g[C]_i, \g[C]_j, \g[C]_k \in \mb{C}$.
\end{enumerate}
\end{lemma}

\begin{proofof}[Lemma \ref{lemma:transform-join-tree-helper}]
Algorithm \ref{alg:join-tree-transformation-helper} terminates by Lemma \ref{lemma:transform-join-tree-helper-terminates}. This allows us to talk about the properties of its output, $\g[T]$.
\begin{enumerate}
    \item[\ref{case1:transformedtree1}]  $\g[T]$ is a join tree by Lemma \ref{lemma:join-tree-single-operation}. 
    \item[\ref{case2:transformedtree1}] Since we do not change any orientations of edges in $\g$ during the course of Algorithm \ref{alg:join-tree-transformation-helper}, $\gamma$ ordering is preserved.
    \item[\ref{case3:transformedtree1}]  By Lemma \ref{lemma:rule-1-join-tree}, if $\langle \g[C]_i, \g[C]_j, \g[C]_k \rangle$ in $\g[T]$ such that $\gamma(\g[C]_i, \g[C]_j)$ but not $\gamma(\g[C]_j, \g[C]_k)$, then either $\Lambda_{ik} = \Lambda_{jk} \subseteq \Lambda_{ij}$ or $\Lambda_{ik} = \Lambda_{ij} \subset \Lambda_{jk}$. However, it is not the case that $\Lambda_{ik} = \Lambda_{jk} \subseteq \Lambda_{ij}$ (otherwise $\mathcal{Q} \neq \varnothing$ in Algorithm \ref{alg:join-tree-transformation-helper}). Therefore, $\Lambda_{ik} = \Lambda_{ij} \subset \Lambda_{jk}$ and $\gamma(\g[C]_i,\g[C]_k)$ holds.
    \item[\ref{case4:transformedtree1}] Suppose for a contradiction that $\g[T]$ does contain a path of the form  $\g[C]_i \to \g[C]_j \leftarrow \g[C]_k$. By case \ref{case3:transformedtree1} above , we then have that  $\Lambda_{ik} = \Lambda_{ij}  \subset \Lambda_{jk},$ and also that $\gamma(\g[C]_i, \g[C]_k)$ holds. 
    But, also, since $\g[C]_k \to \g[C]_j \leftarrow \g[C]_i$,  case \ref{case3:transformedtree1} above leads us to conclude that  $\Lambda_{ik} = \Lambda_{jk}  \subset \Lambda_{ij},$ and  $\gamma(\g[C]_k, \g[C]_i)$ hold in $\g$, which a contradiction.
\end{enumerate}
\end{proofof}

\begin{example}
    \label{ex:l7p111}
 Consider again graph $\g = (\mb{V,E})$ in Figure \ref{fig:jointree11} used in Example \ref{ex:l7p13} above. As discussed in Example \ref{ex:l7p11}, Figure \ref{fig:jointree22} contains three partially directed join trees for $\g$. From top to bottom, these join trees are  $\g[T]_1$, $\g[T]_2$ and $\g[T]_3$. 
 
  Applying Algorithm \ref{alg:join-tree-transformation-helper} to $\g[T]_1$ or to $\g[T]_2$ leads to $\g[T]_3$ as output.
   Note that $\Lambda_{ij} = \{B,D\} = \Lambda_{jk} = \Lambda_{ik}$ and that therefore in $\g[T]_1$, $\Lambda_{ik} = \Lambda_{jk} \subseteq \Lambda_{ij}$, and in $\g[T]_2$, $\Lambda_{ij} = \Lambda_{jk} \subseteq \Lambda_{ik}$. So both $\g[T]_1$ and $\g[T]_2$ satisfy conditions of Lemma \ref{lemma:join-tree-single-operation}.
   
  For $\g[T]_1$, this is since line \ref{line:lemma-call} calls for Lemma \ref{lemma:join-tree-single-operation} to be applied applied to triple $\g[C]_{i} \to \g[C]_j - \g[C]_k$. That is edge $\g[C]_j - \g[C]_k$ is removed from $\g[T]_1$ and edge $\g[C]_i \to \g[C]_k$ is added to create $\g[T]_3$.

  For  $\g[T]_2$, Lemma \ref{lemma:join-tree-single-operation} is applied to $\g[C]_{i} \to \g[C]_k - \g[C]_i$. It removes $\g[C]_k - \g[C]_i$ from $\g[T]_2$ and adds $\g[C]_{i} \to \g[C]_k$ to create $\g[T]_3$.
\end{example}

\begin{example}
    \label{ex:l7p112}
 Consider again graph $\g = (\mb{V,E})$ in Figure \ref{fig:jointree13} used in Example \ref{ex:l7p13} above. As discussed in Example \ref{ex:l7p13}, Figure \ref{fig:jointree23} contains two partially directed join trees for $\g$. From top to bottom, these join trees are  $\g[T]_1$, $\g[T]_2$.
 
  Applying Algorithm \ref{alg:join-tree-transformation-helper} to $\g[T]_1$ leads to $\g[T]_2$ as output. This is because line \ref{line:lemma-call} calls for Lemma \ref{lemma:join-tree-single-operation} to be applied to triple $\g[C]_{i} \to \g[C]_j - \g[C]_k$.  Note that $\Lambda_{ij} = \{B,D\}$, and $\Lambda_{jk} = \{D\} = \Lambda_{ik}$. Therefore in $\g[T]_1$, $\Lambda_{ik} = \Lambda_{jk} \subset \Lambda_{ij}$, so $\g[T]_1$ satisfies conditions of Lemma \ref{lemma:join-tree-single-operation}. That is edge $\g[C]_j - \g[C]_k$ is removed from $\g[T]_1$ and edge $\g[C]_i -\g[C]_k$ is added to create $\g[T]_2$.
\end{example}

\begin{example}
    \label{ex:l7p113}
 Consider again graph $\g = (\mb{V,E})$ in Figure \ref{fig:jointree1} used in Example \ref{ex:l7p13} above. As discussed in Example \ref{ex:l7p13}, Figure \ref{fig:jointree2} contains two partially directed join trees for $\g$. From top to bottom, these join trees are  $\g[T]_1$, $\g[T]_2$.
 
 Applying Algorithm \ref{alg:join-tree-transformation-helper} to $\g[T]_1$ leads to $\g[T]_2$ as output. This is because line \ref{line:lemma-call} calls for Lemma \ref{lemma:join-tree-single-operation} to be applied to triple $\g[C]_{i} \to \g[C]_j \leftarrow \g[C]_k$.    Note that $\Lambda_{ij} = \{B,D\}$,  $\Lambda_{jk} = \{D\} = \Lambda_{ik}$ and $\Lambda_{kl} = \{E\}$. Therefore in $\g[T]_1$, $\Lambda_{ik} = \Lambda_{jk} \subset \Lambda_{ij}$, so $\g[T]_1$ satisfies conditions of Lemma \ref{lemma:join-tree-single-operation}. That is edge $\g[C]_j \leftarrow \g[C]_k$ is removed from $\g[T]_1$ and edge $\g[C]_i \leftarrow \g[C]_k$ is added to create $\g[T]_2$.
\end{example}

In all the examples above there always exists a partially directed join tree $\g[T]$ for a graph $\g$ such that paths $\g[C]_i \to \g[C]_j - \g[C]_k$ and $\g[C]_i \to \g[C]_j \leftarrow \g[C]_k$ do not occur in $\g[T]$. While by case \ref{case4:transformedtree1} of Lemma \ref{lemma:transform-join-tree-helper} it is true that a partially directed join tree without colliders will always exist for an ancestral and chordal partially directed mixed graph $\g$ with no minimal collider paths, the same is not true for paths of the form $\g[C]_i \to \g[C]_j - \g[C]_k$. Example \ref{ex:l7p2} presents one case where all partially directed join trees for  $\g$ contain such paths. 

Lemma \ref{lemma:join-tree-single-operation2} discusses how such paths can be transformed in $\g[T]$, but they will not necessarily disappear entirely from the transformed join tree. Instead, we devise Algorithm \ref{alg:join-tree-transformation} that in addition to colliders, removes all paths of the form $\g[C]_{i_1} \to \g[C]_{i_2} - \dots - \g[C]_{i_k} \leftarrow \g[C]_{i_{k+1}}$ from a partially directed join tree for an ancestral and chordal partially directed mixed graph $\g$ with no minimal collider paths. Additionally, Algorithm \ref{alg:join-tree-transformation} ensures that for a specified maximal clique $\g[C]_0$, no path of the form $\g[C]_{i_1} \to \g[C]_{i_2} - \dots - \g[C]_{i_k} \to \dots \to \g[C]_{r}$ with $\g[C]_r \equiv \g[C]_0$
occurs in the resulting partially directed join tree. We prove these properties in Corollary \ref{cor:algo-transformtree1-result} and demonstrate Algorithm \ref{alg:join-tree-transformation} in Example \ref{ex:algo4-demo}.

\begin{example}\label{ex:l7p2}
     A chordal and ancestral partial mixed graph $\g^{\prime} = (\mb{V,E})$ in Figure \ref{fig:jointree111} has orientations that are   closed under { \ref{R1}, \ref{R2}, \ref{R8}, \ref{R11}}. In fact, the essential ancestral graph of $\g^{\prime}$ as in all previous examples in this section is fully undirected, see Figure \ref{fig:lemma7-issue-p2}(a). 
   
   Four maximal cliques make up $\mb{V}$. These are $\g[C]_i = \{E,F\}, \g[C]_j = \{C,D,F\}$, $\g[C]_k = \{B,C,F\}$, and $\g[C]_l = \{A,B,F\}$. Three partially directed join trees for $\g^{\prime}$ are given in Figure \ref{fig:jointree222}. From top to bottom, these join trees are  $\g[T]_1$, $\g[T]_2$, and $\g[T]_3$. 
   As can be seen from the figure, none of these partially directed join trees have orientations   closed under  \ref{R1}. 
  Based on $\g^{\prime}$, the only valid $\gamma$-relations on the maximal cliques of $\g^{\prime} $ are $\gamma(\g[C]_{i}, \g[C]_j)$, $\gamma(\g[C]_{i}, \g[C]_k)$, and $\gamma(\g[C]_{i}, \g[C]_l)$.
   
   Note that $\Lambda_{ij} = \Lambda_{ik} = \Lambda_{il} = \{F\}$,  $\Lambda_{jk} = \{C,F\}$, and $\Lambda_{kl} = \{B,F\}$. Therefore in $\g[T]_1$, $\Lambda_{ik} = \Lambda_{ij} \subset \Lambda_{jk}$, so applying Algorithm \ref{alg:join-tree-transformation-helper} to $\g[T]_1$ results in $\g[T]_1$ as output. Similarly $\g[T]_2 = \texttt{transformTree}(\g[T]_2)$, and $\g[T]_3 = \texttt{transformTree}(\g[T]_3)$.

   Note also that  since $\Lambda_{jk} \not\subseteq \g[C]_i$, $\g[C]_j \leftarrow \g[C]_i \to \g[C]_k$ cannot be a path in  a valid join tree for $\g$. A similar issue arises with path $\g[C]_k \leftarrow \g[C]_i \to \g[C]_l$. Hence, the list of join trees in Figure \ref{fig:jointree222} is exhaustive for $\g^{\prime}$. This example demonstrates that while Algorithm \ref{alg:join-tree-transformation-helper} deals with some properties of a general partially directed join tree, it is not enough to ensure that the resulting  partially directed join tree for $\g^{\prime} $ has a single root node. 
\end{example}

\begin{lemma}
\label{lemma:join-tree-single-operation2}
Let $\g = (\mathbf{V}, \mathbf{E})$ be an ancestral partial mixed graph with a chordal skeleton and such that $\g$ does not contain minimal collider paths and such that orientations in $\g$ are closed under \ref{R1} and \ref{R11}. Let $\mathcal{T}_0 = (\mathbf{C}, \mathbf{E}_0)$ be a partially directed join tree for $\g$ (Definition \ref{def:partially-directed-join-tree}). Furthermore, suppose that applying Algorithm \ref{alg:join-tree-transformation-helper} to $\g[T]_0$ results in the same tree, that is $\g[T]_0 = \texttt{transformTreeHelper}(\g[T]_0)$. Consider a triple $\langle \g[C]_1, \g[C]_2, \g[C]_3 \rangle$ in $\g[T]_0$ that is of the form $\g[C]_1 \to \g[C]_2 - \g[C]_3$. 
Then the graph $\g[T]$ obtained  from $\g[T]_0$ by removing edge $\langle \g[C]_1, \g[C]_2 \rangle$ and adding edge, $\g[C]_1 \to \g[C]_3$  is still a partially directed join tree  for $\g$.
\end{lemma}

\begin{proofof}[Lemma \ref{lemma:join-tree-single-operation2}]

It is easy to see that $\g[T]$ is a tree: we replace the edge $\langle \g[C]_1,  \g[C]_2 \rangle$ with $\langle \g[C]_1,  \g[C]_3 \rangle$, which will not create any undirected cycles in the graph skeleton since the original graph $\g[T]_0$ did not have any undirected cycles in the graph skeleton.

The nodes of $\g[T]$ are still maximal cliques of $\g$, and by Lemma \ref{lemma:transform-join-tree-helper}, the $\gamma$ property is maintained in $\g[T]$. Hence,  to show that $\g[T]$ is a partially directed join tree for $\g$, we need to show that the running intersection property still holds. Specifically, consider two maximal cliques $\g[C]_i, \g[C]_j$ in $\g$ such that $\Lambda_{ij} \neq \varnothing$ and suppose the unique path between $\g[C]_i$ and $\g[C]_j$ in $\g[T]_0$ is $p$. If $p$ does not contain edge $\langle \g[C]_1, \g[C]_2 \rangle$ then, $p$ also exists in $\g[T]$ and the running intersection holds for this path because $\g[T]_0$ is a join tree.

Suppose that $p$ contains  the subpath $\langle \g[C]_1, \g[C]_2, \g[C]_3 \rangle$  (with $\g[C]_1$ or $\g[C]_3$ possibly being the endpoints). Then the unique path between $\g[C]_i$ and $\g[C]_j$ in $\g[T]$ is $q = p(\g[C]_i, \g[C]_1) \oplus \langle \g[C]_1, \g[C]_3 \rangle \oplus p(\g[C]_3, \g[C]_j)$. Since $\Lambda_{ij} \subseteq \g[C]_1$ and $\Lambda_{ij}\subseteq \g[C]_3$ already holds in $\g[T]_0$, $q$ still satisfies the running intersection property in $\g[T]$. 
A symmetric argument holds if $p$ contains the subpath $\langle \g[C]_3, \g[C]_2, \g[C]_1 \rangle$.

Next, suppose that $p$ contains the edge $\langle \g[C]_1, \g[C]_2 \rangle$ but does not contain node $\g[C]_3$.  Then the unique path between $\g[C]_i$ and $\g[C]_j$ in $\g[T]$ is $q = p(\g[C]_i, \g[C]_1) \oplus \langle \g[C]_1, \g[C]_3, \g[C]_2 \rangle \oplus p(\g[C]_2, \g[C]_j)$.
That is, the path must contain node $\g[C]_3$.  It is sufficient to show that $\Lambda_{ij} \subseteq \g[C]_3$. Since  $\Lambda_{ij} \subseteq \g[C]_1$ and $\Lambda_{ij} \subseteq \g[C]_2$, we have $\Lambda_{ij} \subseteq \Lambda_{12}$. This implies $\Lambda_{ij} \subseteq \Lambda_{23}$ by assumption, and $\Lambda_{23} \subseteq \g[C]_3$. Therefore, $\Lambda_{ij} \subseteq \g[C]_3$, and the running intersection property still holds.
A symmetric argument holds if $p$ contains the subpath $\langle \g[C]_2, \g[C]_1 \rangle$ but not the node $\g[C]_3$.
\end{proofof}

\begin{algorithm}[!t]
\caption{relevantPaths}
\label{alg:paths-for-join-tree-transformation}
\begin{algorithmic}[1]
    \Require Partially directed join tree $\g[T] = (\mathbf{C}, \mathbf{E})$ and node $\g[C]_0 \in \mb{C}$
    \Ensure List of paths $\mathbf{P}$ relevant to Corollary \ref{cor:algo-transformtree1-result}
    \State $\mb{A} \gets \{ \g[C]_1 \to \g[C]_2 - \dots - \g[C]_k \to \dots \to \g[C]_r \mid \langle \g[C]_i, \g[C]_{i+1} \rangle \in \mb{C}, r \geq k > 2, \g[C]_r \equiv \g[C]_0 \}$
    \State $\mb{B} \gets \{ \g[C]_1 \to \g[C]_2 - \dots - \g[C]_{k-1} \leftarrow \g[C]_k \mid \langle \g[C]_i, \g[C]_{i+1} \rangle \in \mb{C}, k > 3 \}$
    \State $\mb{P} \gets \mb{A} \cup \mb{B}$
    \State \Return  $\mb{P}$
\end{algorithmic}
\end{algorithm}

\begin{algorithm}[!t]
\caption{transformTree}
\label{alg:join-tree-transformation}
\begin{algorithmic}[1]
    \Require Partially directed join tree $\g[T] = (\mathbf{C}, \mathbf{E})$, and node $\g[C]_0 \in \mb{C}$ for an ancestral partial mixed graph $\g$ with a chordal skeleton, with edge orientations closed under { \ref{R1}, \ref{R2}, \ref{R8}, \ref{R11}} and such that $\g$ is without minimal collider paths.
    \Ensure Another join tree $\g[T]^{\prime} = (\mathbf{C}, \mathbf{E}^{\prime})$ for $\g$.
    \State $\g[T]^{\prime} \gets \texttt{transformTreeHelper}(\g[T])$
    \State $\mb{P} \gets \texttt{relevantPaths}(\g[T], \g[C]_0)$   \Comment{Algorithm \ref{alg:paths-for-join-tree-transformation}}
    \While {$\mathbf{P} \neq \varnothing$}
        \State $p = \langle \g[C]_1, \dots, \g[C]_k \rangle \in \mathbf{P}$ such that $p = \argmax_{p^{\prime} \in \mathbf{P}} d(\g[C]_1, \g[C]_0)$ \Comment{Definition \ref{def:node-distance}}
        \State $\mathbf{E}^{\prime} \gets ( \mathbf{E}^{\prime} \cup (\g[C]_1 \to \g[C]_3)  ) \setminus ( \g[C]_1 \to \g[C]_2 )$ \Comment{Transform as in Lemma \ref{lemma:join-tree-single-operation2}} \label{line:transform-edge-join-tree}
        \State $\g[T]^{\prime} \gets \texttt{transformTreeHelper}(\g[T]^{\prime})$ \label{line:alg-helper-call-poss-directed}
        \State $\mb{P} \gets \texttt{relevantPaths}(\g[T]^{\prime}, \g[C]_0)$ \Comment{Update paths in $\g[T]^{\prime}$}\label{line:new-paths-p}
    \EndWhile
    
    \State \Return  $\g[T]^{\prime}$
\end{algorithmic}
\end{algorithm}

{As we already discussed, the goal of Algorithm \ref{alg:join-tree-transformation} is to remove all paths of the form $\g[C]_{i_1} \to \g[C]_{i_2} - \dots - \g[C]_{i_k} \leftarrow \g[C]_{i_{k+1}}$ and $\g[C]_{i_1} \to \g[C]_{i_2} - \dots - \g[C]_{i_k} \to \dots \to \g[C]_{r}$ with $\g[C]_r \equiv \g[C]_0$, for a specified maximal clique $\g[C]_0$ in the join tree, thereby making the tree anchored around $\g[C]_0$. The intuition behind this algorithm is repeated application of the operation described in Lemma \ref{lemma:join-tree-single-operation2}. Specifically, we need to be careful about the order in which we apply this operation. Otherwise, we open ourselves to an infinite loop---for instance, in Example \ref{ex:l7p2}, by applying this operation on randomly chosen triples we will traverse the space of the three join trees infinitely. To prevent such infinite loops, we will anchor the two kinds of paths we wish to remove to some node in the tree. When applying the operation in Lemma \ref{lemma:join-tree-single-operation2}, we will always prioritize a path that is \emph{farthest} from this anchor. We will use Definition \ref{def:node-distance} to characterize how far the endpoints from the paths are. For convenience, we will choose $\g[C]_0$ as the anchor (any node in the tree will serve as a valid anchor as long as the tree is connected). We describe the technical details in Lemma \ref{lemma:alg-transform-tree-terminates} and Algorithm \ref{alg:join-tree-transformation}.}

\begin{definition}[Distance between nodes, $d$]
\label{def:node-distance}
For any two nodes, $A, B$ in a graph $\g = (\mb{V}, \mb{E})$, the distance between them along a path $p = \langle A, \dots, B \rangle$ is the number of edges on $p$. We denote this by $d(A, B; p)$. We say $d(A, A) = 0$ and if there is no path from $A$ to $B$, then $d(A, B) = \infty$.
\end{definition}

\begin{remark}
    Observe that in a tree graph, $\g[T] = (\mb{V}, \mb{E})$, there is only one path between $A$ and $B$. Therefore, the distance between $A$ and $B$ is unique and we will refer to this as $d(A, B) := d(A, B; p)$.
\end{remark}

\begin{lemma}
\label{lemma:alg-transform-tree-terminates}
Let $\g = (\mathbf{V}, \mathbf{E})$ be an ancestral partial mixed graph with a chordal skeleton such that $\g$ has no minimal collider paths and such that orientations in $\g$ are closed under \ref{R1} and \ref{R11}. Let $\mathcal{T}_0$ be any join tree for $\g$, $\g[C]_0$ a node in $\g[T]_0$, and $\gamma$ a relation as defined in Definition \ref{def:gamma-order}. Then Algorithm \ref{alg:join-tree-transformation} terminates on input ($\g[T]_0, \g[C]_0)$.
\end{lemma}

\begin{proofof}[Lemma \ref{lemma:alg-transform-tree-terminates}]
By Lemma \ref{lemma:transform-join-tree-helper-terminates}, we know that Algorithm \ref{alg:join-tree-transformation-helper} terminates. Furthermore, Algorithm \ref{alg:paths-for-join-tree-transformation} also terminates since we only consider graphs defined on a finite number of nodes in this manuscript. Therefore, to show the termination of Algorithm \ref{alg:join-tree-transformation}, we only need to show that the set $\mb{P}$ will be empty at some point. Note that every path in $\mb{P}$ starts with a triple of the form $\g[C]_1 \to \g[C]_2 - \g[C]_3$. Hence, for the set $\mb{P}$ to become empty it is enough to show that once a path starting with a triple  $\langle \g[C]_1, \g[C]_2, \g[C]_3 \rangle$ is removed from $\mb{P}$ by applications of Lines \ref{line:transform-edge-join-tree}-\ref{line:new-paths-p}, it will not be added again in a subsequent pass through the while loop. 

For sake of contradiction, assume that Line \ref{line:transform-edge-join-tree} sees a triple  $\langle \g[C]_1, \g[C]_2, \g[C]_3 \rangle$  that was processed in a previous while loop iteration. During the previous encounter of this triple in the while loop, $\g[C]_1 \to  \g[C]_2 -\g[C]_3 $  must have been transformed into $\g[C]_1 \to \g[C]_3 - \g[C]_2$ by Line \ref{line:transform-edge-join-tree}. Observe that since $\Lambda_{13} = \Lambda_{12} \subset \Lambda_{23}$, Algorithm \ref{alg:join-tree-transformation-helper} will not operate on this triple. Therefore, in order to re-encounter the triple $\g[C]_1 \to \g[C]_2 - \g[C]_3$, one of the following must be true:
\begin{enumerate}[label=(\roman*)]
\item\label{case1:algojointree-terminates} there must have been some triple $\g[C]_1 \to \g[C]_{\ell} - \g[C]_2$, $\ell \neq 3$, that got operated on by either Algorithm \ref{alg:join-tree-transformation-helper} or by Line \ref{line:transform-edge-join-tree} to create the edge $\langle \g[C]_1, \g[C]_2 \rangle$.
\item\label{case2:algojointree-terminates} there must have been some path in $\mathbf{P}$ that started with the triple $\langle \g[C]_1, \g[C]_3, \g[C]_2 \rangle$ and therefore got operated on as per Lemma \ref{lemma:join-tree-single-operation2}.
\end{enumerate}
Case \ref{case1:algojointree-terminates} indicates the presence of an undirected cycle in the skeleton of the tree, which leads to a contradiction. Therefore, in the rest of the proof we suppose case \ref{case2:algojointree-terminates} is true.

The fact that we encountered the triple $\langle \g[C]_1, \g[C]_2, \g[C]_3 \rangle$ the first time, in some tree $\g[T]_1$, indicates the presence of one of these two paths:
\begin{enumerate}[label=(A\arabic*)]
\item\label{A1} $\g[C]_1 \to \g[C]_2 - \g[C]_3 - \dots - \g[C]_k \to \dots \to \g[C]_0$ ($\g[C]_3 \equiv \g[C]_0$ or $\g[C]_k \equiv \g[C]_0$ possibly), or
\item\label{A2} $\g[C]_1 \to \g[C]_2 - \g[C]_3 - \dots - \g[C]_{k-1} \leftarrow \g[C]_k$.
\end{enumerate}
Now, when we encounter the triple $\langle \g[C]_1, \g[C]_3, \g[C]_2 \rangle$, later on in some other tree $\g[T]_2$, in Line \ref{line:transform-edge-join-tree}, this indicates the presence of one of these two paths in $\g[T]_2$:
\begin{enumerate}[label=(B\arabic*)]
\item\label{B1} $\g[C]_1 \to \g[C]_3 - \g[C]_2 - \dots - \g[C]_{k^{\prime}} \to \dots \to \g[C]_0$  ($\g[C]_2 \equiv \g[C]_0$ or $\g[C]_{k^{\prime}} \equiv \g[C]_0$ possibly), or
\item\label{B2} $\g[C]_1 \to \g[C]_3 - \g[C]_2 - \g[C]_{i^{\prime}} \dots - \g[C]_{k^{\prime}-1} \leftarrow \g[C]_{k^{\prime}}$.
\end{enumerate}

Clearly if \ref{A1} was true, then \ref{B1} cannot be true
as this indicates the presence of a path from $\g[C]_1$ to $\g[C]_0$ that passes through $\langle \g[C]_1, \g[C]_2, \g[C]_3 \rangle$ in $\g[T]_1$ and another that passes through $\langle \g[C]_1, \g[C]_3, \g[C]_2 \rangle$ in $\g[T]_2$. Observe that after applying Lemma \ref{lemma:join-tree-single-operation2} on path \ref{A1}, the path from $\g[C]_1$ to $\g[C]_0$ does not pass through $\g[C]_2$. For \ref{B1} to be present, there must have already been another path from $\g[C]_2$ to $\g[C]_0$ that does not pass through $\g[C]_3$. This indicates the presence of cycles which contradicts that $\g[T]_1$ is a tree.

Further, possibilities \{\ref{A1}, \ref{B2}\} and \{\ref{A2}, \ref{B1}\} are symmetric. So, without loss of generality, we only consider two cases below----\{\ref{A1}, \ref{B2}\} and \{\ref{A2}, \ref{B2}\}.
In these cases, we rely on the fact that we have a fixed anchor ($\g[C]_0$, here) and that we always choose a path from $\mb{P}$ that starts from a node that is farthest from the anchor (see Definition \ref{def:node-distance} for definition of distance between nodes).

\begin{enumerate}[leftmargin=\parindent,align=left,labelwidth=\parindent,labelsep=5pt]
    \item[\ref{A1} and \ref{B2}.]  Suppose that \ref{A1} was present in $\g[T]_1$ and \ref{B2} is present in $\g[T]_2$. Then, in $\g[T]_2$, the path from $\g[C]_{k^{\prime}}$ to $\g[C]_0$ must pass through $\g[C]_3$. Therefore, this path is longer than the path from $\g[C]_1$ to $\g[C]_0$. Therefore, we would have had to operate on the triple $\langle \g[C]_{k^{\prime}}, \g[C]_{k^{\prime}-1}, \g[C]_{k^{\prime} - 2} \rangle$ before $\langle \g[C]_1, \g[C]_3, \g[C]_2 \rangle$ giving rise to a contradiction.
    \item[\ref{A2} and \ref{B2}.] Now, consider the case where \ref{A2} was present in $\g[T]_1$ and \ref{B2} is present in $\g[T]_2$. Since \ref{A2}  was in $\g[T]_1$ and we operated on $\langle \g[C]_1, \g[C]_2, \g[C]_3 \rangle$ in $\g[T]_1$, it must be that $\g[C]_1$ is farther away from $\g[C]_0$ than $\g[C]_k$. In other words, the path from $\g[C]_1$ to $\g[C]_0$ must pass through some subsequence of \ref{A2}. However, this must imply that, in $\g[T]_2$, the path from $\g[C]_{k^{\prime}}$ to $\g[C]_0$ must pass through $\g[C]_3$. Therefore, $\g[C]_{k^{\prime}}$ is farther away from $\g[C]_0$ than $\g[C]_1$. So we would have had to operate on the triple $\langle \g[C]_{k^{\prime}}, \g[C]_{k^{\prime}-1}, \g[C]_{k^{\prime} - 2} \rangle$ before $\langle \g[C]_1, \g[C]_3, \g[C]_2 \rangle$ giving rise to a contradiction.
\end{enumerate}
\end{proofof}

\begin{corollary}\label{cor:algo-transformtree1-result}

Let $\g = (\mathbf{V}, \mathbf{E})$ be an ancestral partial mixed graph with a chordal skeleton such that $\g$ has no minimal collider paths and such that orientations in $\g$ are closed under \ref{R1} and \ref{R11}. Let $\mathcal{T}_0$ be any join tree for $\g$, $\g[C]_0$ a node in $\g[T]_0$ and let $\g[T]$ be the output of Algorithm \ref{alg:join-tree-transformation}, that is $\g[T] = \texttt{transformTree}(\g[T]_0, \g[C]_0)$.  Then
\begin{enumerate}[label=(\roman*)]
    \item\label{transformtreesound:case1} $\g[T]$ is also a join tree for $\g$, 
    \item\label{transformtreesound:case2} for any pair of cliques, if $\gamma(\g[C]_i, \g[C]_j)$ in $\g[T]_0$, then $\gamma(\g[C]_i, \g[C]_j)$ in $\g[T]$ as well, 
    \item\label{transformtreesound:case3} $\g[T]$ does not contain any colliders, or  paths of the form $\g[C]_{i_1} \to \g[C]_{i_2} - \dots - \g[C]_{i_k} \leftarrow \g[C]_{i_{k+1}}$, $k > 2$, 
    \item\label{transformtreesound:case4} $\g[T]$ is anchored at $\g[C]_{0}$, meaning $\An(\g[C]_0, \g[T]) = \PossAn(\g[C]_0, \g[T])$. 
\end{enumerate}   
\end{corollary}

\begin{proofof}[Corollary \ref{cor:algo-transformtree1-result}]
From Lemmas \ref{lemma:transform-join-tree-helper-terminates} and \ref{lemma:alg-transform-tree-terminates} we know that Algorithm \ref{alg:join-tree-transformation} terminates. Lemmas \ref{lemma:transform-join-tree-helper} and \ref{lemma:join-tree-single-operation2} tell us that cases \ref{transformtreesound:case1} and \ref{transformtreesound:case2} are true. Case \ref{transformtreesound:case3} is true by construction of Algorithm \ref{alg:join-tree-transformation}. For case \ref{transformtreesound:case4} to hold, it is enough to show that $\g[T]$  does not contain  paths of the form $\g[C]_{i_1} \to \g[C]_{i_2} - \dots - \g[C]_{i_k} \to \dots \to \g[C]_{i_r} $, $r \geq k > 2$, where $\g[C]_{i_r} \equiv \g[C]_0$. This clearly holds by construction of Algorithm \ref{alg:join-tree-transformation}.
\end{proofof}

\subsection{Orienting a Partially Directed Join Tree}
\label{sec:orienting-join-tree-and-cliques}

Before we discuss Algorithm \ref{alg:orient-join-tree}, we state and prove a useful set identity. 

\begin{algorithm}[!t]
\caption{orientTree}
\label{alg:orient-join-tree}
\begin{algorithmic}[1]
    \Require Partially directed join tree $\g[T] = (\mathbf{C}, \mathbf{E})$, and node $\g[C]_0 \in \mb{C}$ for an ancestral partial mixed graph $\g$ with a chordal skeleton, with edge orientations closed under { \ref{R1}, \ref{R2}, \ref{R8}, \ref{R11}} and such that $\g$ is without minimal collider paths.
    \Ensure Directed join tree $\g[T]^{\prime} = (\mathbf{C}, \mathbf{E}^{\prime})$.
    
    \State $\g[T]^{\prime} \gets \texttt{transformTree}(\g[T], \g[C]_0)$
    
    \While {an undirected edge is in $\g[T]^{\prime}$}
        \State  Let $p = \langle \g[C]_{j_1}, \dots, \g[C]_{j_k} \rangle, k >1$ be a longest undirected path in $\g[T]^{\prime}$
        \If {$\g[C]_{j_1} \in \An(\g[C]_0, \g[T]^{\prime})$ or $\exists \ \g[C]_{j} \in \mb{C}$, such that  $\g[C]_j \in \Pa(\g[C]_{j_1}, \g[T]^{\prime})$} 
            \State  orient $p$ as $\g[C]_{j_1} \to \dots \to \g[C]_{j_k}$ in $\g[T]^{\prime}$ 
            \Else 
            \State orient $p$ as $\g[C]_{j_1} \leftarrow \dots \leftarrow \g[C]_{j_k}$ in $\g[T]^{\prime}$
        \EndIf
    \EndWhile  
    \State \Return  $\g[T]^{\prime}$
\end{algorithmic}
\end{algorithm}

\begin{proposition}
\label{prop:set-diff-subset}
For any three subsets $A, B, C \subseteq \mb{V}$ of some finite set $\mb{V}$ i.e., $\abs{\mb{V}} < \infty$, we have that
\begin{align*}
    B \setminus A &\subseteq (B \setminus C) \cup (C \setminus A).
\end{align*}
\end{proposition}

\begin{proofof}[Proposition \ref{prop:set-diff-subset}]
Since $\mb{V}$ is finite, set complements are well-defined. Specifically, $C^c \cup C = \mb{V}$. Further, we know that $B \setminus A = B \cap A^c$. Then,
\begin{align*}
    B \setminus A &= (B \setminus A) \cap \mb{V} \\
    &= (B \setminus A) \cap (C^c \cup C) \\
    &= \left((B \setminus A) \cap C^c)\right) \cup \left((B \setminus A) \cap C)\right) \\
    &= \left(B \cap A^c \cap C^c)\right) \cup \left(B \cap A^c \cap C)\right) \\
    &= \left(B \cap C^c \cap A^c)\right) \cup \left(C \cap A^c \cap B)\right) \\
    &= \left((B \setminus C) \cap A^c)\right) \cup \left((C \setminus A) \cap B)\right) \\
    &\subseteq (B \setminus C) \cup (C \setminus A)
\end{align*}
\end{proofof}

\begin{lemma}\label{lem:correct-directed-tree}
    Let $\g$ be an ancestral partial mixed graph with a chordal skeleton such that $\g$ has no minimal collider paths such that the orientations in $\g$ are closed under { \ref{R1}, \ref{R2}, \ref{R8}, \ref{R11}}. Let $\g[T]_0$ be a partially directed join tree for $\g$ as defined in Definition \ref{def:partially-directed-join-tree} and let $\g[C]_0$ be a node in $\g[T]_0$.  Furthermore, let $\g[T]_1 = \texttt{transformTree}(\g[T]_0, \g[C]_0)$ (Algorithm \ref{alg:join-tree-transformation}) and  $\g[T] = \texttt{orientTree}(\g[T]_0, \g[C]_0)$ (Algorithm \ref{alg:orient-join-tree}).  
    Also, let $\pi_{\g[T]}$ be a partial order compatible with $\g[T]$. Then the following hold:
    \begin{enumerate}[label = (\roman*)]
        \item\label{correct-directed-tree:case1} $\g[T]$ is a  directed join tree for $\g$ that does not contain  colliders and $\An(\g[C]_0, \g[T]_1) = \An(\g[C]_0, \g[T])$.
        \item\label{correct-directed-tree:case2} $\pi_{\g[T]}$ induces a edge orientations that are compatible with $\g$. Call this induced graph $\g_{\pi}$ (Definition \ref{def:jointree-inducedorder}).
         \item\label{correct-directed-tree:case4} For any node $A \in \g[C]_0$ there are no new edge marks into $A$ in $\g_{\pi}$ compared to $\g$. Furthermore, for any pair of nodes $A,B \in \g[C]_0$, $\langle A, B \rangle$ is of the same form in $\g$ and $\g_{\pi}.$
        \item\label{correct-directed-tree:case8} If path $\langle A, V_1, \dots, V_k, D \rangle$, $k \ge 1$ is in $\g_{\pi}$ such that $\{A, V_1, \dots, V_k \} \subseteq \g[C]_i$, and  such that $\{V_1, \dots, V_k, D \} \subseteq \g[C]_j$, for some maximal cliques $\g[C]_i, \g[C]_j$ in $\g_{\pi}$, and also $A \notin \Adj(D, \g_{\pi})$, then at least one of the following holds:
        \begin{itemize}
            \item $V_t \to D$ is in $\g_{\pi}$, for all $t \in \{1, \dots, k\}$.
            \item $V_t \to A$  is in $\g_{\pi}$, for all $t \in \{1, \dots, k\}$.
        \end{itemize} 
         \item\label{correct-directed-tree:case51} If $A \bulletarrow B \to C$ is in $\g_{\pi}$  and $A \in \Adj(C, \g_{\pi})$, where $B \to C$ is induced by $\pi_{\g[T]}$, then $A \to C$ is in $\g_{\pi}$.
        \item\label{correct-directed-tree:case52} If $A \to B \bulletarrow C$ is in $\g_{\pi}$ and $A \in \Adj(C, \g_{\pi})$, where $A \to B$ is induced by $\pi_{\g[T]}$, then $A \to C$ is in $\g_{\pi}$.
        \item\label{correct-directed-tree:case3} $\g_{\pi}$ is ancestral, and edge orientations in $\g_{\pi}$ are   closed under  \ref{R2}, and \ref{R8}. Furthermore, $\g_{\pi}$ contains no minimal collider paths and neither does any directed mixed graph $\g[M]$  that is represented by $\g_{\pi}$. 
    \end{enumerate}
\end{lemma}

\begin{proofof}[Lemma \ref{lem:correct-directed-tree}]
\begin{enumerate}

\item[\ref{correct-directed-tree:case1}] We have, $\g[T]_1 = \texttt{transformTree}(\g[T]_0, \g[C]_0)$. By  Corollary \ref{cor:algo-transformtree1-result}  there are no paths in $\g[T]_1$ that are of the forms: 
    \begin{itemize}
    \item  $\g[C]_{i_1} \to \g[C]_{i_2} \leftarrow \g[C]_{i_3}$, or
        \item $\g[C]_{i_1} \to \g[C]_{i_2} - \dots - \g[C]_{i_k} \leftarrow \g[C]_{i_{k+1}}$, $k > 2$, or 
        \item $\g[C]_{i_1} \to \g[C]_{i_2} - \dots - \g[C]_{i_k} \to \dots \to \g[C]_{i_r} $, $r \geq k > 2$, where  $\g[C]_{i_r} \equiv \g[C]_0$.
    \end{itemize}
Orienting paths as in Algorithm \ref{alg:orient-join-tree} will not create colliders in $\g[T]$. Further, we will not create new ancestors for $\g[C]_{0}$ as we always orient paths away from existing ancestors of $\g[C]_0$.
By construction of Algorithm \ref{alg:orient-join-tree}, all ancestors of $\g[C]_0$ in $\g[T]_1$ are also ancestors of $\g[C]_0$ in $\g[T]_0$. Therefore, $\An(\g[C]_0, \g[T]_1) = \An(\g[C]_0, \g[T])$.

\item[\ref{correct-directed-tree:case2}] Note that $\pi_{\g[T]}$ only induces directed edges in $\g_{\pi}$ by Definition \ref{def:jointree-inducedorder}. Hence, to show that edge orientations induced by $\pi_{\g[T]}$ are compatible with $\g$, we need to show that it is possible to orient $A \to B$ for every $A \in \g[C]_i \cap \g[C]_j$, $B \in \g[C]_j \setminus \g[C]_i$ {whenever} $\pi_{\g[T]}(\g[C]_i, \g[C]_j)$ holds in $\g[T]$. 

For any two maximal cliques $\g[C]_i$ and $\g[C]_j$ in $\g$ such that $\pi_{\g[T]}(\g[C]_i, \g[C]_j)$ and $\g[C]_i$, $\g[C]_j$ are adjacent in $\g[T]$,  $\g[C]_i \to \g[C]_j$ is in $\g[T]$ either because $\gamma(\g[C]_i, \g[C]_j)$ holds, or because this edge got oriented by Algorithm \ref{alg:orient-join-tree}. In the former case, the induced orientations in $\g_{\pi}$ are surely compatible with orientations already in $\g$. In the latter case, it must be that $\neg \gamma(\g[C]_i, \g[C]_j)$ and $\neg \gamma(\g[C]_j, \g[C]_i)$. With $\neg \gamma(\g[C]_j, \g[C]_i)$, the contraposition of Lemma \ref{lemma:order-cliques-1} tells us that there is no edge $A \arrowbullet B$, $A \in \g[C]_i \cap \g[C]_j, B \in  \g[C]_j \setminus \g[C]_i$ 
{in $\g$. Therefore, all such edges in $\g$ must be either $A \circbullet B$ or $A \to B$. Thus, it is possible} to orient all such edges as $A \to B$ in {$\g_{\pi}$}.

For any two maximal cliques $\g[C]_i$ and $\g[C]_k$ in $\g$ such that {$\pi_{\g[T]}(\g[C]_i, \g[C]_k)$} and $\g[C]_i \cap \g[C]_k \neq \varnothing$, but $\g[C]_i$ and $\g[C]_k$ are not adjacent in $\g[T]$, there is a path $p = \langle \g[C]_i = \g[C]_{j_1}, \g[C]_{j_2}, \dots, \g[C]_{j_r} = \g[C]_k \rangle$, $r >2$ of the form $\g[C]_i \to \dots \to \g[C]_k$ in $\g[T]$.
We will now prove the rest of this claim by using an induction argument on the length of $p$. For clarity and conciseness, below we will use the following shorthand $\Lambda_{j_{t}j_{s}} \to \g[C]_{j_s} \setminus \g[C]_{j_t}$ for $t,s \in \{1, \dots, r\}, t \neq r$, to say that it is \emph{possible} to orient all edges $\langle A,B \rangle$, such that $A \in  \Lambda_{j_{t}j_{s}}$, $B \in \g[C]_{j_s} \setminus \g[C]_{j_t} $ as $A \to B$ in $\g_{\pi}$. 

For the base of the induction suppose  that $r = 3$ i.e., $p$ is of the form $\g[C]_i \to \g[C]_{j_2} \to \g[C]_k$. If $\Lambda_{ik} = \varnothing$, we are done. Hence, suppose that $\Lambda_{ik} \neq \varnothing$. 

From previous argument for adjacent nodes, we have that $\Lambda_{ij_2} \to \g[C]_{j_2} \setminus \g[C]_i$ and $\Lambda_{j_2k} \to \g[C]_k \setminus \g[C]_{j_2}$.
By the join tree running intersection property, we have that $\Lambda_{ik} \subseteq  \g[C]_{j_2}$. Therefore, $\Lambda_{ik} \subseteq \Lambda_{ij_2}$ and $\Lambda_{ik} \subseteq \Lambda_{j_2k}$. Then, to show that $\Lambda_{ik} \to \g[C]_k \setminus \g[C]_i$ it is enough to show that $\g[C]_k \setminus \g[C]_i \subseteq (\g[C]_{k} \setminus \g[C]_{j_2}) \cup (\g[C]_{j_2} \setminus \g[C]_i)$. 
{This follows from Proposition \ref{prop:set-diff-subset} as the node set $\mb{V}$ is finite.}

For the induction hypothesis suppose that the claim holds for every path of length $t$, $t \ge 3$. We will show that then it also holds for the path of length $t +1.$ Let $r = t+1$ i.e., $p = \langle \g[C]_i, \g[C]_{j_2}, \dots \g[C]_{j_t}, \g[C]_k \rangle$.  If $\Lambda_{ik} = \varnothing$, we have nothing to prove, so suppose $\Lambda_{ik}\neq \varnothing$ The goal is then again to show that $\Lambda_{ik} \to \g[C]_k \setminus \g[C]_i$. 

We know that $\Lambda_{jt_k} \to \g[C]_k \setminus \g[C]_{j_t}$ holds, and from the induction hypothesis, we also know that $\Lambda_{ij_{t}} \to \g[C]_{j_t} \setminus \g[C]_i$ holds. 
By the intersection property, we also have that $\Lambda_{ik} \subseteq C_{j_l},$ for every $l \in \{2, \dots, t\}$. Therefore, $\Lambda_{ik} \subseteq \Lambda_{ij_t}$, and $\Lambda_{ik} \subseteq \Lambda_{j_tk}$. 
Similar to the base case, it is enough to show that $\g[C]_{k} \setminus \g[C]_i \subseteq (\g[C]_k \setminus \g[C]_{j_t}) \cup (\g[C]_{j_t} \setminus \g[C]_i)$. 
{This, of course, follows from Proposition \ref{prop:set-diff-subset} like before.}

\item[\ref{correct-directed-tree:case4}] First, note that by construction, $\g[T]_1$ is a partially directed join tree for $\g$ (Corollary \ref{cor:algo-transformtree1-result}). Hence, case \ref{correct-directed-tree:case2} implies that the only way to obtain new edge marks into $A$ in $\g_{\pi}$ is by adding new ancestors of $\g[C]_0$ in $\g[T]$, compared to $\g[T]_1$. But we know by case  \ref{correct-directed-tree:case1}, that no such edge marks are added. 

For the statement about the form of $\langle A, B \rangle$, note that an edge is of different form in $\g_{\pi}$ compared to $\g$, only if its orientation is induced by $\pi_{\g[T]}$. Also, since $\g[T]_1$ is a partially directed join tree for $\g$, only orientations added to $\g[T]_1$ to create $\g[T]$ would be able to orient $\langle A, B \rangle$ through $\pi_{\g[T]}$.

Since $A,B \in \g[C]_0$, the only way to orient $\langle A, B \rangle$ in some way in $\g_{\pi}$ is if there is a clique $\g[C]_i$, such that $\g[C]_i$ is an ancestor of $\g[C]_0$ in $\g[T]$, but not in $\g[T]_1$. By case \ref{correct-directed-tree:case1}, $\An(\g[C]_0, \g[T]) \setminus \An(\g[C]_0, \g[T]_1) = \varnothing$. Hence, $\langle A, B\rangle$ must be of the same form in both $\g_{\pi}$ and $\g$.

\item[\ref{correct-directed-tree:case8}] Note that the mutually exclusive and collectively exhaustive options for $\g[C]_i$ and $\g[C]_j$ are 
\begin{enumerate}[label = (\alph*)]
    \item $\pi_{\g[T]}(\g[C]_i, \g[C]_j)$: Here, $V_t \to  D$ for all $t \in \{1, \dots, k \}$ by Definition \ref{def:jointree-inducedorder} and cases \ref{correct-directed-tree:case1}, and \ref{correct-directed-tree:case2}.
    \item  $\pi_{\g[T]}(\g[C]_j, \g[C]_i)$: Here, $V_t \to  A$ for all $t \in \{1, \dots, k \}$ by Definition \ref{def:jointree-inducedorder} and cases \ref{correct-directed-tree:case1}, and \ref{correct-directed-tree:case2}.
    \item  $\neg \pi_{\g[T]}(\g[C]_i, \g[C]_j) \land \neg \pi_{\g[T]}(\g[C]_j, \g[C]_i)$: Here, by case \ref{correct-directed-tree:case1} there exists a maximal clique $\g[C]_l$ in $\g_{\pi}$ such that the path between $\g[C]_i$ and $\g[C]_j$ in $\g[T]$ is of the form $\g[C]_i \leftarrow \dots \leftarrow \g[C]_l \to \dots \to \g[C]_j$. By the running intersection property $\{V_1, \dots, V_k \} \subseteq \g[C]_l$. Case \ref{correct-directed-tree:case2} implies that we have that $\pi_{\g[T]}(\g[C]_l, \g[C]_i)$ and $\pi_{\g[T]}(\g[C]_l, \g[C]_j)$. Furthermore, at least one of the nodes $A, D$ is not in $\g[C]_l$ because $A \notin \Adj(D, \g_{\pi})$.
    {Without loss of generality, assume $A \notin \g[C]_l$. Then $\pi_{\g[T]}(\g[C]_l, \g[C]_i)$ implies that $V_t \to A$ is in $\g_{\pi}$ for all $t \in \{1, \dots, k\}$. A symmetric argument holds when $D \notin \g[C]_l$.}
\end{enumerate}

\item[\ref{correct-directed-tree:case51}] By assumption, $A \bulletarrow B \to C$ is in $\g_{\pi}$, $A \in \Adj(C, \g)$ and $B \to C$ is induced by $\pi_{\g[T]}$. Then there are maximal cliques $\g[C]_i, \g[C]_j$, and $\g[C]_k$ in $\g_{\pi}$ such that the following holds:
\begin{itemize}
    \item $\g[C]_i \supseteq \{B\}$, and $C \notin \g[C]_i$,
    \item $\g[C]_j \supseteq \{B,C\}$, and $\pi_{\g[T]}(\g[C]_i, \g[C]_j)$, and 
    \item $\g[C]_k \supseteq \{A,B, C\}$.
\end{itemize}

Next we consider whether $A$ belongs to $\g[C]_i, \g[C]_j$. We have the following cases:  \ref{aj} $A \in \g[C]_j \setminus \g[C]_i$,  \ref{ak} $A \notin \g[C]_i \cup \g[C]_j$, \ref{aall}$ A \in \g[C]_j \cap \g[C]_i$, or \ref{ai} $A \in \g[C]_i \setminus \g[C]_j$.
For the rest of the proof, we show that the cases \ref{aj} and \ref{ak} are in fact not possible, since they lead to a contradiction, while cases \ref{aall} and \ref{ai} lead us to conclude that $A \to C$ is in  $\g_{\pi}$.
\begin{enumerate}[label = (\alph*)]
\item\label{aj} Since $A \bulletarrow B$ is in $\g_{\pi}$, we know that $A$ cannot be in $\g[C]_j \setminus \g[C]_i$.
\item\label{ak} $A \in \g[C]_k \setminus ( \g[C]_i \cup \g[C]_j)$: Since $ B \in \g[C]_k \cap \g[C]_i$ and $A \bulletarrow B$ is in $\g_{\pi}$, we know that $\neg \pi_{\g[T]}(\g[C]_i, \g[C]_k)$ and $\neg \pi_{\g[T]}(\g[C]_j, \g[C]_k)$. Since we also know that $\pi_{\g[T]}(\g[C]_i, \g[C]_j)$, let us consider the options for paths between $\g[C]_i, \g[C]_j$ and $\g[C]_k$. Let $p_{ij}$ be the path from $\g[C]_i$ to $\g[C]_j$ in $\g[T]$, $p_{ik}$ the path from $\g[C]_i$ to $\g[C]_k$ and $p_{jk}$ the path from $\g[C]_j$ to $\g[C]_k$ in $\g[T]$. The only options are that: \ref{correct-directed-tree:case51:1} $\g[C]_i$ is on $p_{jk},$ or that \ref{correct-directed-tree:case51:2} a node from $p_{ij}$ other than $\g[C]_i$ is on $p_{ik}$.
\begin{enumerate}[label = (\arabic*)]
    \item\label{correct-directed-tree:case51:1} Since $\g[C]_{k} \cap \g[C]_j \not\subseteq \g[C]_i$, the running intersection property of $\g[T]$ implies that $\g[C]_i$ cannot be on $p_{jk}$. 
    \item\label{correct-directed-tree:case51:2} $\pi_{\g[T]}(\g[C]_i, \g[C]_j)$ and  $\neg \pi_{\g[T]}(\g[C]_i, \g[C]_k)$ together imply that $\g[C]_k$ is not on $p_{ij}$, and also that no other node from $p_{ij}$ except $\g[C]_i$ is on $p_{ik}$.
\end{enumerate}
\item\label{aall} If $A \in \g[C]_j \cap \g[C]_i$, then $A \to C$ is in $\g_{\pi}$ by $\pi_{\g[T]}(\g[C]_i, \g[C]_j)$. 
\item\label{ai} $A \in (\g[C]_i \cap \g[C]_k) \setminus \g[C]_j$, then as above,  let $p_{ij}$ be the path from $\g[C]_i$ to $\g[C]_j$ in $\g[T]$, $p_{ik}$ the path from $\g[C]_i$ to $\g[C]_k$ and $p_{jk}$ the path from $\g[C]_j$ to $\g[C]_k$ in $\g[T]$. The only options are that: \ref{correct-directed-tree:case51:3} $\g[C]_i$ is on $p_{jk},$ or that \ref{correct-directed-tree:case51:4} a node from $p_{ij}$ other than $\g[C]_i$ is on $p_{ik}$. 
\begin{enumerate}[label = (\arabic*)]
    \item\label{correct-directed-tree:case51:3} Since $\g[C]_{k} \cap \g[C]_j \supseteq \{B,C\} \not\subseteq \g[C]_i$, the running intersection property of $\g[T]$ implies that $\g[C]_i$ cannot be on $p_{jk}$. 
    \item\label{correct-directed-tree:case51:4} Since $\pi_{\g[T]}(\g[C]_i, \g[C]_j)$, having  $\g[C]_k$ on $p_{ij}$, implies $\pi_{\g[T]}(\g[C]_i, \g[C]_k)$ and therefore, $A \to C$ is in $\g_{\pi}.$ Similarly, having any node  from $p_{ij}$ except $\g[C]_i$  on $p_{ik}$ implies the same thing.
\end{enumerate}
\end{enumerate}

\item[\ref{correct-directed-tree:case52}] By assumption, $A \to B \bulletarrow C$ is in $\g_{\pi}$, $A \in \Adj(C, \g)$ and $A \to B$ is induced by $\pi_{\g[T]}$. Then there are maximal cliques $\g[C]_i, \g[C]_j$, and $\g[C]_k$ in $\g_{\pi}$ such that the following holds:
\begin{itemize}
    \item $\g[C]_i \supseteq \{A\}$, and $B \notin \g[C]_i$,
    \item $\g[C]_j \supseteq \{A,B\}$, and $\pi_{\g[T]}(\g[C]_i, \g[C]_j)$, and 
    \item $\g[C]_k \supseteq \{A,B, C\}$.
\end{itemize}

Next we consider whether $C$ belongs to $\g[C]_i, \g[C]_j$. We have the following cases:   \ref{ci} $C \in \g[C]_i \setminus \g[C]_j$, \ref{call}$ C \in \g[C]_j \cap \g[C]_i$, \ref{cj} $C \in \g[C]_j \setminus \g[C]_i$, or  \ref{ck} $C \notin \g[C]_i \cup \g[C]_j$.
For the rest of the proof, we show that the cases \ref{ci} and \ref{call} are in fact not possible, since they lead to a contradiction, while cases \ref{cj} and \ref{ck} lead us to conclude that $A \to C$ is in  $\g_{\pi}$.
\begin{enumerate}[label = (\alph*)]
\item\label{call} If $C \in \g[C]_j \cap \g[C]_i$, then $B \in \g[C]_j \setminus \g[C]_i$,$\pi_{\g[T]}(\g[C]_i, \g[C]_j)$ and $B \bulletarrow C$  together imply a contradiction. 
\item\label{ci} $C \in (\g[C]_i \cap \g[C]_k) \setminus \g[C]_j$. Since $B \in \g[C]_k \setminus \g[C]_i$, and $B \bulletarrow C$ is in $\g_{\pi}$, we have that $\neg \pi_{\g[T]}(\g[C]_i, \g[C]_k)$. Now, let $p_{ij}$ be the path from $\g[C]_i$ to $\g[C]_j$ in $\g[T]$, $p_{ik}$ the path from $\g[C]_i$ to $\g[C]_k$ and $p_{jk}$ the path from $\g[C]_j$ to $\g[C]_k$ in $\g[T]$. The only options are that: \ref{correct-directed-tree:case52:3} $\g[C]_i$ is on $p_{jk},$ or that \ref{correct-directed-tree:case52:4} a node from $p_{ij}$ other than $\g[C]_i$ is on $p_{ik}$. 
\begin{enumerate}[label = (\arabic*)]
    \item\label{correct-directed-tree:case52:3} Since $\g[C]_{k} \cap \g[C]_j \supseteq \{A,B\} \not\subseteq \g[C]_i$, the running intersection property of $\g[T]$ implies that $\g[C]_i$ cannot be on $p_{jk}$. 
    \item\label{correct-directed-tree:case52:4} Since $\pi_{\g[T]}(\g[C]_i, \g[C]_j)$, having  a node from $p_{ij}$ other than $\g[C]_i$ on $p_{ik}$ would imply $\pi_{\g[T]}(\g[C]_i, \g[C]_k)$ which is a contradiction.
\end{enumerate}

\item\label{cj} $C \in (\g[C]_j \cap \g[C]_k) \setminus \g[C]_i$. Since $C \in \g[C]_j \setminus \g[C]_i$ and $A \in \g[C]_i \cap \g[C]_j$, then $\pi_{\g[T]}(\g[C]_i, \g[C]_j)$ implies $A \to C$ is in $\g_{\pi}.$
\item\label{ck} $C \in \g[C]_k \setminus ( \g[C]_i \cup \g[C]_j)$. 
Let us consider the options for paths between $\g[C]_i, \g[C]_j$ and $\g[C]_k$. Let $p_{ij}$ be the path from $\g[C]_i$ to $\g[C]_j$ in $\g[T]$, $p_{ik}$ the path from $\g[C]_i$ to $\g[C]_k$ and $p_{jk}$ the path from $\g[C]_j$ to $\g[C]_k$ in $\g[T]$. The only options are that: \ref{correct-directed-tree:case52:1} $\g[C]_i$ is on $p_{jk},$ or that \ref{correct-directed-tree:case52:2} a node from $p_{ij}$ other than $\g[C]_i$ is on $p_{ik}$.
\begin{enumerate}[label = (\arabic*)]
    \item\label{correct-directed-tree:case52:1} Since $\g[C]_{k} \cap \g[C]_j \supseteq \{A,B\} \not\subseteq \g[C]_i$, the running intersection property of $\g[T]$ implies that $\g[C]_i$ cannot be on $p_{jk}$. 
    \item\label{correct-directed-tree:case52:2} If a node on $p_{ij}$ other than $\g[C]_i$ is on $p_{ik}$,t that implies that $\pi_{\g[T]}(\g[C]_i, \g[C]_k)$. Since $A \in \g[C]_i \cap \g[C]_k$ and $C \in \g[C]_k \setminus \g[C]_i$, we have that $A \to C$ is in $\g_{\pi}$.
\end{enumerate}
\end{enumerate}
\end{enumerate}
\end{proofof}

\subsection{Orienting a Clique} \label{sec:clique}

\begin{lemma}
\label{lemma:existence-spo2}
Suppose an ancestral partial mixed graph $\g = (\mb{V,E})$ with edge orientations   closed under  \ref{R2} and \ref{R8} is a clique that contains no edges of the form $\circarrow$ or $\leftrightarrow$. 
Consider edge $A \circcirc B$ in $\g$ for some $A,B \in \mb{V}$. Then there is are total orderings $\pi_1$ and $\pi_2$ of $\mb{V}$ compatible with $\g$, such that $\g_{\pi_1}$ and $\g_{\pi_2}$ are DAGs and such that  $A \to B$ is in $\g_{\pi_1}$ and $A \leftarrow B$ is in $\g_{\pi_2}$.
\end{lemma}

\begin{proofof}[Lemma \ref{lemma:existence-spo2}]
We will show how to obtain $\pi_1$ using the sink elimination Algorithm of \citet{dor1992simple}. The proof for $\pi_2$ is analogous.

Since $\g$ is ancestral and therefore, acyclic, there will always be at least one node $V$ in $\g$ such that there are no edges out of $V$ in $\g$. This type of node is called a potential sink node according to \citet{dor1992simple} algorithm since $\g$ is a clique. 
    
To obtain $\pi_1$, we consider whether $B$ is a potential sink node in $\g$. 
\begin{enumerate}[label = (\roman*)]

    \item\label{step1} If $B$ is a potential sink, let $\pi^{(1)}$ be a partial ordering that only states that  $\pi^{(1)}(W,B)$ for every node $W \in \mb{V}$.
    Then consider, the induced subgraph $\g_{\mb{V} \setminus \{B\}} = (\mb{V}_{-B}, \mb{E}_{-B})$ where $\mb{V}_{-B} = \mb{V} \setminus \{B\}$ and $\mb{E}_{-B} = \{(S, T) \in \mb{E} \mid S \neq B, T \neq B\}$.
    $\g_{\mb{V} \setminus \{B\}}$ is also a clique that is ancestral and does not contain $\circarrow$ or $\leftrightarrow$ edges. We can then apply the Algorithm of \citet{dor1992simple} to  $\g_{\mb{V} \setminus \{B\}}$ to obtain a total ordering $\pi^{(2)}$ of $\mb{V}\setminus \{B\}$. We can construct $\pi_1$ as follows:
    \begin{align*}
        \pi^{(1)}(V_1, V_2) &\implies \pi_1(V_1, V_2), \\
        \pi^{(2)}(V_1, V_2) &\implies \pi_1(V_1, V_2).
    \end{align*}
    It is easy to see that $\pi_1$ is compatible with $\g$ by construction and $\g_{\pi_1}$ is a DAG with $A \to B$.
        
    \item\label{step2} If $B$ is not a potential sink, then since a potential sink node must exist in $\g$, we only need to show that there is a potential sink node that is different from $A$ in $\g$. Note that since $B$ is not a potential sink there is a node $B \to V_{2}$, for some $V_2 \in \mb{V}$ in $\g$. 
    
    If $A$ was the only potential sink node in $\g$, that would mean that there is a path $B \to V_2 \to \dots \to V_k \to A$, $k \ge 2$ in $\g$. However, since $\g$ is an ancestral clique with edge orientations    closed under  \ref{R2} and \ref{R8}, the successive edges $B \to V_3, \dots B \to V_k$, $B \to A$ are in $\g$. This contradicts $A \circcirc B$ being in $\g$. Hence, there is at least one potential sink node that is different from $A$ in $\g$. 
    
    Suppose this potential sink node in $\g$ that is different from $A$ is called $V$. Let $\pi^{(1)}$ be a partial ordering that only states that  $\pi^{(1)}(W,B)$ for every node $W \in \mb{V}$.  Then consider, the induced subgraph $\g_{\mb{V} \setminus \{V\}}$ (defined like before) which is also an ancestral clique that does not contain $\circarrow$ or $\leftrightarrow$ edges. 
    
    If $B$ is a potential sink in $\g_{\mb{V} \setminus \{V\}}$, we can apply step \ref{step1} to $\g_{\mb{V}\setminus \{V\}}$ to obtain a total ordering $\pi^{(2)}$ of $\mb{V} \setminus \{V\}$ that is compatible with $\g$. Then 
    we extend $\pi^{(1)}$ to $\pi_2$ using $\pi^{(2)}$, as follows 
    \begin{align*}
        \pi^{(1)}(V_1, V_2) &\implies \pi_2(V_1, V_2), \\
        \pi^{(2)}(V_1, V_2) &\implies \pi_2(V_1, V_2).
    \end{align*}
    This is the desired ordering: $\pi_2$ is compatible with $\g$ by construction and $\g_{\pi_2}$ is a DAG with $A \to B$.

    If $B$ is not a potential sink in $\g_{\mb{V} \setminus \{V\}}$, we can apply step \ref{step2} on  $\g_{\mb{V} \setminus \{V\}}$  to obtain $\pi^{(2)}$ and recursively continue obtaining $\pi^{(3)}, \dots, \pi^{(l)}$ until $\g_{\mb{V} \setminus \mb{S}},$ for some $\mb{S} \supset \{V\}$ such that $B$ is a potential sink in $\g_{\mb{V} \setminus \mb{S}}$. Then we apply step \ref{step1} to $\g_{\mb{V} \setminus \mb{S}}$ which gives us partial ordering $\pi^{(l+1)}$. Finally, we construct the desired total ordering $\pi_2$, where for any $V_1, V_2 \in \mb{V}$: 
     \begin{align*}
        \pi^{(j)}(V_1, V_2) &\implies \pi_2(V_1, V_2) \qquad \ \text{for all} \ j \in \{1, \dots, l+1\}.
    \end{align*}
\end{enumerate}  
\end{proofof}

\begin{lemma}
\label{lemma:existence-spo}
Suppose an ancestral partial mixed graph $\g = (\mb{V,E})$ with edge orientations   closed under  \ref{R2} and \ref{R8} is a clique.
Consider the graph $\g^{'}$ obtained from $\g$ in one of the two following ways:
\begin{enumerate}[label=(\alph*)]
    \item\label{MAGorientation1} Orient all variant edge marks ($\circ$) as arrowheads. That is, orient edges of the form $V_i \circcirc V_j$ and of the form $V_i \circarrow V_j$ as $V_i \leftrightarrow V_j$.
    \item\label{MAGorientation2} Choose an edge $A \circbullet B$ in $\g$. Then 
    \begin{enumerate}[label = (\arabic*)]
        \item\label{Clique:case21} orient $A \to B$ in $\g$, and
        \item\label{Clique:case22} for all $C$ in $\g$ such that $B \to C$ is in $\g$, orient $A \to C$, and
        \item\label{Clique:case23} for all $D$ in $\g$ such that $B \circarrow D$ or $B \leftrightarrow D$ is in $\g$, orient the edge mark at $D$ on edge $\langle A , D \rangle$ as an arrowhead, that is, $A \bulletarrow D$ and orient $B \leftrightarrow D$.
    \end{enumerate}
    Then, orient all remaining $V_i \circarrow V_j$ or $V_i \circcirc V_j$ edges in $\g$ as $V_i \leftrightarrow V_j$.
\end{enumerate}
Then $\g^{'}$ is a MAG represented by $\g$.
\end{lemma}

\begin{proofof}[Lemma \ref{lemma:existence-spo}]
Note that for $\g^{'}$ to be a MAG represented by $\g$ it is enough to show that $\g^{'}$ does not contain directed or almost directed cycles of length $3$. 

For case \ref{MAGorientation1}, we only need to worry about creating almost directed cycles in $\g^{'}$. We know these cannot be created in $\g^{'}$ since, $\g$ cannot contain $V_{1} \to V_{2} \to V_{3}$, and $V_1 \circcirc V_3$ for any three nodes $V_1, V_2, V_3$ due to orientations in $\g$ being    closed under  \ref{R2} and \ref{R8}.

For case \ref{MAGorientation2}, note that steps in \ref{Clique:case21}-\ref{Clique:case23} ensure that orientations under \ref{R2} and \ref{R8} are closed after adding $A \to B$. Hence, as long as steps \ref{Clique:case21}-\ref{Clique:case23} can be performed and do not create a directed or almost directed cycle, the remainder of the proof follows by case \ref{MAGorientation1} above. 

By assumption, step \ref{Clique:case21} can be performed. Additionally, step \ref{Clique:case21} cannot in itself create a directed or almost directed cycle since $\g$ is a clique with edge mark orientations closed under \ref{R2} and \ref{R8}.

As for step \ref{Clique:case22}, note that for any $C$ in $\g$, such that $A \circbullet B \to C$ is in $\g$, $A \circbullet C$ must be in $\g$ again, due to edge mark orientations being closed under \ref{R2} and \ref{R8} in $\g$.
Hence, step \ref{Clique:case22} can be performed. 

Furthermore, completing step \ref{Clique:case22} cannot create a directed cycle. To see why, observe that a directed cycle would imply that $C \to E \to A$ was already in $\g$ for some node $E$. This is because in steps \ref{Clique:case21} and \ref{Clique:case22} we do not create any new arrowheads into $A$ and do not orient any edge marks on edges that do not contain $A$.

Since we know $C \to E \to A$ and $A \circbullet C$ cannot both be in $\g$, we know that orienting $A \to C$ does not create a directed cycle. Using a similar reasoning we can conclude that neither $C \to F \bulletarrow A$, nor $C \bulletarrow F \to A$ can be in $\g$, for any node $F$, so orienting $A \to C$ also does not create an almost directed cycle. 

Lastly, consider step \ref{Clique:case23}. We first show that it can be performed, that is that $A \leftarrow D$ cannot occur for the mentioned configuration. Note that if we have $A \circbullet B$ and $B \bulletarrow D$ are in $\g$ we cannot also have  $A \leftarrow D$  in $\g$ as that would imply that edge mark orientations in $\g$ are not   closed under  \ref{R2}.
Hence, it is possible to orient edge $\langle A, D \rangle$ into $D$ i.e., as $A \bulletarrow D$, and by assumption, it is also possible to orient $B \leftrightarrow D$.

Now we only need to show that completing step \ref{Clique:case23} does not create an almost directed cycle. Orienting $B \circarrow D$ as $B \leftrightarrow D$ can only create an almost directed cycle if edge mark orientations in $\g$ are not   closed under  \ref{R8}. 
Additionally, the only other way that completing step \ref{Clique:case23} could create an almost directed cycle, is if in completing step \ref{Clique:case23} we oriented $A \arrowcirc D$ as $A \leftrightarrow D$. But this type of an almost directed cycle would imply that $A \to F \to D$ was already in $\g$ for some $F$, which itself implies an almost directed cycle already exists in $\g$, which is a contradiction. 
\end{proofof}

\begin{example}\label{ex:algo4-demo}
 Consider again the graphs in Figure \ref{fig:lemma7-issue-p2}, the essential ancestral graph $\g$ is in Figure \ref{fig:lemma7-issue-p2}(a), the ancestral partial mixed graph $\g^{\prime} = (\mb{V,E})$ is in  where Figure \ref{fig:jointree111} represents and Figure \ref{fig:jointree222} represents the partially oriented join trees for $\g$ and $\g^{\prime}$. From top to bottom, these join trees are  $\g[T]_1$, $\g[T]_2$, and $\g[T]_3$. 

Suppose that we are interested in finding a MAG that contains a particular orientation of edge $B \circcirc C$. 
Note that  \texttt{transformTree}($\g[T]_1, \g[C]_k$) will return join tree $\g[T]_3,$ and so will \texttt{transformTree}($\g[T]_2, \g[C]_k$), and \texttt{transformTree}($\g[T]_3, \g[C]_k$). Then applying, for instance, $\texttt{orientTree}(\g[T]_1, \g[C]_k)$ (Algorithm \ref{alg:orient-join-tree}) results in the directed join tree $\g[T]$ in Figure \ref{fig:jointree-final1}.
Let $\pi_{\g[T]}$ be the partial ordering compatible with $\g[T]$. Then $\pi_{\g[T]}$ induces edge mark orientations in $\g$ as in Definition \ref{def:jointree-inducedorder} to create graph $\g_{\pi}$ in Figure \ref{fig:jointree-final2}. 
Now, we can use the result of Lemma \ref{lemma:existence-spo} to orient $B\circcirc C$ in $\g_{\pi}$ into any of the three options $B \to C$, $B \leftarrow C$, $B \leftrightarrow C$, thereby resulting in a valid MAG represented by $\g^{\prime}$ of Figure \ref{fig:jointree111}.
\end{example}

\nocite{*}
\bibliography{bib-submission}

\end{document}